\documentclass[11pt]{article}
\usepackage{arxiv}
\usepackage{amsmath,amssymb,mathtools}
\usepackage{booktabs}
\usepackage{longtable,tabularx,array}
\usepackage[numbers,sort&compress]{natbib}
\usepackage{graphicx}
\usepackage{float}
\usepackage{caption}
\usepackage{microtype}
\usepackage{etoolbox}
\usepackage{enumitem}
\usepackage{placeins}
\usepackage{needspace}
\usepackage{xcolor}
\usepackage{hyperref}
\usepackage[nameinlink,noabbrev]{cleveref}
\AtBeginEnvironment{thebibliography}{\raggedright}
\renewcommand{\theHequation}{\thesection.\arabic{equation}}

\definecolor{morandiInk}{HTML}{343A40}
\definecolor{morandiSlate}{HTML}{66727F}
\definecolor{morandiSage}{HTML}{8FA39A}
\definecolor{morandiTerracotta}{HTML}{B77D70}
\definecolor{morandiSand}{HTML}{C8B79D}
\definecolor{morandiLavender}{HTML}{9B8FA6}
\definecolor{morandiBlue}{HTML}{7D9AAA}
\definecolor{morandiMist}{HTML}{F5F2ED}
\definecolor{morandiRose}{HTML}{C7A5A1}

\hypersetup{
  colorlinks=true,
  linkcolor=morandiSlate,
  filecolor=morandiSlate,
  citecolor=morandiTerracotta,
  urlcolor=morandiBlue,
  pdftitle={Shared relations, specific choices: conditional sensitivity and competing pathways in language models},
  pdfauthor={Kaixiang Shu}
}

\AtBeginDocument{\setlength{\headheight}{26pt}}

\setlist{itemsep=0.2em,topsep=0.3em}
\renewcommand{\topfraction}{0.92}
\renewcommand{\bottomfraction}{0.85}
\renewcommand{\textfraction}{0.07}
\renewcommand{\floatpagefraction}{0.75}
\makeatletter
\patchcmd{\@maketitle}
  {\begin{tabular}[t]{c}\bf\rule{\z@}{24\p@}\@author\end{tabular}}
  {\ifx\@author\@empty\else\begin{tabular}[t]{c}\bf\rule{\z@}{24\p@}\@author\end{tabular}\fi}{}{}
\patchcmd{\@maketitle}
  {\vskip 0.4in \@minus 0.1in \center{\@date}   \vskip 0.2in}
  {\ifx\@date\@empty\vskip .14in\else\vskip .2in\center{\@date}\vskip .14in\fi}{}{}
\makeatother
\pretocmd{\section}{\Needspace{7\baselineskip}}{}{}
\pretocmd{\subsection}{\Needspace{6\baselineskip}}{}{}
\pretocmd{\subsubsection}{\Needspace{6\baselineskip}}{}{}
\graphicspath{{figures/}{figures/generated/}}

\newcommand{\WienerG}{\ensuremath{\mathsf G}}

\renewcommand{\shorttitle}{Shared relations, specific choices}
\title{Shared relations, specific choices: conditional sensitivity and competing pathways in language models}
\author{Kaixiang Shu\\{\normalfont Independent Researcher}}
\date{}
\let\supplementarymaketitle\maketitle
\begin{document}
\maketitle
\begin{abstract}
How does a relational representation shared across queries drive a choice specific to the current question? We show that shared relational structure acquires its effect on selection through conditional sensitivity and competing pathways. Controlled entity-binding experiments with GPT-2 Small, Qwen2.5-1.5B and Llama-3.2-1B reveal a low-dimensional relational background that is predominantly shared across query roles and transfers to new names. Strong same-domain interactions persist even when another attribute domain is queried. After removing the shared component, restoring independently trained relation directions brings output contrasts closer to their unedited values than sign-reversed or equal-norm random controls in all three models. Yet a transferable relational representation need not have a transferable functional sign. In Qwen2.5, fixing model weights, the training-mean relation vector and edit dose yields opposite selection effects across binary and lexical-mapping prompts with different answer vocabularies. We derive how differences in downstream sensitivity across attribute--query combinations determine the shared direction's local signed selection gain. In a separate eight-condition protocol, mean local selection gain is positive in Qwen2.5 and negative in Llama3.2. Late-component clamps resolve opposing pathway contributions to these gains. In Llama3.2, negative attention-only transmission is partially offset by positive transmission through paths traversing both selected attention and MLP sets. Across distributed inputs, source-coordinate analyses establish distributed factor coverage; hidden-receiver readouts reveal layout-dependent combination; and instance-geometry interventions separate response direction from amplitude. These findings identify how a transferable relational background supports specific choices: its functional sign emerges from conditional sensitivity and the balance of competing downstream paths.
\end{abstract}
\section{Introduction}
\label{sec:introduction}

Consider two facts: Lina is red and Niko is blue. Asking about Lina or Niko changes the required answer while leaving both facts intact. A language model must therefore maintain the attributes distributed across the text and select the one associated with the queried entity. This creates a distinction between the relations the model represents and the relations it currently needs to use. Does its internal organization track only the requested answer, or does it retain a shared relational structure that different queries can access?

Understanding relational computation requires connecting the structure encoded in hidden states to the operations that use it. Probes identify variables decodable from hidden states~\citep{belinkov2022probing}. Binding studies characterize entity--attribute associations~\citep{feng2024binding}, while linear relation decoding links subject representations to object predictions~\citep{hernandez2024linearity}. Direction-based interventions use activation differences or learned concept directions to steer model behavior~\citep{turner2023activation,zou2023representation,li2023iti}. Their reliability varies across inputs~\citep{tan2024steering}, and a direction's discriminative orientation can oppose its causal control effect~\citep{torop2026inverted}. We examine how a relation shared across queries acquires its signed effect on selection, connecting role-shared geometry to conditional sensitivity and competing pathways.

We investigate this question in GPT-2 Small, Qwen2.5-1.5B and Llama-3.2-1B using controlled entity-binding materials, with subject-number agreement as a contrasting task. Independent manipulations of two attributes and the queried entity reveal a low-dimensional relational component at middle depth. This $AB$ component is predominantly shared across query roles and transfers to held-out names. In two-domain materials, strong same-domain interactions persist even when the current question targets the other domain, whereas cross-domain interactions remain small. The shared $AB$ direction offers a concrete target for testing how relational structure participates in selection.

Independent training donors establish a transferable functional effect. After removing a test group's $AB$ component, restoring a donor direction brings the output contrast closer to its original value than sign-reversed or equal-norm random controls in all three models. Yet a transferable relational representation need not have a transferable functional sign. Holding Qwen2.5's weights, training-mean relation vector and edit dose constant produces opposite selection effects in binary prompts with orange/purple answers and lexical-mapping prompts with red/blue answers. The direction's signed effect depends on the computation that receives it and the output contrast being measured.

Conditional sensitivity provides the link between shared structure and specific choices. With equal directional sensitivity in every attribute--query condition, the shared edit would have zero first-order effect on role selection. The factorial contrast of unequal local sensitivities determines its signed selection gain. In a separate eight-condition intervention protocol, mean local selection gain is positive in Qwen2.5 and negative in Llama3.2. Late-component clamps resolve opposing pathway contributions to these gains. In Llama3.2, negative transmission through selected attention-only paths is partially offset by positive transmission through paths traversing both selected attention and MLP sets. Position-level interventions add a complementary account: shared receiving windows strengthen non-additive combination, and candidate-source contributions support the later fact relative to the earlier one.

Complementary source-coordinate measurements locate the factors carried across input positions (Fig.~\ref{fig:framework}). A nonterminal hidden feature depends only on its visible prefix, whereas its source estimate combines that feature with full-instance geometry and calibrated statistics to expose structure at later positions. These measurements distinguish attributes recovered in source spans from relational contrasts exposed directly at hidden receivers. Direct hidden-receiver readouts show how layout changes the benefit of combining positions. Instance-geometry analyses show that better source alignment can accompany poorer downstream alignment. In response prediction, improved directional alignment can coexist with excess amplitude. Together, these results explain how shared relations support specific choices: conditional sensitivity and competing pathways give transferable structure its effect on selection.

\section{Unified measurement framework and controlled experimental design}
\label{sec:methods}

Connecting relational structure to selection and source-coordinate observations requires explicit definitions of the controlled factors, output contrasts and intervention sites.

\subsection{Controlled factors, query roles and output coordinates}
\label{sec:methods-design}
We studied the pretrained base models GPT-2 Small~\citep{radford2019language}, Qwen2.5-1.5B~\citep{qwen2024technical} and Llama-3.2-1B~\citep{meta2024llama32}. Shallow, middle and deep anchors align their representational trajectories (Table~\ref{tab:design}). Layer numbers are one-based and token positions are zero-based. Unless an intervention specifies another node, the hidden receiver is the attention output after the output projection and before residual addition. All model parameters remain fixed.

The materials progressively separate attribute effects, query roles and domain structure. Four-corner groups independently vary two factors, $a,b\in\{-1,+1\}$, while holding the surrounding context fixed. In binding, these are the attributes of two entities; in syntax, they are grammatical-number factors. Subject-number agreement provides a structure-sensitive comparison with established behavioral and causal analyses~\citep{linzen2016syntax,finlayson2021agreement}. Four-corner binding groups hold the queried entity fixed. Crossing each attribute pair with both query roles, $r\in\{-1,+1\}$, yields eight conditions and separates role-shared relations from query-dependent modulation. Disjoint sets of 128 training, 32 development and 64 test name groups separate axis fitting from evaluation. Further panels test whether relational structure extends beyond binary values, persists across different lexical mappings and remains present when the query targets another attribute domain (Fig.~\ref{fig:framework}A,C; Supplementary Section~\ref{sup:s1-design}).

For each eight-condition group, let $m(a,b,r)$ be the logit of candidate 1 minus that of candidate 0, with candidate order fixed across conditions. Its factorial coefficients and the two output contrasts are
\begin{equation}
 m_X=\frac{1}{8}\sum_{a,b,r}\chi_X(a,b,r)m(a,b,r),\qquad
 C=m_A+m_B,\qquad G=m_{AR}-m_{BR},
 \label{eq:output-contrasts}
\end{equation}
where $\chi_X$ is the product of the signs indexed by $X$. We assign $r=+1$ to the query for entity $A$ and $r=-1$ to the query for entity $B$. The common-property response $C$ measures the shared attribute response; $G$ is the symmetric target-directed margin on conflicting-attribute conditions. Keeping the raw candidate contrast before factorial decomposition preserves the distinction between these two responses.

Causal tracing and circuit analysis localize components involved in a behavior~\citep{meng2022locating,conmy2023automated}. Interchange interventions and path patching test how specified variables and routes contribute to that behavior~\citep{geiger2021causal,wang2023ioi,goldowskydill2023pathpatching}. Activation-patching analyses identify the intervention and response metric as determinants of the measured effect~\citep{zhang2024patching}. Our forward interventions therefore specify both the operation and its output contrast. Direction edits modify the middle-layer query state and resume the native forward pass. Donor restoration replaces a removed component with an independently trained direction at a matched norm. A clean clamp replaces the selected component's prediction-position write with its value in the same unedited condition. Source-branch closure instead sets selected post-softmax value contributions to zero. These operations isolate directional sensitivity, component transmission and source-position contributions, respectively.

\begin{table}[!htbp]
\centering\small
\caption{Models and controlled comparisons. Layer anchors identify shallow/middle/deep attention outputs. Complete counterfactual groups are the sampling unit for controlled materials. Architecture, tokenization and material details are in Supplementary Table~\ref{tab:supp-design}.}
\label{tab:design}
\setlength{\tabcolsep}{4pt}
\begin{tabularx}{\linewidth}{lXXX}
\toprule
 & GPT-2 & Qwen2.5 & Llama3.2\\
\midrule
Blocks / channels & 12 / 768 & 28 / 1,536 & 16 / 2,048\\
Query / KV heads & 12 / 12 & 12 / 2 & 32 / 8\\
Normalization & LayerNorm & RMSNorm & RMSNorm\\
Position encoding & Absolute & RoPE & RoPE\\
Layer anchors & 2 / 6 / 10 & 5 / 14 / 23 & 3 / 8 / 13\\
\midrule
\multicolumn{4}{l}{\textit{Material and intervention contrasts}}\\
Four-corner factors & \multicolumn{3}{p{.69\linewidth}}{Two independent factors; syntax and binding; 64 groups per task for direction interventions.}\\
Eight-condition roles & \multicolumn{3}{p{.69\linewidth}}{$A\times B\times R$; 128/32/64 training/development/test name groups; 15-token test sentences.}\\
Domain / lexical / value & \multicolumn{3}{p{.69\linewidth}}{Color--shape query controls; matched different-lexeme mappings; complete three-value combinations.}\\
Signed transmission & \multicolumn{3}{p{.69\linewidth}}{64-group finite edits; first 32 groups for initial derivatives and direct-branch clamps; late-component extension to all 64 groups in Qwen2.5/Llama3.2.}\\
Source and geometry & \multicolumn{3}{p{.69\linewidth}}{Document-paired source measurements; source-span readout, hidden-receiver layout adaptation and representation--geometry response comparisons.}\\
\bottomrule
\end{tabularx}
\end{table}

\subsection{A source-coordinate interface for distributed factors}
\label{sec:methods-coordinates}
To examine how selected features express input-related factors across source positions, we use the source-grounded Wiener interface~\citep{shu2026source}. It combines a selected forward feature, the model's local derivatives and independently calibrated statistics. For a target attention output at layer $d$, the inverse chain retains the input embedding, complete blocks $1,\ldots,d-1$ and the target output. Each of the resulting $d$ adjacent edges has its own frequency-specific, full-channel maps (Fig.~\ref{fig:framework}D).

The causal mask determines which inputs can influence the native feature $H_q$: $\partial H_q/\partial x_s=0$ for positions $s>q$. The raw vector--Jacobian product (VJP) includes all causal paths to that feature and has the same source-support restriction. Source inversion instead estimates a full source state,
\[
 \widehat x^{(q)}=\mathcal I(P_qH,J(x);\Theta_{\rm cal}),
\]
where $P_q$ retains position $q$ in the initial target, $J(x)$ collects the full-position local derivatives along the retained chain, and $\Theta_{\rm cal}$ denotes the calibrated maps. The estimate therefore conditions on the actual full instance as well as the selected feature. Its later-position coordinates describe inferred source structure. Frequency-specific maps mix positions, and the full-position consistency correction couples source positions through their shared descendants (Supplementary Section~\ref{sup:s2-support}).

At edge $v$, let $H_v$ be the parent state, $J_v$ its local child-state derivative and $R_v$ the activation-weighted mean-seed adjoint signal. Primes denote an orthonormal real Fourier transform along token position, and $S_{XY,v}(\omega)=\mathbb E[X'_v(\omega)Y'_v(\omega)^*]$ denotes an uncentered calibration moment. The first stage estimates the source state from its reverse signal:
\begin{equation}
 \mathsf G_v(\omega)=S_{HR,v}(\omega)[S_{RR,v}(\omega)+\lambda_{G,v}I]^{-1},
 \qquad \widehat H_v^{(G)}=\mathcal F^{-1}[\mathsf G_vR'_v].
 \label{eq:methods-g}
\end{equation}
The second stage corrects the local state residual using a forward-consistency cotangent,
\begin{align}
 c_v&=J_v^*(H_{c(v)}^{\rm on}-J_v\widehat H_v^{(G)}),\qquad
 \Delta H_v=H_v-\widehat H_v^{(G)},\\
 D_v(\omega)&=S_{\Delta H,c,v}(\omega)[S_{cc,v}(\omega)+\lambda_{D,v}I]^{-1},\qquad
 \widehat H_v=\widehat H_v^{(G)}+\mathcal F^{-1}[D_vc'_v].
 \label{eq:methods-d}
\end{align}
At the first edge, $H_{c(v)}^{\rm on}=P_qH$; subsequent edges receive the repaired full-position child state supplied by the reverse pass. Both regressions have zero intercept and preserve full channel coupling. The $D$ target is the local state residual; the consistency cotangent is its input. Repaired states propagate through one finite reverse pass, with the position selection applied only at the initial target.

The same matrix Wiener construction applies to spatial coordinates in vision and token positions in language: the declared coordinate axes are Fourier-transformed, and full-channel maps act at the stored frequencies~\citep{shu2026source}. Earlier GPT-2 experiments used 4,096 calibration windows and tied the maps across token-frequency bins, reducing calibration degrees of freedom. Here we fit independent per-frequency maps at every adjacent edge of the retained inverse chain. Calibration used 65,536 natural windows per GPT-2 depth and 131,072 per Qwen2.5/Llama3.2 depth, with three recorded map seeds and an additional 131,072-window GPT-2 middle-layer comparison. Supplementary Section~\ref{sup:s2-wiener} specifies the frequency constraints, numerical settings and paired comparisons of inverse-chain topology and calibration budget.

\subsection{Complementary measurements and statistical units}
\label{sec:methods-boundary}
\begin{figure}[!htbp]
\centering
\includegraphics[width=\linewidth]{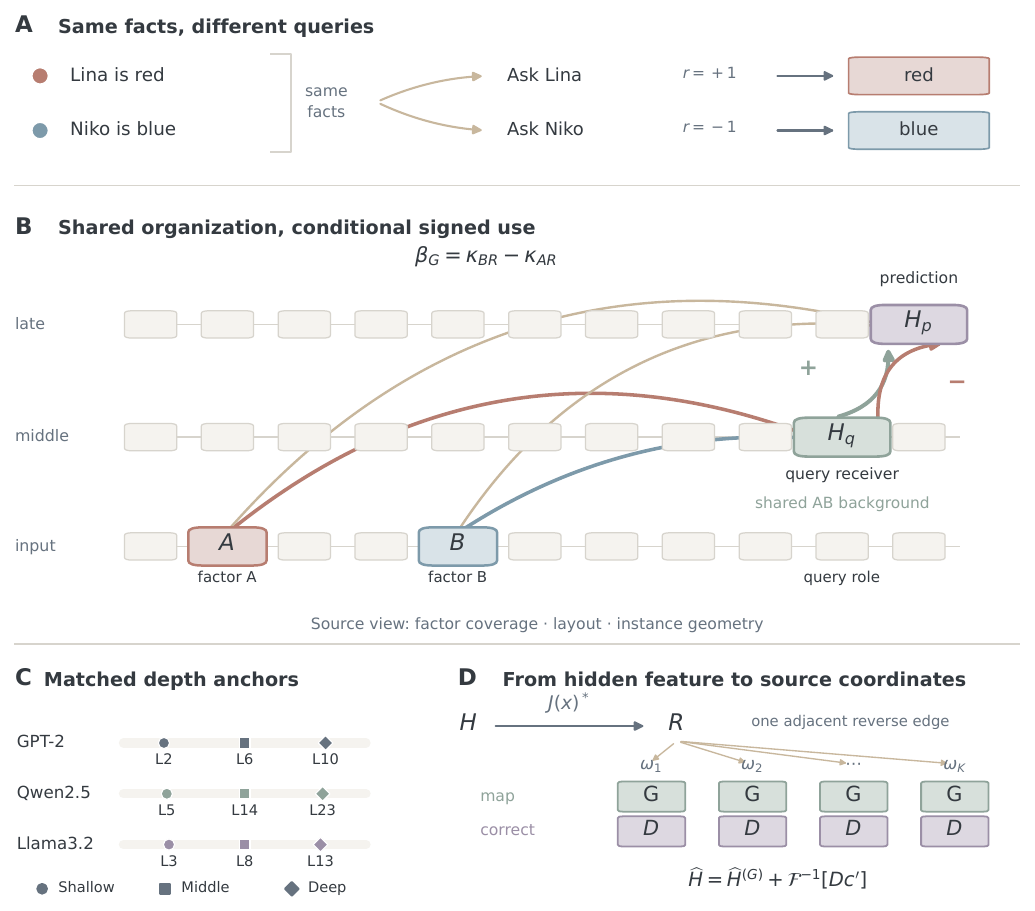}
\caption{\textbf{Separating shared relations from their conditional use.} \textbf{A}, The same facts support different targets when the queried entity changes. \textbf{B}, A token-by-depth schematic depicts internal representations and conditional readout: distributed factors, a middle-query relational background and signed reading at the prediction receiver. Colored routes illustrate factor contributions and opposing downstream gains; their widths and positions are schematic. Source-coordinate observations and instance-geometry tests provide complementary evidence about shared structure and its conditional use, through factor coverage, layout and response direction/amplitude. \textbf{C}, Shallow, middle and deep anchors in the three models. \textbf{D}, Each adjacent inverse edge uses local geometry to construct a reverse signal, then combines frequency-specific state estimates and forward-consistency corrections.}
\label{fig:framework}
\end{figure}

The measurements address successive questions about the mechanism (Fig.~\ref{fig:framework}B). Factor readout measures available information, while forward interventions test its contribution to the specified output~\citep{ravichander2021probing}. Factorial geometry identifies the organization shared across conditions. Forward edits and clamps measure the output effects of specified directions and components. Raw VJPs measure native first-order sensitivity; source-coordinate readouts locate factors and relationships in the estimate conditioned on the feature, instance geometry and calibration. Synthetic module-derivative substitutions test how instance geometry affects reinjected responses, while local predictors test the predictability of held-out descendant responses. Source-reference NMSE, factor accuracy, directional cosine and output margin retain their respective denominators and interpretations.

Uncertainty is estimated over complete source groups or natural documents, after aggregating repeated conditions and map seeds within each unit. The 21 original comparisons retain four separate Holm-corrected families. Later role, domain, layout, scale and late-clamp analyses retain their own sampling and testing procedures. Training-frozen axes are evaluated on held-out groups; the late-clamp study extends the original 64-group test set after fixing its selected nodes. Supplementary Table~\ref{tab:supp-primary-tests} reports the original effect sizes, confidence intervals and corrected probabilities, and the relevant supplementary sections specify subsequent protocols.

% Shared relational organization: final v7 structure.
\section{A shared low-dimensional workspace supports same-domain relational backgrounds}
\label{sec:results1}

\subsection{Depth-dependent reorganization reveals shared task geometry}
\label{sec:results1-depth}
Entity binding changed how the two property factors were organized across depth. We compared balanced four-corner conditions at shallow, middle and deep attention outputs in GPT-2, Qwen2.5 and Llama3.2. For each source group, the decomposition $h_{ab}=\mu+aA+bB+abI$, with $a,b\in\{-1,+1\}$, separates the two main effects from their interaction. The conditional directions are $\delta_A^{\pm}=2(A\pm I)$, with analogous directions for $B$, and the interaction-energy fraction is
\begin{equation}
 \eta_I=\frac{\lVert I\rVert^2}{\lVert A\rVert^2+\lVert B\rVert^2+\lVert I\rVert^2}.
 \label{eq:interaction-energy}
\end{equation}
All fractions and cosines were computed within source groups before averaging. These four-corner comparisons used the query receiver at position $q14$ in the 16-token materials.

\begin{figure}[p]
 \centering
 \includegraphics[width=\linewidth,height=.82\textheight,keepaspectratio]{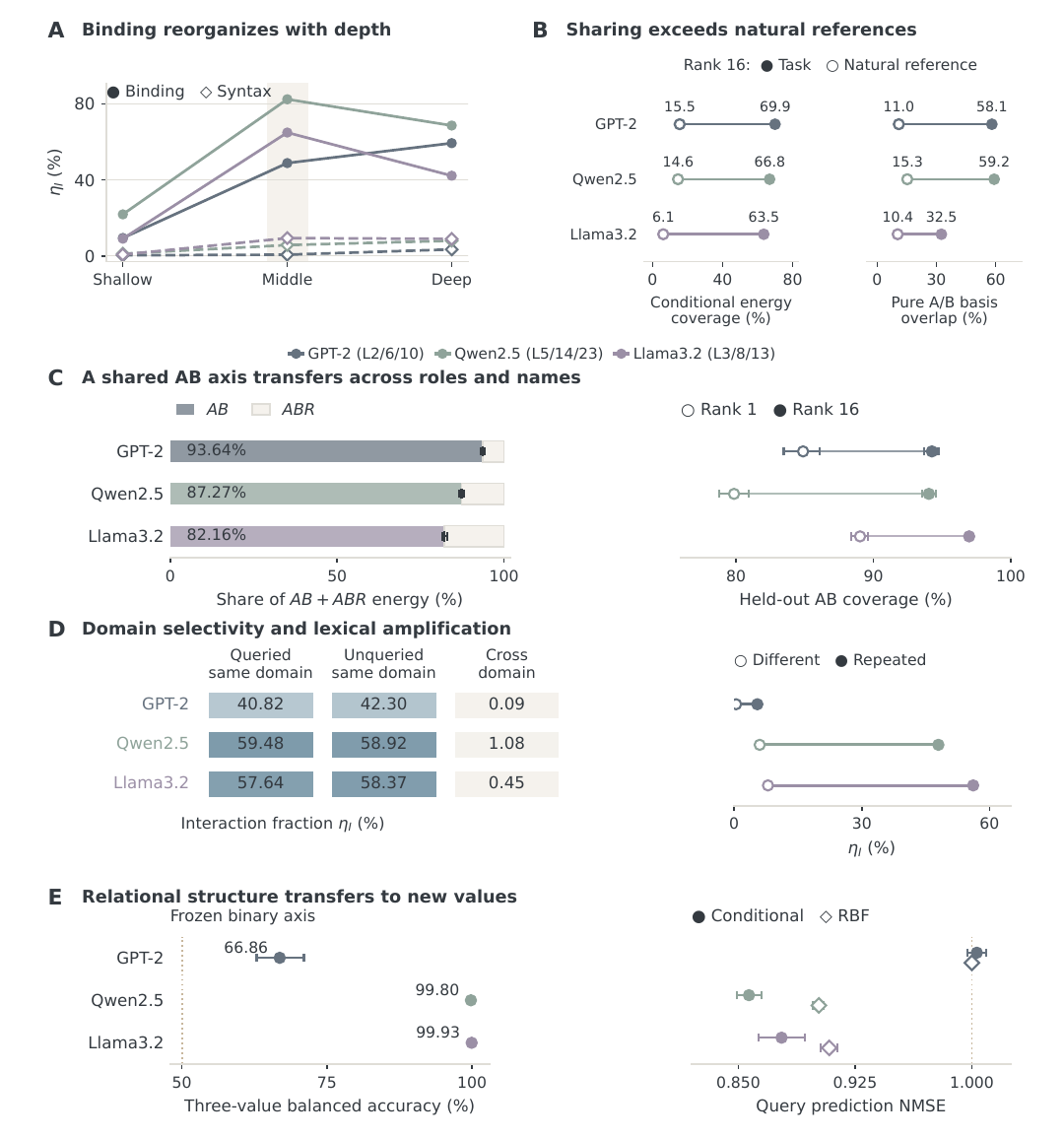}
 \caption{Shared task geometry supports a predominantly role-independent relational background. (A) Binding and syntax interaction fractions across the three measured depths. (B) Paired conditional-$A$ energy coverage and centered pure-main-effect basis overlap, each compared with its corresponding natural reference. (C) Composition of $AB+ABR$ energy: model-colored segments show $AB$, warm-grey segments show $ABR$, and boundary whiskers show the $AB$ fraction interval. The right-hand plot gives coverage of held-out-name $AB$ energy by independent training rank-1 and rank-16 bases. (D) Same-domain versus cross-domain interaction, followed by the length-matched lexical comparison. (E) Complete-group-centered three-value relation classification using the original binary interaction axis $I$, and held-out query displacement prediction in a fixed rank-16 basis. Classification and prediction are separate protocols. Intervals in C, D (right) and E are 95\% source-group bootstrap intervals; the domain comparison uses 32 groups and the role, lexical and three-value comparisons use 64. Dotted reference lines denote 50\% balanced accuracy and NMSE 1. Model colors are consistent throughout. Pale bands mark the middle checkpoint.}
 \label{fig:relational-geometry}
\end{figure}

Binding conditional directions were positively aligned at shallow depth and negatively aligned at middle depth in all three models (Supplementary Fig.~\ref{fig:supp-v7-depth}A--C). At the middle checkpoints (layers 6, 14 and 8), the factor-$A/B$ cosines were $-0.379/-0.260$, $-0.818/-0.809$ and $-0.603/-0.597$, respectively. Syntax retained positive cosines at the same checkpoints: $+0.976/+0.964$, $+0.794/+0.754$ and $+0.701/+0.586$. Binding interaction fractions rose from $9.49\%$, $21.93\%$ and $9.13\%$ at shallow depth to $48.91\%$, $82.49\%$ and $64.97\%$ at middle depth (Fig.~\ref{fig:relational-geometry}A). Because $\delta_A^+\cdot\delta_A^-=4(\lVert A\rVert^2-\lVert I\rVert^2)$, this directional transition identifies the increasing contribution of relational interaction relative to the individual main effects. The deep checkpoints retained substantial binding interaction ($59.39\%$, $68.66\%$ and $42.25\%$), whereas syntax interaction remained below $10\%$ throughout the sampled depths (Supplementary Fig.~\ref{fig:supp-v7-depth} and Table~\ref{tab:supp-v7-depth}).

We next asked whether the two factors occupied a shared low-dimensional workspace beyond their common interaction and natural-state covariance. Averaging equally over the two conditioning values yields
\begin{equation}
 M_{\delta_A}=4(M_A+M_I),\qquad M_{\delta_B}=4(M_B+M_I),
 \label{eq:four-corner-moment}
\end{equation}
where $M_X=\mathbb E_g[X_gX_g^{\mathsf T}]$ is the uncentered second moment across source groups. The conditional moments additionally average over the conditioning factor. This identity separates the algebraically shared interaction from the empirical organization of the main effects.

Two comparisons identified task-specific sharing. First, in the same middle-layer hidden coordinates, a task rank-16 basis covered $69.90\%$, $66.80\%$ and $63.47\%$ of paired binding conditional-$A$ energy; the leading 16 natural-hidden-state principal components covered only $15.51\%$, $14.56\%$ and $6.14\%$ (Fig.~\ref{fig:relational-geometry}B). Task--natural basis overlaps were $16.31\%$, $8.76\%$ and $7.02\%$. Second, after removing $I$ and centering each pure main effect on its training mean, the $A/B$ rank-16 basis overlaps remained $58.06\%$, $59.22\%$ and $32.53\%$, compared with $11.01\%$, $15.28\%$ and $10.43\%$ between independent natural-covariance bases. Confirmation-set coverage in the opposite factor's basis exceeded coverage in the natural-random reference for both factors in all six model-by-task panels. The binding excesses and their intervals are reported in Table~\ref{tab:results1-background}; Supplementary Fig.~\ref{fig:supp-v7-natural} and Table~\ref{tab:supp-v7-natural} give the complete comparisons. Together, the depth transition and the two reference comparisons identify low-dimensional task organization shared by the factors beyond their common interaction and ambient covariance.

\subsection{Query-role decomposition identifies a transferable AB background}
\label{sec:results1-d1}
The shared interaction could reflect either a relation common to both queries or a signal specific to the currently queried entity. We separated these possibilities with the 15-token eight-condition design, independently crossing the two properties with query role $r\in\{-1,+1\}$:
\begin{equation}
 h_{abr}=\mu+aA+bB+rR+abI_{AB}+arI_{AR}+brI_{BR}+abrI_{ABR}.
 \label{eq:eight-condition-background}
\end{equation}
At its middle-layer query receiver ($q13$), the two role-conditioned interactions satisfy $I^{(\pm)}=I_{AB}\pm I_{ABR}$. The role-independent $AB$ term accounted for $93.64\%$, $87.27\%$ and $82.16\%$ of $AB+ABR$ energy, with role-swap cosines of $0.8803$, $0.7691$ and $0.6527$ (Fig.~\ref{fig:relational-geometry}C). Thus the dominant interaction persisted across the identity of the queried entity alongside smaller, separately resolved role-modulated components.

We then tested whether this relation axis extended to new entity names. The 128 training, 32 development and 64 test groups used disjoint names with shared colors and templates. A rank-1 $AB$ axis frozen on training names covered $84.88\%$, $79.86\%$ and $89.03\%$ of test-group $AB$ energy; rank-16 coverage reached $94.26\%$, $94.04\%$ and $96.97\%$. Table~\ref{tab:results1-background} reports the source-group intervals. The all-layer decomposition and the anchored query/prediction comparisons resolve the accompanying $R$, $AR$, $BR$ and $ABR$ terms (Supplementary Fig.~\ref{fig:supp-v7-role} and Table~\ref{tab:supp-v7-role}). This combination of role stability and held-out-name coverage identifies a transferable relational background whose dominant direction is shared across queries.

\subsection{Domain, lexical structure and value range delimit relational organization}
\label{sec:results1-domain}
We next asked whether the background requires its attribute domain to be queried. A 64-condition color--shape design crossed the facts, queried entity and queried domain. At the middle query state, queried and unqueried same-domain interaction fractions were comparable: $40.82/42.30\%$ in GPT-2, $59.48/58.92\%$ in Qwen2.5 and $57.64/58.37\%$ in Llama3.2. Cross-domain fractions were $0.09\%$, $1.08\%$ and $0.45\%$. Mapping both domains onto the same binary answer code kept cross-domain interaction low ($0.058\%$, $0.268\%$ and $0.239\%$). These query-conditioned states therefore retained same-domain relational organization independently of whether that domain was the current target (Fig.~\ref{fig:relational-geometry}D; Supplementary Fig.~\ref{fig:supp-v7-generalization} and Table~\ref{tab:supp-v7-generalization}).

Lexical structure strongly modulated its magnitude. Strictly length-matched different-lexeme mappings reduced interaction fractions from $5.45/47.99/56.16\%$ to $0.45/6.00/7.94\%$ across the three models. To assess the surviving direction, we projected states onto the frozen binary interaction axis $I$ after subtracting the unlabeled mean of each complete counterfactual group. This group-supported diagnostic gave same/different balanced accuracies of $50.20\%$, $57.81\%$ and $64.06\%$ (Table~\ref{tab:results1-background}). Qwen2.5 and Llama3.2 retained partial cross-lexeme category alignment, while GPT-2 was near chance. Lexical amplification and transferable relational direction therefore contributed differently across models.

A separate three-value diagnostic tested whether the original binary interaction axis $I$ could classify equality across the complete $3\times3\times R$ design, using the same complete-group centering. Balanced accuracies were $66.86\%$, $99.80\%$ and $99.93\%$ (Fig.~\ref{fig:relational-geometry}E, left; Table~\ref{tab:results1-background}). This tests value transfer of the original four-corner relation axis, whereas the preceding eight-condition analysis tested new-name transfer of $AB$.

Finally, response prediction tested unseen combinations within a common rank-16 basis. We withheld combinations $(0,2)$, $(1,0)$ and $(2,1)$ and the test vocabulary, while providing each predictor with the same within-group $(0,0,R0)$ reference state. At the middle query receiver, conditional-linear prediction yielded lower NMSE than local RBF in Qwen2.5 ($0.85685$ versus $0.90176$) and Llama3.2 ($0.87779$ versus $0.90837$). GPT-2 remained near the zero-response reference of 1 ($1.00318$ versus $0.99999$; Fig.~\ref{fig:relational-geometry}E, right). Supplementary Table~\ref{tab:supp-v7-prediction} reports both receivers, all predictors and development-fitted amplitude controls. These complementary tests identify shared relational organization across roles and names, same-domain preservation across query targets, and model-dependent lexical and value transfer. The next section asks how this shared direction acquires a signed effect on role selection.

\begin{table}[!htbp]
\centering\footnotesize
\setlength{\tabcolsep}{4pt}
\caption{Core geometry and generalization measurements. Brackets are 95\% source-group bootstrap intervals. Energy and coverage entries are percentages; excess coverage is in percentage points. Relation classification uses the original binary axis after unlabeled centering within complete counterfactual groups. The three-value classifier and held-out-combination predictor are separate tests. Except for the 32-group domain experiment, confirmation/test sets contain 64 source groups.}
\label{tab:results1-background}
\begin{tabular}{lrrr}
\toprule
Measurement & GPT-2 & Qwen2.5 & Llama3.2 \\
\midrule
\multicolumn{4}{l}{\emph{Pure-main-effect sharing in binding}} \\
Basis overlap (\%) & 58.06 & 59.22 & 32.53 \\
Natural-pair overlap (\%) & 11.01 & 15.28 & 10.43 \\
$A$-on-$B$ excess (pp) & 37.04 [33.97, 40.02] & 20.74 [16.88, 24.71] & 19.75 [17.54, 22.03] \\
$B$-on-$A$ excess (pp) & 36.08 [31.64, 40.59] & 21.70 [18.04, 25.58] & 16.37 [13.66, 19.09] \\
\addlinespace\multicolumn{4}{l}{\emph{Role-decoupled background and new-name coverage}} \\
$AB/(AB+ABR)$ energy (\%) & 93.64 [93.07, 94.19] & 87.27 [86.65, 87.89] & 82.16 [81.43, 82.90] \\
Role-swap cosine & 0.880 [0.869, 0.891] & 0.769 [0.759, 0.779] & 0.653 [0.639, 0.667] \\
Frozen rank-1 coverage (\%) & 84.88 [83.47, 86.12] & 79.86 [78.78, 80.93] & 89.03 [88.39, 89.58] \\
Frozen rank-16 coverage (\%) & 94.26 [93.71, 94.73] & 94.04 [93.57, 94.52] & 96.97 [96.69, 97.22] \\
\addlinespace\multicolumn{4}{l}{\emph{Domain preservation and group-supported transfer}} \\
Queried same-domain $\eta_I$ (\%) & 40.82 [39.49, 42.24] & 59.48 [58.57, 60.35] & 57.64 [57.05, 58.22] \\
Unqueried same-domain $\eta_I$ (\%) & 42.30 [41.08, 43.51] & 58.92 [57.92, 59.86] & 58.37 [57.61, 59.10] \\
Cross-domain $\eta_I$ (\%) & 0.090 [0.085, 0.095] & 1.076 [1.050, 1.104] & 0.449 [0.433, 0.468] \\
Different-lexeme accuracy (\%) & 50.20 [48.83, 51.56] & 57.81 [55.66, 60.16] & 64.06 [62.30, 66.02] \\
Three-value accuracy (\%) & 66.86 [62.89, 71.03] & 99.80 [99.48, 100.00] & 99.93 [99.80, 100.00] \\
\addlinespace\multicolumn{4}{l}{\emph{Held-out query displacement prediction, common rank-16 basis}} \\
Conditional-linear NMSE & 1.003 [0.997, 1.009] & 0.857 [0.849, 0.865] & 0.878 [0.863, 0.893] \\
Local-RBF NMSE & 1.000 [1.000, 1.000] & 0.902 [0.898, 0.905] & 0.908 [0.903, 0.914] \\
\bottomrule\end{tabular}\end{table}

\FloatBarrier

\section{Conditional gain and multi-pathway signed readout}
\label{sec:results2}
A relation shared across query roles can influence a specific answer if the downstream computation responds differently to it across conditions. We tested this link by separating unsigned receiver geometry, transferable relational perturbations, and the paths that transmit their signed effects.

\subsection{Receiver energy and internal angles leave the functional sign unresolved}
\label{sec:results2-receivers}
\begin{figure}[p]
\centering
\includegraphics[width=\linewidth,height=.80\textheight,keepaspectratio]{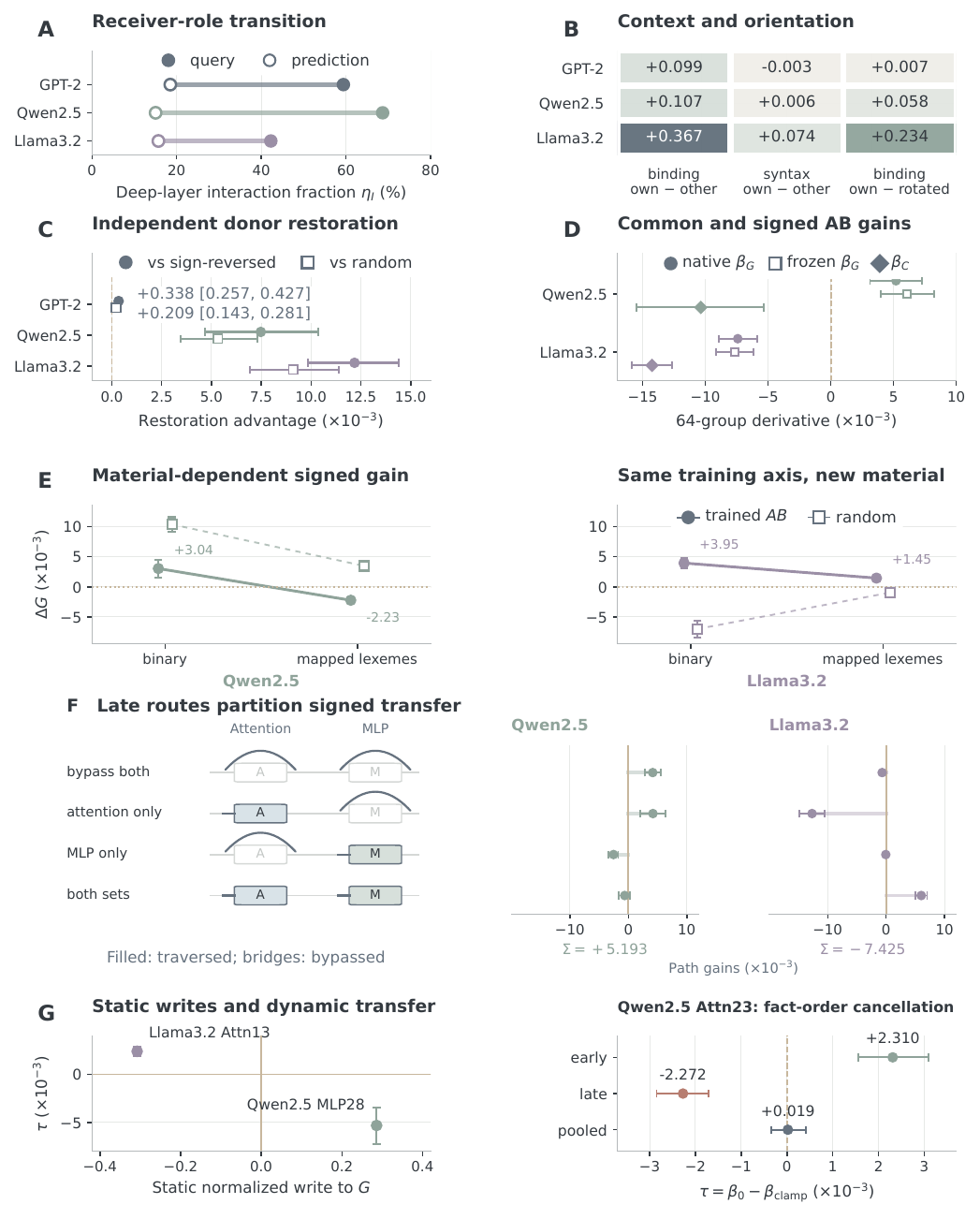}
\caption{Shared relational directions acquire conditional signed effects. (A) Deep query/prediction interaction fractions in four-corner binding materials. (B) Matching-context and common-rotation contrasts in response cosine. (C) Independent training-donor advantages over sign-reversed and equal-norm random refills. (D) Protocol-specific $AB$ derivatives, including fixed final-denominator controls. (E) A fixed training-mean direction across binary and mapped-lexeme materials, together with equal-norm random controls. The output pairs are orange/purple and red/blue, respectively; category and query-role coding are shared. (F) Four tangent-path sums with 95\% intervals, classified by traversal of the selected attention ($\mathcal A$) and MLP ($\mathcal M$) sets; their sum equals native $\beta_G$. (G) Static normalized writes and dynamic net transmission in separate coordinates, including fact-order cancellation at Qwen2.5 Attention~23. Error bars are 95\% source-group bootstrap intervals where shown. The derivative, finite-dose and four-corner experiments retain their respective protocols.}
\label{fig:signed-transfer}
\end{figure}

Relational organization changed sharply between adjacent receiving positions. In the 16-token binding materials, deep query receivers retained interaction fractions of $59.39\%$, $68.66\%$ and $42.25\%$ in GPT-2, Qwen2.5 and Llama3.2, whereas the adjacent prediction receivers had $18.50\%$, $15.03\%$ and $15.69\%$ (Fig.~\ref{fig:signed-transfer}A). The prediction receiver's own middle-to-deep declines were $39.08$, $61.51$ and $64.71$ percentage points. Absolute interaction energy still grew over this transition in GPT-2 and Qwen2.5; stronger growth of attribute main effects reduced its relative share. Thus prediction involved a redistribution of factor energy within a state that retained relational structure.

The role-balanced eight-condition materials then separated energy from signed choice. We compared $G$ with total attribute energy $T=\|A\|^2+\|B\|^2$ and fixed-entity asymmetry $S_{\rm entity}=(\|A\|^2-\|B\|^2)/T$. For role alignment, $A_r=A+rI_{AR}$ and $B_r=B+rI_{BR}$ defined the queried and distractor contributions; their normalized energy difference gave $S_{\rm target}$. Its deep prediction-receiver Spearman correlation with $G$ was $+0.796$ in GPT-2, but $+0.011$ in Qwen2.5 and $+0.007$ in Llama3.2, with the latter intervals spanning zero. Unsigned energy therefore supplied a model-dependent correlate of signed selection (Supplementary Table~\ref{tab:supp-role-correlations}).

Forward edits identified an additional determinant: orientation relative to the downstream computation. In the four-corner materials, equal-norm factor directions from the matching context improved descendant-response cosine over directions from the alternative context by $+0.0994$, $+0.1071$ and $+0.3671$ in binding, with positive intervals in all three models. Syntax contrasts were $-0.0029$, $+0.0057$ and $+0.0745$; the first two intervals spanned zero. A common orthogonal rotation preserved internal angles but reduced binding response matching in Qwen2.5 and Llama3.2 by $0.0581$ $[0.0176,0.1000]$ and $0.2336$ $[0.1767,0.2915]$, respectively (Fig.~\ref{fig:signed-transfer}B). Scaling doses from $0.1$ to $0.25$ preserved response-direction cosine $\geq0.9995$, while norm ratios relative to the expected $2.5$-fold change lay between $0.99939$ and $1.00057$. These factor-direction experiments establish locally proportional, orientation-sensitive transfer and motivate a directional-sensitivity analysis of the separately defined $AB$ edit (Supplementary Fig.~\ref{fig:supp-v7-reading}, Table~\ref{tab:supp-dose-proportionality}).

\subsection{A transferable relational component acquires gain from conditional sensitivity}
\label{sec:results2-ab-causal}
The shared $AB$ component had a transferable functional effect. After removing each test group's rank-16 $AB$ projection, we refilled it with a direction estimated from independent training names. This donor brought $G$ closer to its unedited value than either an equal-norm random direction or the sign-reversed donor. Advantages over random refill were $0.209$ $[0.143,0.281]$, $5.319$ $[3.468,7.316]$ and $9.099$ $[6.946,11.399]$ $\times10^{-3}$; advantages over reversed refill were $0.338$ $[0.257,0.427]$, $7.473$ $[4.684,10.356]$ and $12.181$ $[9.845,14.388]$ $\times10^{-3}$, respectively. All six intervals were positive, establishing transfer of a functional direction even for GPT-2, whose mean local $G$ slope was near zero (Fig.~\ref{fig:signed-transfer}C; Supplementary Table~\ref{tab:supp-donor-restore}).

We connected this functional effect to the output contrasts defined in Section~\ref{sec:methods}. For each balanced condition, the middle-query edit was
\begin{equation}
 h_{abr}(\epsilon)=h_{abr}+ab\epsilon v,\qquad
 \kappa(a,b,r)=\nabla_hm(a,b,r)^{\mathsf T}v,
 \label{eq:ab-direction}
\end{equation}
where $v$ was fixed across the eight conditions within a source group. Under the projection protocol, $v$ was that group's $AB$ coefficient projected onto the training-frozen rank-16 basis. Writing $\kappa_X=\mathbb E[\chi_X\kappa]$ gives
\begin{equation}
 \beta_C=\frac{\partial C}{\partial\epsilon}=\kappa_A+\kappa_B,
 \qquad
 \beta_G=\frac{\partial G}{\partial\epsilon}
 =\mathbb E[(ar-br)ab\kappa]=\kappa_{BR}-\kappa_{AR}.
 \label{eq:conditional-gain}
\end{equation}
A context-invariant linear readout makes $\kappa$ constant and hence $\beta_G=0$. The shared direction changes role selection when downstream sensitivity differs across the attribute--query combinations.

The first 32 test groups gave analytic $\beta_G=-0.0175$, $+6.505$ and $-7.273$ $\times10^{-3}$; GPT-2's interval spanned zero, while the Qwen2.5 and Llama3.2 intervals had opposite signs (Table~\ref{tab:results2-core}). The late-clamp extension reused the same 64 test groups and training basis, extending derivatives to all groups and adding component interventions. Its native $\beta_G$ was $+5.193$ and $-7.425$ $\times10^{-3}$, while $\beta_C$ was negative in both models ($-10.371$ and $-14.262$ $\times10^{-3}$). Freezing each condition's final RMSNorm denominator preserved the $G$ signs, giving $+6.065$ and $-7.655$ $\times10^{-3}$ (Fig.~\ref{fig:signed-transfer}D; Supplementary Table~\ref{tab:supp-clamp-numerics}). The final scale adjusted gain magnitude; the two output contrasts selected different combinations of conditional sensitivities. Finite removal of half the $AB$ projection likewise increased $C$ in Qwen2.5 and Llama3.2, consistent with their negative local slopes (Supplementary Table~\ref{tab:supp-d2-protocols}).

A fixed direction could also change its output sign within one model. Holding the Qwen2.5 weights, training-mean $AB$ vector and dose $+0.5abv$ fixed gave $\Delta G=+3.04$ $[1.56,4.50]$ $\times10^{-3}$ in binary materials and $-2.23$ $[-2.55,-1.91]$ $\times10^{-3}$ in mapped-lexeme materials. Llama3.2 remained positive, with effects of $+3.95$ and $+1.45$ $\times10^{-3}$ (Fig.~\ref{fig:signed-transfer}E; Table~\ref{tab:results2-core}). Category and query-role coding were identical, while the prompts and output pairs changed from orange/purple to red/blue. The comparison therefore measures transfer of one numerical injection axis across complete material and readout working points; the projection protocol above fixes its direction within each group. Signed gain depends on the relation direction together with the computation that receives and reads it (Supplementary Section~\ref{sup:s4-material-protocol}, Fig.~\ref{fig:supp-v7-ab-transfer}, Table~\ref{tab:supp-donor-restore}).

\begin{table}[!htbp]
\centering\small
\setlength{\tabcolsep}{4pt}
\caption{Relational restoration, signed effects and tangent paths by protocol. Entries are estimates [95\% source-group bootstrap intervals] in $10^{-3}$ units. Donor advantage measures greater reduction in absolute distance from native $G$ than the stated refill control. Projection rows use each group's $AB$ coefficient in the frozen training rank-16 basis; material-transfer rows use one fixed training-mean vector, with output pairs orange/purple (binary) and red/blue (mapped). GPT-2 material effects use an unedited baseline recomputed on the same CPU in batches of 16. The 32-group derivative subset belongs to the 64-group projection test set. Path sums use unrounded values; dashes mark comparisons not included in the late-clamp protocol.}
\label{tab:results2-core}
\begin{tabular}{llll}
\toprule
Quantity and protocol & GPT-2 & Qwen2.5 & Llama3.2 \\
\midrule
\multicolumn{4}{l}{\emph{Independent training-donor restoration, $n=64$}} \\
Donor $-$ reversed & \shortstack{$+0.338$\\$[+0.257,+0.427]$} & \shortstack{$+7.473$\\$[+4.684,+10.356]$} & \shortstack{$+12.181$\\$[+9.845,+14.388]$} \\
Donor $-$ random & \shortstack{$+0.209$\\$[+0.143,+0.281]$} & \shortstack{$+5.319$\\$[+3.468,+7.316]$} & \shortstack{$+9.099$\\$[+6.946,+11.399]$} \\
\addlinespace
\multicolumn{4}{l}{\emph{Group-specific projection: derivatives and finite edits}} \\
$\beta_G$, analytic, $n=32$ & \shortstack{$-0.018$\\$[-0.141,+0.105]$} & \shortstack{$+6.505$\\$[+3.898,+9.111]$} & \shortstack{$-7.273$\\$[-9.393,-5.086]$} \\
$\Delta G$, dose $-0.5$, $n=64$ & \shortstack{$+0.021$\\$[-0.018,+0.058]$} & \shortstack{$-1.887$\\$[-2.919,-0.867]$} & \shortstack{$+3.678$\\$[+2.892,+4.372]$} \\
$\Delta C$, dose $-0.5$, $n=64$ & \shortstack{$-0.432$\\$[-0.540,-0.321]$} & \shortstack{$+5.546$\\$[+2.988,+8.114]$} & \shortstack{$+7.092$\\$[+6.280,+7.928]$} \\
\addlinespace
\multicolumn{4}{l}{\emph{Fixed training-mean vector: dose $+0.5$, $n=64$}} \\
$\Delta G$, binary & \shortstack{$+0.039$\\$[+0.021,+0.057]$} & \shortstack{$+3.040$\\$[+1.560,+4.500]$} & \shortstack{$+3.950$\\$[+3.160,+4.750]$} \\
$\Delta G$, mapped & \shortstack{$+0.065$\\$[+0.051,+0.079]$} & \shortstack{$-2.230$\\$[-2.550,-1.910]$} & \shortstack{$+1.450$\\$[+1.120,+1.770]$} \\
\addlinespace
\multicolumn{4}{l}{\emph{Late-clamp extension: local derivatives, $n=64$}} \\
Native $\beta_G$ & --- & \shortstack{$+5.193$\\$[+3.119,+7.281]$} & \shortstack{$-7.425$\\$[-8.930,-5.902]$} \\
Native $\beta_C$ & --- & \shortstack{$-10.371$\\$[-15.498,-5.346]$} & \shortstack{$-14.262$\\$[-15.875,-12.637]$} \\
Frozen-denominator $\beta_G$ & --- & \shortstack{$+6.065$\\$[+3.948,+8.246]$} & \shortstack{$-7.655$\\$[-9.138,-6.141]$} \\
\addlinespace
\multicolumn{4}{l}{\emph{Tangent paths classified by the selected late-node sets, $n=64$}} \\
Bypass both, $b_{\varnothing}$ & --- & \shortstack{$+4.165$\\$[+2.809,+5.581]$} & \shortstack{$-0.685$\\$[-0.871,-0.509]$} \\
Attention only, $b_{A\bar M}$ & --- & \shortstack{$+4.200$\\$[+2.061,+6.379]$} & \shortstack{$-12.683$\\$[-14.915,-10.488]$} \\
MLP only, $b_{\bar A M}$ & --- & \shortstack{$-2.540$\\$[-3.384,-1.731]$} & \shortstack{$-0.065$\\$[-0.231,+0.101]$} \\
Both sets, $b_{AM}$ & --- & \shortstack{$-0.633$\\$[-1.610,+0.326]$} & \shortstack{$+6.008$\\$[+5.074,+7.003]$} \\
\bottomrule
\end{tabular}
\end{table}

\subsection{Late attention and MLP paths transmit and compensate signed gain}
\label{sec:results2-clamp}
Relational effects extended beyond the designated direct query and candidate contributions. In the 32-group direct-path subset, clamping both contributions to their clean values during half-projection removal retained $\Delta G=-2.288$ $[-3.460,-1.109]$ and $+4.673$ $[3.461,5.927]$ $\times10^{-3}$ in Qwen2.5 and Llama3.2, respectively (Supplementary Table~\ref{tab:supp-d2-path-residual}). We next measured how late components transmit the $AB$ effect in the native graph, using prediction-position clean clamps on fixed late sets selected from the preceding baseline attribution: attention layers $\mathcal A=\{21,23,25,27\}$ and MLP layers $\mathcal M=\{24,28\}$ in Qwen2.5, and $\mathcal A=\{12,13,15,16\}$ and $\mathcal M=\{15,16\}$ in Llama3.2. Each clean clamp substituted the same condition's unedited write.

Let $\beta_0$ be native gain, and $\beta_A$, $\beta_M$, $\beta_{AM}$ the gains after clamping attention, MLP or both sets. Because the graphs share the zero-edit state, their local derivatives partition by traversal of these sets:
\begin{align}
 b_{\varnothing}&=\beta_{AM}, & b_{A\bar M}&=\beta_M-\beta_{AM},\\
 b_{\bar A M}&=\beta_A-\beta_{AM}, & b_{AM}&=\beta_0-\beta_A-\beta_M+\beta_{AM}.
 \label{eq:tangent-paths}
\end{align}
The four path totals sum exactly to $\beta_0$ within each group. Llama3.2's negative attention-only total ($-12.683\times10^{-3}$) was partially offset by paths traversing both selected sets ($+6.008\times10^{-3}$). Qwen2.5 instead had positive bypass and attention-only totals ($+4.165$ and $+4.200\times10^{-3}$), partly offset by the MLP-only total ($-2.540\times10^{-3}$); its overlap interval spanned zero (Fig.~\ref{fig:signed-transfer}F; Table~\ref{tab:results2-core}). Equal-norm random directions also traversed the selected attention sets. In Llama3.2, the selected MLP set transmitted an absolute $AB$ shift exceeding the random reference by $2.527$ $[1.596,3.431]$ $\times10^{-3}$, identifying direction-dependent use of general downstream channels (Supplementary Figs.~\ref{fig:supp-v7-clamp}--\ref{fig:supp-v7-controls}, Tables~\ref{tab:supp-clamp-all}--\ref{tab:supp-clamp-numerics}).

Static write composition and dynamic transmission resolved different properties of these components. With net dynamic transmission defined as $\tau_k=\beta_0-\beta_{\mathrm{clamp}(k)}$, Qwen2.5 MLP~28 had a positive static write to $G$ ($+0.286$) but negative $\tau$ ($-5.307\times10^{-3}$). Llama3.2 Attention~13 had a negative static write ($-0.308$) but positive $\tau$ ($+2.373\times10^{-3}$). Qwen2.5 Attention~23 further separated condition-specific effects from their mean: early and late net transfers of $+2.310$ and $-2.272\times10^{-3}$ nearly canceled (Fig.~\ref{fig:signed-transfer}G). These measurements distinguish the assembly of the baseline response from the propagation of a relational perturbation.

\subsection{Shared receivers and fact order organize position-level competition}
\label{sec:receiving}
\label{sec:receiving-order}
\begin{figure}[!htbp]
\centering
\includegraphics[width=\linewidth,height=.78\textheight,keepaspectratio]{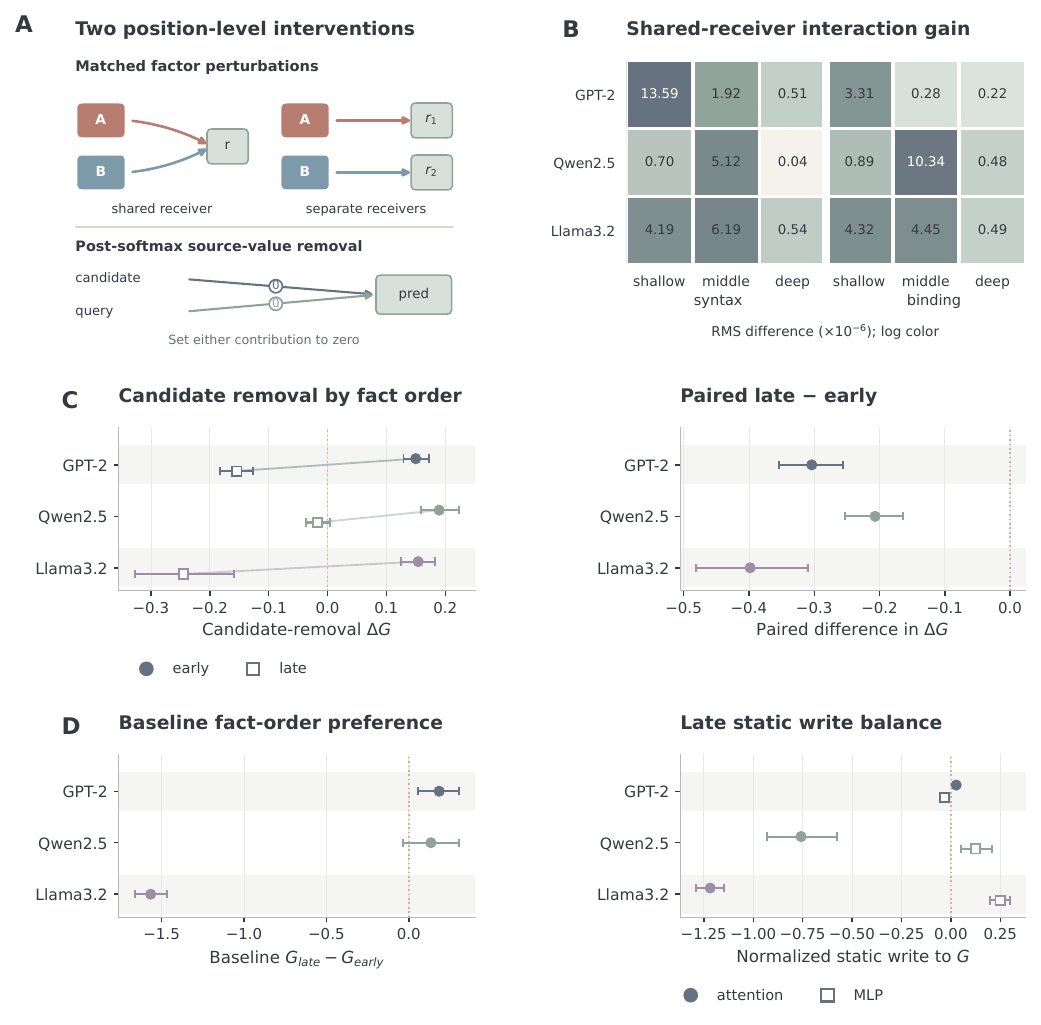}
\caption{Receiving arrangements and fact order organize functional competition. (A) Two position-level operations: shared or separated receivers for matched factor perturbations, and removal of a designated deep post-softmax source-value contribution at the prediction receiver. Source removal preserves the other attention probabilities. (B) Shared-minus-separated interaction RMS in all 18 task--model--depth panels; each cell reports its original value, with a logarithmic color scale. (C) Candidate-removal effects for early and late queried facts and their paired difference. (D) Baseline fact-order preference alongside late attention/MLP static writes to $G$, in separate coordinates. Error bars indicate 95\% source-group bootstrap intervals. The receiving comparison uses route-wise RMS averaging in GPT-2 and pooled-route RMS in Qwen2.5 and Llama3.2. Source removal and static write attribution retain their respective operations and scales.}
\label{fig:receiving-competition}
\end{figure}

Position-level interventions provided a complementary test of how contributions combine. We delivered the same factor directions to shared or separated receiving positions and measured the four-corner interaction in candidate margin. Shared receivers increased its RMS in all 18 model--task--depth comparisons, with positive paired intervals (Fig.~\ref{fig:receiving-competition}A--B). This establishes stronger non-additive combination in the tested receiving configurations. Each model retains its original RMS estimator: route-wise averaging for GPT-2 and pooled-route evaluation for Qwen2.5 and Llama3.2 (Supplementary Fig.~\ref{fig:supp-v7-receiving}, Table~\ref{tab:sup-c6-values}).

In the separate eight-condition materials, removing the designated post-softmax candidate or query value contribution changed common-property and role-selection responses. These interventions set a source contribution to zero while retaining the other attention probabilities. Candidate removal had pooled $\Delta G=-0.002177$, $+0.086206$ and $-0.045159$ in GPT-2, Qwen2.5 and Llama3.2; the GPT-2 interval spanned zero. Both routes reached signed selection, and their effects depended on the surrounding computation (Supplementary Table~\ref{tab:supp-source-order}).

Pairing early and late queried facts within the same source groups revealed a common order-conditioned effect beneath the different pooled signs. Candidate removal had positive early means and negative late means in all three models; paired late-minus-early differences were $-0.30389$ $[-0.35474,-0.25643]$, $-0.20683$ $[-0.25314,-0.16410]$ and $-0.39826$ $[-0.48122,-0.31030]$ (Fig.~\ref{fig:receiving-competition}C). Qwen2.5's late effect alone had an interval spanning zero, while its paired difference remained negative. The candidate route therefore provided relative support for the later fact. In Llama3.2 this support opposed a baseline late-minus-early margin of $-1.56696$ $[-1.66287,-1.47040]$.

Normalization-aware write attribution located the baseline competition. Llama3.2 attention layers 12--16 contributed $-1.21788$ to $G$, opposed by an MLP contribution of $+0.24844$; their contributions to the baseline late-minus-early contrast were $-1.94321$ and $+0.43123$. Qwen2.5 attention layers 21--28 likewise contributed negatively to $G$ ($-0.75834$), with a positive MLP sum ($+0.12471$) and a prominent final MLP~28 write ($+0.28616$; Fig.~\ref{fig:receiving-competition}D). Static write sums describe the baseline balance, while source removal measures the downstream effect of a specified contribution. Their joint interpretation identifies relative late-fact support operating within competing residual writes (Supplementary Fig.~\ref{fig:supp-v7-order}, Table~\ref{tab:supp-source-order}).

The interventions establish how a shared hidden relation influences selection through conditional sensitivity, signed paths and position-level competition. Source-coordinate measurements next locate the input-related factors expressed by the selected features; direct hidden-receiver readouts test how combining positions depends on layout.

\section{Source-factor carrying and hidden-receiver complementarity}
\label{sec:source-carrying}
\label{sec:results3}

The preceding interventions identified how hidden relational directions influence downstream selection. We now examine a complementary question: which source factors become observable when the same computations are expressed in input coordinates? The calibrated source map combines a selected hidden feature, the instance geometry and data statistics. We compare these source-coordinate observations with hidden-receiver readouts, then test how layout changes the benefit of combining hidden receivers across positions.

\subsection{Non-terminal source structure separates statistical carrying from transport background}
\label{sec:source-statistics}
\begin{figure}[!htbp]
\centering
\includegraphics[width=\linewidth,height=.79\textheight,keepaspectratio]{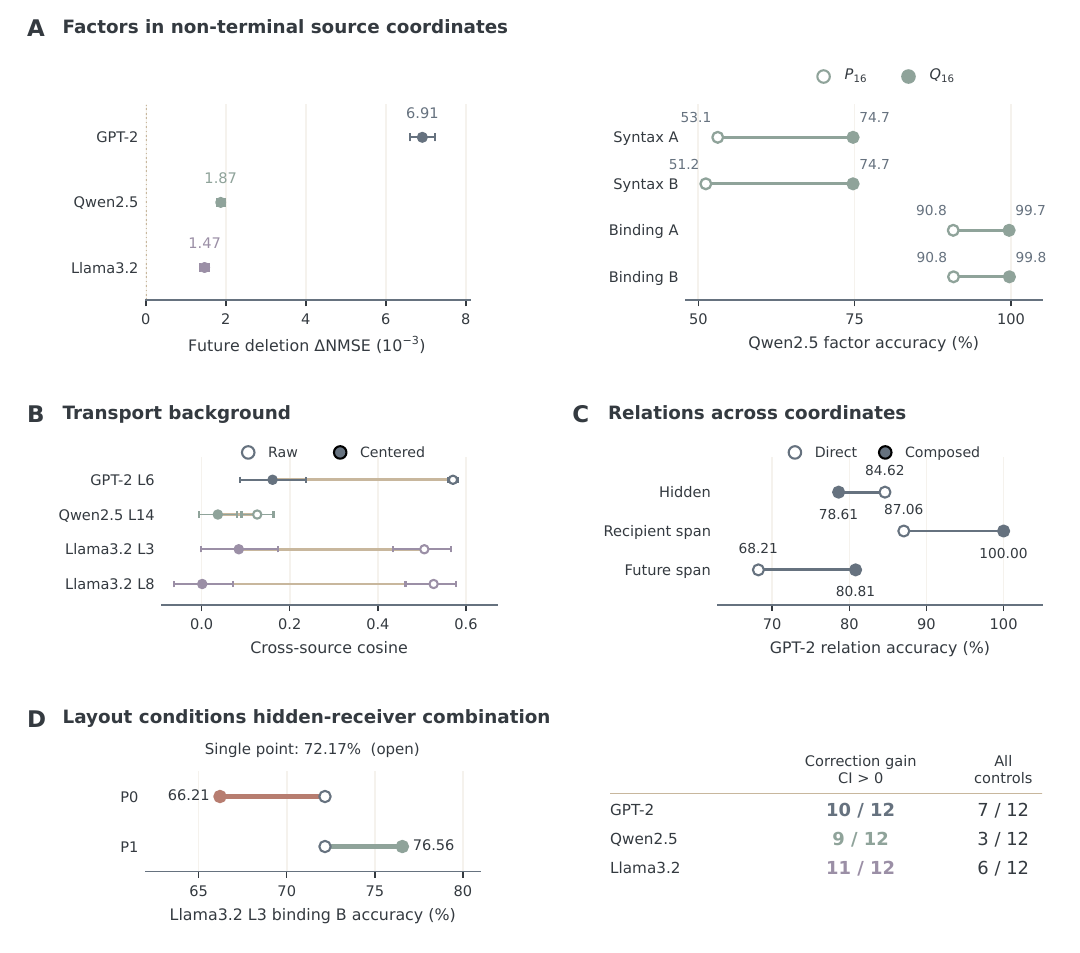}
\caption{Source-factor carrying and hidden-receiver complementarity. \textbf{A}, Middle-layer deletion effects for q7's estimated source coordinates at positions $s>7$, subtracted from the fixed full-query inverse sum; intervals are 95\% document-bootstrap intervals. Adjacent factor accuracies use Qwen2.5 prefix-source spans in principal and complementary statistical coordinates. \textbf{B}, Cross-source cosine at the same q14 and input node, before and after subtraction of an independently estimated inverse-state mean; intervals describe fixed mismatched pairs resampled by source document. \textbf{C}, Direct and attribute-composed relation accuracy in 512 matched GPT-2 binding groups. Composition is performed by the evaluator after separately decoding both attributes. \textbf{D}, Combining hidden-receiver features at q12--q15 yields opposite accuracy changes in two Llama3.2 layouts despite equal single-point accuracies. The adjacent fractions summarize 12 P0 panels per model: correction gains require positive 95\% intervals for correction adaptation with the new single-point baseline fixed; all-controls success requires positive 95\% intervals against the single-point, repeated-point and shuffled-position controls. These criteria are met in 30/36 and 16/36 panels, respectively.}
\label{fig:source-carrying}
\end{figure}

A nonterminal feature has prefix-limited forward sensitivity, while its source estimate can extend across the sequence through calibrated statistics and full-instance geometry. We call the estimated coordinates after the selected position its \emph{future component}. For q7, we subtracted this component from the same source's fixed full-query inverse sum. Input-reference NMSE increased in all nine model-by-depth panels. Middle-layer increases were $+0.006913$ in GPT-2, $+0.001873$ in Qwen2.5 and $+0.001470$ in Llama3.2 (Fig.~\ref{fig:source-carrying}A; Table~\ref{tab:source-carrying}). This deletion measures the component's contribution to that source-reference distance. Factor readouts independently establish which controlled variables the source coordinates carry.

Both principal and complementary source coordinates exposed task factors. In Qwen2.5 prefix source spans, the rank-16 statistical principal projection $P_{16}$ supported syntax accuracies of $53.11\%$ and $51.20\%$ for factors $A$ and $B$, whereas its orthogonal complement $Q_{16}$ supported $74.75\%$ for both. Binding readouts were high in both spaces: $90.77/90.82\%$ in $P_{16}$ and $99.71/99.76\%$ in $Q_{16}$. These projections partition the measured inverse representation; the natural hidden-state principal directions in Section~\ref{sec:results1} describe a different space. The high-dimensional complement thus carries directly readable factors alongside the principal component (Supplementary Table~\ref{tab:sup-source-readouts}).

Following the inverse chain distinguished this factor carrying from a shared natural background. At fixed q14 and input node 0, subtracting an inverse-state mean estimated from independent documents reduced cross-source cosine from $0.5705$ to $0.1610$ for GPT-2 L6, from $0.1261$ to $0.0368$ for Qwen2.5 L14, and from $0.5055/0.5266$ to $0.0843/0.0013$ for Llama3.2 L3/L8 (Fig.~\ref{fig:source-carrying}B). For the same Llama3.2 natural windows across L3 and L8, centered input vectors retained cosine $0.7391$ $[0.6697,0.8063]$ at q14; the all-query observation gave $0.9764$ $[0.9603,0.9869]$. The controlled binding contrasts had lower mean signed cross-depth cosine than the matched natural-window vectors (Supplementary Fig.~\ref{fig:sup-source-transport}E). Shared background, same-sample transport and controlled-factor redirection are therefore distinct properties of the source representation (Supplementary Fig.~\ref{fig:sup-source-transport}; Table~\ref{tab:sup-d5-input}).

\subsection{Hidden receivers and source spans expose different readable forms of a relation}
\label{sec:source-observation}
The extracted factors can support relation readout through different observation coordinates. We compared direct linear relation decoding with a two-step readout that separately decoded both attributes and then evaluated their equality. In 512 matched GPT-2 L6 binding groups, hidden receivers favored direct decoding: $84.62\%$ versus $78.61\%$ for attribute composition. Recipient source spans showed the reverse ordering, $87.06\%$ versus $100\%$, as did future source spans, $68.21\%$ versus $80.81\%$ (Fig.~\ref{fig:source-carrying}C). The paired direct-minus-composed differences were $+6.01$ $[3.86,8.11]$, $-12.94$ $[-14.65,-11.23]$ and $-12.60$ $[-14.36,-10.84]$ percentage points, respectively. Under the same readout budget, hidden receivers expose a directly readable relation while source spans particularly expose its constituent attributes (Supplementary Table~\ref{tab:sup-relation-domains}).

A fixed-prefix control directly separated forward support from source-estimate structure. In GPT-2, permuting the suffix while holding the first eight tokens fixed left $H_7$ and the first raw VJP unchanged, with zero raw energy after q7. Each variant supplied its own actual geometry to the fixed calibrated maps. Undoing the permutation raised cosine between the estimated suffixes from $0.4900$ to $0.8812$: the source estimate tracked suffix order while the native feature retained its prefix dependence. Qwen2.5 and Llama3.2 also showed positive permutation-alignment gains (Supplementary Section~\ref{sup:s7-order}). Role-based source slicing uses the known factor positions, while composed relation accuracy measures the evaluator's use of separately decoded attributes. Together, these measurements identify complementary readable structures in hidden and source coordinates.

\subsection{Cross-position complementarity depends on layout-adapted combination weights}
\label{sec:source-layout}
We tested cross-position combination using hidden-receiver features at q12--q15, with readouts fitted directly to the attention outputs before residual addition. Readable factors at several receivers do not by themselves determine whether combining them improves a readout. Llama3.2 L3 binding factor $B$ made this distinction explicit: the best single-point accuracy was $72.17\%$ in both layouts, but adding positions changed accuracy to $66.21\%$ in P0 and $76.56\%$ in P1 (Fig.~\ref{fig:source-carrying}D). Equal single-point accuracy therefore coexisted with opposite combination outcomes, although the two layouts retained different single-point cross-entropies.

We separated adaptation of the single-point readout from adaptation of its cross-position correction. With the new single-point baseline held fixed, retraining the correction at the same budget produced positive 95\% intervals in 30 of 36 P0 panels. Sixteen panels simultaneously exceeded the new single point, repeated-point controls and shuffled-extra-position controls; five panels retained negative adapted increments. The relative role ordering of q12--q15 was identical between layouts. The measured change therefore resides in the layout-conditioned combination weights (Supplementary Fig.~\ref{fig:sup-source-layout}; Table~\ref{tab:sup-d4-counts}). Distributed factor coverage in hidden receivers thus supports conditional complementarity through the readout that combines their features.

\begin{table}[!htbp]
\centering\small
\caption{Source-coordinate observations and hidden-receiver readouts. Source-deletion effects subtract q7's estimated coordinates at $s>7$ from the same source's fixed full-query inverse sum; changes in input-reference NMSE have 95\% document-bootstrap intervals. Relation contrasts compare hidden receivers and inverse source spans, using paired percentage-point differences in 512 source groups. Adaptation counts concern direct hidden-receiver features at q12--q15 and require positive 95\% group-bootstrap intervals; each model contributes 12 panels.}
\label{tab:source-carrying}
\begin{tabular}{llr}
\toprule
Measurement & Model or view & Estimate [95\% interval]\\
\midrule
\multicolumn{3}{l}{\textit{Future-component deletion at the middle layer}}\\
$\Delta$NMSE & GPT-2 & $+0.006913$ $[0.006611,0.007233]$\\
 & Qwen2.5 & $+0.001873$ $[0.001782,0.001970]$\\
 & Llama3.2 & $+0.001470$ $[0.001351,0.001583]$\\
\midrule
\multicolumn{3}{l}{\textit{Direct minus attribute-composed relation accuracy (pp)}}\\
GPT-2 binding & Hidden receivers & $+6.01$ $[3.86,8.11]$\\
 & Recipient source spans & $-12.94$ $[-14.65,-11.23]$\\
 & Future source spans & $-12.60$ $[-14.36,-10.84]$\\
\midrule
\multicolumn{3}{l}{\textit{Hidden receivers: correction adaptation / all three controls}}\\
P0 panels & GPT-2 & $10/12\quad /\quad 7/12$\\
 & Qwen2.5 & $9/12\quad /\quad 3/12$\\
 & Llama3.2 & $11/12\quad /\quad 6/12$\\
\bottomrule
\end{tabular}
\end{table}

The source-coordinate comparisons establish factor carrying and transport background, while the hidden-receiver experiment identifies layout-dependent combination. We next change the hidden-feature/geometry pairing itself to determine how the extracted source directions relate to downstream responses.

\section{Instance geometry, module specialization and local response prediction}
\label{sec:instance-geometry}

Source-coordinate alignment describes an inferred direction; reinjection tests its downstream effect. We therefore distinguish source-direction alignment, descendant-response direction and response amplitude while varying the selected feature and its instance geometry.

\subsection{Joint feature--geometry changes separate source alignment from response alignment}
\label{sec:geometry-joint}
A controlled source change alters both the selected feature $H$ and its instance geometry $J$. With the statistical maps fixed, let $\mathcal A(J)$ denote the instantiated source map. We compared
\begin{equation}
d_H=\mathcal A(J_0)(H_1-H_0),\qquad
d_T=\mathcal A(J_1)H_1-\mathcal A(J_0)H_0,
\label{eq:joint-source-change}
\end{equation}
which isolate a feature change at fixed geometry and a joint endpoint change, respectively. Reinjection retained only the changed source span and applied dose $0.25$ to both the inferred direction and the actual source-change reference. Matching the actual feature and geometry mattered for source extraction: GPT-2 source-reference NMSE was approximately $0.4790$ for the true pairing and $0.8922/0.8946$ after independent/whole-group geometry substitution.

\begin{figure}[!htbp]
\centering
\includegraphics[width=\linewidth,height=.79\textheight,keepaspectratio]{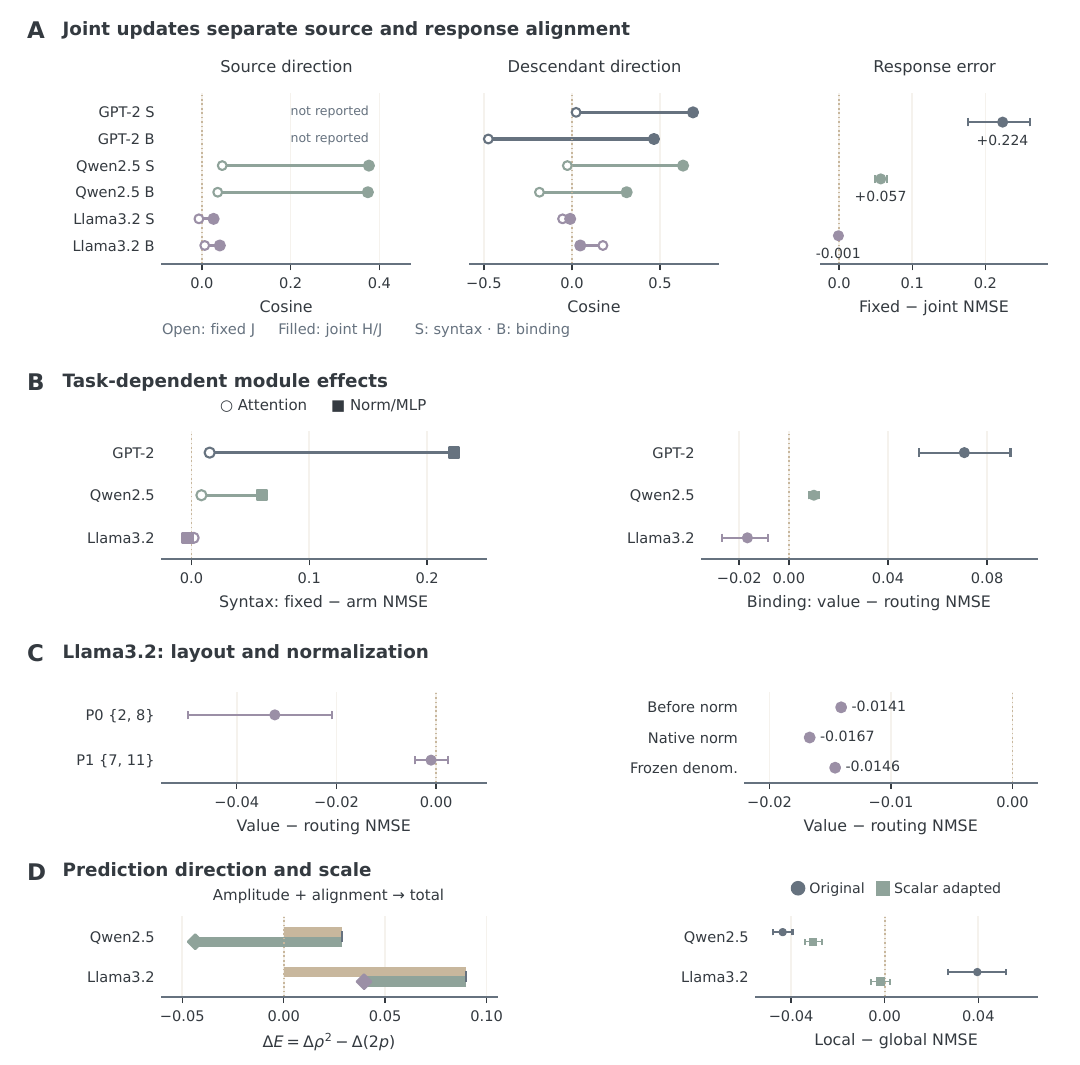}
\caption{Instance geometry separates source direction, response direction and amplitude. \textbf{A}, Original endpoint reinjection: fixed-$J$ (open) and joint-$H/J$ (filled) cosines, with response-NMSE improvement on a separate axis. This protocol reports source cosines for Qwen2.5/Llama3.2; GPT-2 source cosines were not separately reported. \textbf{B}, Synthetic endpoint updates: syntax module effects and binding value-minus-routing NMSE. \textbf{C}, Llama3.2 layout contrasts and persistence of the aggregate sign before final normalization. \textbf{D}, Syntax prediction error decomposed into norm and projection terms: sand segments show $\Delta\rho^2$, sage segments show $-\Delta(2p)$, and diamonds mark the total $\Delta E$. The adjacent panel shows five-fold scalar adaptation with prediction directions fixed. Points are measured means; error bars are 95\% source-group bootstrap intervals.}
\label{fig:instance-geometry}
\end{figure}

Reinjecting these source directions into native forward computation exposed a model-dependent relation between source and response alignment (Fig.~\ref{fig:instance-geometry}A). At natural amplitude, the fixed-minus-joint descendant-response NMSE difference was $+0.223557$ $[0.176577,0.260912]$ in GPT-2 and $+0.056955$ $[0.049063,0.065128]$ in Qwen2.5, whereas Llama3.2 gave $-0.000894$ $[-0.003550,0.001688]$. Joint updates strongly improved response cosine in GPT-2 and Qwen2.5 for both tasks. In Llama3.2, syntax cosine increased from $-0.0534$ to $-0.0102$, with an uncertain paired difference, $+0.0431$ $[-0.0164,0.1035]$; binding cosine decreased from $0.1756$ to $0.0470$, with difference $-0.1285$ $[-0.1798,-0.0808]$. Source-direction cosine increased in both Qwen2.5 and Llama3.2, establishing that better source alignment and better downstream alignment can separate (Supplementary Table~\ref{tab:sup-r1-source}).

To separate response direction from amplitude, we express the error of each predicted or reinjected response $\widehat r$ relative to its native reference $r$ as
\begin{equation}
E=\mathrm{NMSE}=1+a^2-2ac,\qquad
a=\frac{\lVert\widehat r\rVert}{\lVert r\rVert},\quad c=\cos(\widehat r,r).
\label{eq:nmse-decomposition}
\end{equation}
The identity is evaluated before averaging samples or factor flips. It separates a directional match from the response scale at which that direction is expressed.

\subsection{Module contributions depend on task, layout and response direction}
\label{sec:geometry-modules}
The residual-stream framework distinguishes attention's routing computation from the information carried by its value outputs~\citep{elhage2021framework}. We localized geometry dependence by updating selected endpoint derivatives inside the source map while holding $\WienerG/D$ fixed. Attention and normalization/MLP components were separated, and attention was further decomposed as
\begin{equation}
\delta(PV)=(\delta P)V+P(\delta V).
\label{eq:routing-value}
\end{equation}
The routing arm updated the complete $(\delta P)V$ endpoint, including its $V$ coefficient; the value arm updated the complete $P(\delta V)$ endpoint, including its $P$ coefficient. Native forward passes measured the responses to these synthetic source directions in 32 syntax and 32 binding groups per model (Supplementary Fig.~\ref{fig:sup-instance}; Table~\ref{tab:sup-d6-directions}).

The strongest syntax benefit in GPT-2 and Qwen2.5 came from normalization/MLP updates, with fixed-minus-arm NMSE differences of $+0.2226$ and $+0.0596$. Binding instead distinguished the attention terms: value-minus-routing NMSE was $+0.070782$ $[0.052651,0.089391]$ for GPT-2, $+0.010108$ $[0.008041,0.012168]$ for Qwen2.5 and $-0.016694$ $[-0.026928,-0.008398]$ for Llama3.2 (Fig.~\ref{fig:instance-geometry}B). A symmetric decomposition of the last contrast assigned $-0.014037$ to direction and $-0.002657$ to amplitude. The direction term accounted for approximately $84\%$ of the mean error difference in this algebraic decomposition.

Layout then resolved much of the Llama3.2 reversal. The value-minus-routing difference was $-0.032367$ $[-0.049873,-0.020976]$ in P0, with source positions $\{2,8\}$, and $-0.001020$ $[-0.004324,0.002405]$ in P1, with positions $\{7,11\}$; their interaction was $-0.031347$ $[-0.049146,-0.019308]$ (Fig.~\ref{fig:instance-geometry}C). Each layout contained 16 distinct source groups with balanced early/late queries. The aggregate difference was already $-0.014103$ before final normalization. These source-group strata identify a layout-conditioned division of the synthetic geometry updates, encompassing position, spacing and context. Their source-direction and descendant-direction orderings also differ in P1, reinforcing the three-metric distinction (Table~\ref{tab:working-points}; Supplementary Table~\ref{tab:sup-d6-directions}).

\subsection{Local feature--geometry neighborhoods improve response prediction through direction and scale}
\label{sec:geometry-local}
An independent prediction analysis compared global response models with local neighborhoods selected using $H$, $J$ or $H+J$. The model-specific predictors used standardized observation distances and development-selected fitting configurations to predict held-out descendant changes (Supplementary Section~\ref{sup:s8-geometry}). In binding, the joint neighborhood reduced raw NMSE relative to the global model by $0.0272$, $0.1383$ and $0.0651$ in GPT-2, Qwen2.5 and Llama3.2, respectively. The full predictor comparison retains the task-specific zero and negative results (Supplementary Fig.~\ref{fig:sup-local-prediction}; Table~\ref{tab:sup-c4-scale}).

Syntax supplied a direct separation of directional gain from error reduction. Rewriting equation~\eqref{eq:nmse-decomposition} as $E=1+\rho^2-2p$, where $p=\widehat y^{\mathsf T}y/\lVert y\rVert^2$, the local-minus-global differences in Qwen2.5 were $\Delta E=-0.043591$, $\Delta\rho^2=+0.028752$ and $\Delta(2p)=+0.072342$. Llama3.2 gave $+0.039569$, $+0.089924$ and $+0.050355$, respectively (Fig.~\ref{fig:instance-geometry}D). Its cosine nevertheless improved by $+0.022817$ $[0.014470,0.031320]$: the larger squared-norm term outweighed the stronger projection onto the true response.

To test this scale contribution directly, we held each prediction direction fixed and fitted one nonnegative scalar separately for the global and local arms. Five-fold cross-fitting kept complete source groups together, fitted scales on the other four folds and refitted them within bootstrap samples. Llama3.2 syntax changed from a clear local-model disadvantage to $\Delta E=-0.001754\;[-0.005966,0.002130]$. This supervised scale control removes the original directional-plus-amplitude disadvantage while leaving the relative total error unresolved (Table~\ref{tab:working-points}). Across these experiments, instance geometry changes the source directions exposed by inversion, and their downstream consequences depend separately on module composition, layout and response scale.

\section{Discussion and conclusion}
\label{sec:discussion}

\subsection{Shared relational organization and conditional use}
\label{sec:discussion-mechanism}
The results connect two properties that can appear contradictory: a relation can be shared across query roles, yet have a strongly context-dependent effect on selection. In the tested materials, middle-layer binding contrasts concentrate in a shared low-dimensional workspace and contain a predominantly role-independent $AB$ component. Independent donors transfer part of that component's functional effect across groups. The downstream network supplies the conditional sensitivity that converts the shared perturbation into a role-specific output contrast. The identity $\beta_G=\kappa_{BR}-\kappa_{AR}$ makes this connection explicit: a common direction can generate a selection effect because its local readout changes across attribute--query conditions.

This account extends the distinction between representing a binding and using it. Binding-ID analyses identify mechanisms that associate entities and attributes in context~\citep{feng2024binding}; linear relation decoding identifies useful maps between subject and object representations~\citep{hernandez2024linearity}. Here, factorial decomposition isolates the component shared across query roles, and forward intervention measures how that component enters selection. Its functional sign belongs to the direction together with its downstream working point and output contrast.

Linear-representation theory connects concept directions to causal interventions~\citep{park2024linear}. Steering studies document variability across inputs~\citep{tan2024steering}, opposite effects on individual held-out examples~\citep{braun2025unreliability}, and inversions between discriminative orientation and causal control~\citep{torop2026inverted}. Context-aware steering fields use gradients of a learned concept score to adapt intervention directions to the current activation~\citep{li2026vectorfields}. Our controlled materials connect these questions to role selection. Qwen2.5's fixed-direction reversal across prompt and answer vocabularies demonstrates that a transferable relation can change its signed effect within one model. The factorial identity specifies that dependence through a contrast of native downstream sensitivities. Separate late-component interventions measure the opposing paths that compose the local gain in the eight-condition protocol.

Position-level analyses connect this conditional use to distributed inputs. Studies of factual recall describe attribute enrichment at subject positions and subsequent attention-mediated transport to prediction positions~\citep{geva2023dissecting}. In our controlled materials, shared receiving windows increase non-additive combination. Removing selected post-softmax value contributions while retaining the other attention weights shows that candidate sources support later facts relative to earlier ones.

Component compensation provides a complementary perspective on pathway balance. The Hydra Effect measures downstream self-repair after component ablation~\citep{mcgrath2023hydra}, and conditional co-ablation identifies backups whose contributions emerge when another component is removed~\citep{gong2026coablation}. Our clean clamps preserve the same unedited forward state and partition the local response to the $AB$ perturbation. The resulting decomposition assigns signed transmission to paths according to their traversal of selected attention and MLP sets. Static normalized writes describe the composition of the original output; dynamic slopes describe how an edited relation propagates through it. Their opposite signs at individual nodes distinguish baseline output composition from perturbation transmission, while the path decomposition quantifies partial cancellation.

Source coordinates add a second level to the explanation (Fig.~\ref{fig:framework}B). The causal mask restricts native sensitivity to the visible prefix, while calibrated cross-position statistics and full-position consistency allow a broader source estimate. With $H_7$ fixed, inferred suffixes track source order through the actual instance geometry. Hidden retention of unqueried attributes and same-domain relations occurs among visible facts; source estimation combines the selected feature with geometry and calibration from the full instance. Both levels distinguish available structure from current selection. Source readouts expose recovered attributes and direct hidden-receiver readouts reveal relations; combining q12--q15 hidden receivers additionally exposes layout-dependent factor readout. Instance geometry further separates source alignment, descendant direction and response amplitude: improving one can worsen another. These comparisons connect factor coverage, relational organization and signed reading to their source-coordinate and downstream expression.

Table~\ref{tab:working-points} connects four cross-model differences to the comparisons that explain their measured dependencies. Together, they distinguish signed relation gain, fact-order balance, layout-dependent derivative specialization and response scale.

\begin{table}[!htbp]
\centering\footnotesize
\caption{\textbf{Cross-model differences and the comparisons that resolve their measured dependencies.} Triplets follow GPT-2, Qwen2.5 and Llama3.2 order. Relation derivatives are from the two-model 64-group late-clamp protocol; removal effects and NMSE differences retain their own units. The final interval is a 95\% source-group bootstrap interval. Full estimates and intervals are in Tables~\ref{tab:results2-core}, \ref{tab:supp-source-order}, \ref{tab:sup-d6-directions} and \ref{tab:sup-c4-scale}.}
\label{tab:working-points}
\setlength{\tabcolsep}{4pt}
\begin{tabularx}{\linewidth}{>{\raggedright\arraybackslash}p{.30\linewidth}>{\raggedright\arraybackslash}p{.32\linewidth}>{\raggedright\arraybackslash}X}
\toprule
Observed difference & Comparison and evidence scope & Mechanistic interpretation\\
\midrule
Relation gain: native $\beta_G$ is $+5.193$ in Qwen2.5 and $-7.425$ in Llama3.2 ($\times10^{-3}$). & Native edits and fixed final denominators; clean clamps of selected late components in 64 eight-condition groups. & Conditional sensitivities and the measured balance of selected tangent paths determine signed transfer.\\[5pt]
Candidate removal, pooled $\Delta G$: $-0.00218$, $+0.08621$, $-0.04516$. The GPT-2 interval spans zero. & Post-softmax removal, paired within 64 groups: late-minus-early $\Delta G$ is $-0.3039$, $-0.2068$, $-0.3983$, with negative intervals. & Candidate contributions support later facts relative to earlier ones across models, despite different pooled signs.\\[5pt]
Binding value-minus-routing NMSE: $+0.0708$, $+0.0101$, $-0.0167$. & Synthetic derivative endpoints at natural amplitude in 32 binding groups per model; direction/amplitude decomposition and 16 distinct groups per layout. & Llama3.2's value advantage is predominantly directional and concentrated in P0; layout strata vary position, spacing and context together.\\[5pt]
Syntax local-minus-global $H+J$ NMSE: $-0.0690$, $-0.0436$, $+0.0396$. & Held-out prediction with model-specific fits; norm/projection decomposition and five-fold scalar fitting at fixed prediction directions. & Llama3.2 combines better direction with excess amplitude. Calibrated local-minus-global error is $-0.00175$ [$-0.00597,+0.00213$], leaving the difference unresolved.\\
\bottomrule
\end{tabularx}
\end{table}

\subsection{Mechanistic implications and testable extensions}
\label{sec:discussion-hypotheses}
The source-grounded interface complements single-state probes by retaining input-position coordinates and the actual local derivatives associated with the selected feature. Forward interventions then test specific directional and component effects. Combining these views offers a practical way to study distributed computation: establish which factors are available, identify the relational contrasts they form, measure the conditional gains that read those contrasts, and examine how source geometry changes their expression.

The architecture measurements in Supplementary Table~\ref{tab:sup-architecture} motivate targeted interventions. Small bidirectional perturbations of existing query and key biases in fixed GPT-2 and Qwen2.5 checkpoints could test their effects on routing/value specialization and signed relation gain. GPT-2's shared key bias supplies a predicted negative control because its common logit offset cancels in softmax. These interventions identify contributions within trained circuits; matched training comparisons would test architectural causes. Separate controlled comparisons of inter-fact spacing, prefix padding and absolute token position using the same lexical facts would distinguish positional dependencies bundled into the present layout contrast. Expanding attribute domains, value sets and model families would determine how far shared relation axes transfer and when new axes are required.

Across the three models, distributed factors form a shared relational background whose use depends on conditional sensitivity and pathway competition. Source transport and instance geometry shape its input-coordinate presentation and downstream response. Layout further determines how hidden receivers can be combined to read its factors.

\FloatBarrier
\clearpage
\section*{Code availability}
The source-grounded feature inversion implementation is publicly available at \url{https://github.com/k-Jayus/source-grounded-feature-inversion}.

\bibliographystyle{plainnat}
\bibliography{references}

@article{shu2026source,
  author = {Shu, Kaixiang},
  title = {From Preimage Search to Source-Grounded Feature Inversion},
  journal = {arXiv preprint arXiv:2607.12526},
  year = {2026},
  doi = {10.48550/arXiv.2607.12526},
  url = {https://arxiv.org/abs/2607.12526}
}

@techreport{radford2019language,
 author = {Radford, Alec and Wu, Jeffrey and Child, Rewon and Luan, David and Amodei, Dario and Sutskever, Ilya},
 title = {Language Models are Unsupervised Multitask Learners},
 institution = {OpenAI},
 year = {2019},
 url = {https://cdn.openai.com/better-language-models/language_models_are_unsupervised_multitask_learners.pdf}
}

@article{qwen2024technical,
 author = {{Qwen Team}},
 title = {{Qwen2.5} Technical Report},
 journal = {arXiv preprint arXiv:2412.15115},
 year = {2024},
 doi = {10.48550/arXiv.2412.15115},
 url = {https://arxiv.org/abs/2412.15115}
}

@misc{meta2024llama32,
 author = {{Meta}},
 title = {{Llama 3.2} Model Card},
 year = {2024},
 url = {https://github.com/meta-llama/llama-models/blob/main/models/llama3_2/MODEL_CARD.md}
}

@article{belinkov2022probing,
 author = {Belinkov, Yonatan},
 title = {Probing Classifiers: Promises, Shortcomings, and Advances},
 journal = {Computational Linguistics},
 volume = {48}, number = {1}, pages = {207--219},
 year = {2022}, doi = {10.1162/coli_a_00422},
 url = {https://doi.org/10.1162/coli_a_00422}
}

@inproceedings{feng2024binding,
 author = {Feng, Jiahai and Steinhardt, Jacob},
 title = {How do Language Models Bind Entities in Context?},
 booktitle = {The Twelfth International Conference on Learning Representations},
 year = {2024},
 url = {https://openreview.net/forum?id=zb3b6oKO77}
}

@inproceedings{hernandez2024linearity,
 author = {Hernandez, Evan and Sharma, Arnab Sen and Haklay, Tal and Meng, Kevin and Wattenberg, Martin and Andreas, Jacob and Belinkov, Yonatan and Bau, David},
 title = {Linearity of Relation Decoding in Transformer Language Models},
 booktitle = {The Twelfth International Conference on Learning Representations},
 year = {2024},
 url = {https://openreview.net/forum?id=w7LU2s14kE}
}

@inproceedings{meng2022locating,
 author = {Meng, Kevin and Bau, David and Andonian, Alex and Belinkov, Yonatan},
 title = {Locating and Editing Factual Associations in {GPT}},
 booktitle = {Advances in Neural Information Processing Systems},
 volume = {35}, pages = {17359--17372},
 publisher = {Curran Associates, Inc.},
 year = {2022}, doi = {10.52202/068431-1262},
 url = {https://proceedings.neurips.cc/paper_files/paper/2022/hash/6f1d43d5a82a37e89b0665b33bf3a182-Abstract-Conference.html}
}

@inproceedings{wang2023ioi,
 author = {Wang, Kevin Ro and Variengien, Alexandre and Conmy, Arthur and Shlegeris, Buck and Steinhardt, Jacob},
 title = {Interpretability in the Wild: a Circuit for Indirect Object Identification in {GPT-2} Small},
 booktitle = {The Eleventh International Conference on Learning Representations},
 year = {2023},
 url = {https://openreview.net/forum?id=NpsVSN6o4ul}
}

@inproceedings{conmy2023automated,
 author = {Conmy, Arthur and Mavor-Parker, Augustine and Lynch, Aengus and Heimersheim, Stefan and Garriga-Alonso, Adri{\`a}},
 title = {Towards Automated Circuit Discovery for Mechanistic Interpretability},
 booktitle = {Advances in Neural Information Processing Systems},
 volume = {36}, pages = {16318--16352},
 publisher = {Curran Associates, Inc.},
 year = {2023}, doi = {10.52202/075280-0719},
 url = {https://proceedings.neurips.cc/paper_files/paper/2023/hash/34e1dbe95d34d7ebaf99b9bcaeb5b2be-Abstract-Conference.html}
}

@inproceedings{park2024linear,
 author = {Park, Kiho and Choe, Yo Joong and Veitch, Victor},
 title = {The Linear Representation Hypothesis and the Geometry of Large Language Models},
 booktitle = {Proceedings of the 41st International Conference on Machine Learning},
 series = {Proceedings of Machine Learning Research},
 volume = {235}, pages = {39643--39666},
 publisher = {PMLR},
 year = {2024},
 url = {https://proceedings.mlr.press/v235/park24c.html}
}

@inproceedings{geiger2021causal,
 author = {Geiger, Atticus and Lu, Hanson and Icard, Thomas and Potts, Christopher},
 title = {Causal Abstractions of Neural Networks},
 booktitle = {Advances in Neural Information Processing Systems},
 volume = {34}, pages = {9574--9586},
 publisher = {Curran Associates, Inc.},
 year = {2021},
 url = {https://proceedings.neurips.cc/paper_files/paper/2021/hash/4f5c422f4d49a5a807eda27434231040-Abstract.html}
}

@inproceedings{zhang2024patching,
 author = {Zhang, Fred and Nanda, Neel},
 title = {Towards Best Practices of Activation Patching in Language Models: Metrics and Methods},
 booktitle = {The Twelfth International Conference on Learning Representations},
 year = {2024},
 url = {https://openreview.net/forum?id=Hf17y6u9BC}
}

@article{su2024roformer,
 author = {Su, Jianlin and Ahmed, Murtadha and Lu, Yu and Pan, Shengfeng and Bo, Wen and Liu, Yunfeng},
 title = {{RoFormer}: Enhanced Transformer with Rotary Position Embedding},
 journal = {Neurocomputing},
 volume = {568}, pages = {127063},
 year = {2024}, doi = {10.1016/j.neucom.2023.127063},
 url = {https://doi.org/10.1016/j.neucom.2023.127063}
}

@inproceedings{zhang2019rmsnorm,
 author = {Zhang, Biao and Sennrich, Rico},
 title = {Root Mean Square Layer Normalization},
 booktitle = {Advances in Neural Information Processing Systems},
 volume = {32},
 publisher = {Curran Associates, Inc.},
 year = {2019},
 url = {https://proceedings.neurips.cc/paper_files/paper/2019/hash/1e8a19426224ca89e83cef47f1e7f53b-Abstract.html}
}

@inproceedings{ainslie2023gqa,
 author = {Ainslie, Joshua and Lee-Thorp, James and de Jong, Michiel and Zemlyanskiy, Yury and Lebron, Federico and Sanghai, Sumit},
 title = {{GQA}: Training Generalized Multi-Query Transformer Models from Multi-Head Checkpoints},
 booktitle = {Proceedings of the 2023 Conference on Empirical Methods in Natural Language Processing},
 pages = {4895--4901},
 publisher = {Association for Computational Linguistics},
 year = {2023}, doi = {10.18653/v1/2023.emnlp-main.298},
 url = {https://aclanthology.org/2023.emnlp-main.298/}
}

@article{shazeer2020glu,
 author = {Shazeer, Noam},
 title = {{GLU} Variants Improve Transformer},
 journal = {arXiv preprint arXiv:2002.05202},
 year = {2020}, doi = {10.48550/arXiv.2002.05202},
 url = {https://arxiv.org/abs/2002.05202}
}

@article{linzen2016syntax,
    title = "Assessing the Ability of {LSTM}s to Learn Syntax-Sensitive Dependencies",
    author = "Linzen, Tal  and
      Dupoux, Emmanuel  and
      Goldberg, Yoav",
    editor = "Lee, Lillian  and
      Johnson, Mark  and
      Toutanova, Kristina",
    journal = "Transactions of the Association for Computational Linguistics",
    volume = "4",
    year = "2016",
    address = "Cambridge, MA",
    publisher = "MIT Press",
    url = "https://aclanthology.org/Q16-1037/",
    doi = "10.1162/tacl_a_00115",
    pages = "521--535"
}

@inproceedings{finlayson2021agreement,
    title = "Causal Analysis of Syntactic Agreement Mechanisms in Neural Language Models",
    author = "Finlayson, Matthew  and
      Mueller, Aaron  and
      Gehrmann, Sebastian  and
      Shieber, Stuart  and
      Linzen, Tal  and
      Belinkov, Yonatan",
    editor = "Zong, Chengqing  and
      Xia, Fei  and
      Li, Wenjie  and
      Navigli, Roberto",
    booktitle = "Proceedings of the 59th Annual Meeting of the {A}ssociation for {C}omputational {L}inguistics and the 11th International Joint Conference on Natural Language Processing (Volume 1: Long Papers)",
    month = aug,
    year = "2021",
    address = "Online",
    publisher = "Association for Computational Linguistics",
    url = "https://aclanthology.org/2021.acl-long.144/",
    doi = "10.18653/v1/2021.acl-long.144",
    pages = "1828--1843"
}

@inproceedings{ravichander2021probing,
    title = "Probing the Probing Paradigm: Does Probing Accuracy Entail Task Relevance?",
    author = "Ravichander, Abhilasha  and
      Belinkov, Yonatan  and
      Hovy, Eduard",
    editor = "Merlo, Paola  and
      Tiedemann, Jorg  and
      Tsarfaty, Reut",
    booktitle = "Proceedings of the 16th Conference of the {E}uropean Chapter of the {A}ssociation for {C}omputational {L}inguistics: Main Volume",
    month = apr,
    year = "2021",
    address = "Online",
    publisher = "Association for Computational Linguistics",
    url = "https://aclanthology.org/2021.eacl-main.295/",
    doi = "10.18653/v1/2021.eacl-main.295",
    pages = "3363--3377"
}

@inproceedings{geva2023dissecting,
    title = "Dissecting Recall of Factual Associations in Auto-Regressive Language Models",
    author = "Geva, Mor  and
      Bastings, Jasmijn  and
      Filippova, Katja  and
      Globerson, Amir",
    editor = "Bouamor, Houda  and
      Pino, Juan  and
      Bali, Kalika",
    booktitle = "Proceedings of the 2023 Conference on Empirical Methods in Natural Language Processing",
    month = dec,
    year = "2023",
    address = "Singapore",
    publisher = "Association for Computational Linguistics",
    url = "https://aclanthology.org/2023.emnlp-main.751/",
    doi = "10.18653/v1/2023.emnlp-main.751",
    pages = "12216--12235"
}

@inproceedings{li2023iti,
 author = {Li, Kenneth and Patel, Oam and Vi\'{e}gas, Fernanda and Pfister, Hanspeter and Wattenberg, Martin},
 booktitle = {Advances in Neural Information Processing Systems},
 doi = {10.52202/075280-1797},
 editor = {A. Oh and T. Naumann and A. Globerson and K. Saenko and M. Hardt and S. Levine},
 pages = {41451--41530},
 publisher = {Curran Associates, Inc.},
 title = {Inference-Time Intervention: Eliciting Truthful Answers from a Language Model},
 url = {https://proceedings.neurips.cc/paper_files/paper/2023/file/81b8390039b7302c909cb769f8b6cd93-Paper-Conference.pdf},
 volume = {36},
 year = {2023}
}

@inproceedings{tan2024steering,
 author = {Tan, Daniel and Chanin, David and Lynch, Aengus and Paige, Brooks and Kanoulas, Dimitrios and Garriga-Alonso, Adri\`{a} and Kirk, Robert},
 booktitle = {Advances in Neural Information Processing Systems},
 doi = {10.52202/079017-4417},
 editor = {A. Globerson and L. Mackey and D. Belgrave and A. Fan and U. Paquet and J. Tomczak and C. Zhang},
 pages = {139179--139212},
 publisher = {Curran Associates, Inc.},
 title = {Analysing the Generalisation and Reliability of Steering Vectors},
 url = {https://proceedings.neurips.cc/paper_files/paper/2024/file/fb3ad59a84799bfb8d700e56d19c231b-Paper-Conference.pdf},
 volume = {37},
 year = {2024}
}

@article{turner2023activation,
  author = {Turner, Alexander Matt and Thiergart, Lisa and Leech, Gavin and Udell, David and Vazquez, Juan J. and Mini, Ulisse and MacDiarmid, Monte},
  title = {Steering Language Models With Activation Engineering},
  journal = {arXiv preprint arXiv:2308.10248},
  year = {2023},
  doi = {10.48550/arXiv.2308.10248},
  url = {https://arxiv.org/abs/2308.10248v5}
}

@article{zou2023representation,
  author = {Zou, Andy and Phan, Long and Chen, Sarah and Campbell, James and Guo, Phillip and Ren, Richard and Pan, Alexander and Yin, Xuwang and Mazeika, Mantas and Dombrowski, Ann-Kathrin and Goel, Shashwat and Li, Nathaniel and Byun, Michael J. and Wang, Zifan and Mallen, Alex and Basart, Steven and Koyejo, Sanmi and Song, Dawn and Fredrikson, Matt and Kolter, J. Zico and Hendrycks, Dan},
  title = {Representation Engineering: A Top-Down Approach to {AI} Transparency},
  journal = {arXiv preprint arXiv:2310.01405},
  year = {2023},
  doi = {10.48550/arXiv.2310.01405},
  url = {https://arxiv.org/abs/2310.01405v4}
}

@article{braun2025unreliability,
  author = {Braun, Joschka and Eickhoff, Carsten and Krueger, David and Bahrainian, Seyed Ali and Krasheninnikov, Dmitrii},
  title = {Understanding (Un)Reliability of Steering Vectors in Language Models},
  journal = {arXiv preprint arXiv:2505.22637},
  year = {2025},
  doi = {10.48550/arXiv.2505.22637},
  url = {https://arxiv.org/abs/2505.22637v1}
}

@article{torop2026inverted,
  author = {Torop, Max and Masoomi, Aria and Dy, Jennifer},
  title = {Inverted Detection and Control in Steering Vectors},
  journal = {arXiv preprint arXiv:2608.02957},
  year = {2026},
  doi = {10.48550/arXiv.2608.02957},
  url = {https://arxiv.org/abs/2608.02957v1}
}

@article{li2026vectorfields,
  author = {Li, Jiaqian and Li, Yanshu and Huang, Kuan-Hao},
  title = {Steering Vector Fields for Context-Aware Inference-Time Control in Large Language Models},
  journal = {arXiv preprint arXiv:2602.01654},
  year = {2026},
  doi = {10.48550/arXiv.2602.01654},
  url = {https://arxiv.org/abs/2602.01654v2}
}

@article{mcgrath2023hydra,
  author = {McGrath, Thomas and Rahtz, Matthew and Kramar, Janos and Mikulik, Vladimir and Legg, Shane},
  title = {The Hydra Effect: Emergent Self-repair in Language Model Computations},
  journal = {arXiv preprint arXiv:2307.15771},
  year = {2023},
  doi = {10.48550/arXiv.2307.15771},
  url = {https://arxiv.org/abs/2307.15771v1}
}

@article{goldowskydill2023pathpatching,
  author = {Goldowsky-Dill, Nicholas and MacLeod, Chris and Sato, Lucas and Arora, Aryaman},
  title = {Localizing Model Behavior with Path Patching},
  journal = {arXiv preprint arXiv:2304.05969},
  year = {2023},
  doi = {10.48550/arXiv.2304.05969},
  url = {https://arxiv.org/abs/2304.05969v2}
}

@article{merity2016pointer,
  author = {Merity, Stephen and Xiong, Caiming and Bradbury, James and Socher, Richard},
  title = {Pointer Sentinel Mixture Models},
  journal = {arXiv preprint arXiv:1609.07843},
  year = {2016},
  doi = {10.48550/arXiv.1609.07843},
  url = {https://arxiv.org/abs/1609.07843v1}
}

@article{gong2026coablation,
  author = {Gong, Zhiren and Lu, He and Wang, Tiantong and Zhang, Yichi and Wang, Yixin and Zeng, Zihao and Xiao, Ming and Yuen, Chau and Lim, Wei Yang Bryan},
  title = {Conditional Co-Ablation: Recovering Self-Repair Backups in Transformer Circuits},
  journal = {arXiv preprint arXiv:2607.01940},
  year = {2026},
  doi = {10.48550/arXiv.2607.01940},
  url = {https://arxiv.org/abs/2607.01940v3}
}

@article{elhage2021framework,
   title={A Mathematical Framework for Transformer Circuits},
   author={Elhage, Nelson and Nanda, Neel and Olsson, Catherine and Henighan, Tom and Joseph, Nicholas and Mann, Ben and Askell, Amanda and Bai, Yuntao and Chen, Anna and Conerly, Tom and DasSarma, Nova and Drain, Dawn and Ganguli, Deep and Hatfield-Dodds, Zac and Hernandez, Danny and Jones, Andy and Kernion, Jackson and Lovitt, Liane and Ndousse, Kamal and Amodei, Dario and Brown, Tom and Clark, Jack and Kaplan, Jared and McCandlish, Sam and Olah, Chris},
   year={2021},
   journal={Transformer Circuits Thread},
   url={https://transformer-circuits.pub/2021/framework/index.html}
}

% Supplementary Materials, compiled in the same document.
\clearpage
\pdfbookmark[0]{Supplementary materials}{supplementary-materials}
\renewcommand{\shorttitle}{Supplementary Materials: shared relations, specific choices}
\title{Supplementary Materials: Shared relations, specific choices: conditional sensitivity and competing pathways in language models}
\author{Kaixiang Shu\\{\normalfont Independent Researcher}}
\date{}
\setcounter{topnumber}{4}
\setcounter{bottomnumber}{3}
\setcounter{totalnumber}{6}
\renewcommand{\topfraction}{0.92}
\renewcommand{\bottomfraction}{0.85}
\renewcommand{\textfraction}{0.06}
\renewcommand{\floatpagefraction}{0.72}
\raggedbottom
\supplementarymaketitle
\appendix
\renewcommand{\thesection}{S\arabic{section}}
\renewcommand{\thesubsection}{\thesection.\arabic{subsection}}
\renewcommand{\theHsection}{supp.\arabic{section}}
\renewcommand{\theHsubsection}{supp.\arabic{section}.\arabic{subsection}}
\setcounter{section}{0}
\renewcommand{\thefigure}{S\arabic{figure}}
\renewcommand{\thetable}{S\arabic{table}}
\renewcommand{\theequation}{S\arabic{equation}}
\renewcommand{\theHfigure}{supp.\arabic{figure}}
\renewcommand{\theHtable}{supp.\arabic{table}}
\renewcommand{\theHequation}{supp.\arabic{equation}}
\setcounter{figure}{0}
\setcounter{table}{0}
\setcounter{equation}{0}
\section{Models and controlled materials}
\label{sup:s1-design}

The controlled designs separate attribute organization, query-role dependence and the conditions under which a relation is retained. We use pretrained base checkpoints and preserve each model's native attention, positional encoding and normalization. Model forward computation runs in evaluation mode with FP32 parameters, eager attention and caching disabled. The recorded implementation uses PyTorch 2.5.1 and Transformers 4.46.3. Layer indices are one-based; sequence indices are zero-based.

\subsection{Architecture and receiver alignment}
\begin{figure}[!htbp]
\centering
\includegraphics[width=\linewidth]{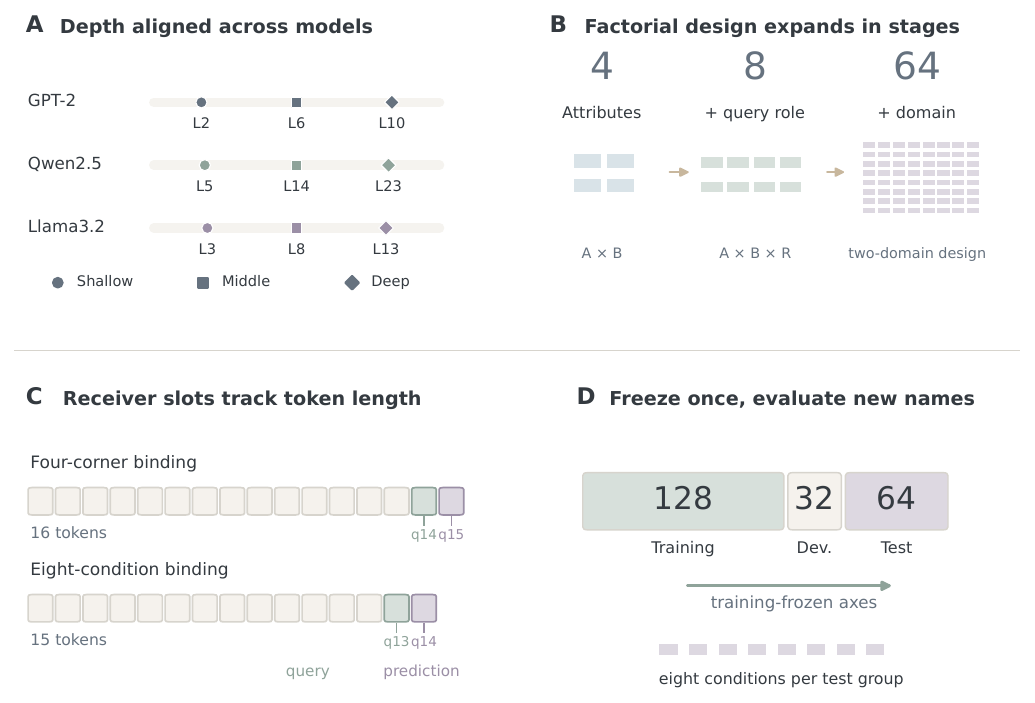}
\caption{\textbf{Controlled design and receiver alignment.} \textbf{A}, Shallow, middle and deep anchors, positioned by layer index divided by total model depth. \textbf{B}, Related controlled designs cross two attributes (four conditions), attributes and query role (eight conditions), or color, shape, queried entity and queried domain (64 conditions). Each design uses its own complete source groups. \textbf{C}, The query and prediction slots differ between the 16-token and 15-token binding families. \textbf{D}, The role-balanced design uses disjoint training, development and test names; projections and interventions reuse training-frozen bases.}
\label{fig:supp-design}
\end{figure}

The middle anchors lie halfway through each model's blocks (Table~\ref{tab:supp-design}; Fig.~\ref{fig:supp-design}). A hidden receiver is the attention output after output projection and before residual addition. Query-entity and prediction-token receivers are adjacent in the binding materials, but the slot indices depend on the material family. This distinction permits comparisons of receiver function without conflating sentence lengths.

\begin{table}[!htbp]
\centering\small
\caption{\textbf{Architecture and material alignment.} Panel A records the actual model configurations. Panel B records the counterfactual groups used for the principal controlled designs; training, development and test counts refer to complete source groups.}
\label{tab:supp-design}
\setlength{\tabcolsep}{4pt}
\begin{tabularx}{\linewidth}{lXXX}
\toprule
\multicolumn{4}{l}{\textit{A. Model configuration}}\\
Attribute & GPT-2 & Qwen2.5 & Llama3.2\\
\midrule
Blocks / channels & 12 / 768 & 28 / 1,536 & 16 / 2,048\\
Query / KV heads & 12 / 12 & 12 / 2 & 32 / 8\\
Head dimension & 64 & 128 & 64\\
Unique parameters & 124,439,808 & 1,543,714,304 & 1,235,814,400\\
Normalization & LayerNorm & RMSNorm & RMSNorm\\
Positional encoding & Absolute & RoPE & RoPE\\
Q/K/V projection bias & Yes / yes / yes & Yes / yes / yes & No / no / no\\
Shallow/middle/deep & L2/L6/L10 & L5/L14/L23 & L3/L8/L13\\
\midrule
\multicolumn{4}{l}{\textit{B. Material families and pairing}}\\
Family & \multicolumn{1}{l}{Groups} & \multicolumn{2}{l}{Controlled variation and slots}\\
\midrule
Four-corner directions & 64 per task/model & \multicolumn{2}{p{.46\linewidth}}{Two factors. Binding length 16; query/prediction positions 14/15. Shared-receiver output positions: syntax 9, binding 15.}\\
Role decomposition & 128/32/64 & \multicolumn{2}{p{.46\linewidth}}{Eight $A\times B\times R$ conditions. Test length 15; query/prediction positions 13/14.}\\
Finite component edits & 64 test & \multicolumn{2}{p{.46\linewidth}}{Training-frozen bases; test-group projection amplitudes. Initial analytic and direct-path comparisons use the first 32 groups.}\\
Three values / lexemes & 32/16/64 & \multicolumn{2}{p{.46\linewidth}}{Complete counterfactual groups. Lexical mapping controls match length and query position.}\\
Two domains & 32/16/32 & \multicolumn{2}{p{.46\linewidth}}{64 conditions: color and shape assignments, queried entity and queried domain. Direct and shared binary answer codes.}\\
Late-component clamps & Same 64 test & \multicolumn{2}{p{.46\linewidth}}{Qwen2.5/Llama3.2; original rank-16 directions; six selected late nodes and causal-order controls.}\\
\bottomrule
\end{tabularx}
\end{table}

\subsection{Complete groups separate factors from query roles}
In a four-corner group, all four combinations of two binary factors share the remaining context. The eight-condition design adds the independently varied queried entity. Its training, development and test sets use disjoint names with shared colors and templates. Each test set contains 512 sentences, and the 15-token lengths and role spans agree across the three tokenizers. Development materials retain their own tokenization records, so test-slot indices apply specifically to the test materials.

For the three-value and different-lexeme relation classifiers, each complete counterfactual group's unlabeled mean is subtracted before projection onto the frozen original relation axis. Reported balanced accuracies therefore measure a relation within the group-supported coordinate. The eight-condition $AB$ axis used for held-out-name transfer is fitted separately. The two-domain design crosses the queried domain with the property assignments, allowing queried, unqueried same-domain and cross-domain interactions to be evaluated on matched facts. Its 64 conditions include four binary property assignments and two binary query factors.

\subsection{Stimulus templates, vocabularies and split rules}
Supplementary Data (\texttt{materials/}) provides every frozen prompt, condition label, source-group ID, split and recorded tokenization for the four-corner and layout families, together with the full eight-condition and lexical/domain/value materials. The accompanying manifests list the actual retained word pools and template assignments; the original material-generation routines specify punctuation, neutral prefixes and fact ordering. Four-corner syntax varies subject and attractor number in a common sentence skeleton with 12 relation words; binding varies red/blue attributes with 12 predicate words. Training uses template indices 0--5 and layouts 0--1; development uses templates 6--7 and layout 2; unseen confirmation conditions use templates 8--11 and layout 3. Confirmation crosses template, position and vocabulary familiarity in eight cells. These are the stored lexical-template families and source-group partitions.

The eight-condition design uses shuffled ordered name pairs (seed 2026102907), the color pairs red/blue, green/black, gold/silver and yellow/purple, and four prefixes: ``In this town,'' ``In the story,'' ``At the school,'' and ``For this game.'' Each prompt joins two ``[name] wears [color].'' facts and ``Now [queried name] wears''; fact order alternates in four-group blocks. The transfer materials use separate split-specific names and color vocabularies, explicit lexical-to-category dictionaries, and the full attribute/query factorials. Their readable protocol and word inventories are in \texttt{materials/README.md}; the fixed-axis comparison's exact prompts, output words and coding are shown in Section~\ref{sup:s4-material-protocol}.

\subsection{Position, direction and component operations are distinct comparisons}
Layout experiments compare readouts trained on old and new positions, then hold the new single-position predictor fixed while adapting the additional-position correction. Copied and shuffled extra positions test the contribution of complementary observations. In the forward graph, direction edits alter the middle query receiver; clean clamps restore prediction-position component writes to the same condition's baseline; source closure removes designated post-softmax value contributions. Each experiment pairs its intervention with a clean forward pass under the same model and numerical backend.

\FloatBarrier
\section{Source-coordinate estimation and statistical inference}
\label{sup:s2-wiener}

\begin{figure}[!htbp]
\centering
\includegraphics[width=\linewidth]{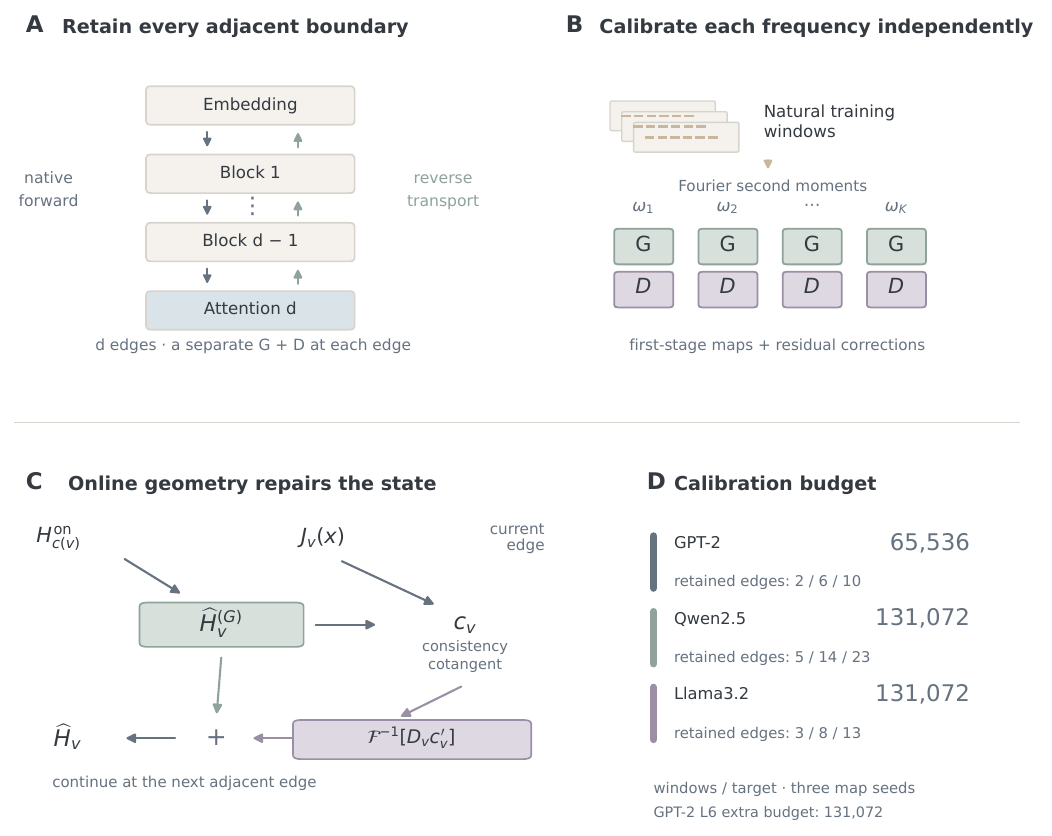}
\caption{\textbf{Calibration and online transport have distinct roles.} \textbf{A}, A target at depth $d$ retains every adjacent boundary up to its attention output. \textbf{B}, Independent natural windows fit full-channel, frequency-specific first-stage maps and residual corrections. \textbf{C}, The selected feature, local derivative and repaired online child determine each reverse step. \textbf{D}, Calibration budgets and retained edge counts vary by model and depth; repeated map seeds are aggregated within the source sampling unit.}
\label{fig:supp-wiener}
\end{figure}

The source-coordinate interface transports the structure associated with a selected feature through the model's actual local derivatives. Calibration supplies frequency-specific statistical maps; the online reverse pass supplies the feature and instance geometry. Figure~\ref{fig:supp-wiener} separates these two stages.

\subsection{Coordinate and operator definitions}
At parent node $v$, $H_v\in\mathbb R^{L\times C_v}$ is the forward state and $R_v$ is the local reverse signal. Under the full-channel mean-seed convention, the target cotangent is activation weighted and divided by channel count; the corresponding channel scaling is retained at reverse boundaries. The local derivative $J_v$ and adjoint $J_v^*$ are evaluated at the actual example. An orthonormal real Fourier transform acts on token position, preserving all channels. A 16-token window has nine stored frequency bins, including DC and Nyquist.

\begin{table}[!htbp]
\centering\small
\caption{\textbf{Source-coordinate operators.} Stars denote complex-conjugate transpose in Fourier moments and the adjoint for local derivatives. The moments are uncentered because both regression stages have zero intercept.}
\label{tab:supp-operators}
\begin{tabularx}{\linewidth}{lX}
\toprule
Symbol & Definition\\
\midrule
$H_v,R_v\in\mathbb R^{L\times C_v}$ & Parent state and activation-weighted reverse signal.\\
$J_v$ & Local derivative from parent state to retained child state.\\
$X'_v=\mathcal F X_v$ & Orthonormal transform along token position.\\
$S_{XY,v}(\omega)$ & $\mathbb E[X'_v(\omega)Y'_v(\omega)^*]$ on calibration windows.\\
$\mathsf G_v(\omega)$ & $S_{HR,v}[S_{RR,v}+\lambda_{G,v}I]^{-1}$, an independent full-channel map for each bin.\\
$\widehat H_v^{(G)}$ & $\mathcal F^{-1}[\mathsf G_vR'_v]$.\\
$c_v$ & $J_v^*(H_{c(v)}^{\rm on}-J_v\widehat H_v^{(G)})$, the forward-consistency cotangent.\\
$\Delta H_v$ & $H_v-\widehat H_v^{(G)}$, the second-stage regression target.\\
$D_v(\omega)$ & $S_{\Delta H,c,v}[S_{cc,v}+\lambda_{D,v}I]^{-1}$.\\
$\widehat H_v$ & $\widehat H_v^{(G)}+\mathcal F^{-1}[D_vc'_v]$.\\
\bottomrule
\end{tabularx}
\end{table}

For target attention at layer $d$, the retained chain is embedding, complete blocks $1,\ldots,d-1$, then the target attention output before residual addition. There are exactly $d$ adjacent edges, each with an independently fitted $\mathsf G/D$ pair. The repaired child state becomes the online input to the next upstream correction. The second stage regresses a state residual from a consistency cotangent, and its full-channel matrices are estimated independently across frequency bins.

\subsection{Causal support and the support of a source estimate}
\label{sup:s2-support}
Let $P_q$ retain only position $q$ in the target attention-output tensor. In a causally masked forward graph, $\partial H_q/\partial x_s=0$ for $s>q$, so the raw adjoint of this selected feature is zero at later source positions. This derivative includes every computational path from the input to the selected feature. The source estimator takes $P_qH$, the actual local derivatives over all positions on the retained chain, and the calibrated maps as its inputs. Here a parent state is upstream in depth and contains all sequence positions; it is the state tensor being estimated.

Two operations give that estimate a broader support. First, applying frequency-specific channel maps between the position Fourier transform and its inverse mixes source positions. A raw signal confined to the prefix can therefore produce a first-stage estimate with later-position entries. Second, the consistency correction uses the full child tensor:
\[
 c_v=J_v^*H_{c(v)}^{\rm on}-J_v^*J_v\widehat H_v^{(G)}.
\]
Although $J_v$ is block lower triangular in position, $J_v^*J_v$ couples parent positions that share a child position. For a two-position scalar illustration, take
\[
 J=\begin{pmatrix}1&0\\a&1\end{pmatrix},\qquad
 H^{\rm on}=\widehat H^{(G)}=\begin{pmatrix}h\\0\end{pmatrix}.
\]
Then the raw adjoint $J^*H^{\rm on}=(h,0)^{\mathsf T}$ has no second-position entry, whereas the consistency cotangent is
\[
 J^*(H^{\rm on}-J\widehat H^{(G)})
 =\begin{pmatrix}-a^2h\\-ah\end{pmatrix}.
\]
The later entry arises through the shared descendant under a causal forward operator. This example establishes the allowed support of the correction. Its empirical size is determined by the actual geometry and fitted maps.

The position projection is applied at the initial target only. Subsequent reverse edges receive repaired full-position states, so both statistical and geometric coupling can propagate through the finite reverse pass. The resulting source estimate is conditional on the complete observed instance. The fixed-prefix experiment in Supplementary Section~\ref{sup:s7-order} tests this distinction while keeping the native feature and its first raw adjoint unchanged.

\subsection{Calibration and historical comparisons}
Calibration windows were cut within individual articles, using each model's tokenizer with no added special tokens, a length and stride of 16, and incomplete tails discarded. GPT-2 used the official WikiText-2 raw training split~\citep{merity2016pointer} (revision \texttt{b08601e}): 149,297 eligible windows from 600 articles. Article text, whitespace-normalized body and normalized-title hashes were checked across the training, validation and test splits. A seed-specific permutation sampled 65,536 windows without replacement; the 131,072-window comparison extended the same permutations and retained their original 65,536-window prefixes. The original validation and test pools each contained 512 windows selected by shuffled document round-robin from 60 articles; the first 128 validation windows supplied calibration diagnostics.

For Qwen2.5/Llama3.2, 64 WikiText-2 training articles were first reserved for the natural reference pool. The remaining 536 articles were combined with 128 distinct articles from the first WikiText-103 raw training shard. Each tokenizer independently produced a 131,072-window calibration pool by shuffling nonoverlapping windows within articles and traversing articles round-robin. Reference, development and confirmation pools contained 8,192, 256 and 512 windows from 64, 60 and 64 articles, respectively; their normalized-title identities were disjoint from calibration. Development used WikiText-2 validation articles, and confirmation reused the 64 WikiText-103 articles selected for the GPT-2 natural confirmation analysis. At the full Qwen2.5/Llama3.2 budget, seeds permute the same window pool. The source package records corpus hashes, document memberships, window-selection rules and saved calibration identities in \texttt{methods\_support/calibration\_prediction\_protocol.json}.

The relative ridge coefficient is 0.01 for both stages. Within each stage and edge, its effective scale is the mean real diagonal of the Fourier input second moment, averaged over channels and stored bins and bounded below by $10^{-30}$. Sharing this scalar scale preserves independent matrix solves across frequency bins. Table~\ref{tab:supp-calibration-settings} records the calibration budgets and numerical representation.

The earlier GPT-2 configuration imposed $\WienerG_v(\omega)=\overline{\WienerG}_v$ and $D_v(\omega)=\overline D_v$ across all nine stored frequency bins~\citep{shu2026source}. For each stage and inverse edge, second moments were pooled with equal weights over the stored rFFT bins before solving for a shared full-channel matrix. This parameter tying reduced calibration degrees of freedom under the 4,096-window budget. The present configuration fits each frequency independently within the same two-stage estimator family. Table~\ref{tab:supp-calibration-comparisons} records the historical frequency constraint and reports separate numerical comparisons of inverse-chain topology and calibration budget.

\begin{table}[!htbp]
\centering\small
\caption{\textbf{Calibration configuration.} Window counts apply separately at each target depth. The Qwen2.5/Llama3.2 seeds permute the same full calibration pool.}
\label{tab:supp-calibration-settings}
\begin{tabularx}{\linewidth}{lXXX}
\toprule
 & GPT-2 & Qwen2.5 & Llama3.2\\
\midrule
Target depth / edges & 2/6/10 & 5/14/23 & 3/8/13\\
Windows per depth & 65,536 & 131,072 & 131,072\\
Additional budget & L6: 131,072 & --- & ---\\
Map seeds & 123/321/567 & 123/321/567 & 123/321/567\\
Window length / bins & 16 / 9 & 16 / 9 & 16 / 9\\
Relative ridge, $\WienerG/D$ & 0.01 / 0.01 & 0.01 / 0.01 & 0.01 / 0.01\\
Forward / maps & FP32 / complex64 & FP32 / complex64 & FP32 / complex64\\
\bottomrule
\end{tabularx}
\end{table}

\begin{table}[!htbp]
\centering\small
\caption{\textbf{Separating estimator constraints, topology and calibration budget.} The quantitative rows compare source-reference NMSE on paired materials. Each row identifies the factor changed; these distances characterize the specified source coordinates.}
\label{tab:supp-calibration-comparisons}
\begin{tabularx}{\linewidth}{>{\raggedright\arraybackslash}p{.27\linewidth}>{\raggedright\arraybackslash}X>{\raggedright\arraybackslash}X}
\toprule
Comparison & Reference & Full adjacent-edge result\\
\midrule
Frequency constraint (configuration) & Earlier GPT-2: shared full-channel maps across frequency bins; 4,096 calibration windows & Independent full-channel maps at each of the nine stored calibration frequencies\\
GPT-2 L6, 65,536 windows & Three retained edges: 0.5863 & Six adjacent edges: 0.4718\\
GPT-2 L10, 65,536 windows & Historical sparse boundaries: 0.6874 & Ten adjacent edges: 0.4563\\
Qwen2.5 L14, 131,072 & Historical sparse boundaries: 0.8961 & Fourteen adjacent edges: 0.8498\\
Llama3.2 L3, 131,072 & Historical sparse boundaries: 0.8038 & Three adjacent edges: 0.8592\\
Llama3.2 L8, 131,072 & Historical sparse boundaries: 0.9013 & Eight adjacent edges: 0.8522\\
Llama3.2 L13, 131,072 & Historical sparse boundaries: 0.9052 & Thirteen adjacent edges: 0.8551\\
GPT-2 L6, adjacent chain & 65,536 windows: 0.4718 & 131,072 windows: 0.4711\\
\bottomrule
\end{tabularx}
\end{table}

\subsection{Sampling, uncertainty and testing}
All conditions, factor flips, positions and map seeds belonging to a source group are aggregated before resampling that group. Natural-window analyses resample documents. The original comparisons use 2,000 bootstrap draws for 95\% effect-size intervals. Six GPT-2 source comparisons use 9,999 centered-bootstrap null draws: the source-response, nested-readout, local-prediction and auxiliary-depth comparisons are one-sided greater tests, while statistical-direction selectivity and shared-receiver interaction are two-sided. The three GPT-2 geometry/direction comparisons and six comparisons in each other model use 9,999 two-sided group sign flips. Holm correction is applied separately to these four families (Table~\ref{tab:supp-primary-tests}).

The later component, domain, layout and response-scale experiments retain their own source-group intervals. The late-clamp extension reuses the existing 64 groups and training-frozen injection directions, with selected nodes fixed before execution. Its intervals use 10,000 paired bootstrap draws; 50,000 two-sided sign flips support separate Holm families for 12 single-node effects, 12 matched random-direction contrasts and two final-denominator effects. Joint-node operations, early/late stratification and the common-property contrast are additional estimates. This preserves the distinction between the original frozen tests and subsequent mechanistic analyses.

\begingroup
\footnotesize
\setlength{\tabcolsep}{3pt}
\begin{longtable}{p{.33\linewidth}rrp{.23\linewidth}rr}
\caption{\textbf{Original frozen comparisons.} Effect sizes retain their original metrics and comparison order. Confidence intervals are 95\% group/document bootstrap intervals. Holm correction is within each of the four named families. $n$ counts complete source groups except the two natural-document deletion comparisons. The source-response and direction interventions use dose 0.25; statistical-direction selectivity uses a rank-16 fixed-norm direction.}\label{tab:supp-primary-tests}\\
\toprule
Comparison & $n$ & Effect & 95\% interval & $p$ & Holm $p$\\
\midrule
\endfirsthead
\multicolumn{6}{l}{\textit{Table S5 continued}}\\
\toprule
Comparison & $n$ & Effect & 95\% interval & $p$ & Holm $p$\\
\midrule
\endhead
\bottomrule
\endfoot
\multicolumn{6}{l}{\textit{GPT-2: source comparisons}}\\[2pt]
Source-response NMSE: random minus estimated direction & 256 & $-0.0192163$ & $[-0.0257325,-0.0130746]$ & $1$ & $1$\\[4pt]
Syntax nested readout: single minus combined CE & 512 & $0.0439348$ & $[0.0356704,0.0518931]$ & $0.0001$ & $0.0006$\\[4pt]
Syntax principal-direction selectivity minus random & 128 & $2.628\!\times\!10^{-5}$ & $[1.606\!\times\!10^{-5},3.582\!\times\!10^{-5}]$ & $0.0001$ & $0.0006$\\[4pt]
Syntax response prediction: global minus local NMSE & 512 & $0.0689937$ & $[0.06365,0.0743455]$ & $0.0001$ & $0.0006$\\[4pt]
Syntax shared minus split receiver RMS & 128 & $1.920\!\times\!10^{-6}$ & $[1.510\!\times\!10^{-6},2.376\!\times\!10^{-6}]$ & $0.0001$ & $0.0006$\\[4pt]
Auxiliary-depth response: random minus estimated-direction NMSE & 256 & $0.0347568$ & $[0.0294633,0.0397611]$ & $0.0001$ & $0.0006$\\[4pt]
\multicolumn{6}{l}{\textit{GPT-2: geometry and direction}}\\[2pt]
Natural-dose response: fixed-geometry minus joint NMSE & 128 & $0.223557$ & $[0.176577,0.260912]$ & $0.0001$ & $0.0003$\\[4pt]
New-layout readout: old minus adapted CE & 64 & $1.25874$ & $[1.08413,1.44514]$ & $0.0001$ & $0.0003$\\[4pt]
Direction response: own minus other cosine & 128 & $0.0482603$ & $[0.0120938,0.0842114]$ & $0.0124$ & $0.0124$\\[4pt]
\multicolumn{6}{l}{\textit{Qwen2.5}}\\[2pt]
Full-future deletion: source NMSE increase & 64 & $0.00187316$ & $[0.00178167,0.00197038]$ & $0.0001$ & $0.0006$\\[4pt]
Syntax nested readout: single minus combined CE & 499 & $0$ & $[0,0]$ & $1$ & $1$\\[4pt]
Natural-dose response: fixed-geometry minus joint NMSE & 128 & $0.0569546$ & $[0.0490629,0.0651278]$ & $0.0001$ & $0.0006$\\[4pt]
Syntax response prediction: global minus local NMSE & 499 & $0.0435905$ & $[0.0395619,0.0478223]$ & $0.0001$ & $0.0006$\\[4pt]
Direction response: own minus other cosine & 128 & $0.0564148$ & $[0.0239076,0.0900777]$ & $0.0012$ & $0.0024$\\[4pt]
Syntax shared minus separate receiver RMS & 64 & $5.119\!\times\!10^{-6}$ & $[3.155\!\times\!10^{-6},7.260\!\times\!10^{-6}]$ & $0.0001$ & $0.0006$\\[4pt]
\multicolumn{6}{l}{\textit{Llama3.2}}\\[2pt]
Full-future deletion: source NMSE increase & 64 & $0.00146978$ & $[0.00135138,0.00158291]$ & $0.0001$ & $0.0006$\\[4pt]
Syntax nested readout: single minus combined CE & 499 & $0$ & $[0,0]$ & $1$ & $1$\\[4pt]
Natural-dose response: fixed-geometry minus joint NMSE & 128 & $-0.000894457$ & $[-0.00355015,0.00168777]$ & $0.5081$ & $1$\\[4pt]
Syntax response prediction: global minus local NMSE & 499 & $-0.0395685$ & $[-0.0522253,-0.0272242]$ & $0.0001$ & $0.0006$\\[4pt]
Direction response: own minus other cosine & 128 & $0.220809$ & $[0.178699,0.266485]$ & $0.0001$ & $0.0006$\\[4pt]
Syntax shared minus separate receiver RMS & 64 & $6.191\!\times\!10^{-6}$ & $[4.346\!\times\!10^{-6},8.079\!\times\!10^{-6}]$ & $0.0001$ & $0.0006$\\[4pt]
\end{longtable}
\endgroup

\FloatBarrier
\section{Low-dimensional geometry and relational background}
\label{sup:s3-geometry}

\subsection{Depth-dependent directional organization}
\begin{figure}[!htbp]\centering
\includegraphics[width=\linewidth,height=.72\textheight,keepaspectratio]{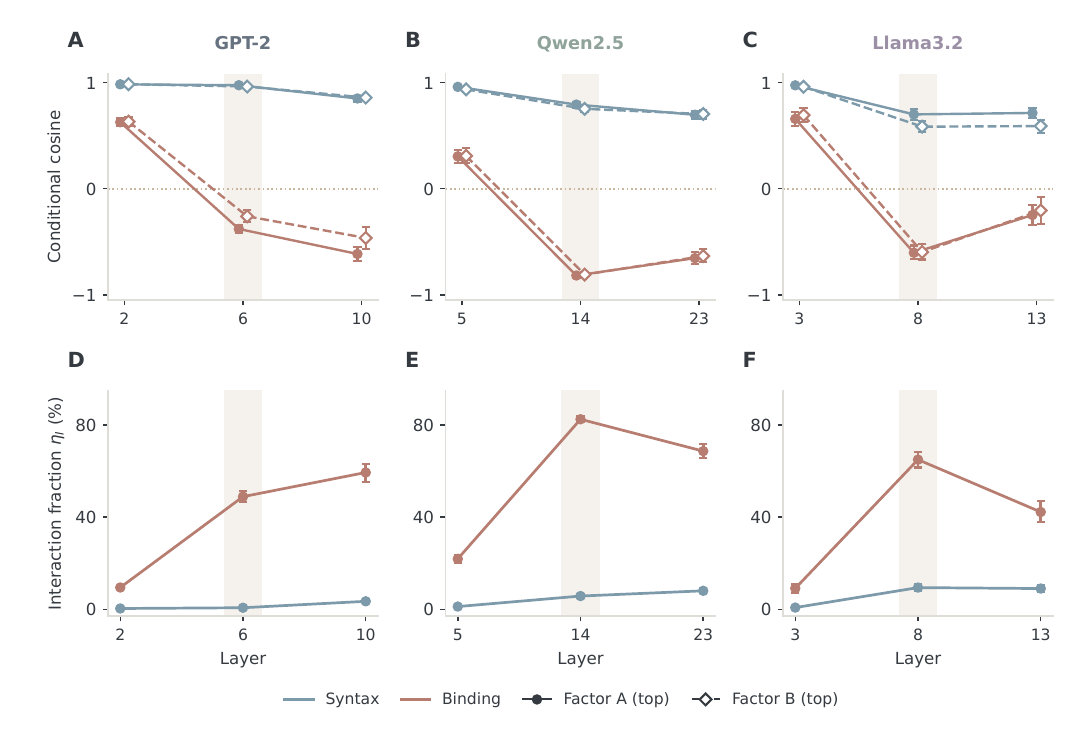}
\caption{Conditional directions and interaction across the anchored depths. A--C, syntax and binding conditional cosines share one coordinate system within each model; filled circles with solid lines identify factor $A$, and open diamonds with dashed lines identify factor $B$. D--F, interaction-energy fractions. Blue denotes syntax and terracotta denotes binding in both rows. Columns correspond to GPT-2, Qwen2.5 and Llama3.2. Lines join the three measured checkpoints; whiskers show 95\% source-group bootstrap intervals ($n=64$). Pale bands mark the middle checkpoint.}\label{fig:supp-v7-depth}
\end{figure}

The balanced four-corner design represented each source group as $h_{ab}=\mu+aA+bB+abI$. Its conditional differences are $\delta_A^{\pm}=2(A\pm I)$ and $\delta_B^{\pm}=2(B\pm I)$. We measured their contextual cosines and the interaction-energy fraction $\eta_I=\|I\|^2/(\|A\|^2+\|B\|^2+\|I\|^2)$ at the three anchored attention-output depths. The receiver in this comparison is zero-based position 14 in the 16-token four-corner materials. The dimensionless interaction ratio is
\begin{equation}
 \rho_I=\frac{\|\delta_A^+-\delta_A^-\|}{\tfrac14\sum_{X\in\{A,B\}}\sum_{s\in\{-,+\}}\|\delta_X^s\|}.
 \label{eq:supp-v7-interaction-ratio}
\end{equation}
It ranges from 0 for an additive response to 2 for an interaction-only response. All three quantities are evaluated within complete source groups and then averaged. Binding directions change from positive shallow alignment to negative middle alignment, with increasing interaction energy, whereas syntax directions remain positively aligned (Fig.~\ref{fig:supp-v7-depth}; Table~\ref{tab:supp-v7-depth}). The depth sweep comprises the three prespecified checkpoints. Reported intervals retain the source-group bootstrap estimates for their respective analyses.
\begin{table}[!htbp]
\centering\footnotesize\setlength{\tabcolsep}{3.5pt}
\caption{Anchored four-corner geometry at position $q14$. Cosines and $\rho_I$ are dimensionless; $\eta_I$ is a percentage of $A+B+I$ energy. Brackets are 95\% source-group bootstrap intervals ($n=64$).}\label{tab:supp-v7-depth}
\begin{tabular}{llrrrr}
\toprule
Model / task & Layer & Cosine $A$ & Cosine $B$ & $\eta_I$ (\%) & $\rho_I$ \\
\midrule
GPT-2 / Syntax & 2 & 0.985 [0.983, 0.987] & 0.986 [0.981, 0.989] & 0.38 [0.31, 0.46] & 0.167 [0.154, 0.180] \\
GPT-2 / Syntax & 6 & 0.976 [0.972, 0.980] & 0.964 [0.960, 0.968] & 0.71 [0.64, 0.78] & 0.235 [0.223, 0.247] \\
GPT-2 / Syntax & 10 & 0.852 [0.821, 0.878] & 0.860 [0.839, 0.879] & 3.49 [3.03, 4.02] & 0.509 [0.479, 0.542] \\
GPT-2 / Binding & 2 & 0.628 [0.589, 0.665] & 0.634 [0.589, 0.676] & 9.49 [8.68, 10.31] & 0.829 [0.794, 0.862] \\
GPT-2 / Binding & 6 & -0.379 [-0.416, -0.339] & -0.260 [-0.315, -0.200] & 48.91 [46.51, 51.27] & 1.617 [1.587, 1.643] \\
GPT-2 / Binding & 10 & -0.615 [-0.677, -0.547] & -0.464 [-0.565, -0.362] & 59.39 [55.40, 63.27] & 1.731 [1.692, 1.772] \\
Qwen2.5 / Syntax & 5 & 0.961 [0.955, 0.966] & 0.937 [0.930, 0.944] & 1.21 [1.06, 1.38] & 0.305 [0.288, 0.324] \\
Qwen2.5 / Syntax & 14 & 0.794 [0.765, 0.821] & 0.754 [0.725, 0.782] & 5.78 [5.08, 6.53] & 0.653 [0.620, 0.687] \\
Qwen2.5 / Syntax & 23 & 0.698 [0.655, 0.738] & 0.705 [0.661, 0.743] & 8.05 [6.97, 9.27] & 0.752 [0.707, 0.800] \\
Qwen2.5 / Binding & 5 & 0.305 [0.241, 0.368] & 0.310 [0.239, 0.383] & 21.93 [20.13, 23.75] & 1.215 [1.163, 1.262] \\
Qwen2.5 / Binding & 14 & -0.818 [-0.844, -0.790] & -0.809 [-0.828, -0.790] & 82.49 [81.23, 83.70] & 1.903 [1.895, 1.911] \\
Qwen2.5 / Binding & 23 & -0.654 [-0.710, -0.596] & -0.634 [-0.693, -0.572] & 68.66 [65.50, 71.67] & 1.804 [1.778, 1.829] \\
Llama3.2 / Syntax & 3 & 0.975 [0.973, 0.978] & 0.962 [0.958, 0.966] & 0.77 [0.71, 0.84] & 0.247 [0.237, 0.257] \\
Llama3.2 / Syntax & 8 & 0.701 [0.645, 0.750] & 0.586 [0.531, 0.637] & 9.44 [7.96, 11.09] & 0.800 [0.749, 0.855] \\
Llama3.2 / Syntax & 13 & 0.715 [0.663, 0.761] & 0.591 [0.527, 0.652] & 9.03 [7.55, 10.69] & 0.790 [0.738, 0.847] \\
Llama3.2 / Binding & 3 & 0.659 [0.594, 0.725] & 0.694 [0.631, 0.762] & 9.13 [7.16, 11.15] & 0.719 [0.634, 0.800] \\
Llama3.2 / Binding & 8 & -0.603 [-0.666, -0.535] & -0.597 [-0.667, -0.519] & 64.97 [61.55, 68.14] & 1.772 [1.741, 1.800] \\
Llama3.2 / Binding & 13 & -0.248 [-0.341, -0.154] & -0.205 [-0.331, -0.075] & 42.25 [37.74, 46.95] & 1.510 [1.445, 1.570] \\
\bottomrule
\end{tabular}
\end{table}

% Keep the depth table intact and allow the next evidence block to fill its page.
\FloatBarrier
\subsection{Algebraic mixing and independent natural references}
Let $M_X=\mathbb E_g[X_gX_g^{\mathsf T}]$ be an uncentered second moment across source groups. Equal weighting of both conditioning values cancels the cross terms and gives $M_{\delta_A}=4(M_A+M_I)$ and $M_{\delta_B}=4(M_B+M_I)$. Thus conditional-difference sharing includes a common $I$ contribution. The interaction accounted for 77.46\%, 96.17\% and 92.23\% of the energy captured in the opposite factor's rank-16 conditional basis in GPT-2, Qwen2.5 and Llama3.2.

We distinguished three reference constructions. Natural hidden-state PCA used independent natural-text activations at the same layer and receiver; its coverage is projected paired conditional-difference energy divided by total paired energy. Basis overlap is $\|U_1^{\mathsf T}U_2\|_F^2/16$, with a separate denominator. For the pure-main-effect comparison, we removed $I$, centered $A$ and $B$ on their respective training means, and generated rank-16 Gaussian sketches from an independent natural-activation covariance without using task-factor vectors. The intervals of the task-minus-natural coverage contrast quantify variation across confirmation source groups conditional on the fitted bases. The pooled task-difference sketch is a third construction: it uses the pooled $A/B$ conditional differences, with coverage 60.37\% and 56.49\% in Qwen2.5 and Llama3.2. Its task-conditioned covariance differs from both natural references.

\begin{figure}[!htbp]\centering
\includegraphics[width=\linewidth,height=.72\textheight,keepaspectratio]{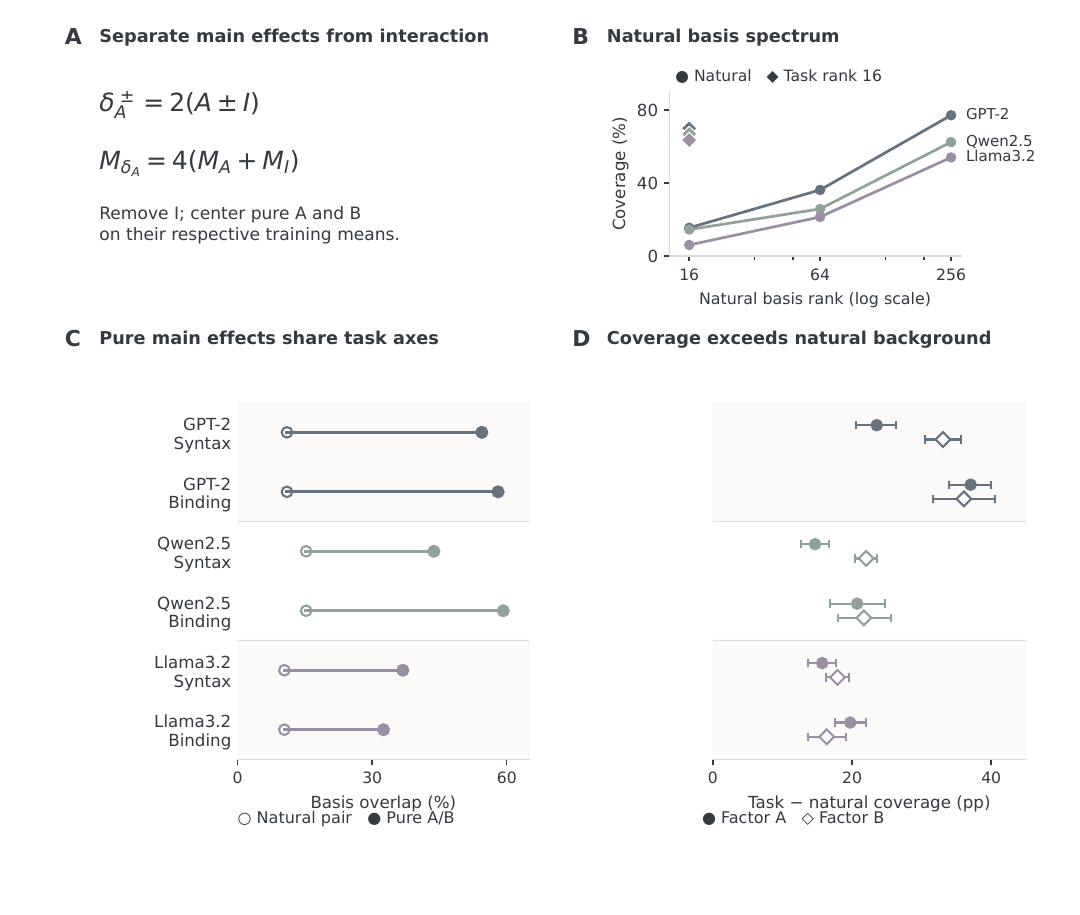}
\caption{Distinct controls for low-dimensional factor sharing. A, the balanced second-moment identity separates main effects from interaction mixing. B, paired conditional-$A$ energy coverage in natural-hidden-state bases at ranks 16, 64 and 256 on a logarithmic rank axis; diamonds mark task rank-16 coverage. C, centered pure-main-effect basis overlap versus the mean overlap of independent natural-covariance sketch pairs. D, paired confirmation coverage excess in the opposite-factor task basis over the natural-random basis for factors $A$ and $B$. Intervals in D resample 64 confirmation source groups. Coverage, overlap and excess are plotted on their respective axes.}\label{fig:supp-v7-natural}
\end{figure}

The same-space rank comparison, pure-main-effect overlaps and factor-wise confirmation coverages appear together in Fig.~\ref{fig:supp-v7-natural} and Table~\ref{tab:supp-v7-natural}. Every model-by-task panel retained positive cross-factor excess coverage over the specified independent natural reference.
\begin{table}[!htbp]
\centering\footnotesize\setlength{\tabcolsep}{3.5pt}
\caption{Natural-reference comparisons. Block A uses paired conditional-$A$ binding energy and natural-hidden PCA. Block B uses centered pure main effects and natural-covariance sketch pairs. Block C gives confirmation coverage of each factor in the opposite-factor basis and in natural-random bases, with paired excess. Coverage and overlap are percentages; excess is in percentage points. Block C intervals use 2,000 bootstrap resamples of 64 confirmation source groups.}\label{tab:supp-v7-natural}
\begin{tabular}{llrrrr}
\toprule
Model & Task / factor & Task / opposite & Natural & Rank 256 / excess & Basis overlap \\
\midrule

\multicolumn{6}{l}{\emph{A. Conditional-$A$ energy: task rank 16; natural ranks 16, 64, 256; task--natural rank-16 overlap}}\\
Model & Task r16 & Natural r16 & Natural r64 & Natural r256 & Overlap \\
GPT-2 & 69.90 & 15.51 & 36.23 & 77.03 & 16.31 \\
Qwen2.5 & 66.80 & 14.56 & 25.84 & 62.36 & 8.76 \\
Llama3.2 & 63.47 & 6.14 & 21.44 & 53.98 & 7.02 \\
\addlinespace\multicolumn{6}{l}{\emph{B. Centered pure-main-effect and independent natural-pair overlaps}}\\
Model & Task & Pure $A/B$ & Natural pair &  &  \\
GPT-2 & Syntax & 54.44 & 11.01 &  &  \\
GPT-2 & Binding & 58.06 & 11.01 &  &  \\
Qwen2.5 & Syntax & 43.75 & 15.28 &  &  \\
Qwen2.5 & Binding & 59.22 & 15.28 &  &  \\
Llama3.2 & Syntax & 36.84 & 10.43 &  &  \\
Llama3.2 & Binding & 32.53 & 10.43 &  &  \\
\addlinespace\multicolumn{6}{l}{\emph{C. Factor-wise confirmation coverage and its paired excess}}\\
Model & Task / factor & Opposite factor & Natural random & Excess (pp) &  \\
GPT-2 & Syntax / $A$ & 32.27 [29.15, 35.28] & 8.71 [8.16, 9.31] & 23.56 [20.57, 26.38] &  \\
GPT-2 & Syntax / $B$ & 42.15 [39.62, 44.84] & 9.07 [8.43, 9.86] & 33.07 [30.43, 35.72] &  \\
GPT-2 & Binding / $A$ & 46.29 [43.08, 49.43] & 9.25 [8.40, 10.18] & 37.04 [33.97, 40.02] &  \\
GPT-2 & Binding / $B$ & 44.76 [40.30, 49.04] & 8.69 [7.80, 9.70] & 36.08 [31.64, 40.59] &  \\
Qwen2.5 & Syntax / $A$ & 23.38 [21.00, 25.78] & 8.68 [8.06, 9.37] & 14.70 [12.64, 16.67] &  \\
Qwen2.5 & Syntax / $B$ & 31.85 [30.07, 33.65] & 9.81 [9.40, 10.25] & 22.03 [20.43, 23.64] &  \\
Qwen2.5 & Binding / $A$ & 26.17 [22.44, 29.99] & 5.43 [5.04, 5.86] & 20.74 [16.88, 24.71] &  \\
Qwen2.5 & Binding / $B$ & 26.93 [23.07, 30.90] & 5.23 [4.90, 5.57] & 21.70 [18.04, 25.58] &  \\
Llama3.2 & Syntax / $A$ & 21.81 [19.72, 23.92] & 6.09 [5.82, 6.35] & 15.72 [13.67, 17.75] &  \\
Llama3.2 & Syntax / $B$ & 23.72 [22.01, 25.47] & 5.80 [5.53, 6.08] & 17.92 [16.32, 19.57] &  \\
Llama3.2 & Binding / $A$ & 24.08 [21.82, 26.45] & 4.33 [4.15, 4.50] & 19.75 [17.54, 22.03] &  \\
Llama3.2 & Binding / $B$ & 21.08 [18.33, 23.88] & 4.71 [4.51, 4.92] & 16.37 [13.66, 19.09] &  \\
\bottomrule
\end{tabular}
\end{table}

\FloatBarrier
\subsection{Query-role decomposition and training-frozen transfer}
For the separate 15-token eight-condition materials, $a,b,r\in\{-1,+1\}$ vary independently and
\begin{equation}
 h_{abr}=\mu+aA+bB+rR+abI_{AB}+arI_{AR}+brI_{BR}+abrI_{ABR}.
 \label{eq:supp-eight-condition}
\end{equation}
The query and prediction receivers are at zero-based positions 13 and 14. At a fixed role, $I^{(\pm)}=I_{AB}\pm I_{ABR}$; the role-independent fraction is $\|I_{AB}\|^2/(\|I_{AB}\|^2+\|I_{ABR}\|^2)$. The 128 training, 32 development and 64 test groups use disjoint names with shared colors and templates. Training-frozen rank-1 and rank-16 coverage is the fraction of each test group's $AB$ energy captured by that basis, averaged across test groups.

\begin{figure}[!htbp]\centering
\includegraphics[width=\linewidth,height=.80\textheight,keepaspectratio]{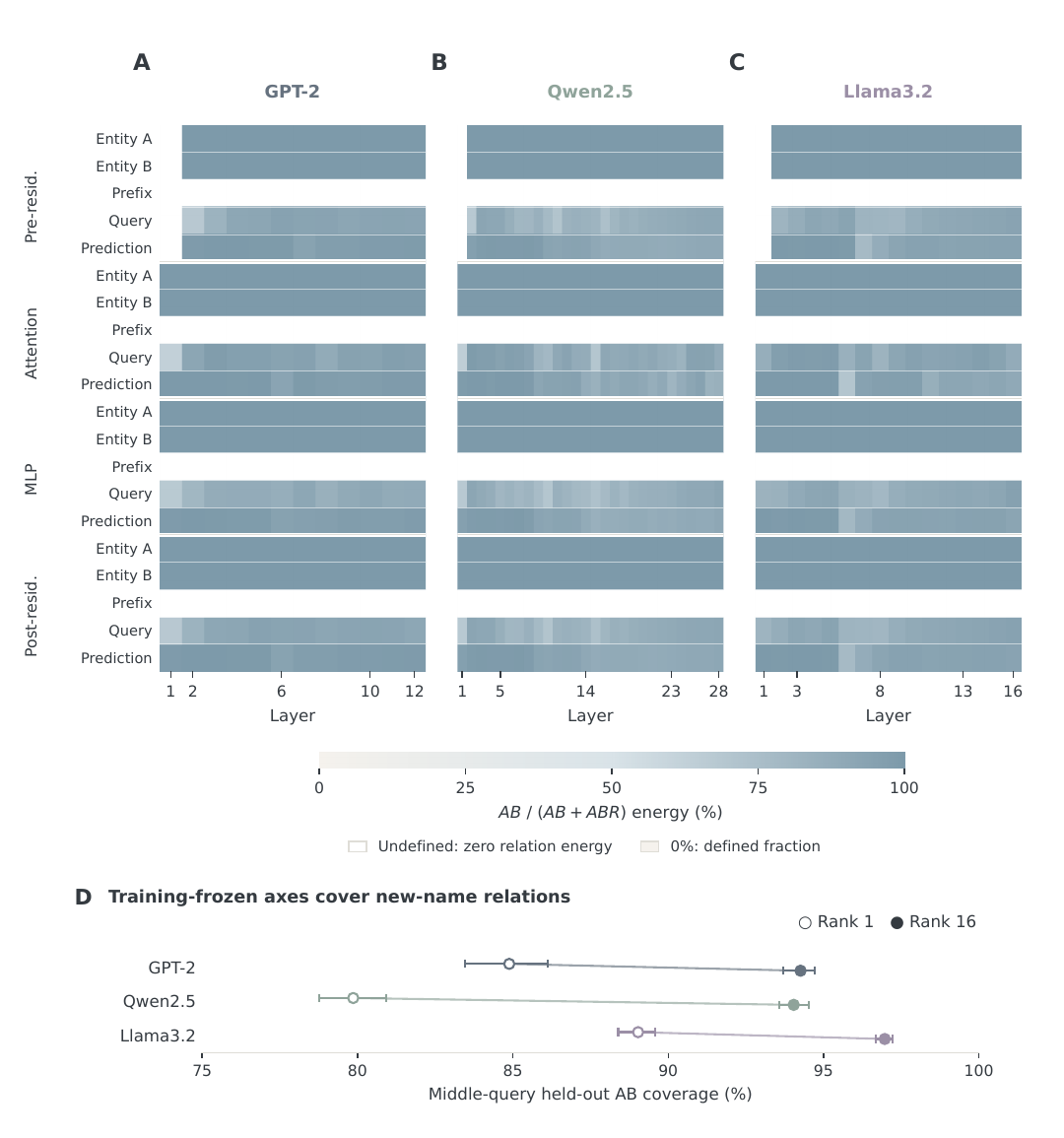}
\caption{Role-independent interaction across layers, nodes and receivers. A--C, $AB/(AB+ABR)$ energy fraction at residual inputs, attention writes, MLP writes and residual outputs for the entity-A, entity-B, prefix, query and prediction positions in each model. All models use the same linear 0--100\% color scale. White cells indicate undefined zero-relation-energy fractions; warm ivory is the zero point for a defined fraction. D, training-frozen rank-1 and rank-16 coverage of held-out-name $AB$ energy at the middle query receiver, with 95\% source-group bootstrap intervals ($n=64$). The associated source-data file retains all seven factor-energy shares for the complete layer/node/role grid.}\label{fig:supp-v7-role}
\end{figure}

Figure~\ref{fig:supp-v7-role} spans every layer, four node types and five receiving roles. At positions preceding the query, the role-dependent coefficients are exactly zero under the causal mask; cells whose relation energy is zero are left blank. Table~\ref{tab:supp-v7-role} gives all seven non-mean component shares at the anchored attention outputs and the separate middle-query relation and transfer metrics. Its seven-component denominator resolves the large $R$ contribution separately from the $AB+ABR$ denominator used to assess role independence.
\begingroup\footnotesize\setlength{\tabcolsep}{3.5pt}
\begin{longtable}{llrrrrrrr}
\caption{Eight-condition energies and training-frozen transfer. Block A lists all seven non-mean energy shares at the anchored attention outputs; each row is normalized within group before averaging over 64 test groups. Block B uses the separate relation-energy denominator and test-group $AB$ energy. Brackets denote 95\% source-group intervals.}\label{tab:supp-v7-role}\\
\toprule
Model & Layer / role & $A$ & $B$ & $R$ & $AB$ & $AR$ & $BR$ & $ABR$ \\
\midrule\endfirsthead
\multicolumn{9}{l}{\textit{Continued}}\\\toprule
Model & Layer / role & $A$ & $B$ & $R$ & $AB$ & $AR$ & $BR$ & $ABR$ \\
\midrule\endhead
\bottomrule\endfoot
\multicolumn{9}{l}{\emph{A. Seven-factor energy shares (\%)}}\\
GPT-2 & L2 / query & 3.71 & 3.39 & 91.12 & 0.50 & 0.63 & 0.59 & 0.05 \\
GPT-2 & L2 / pred & 26.46 & 26.50 & 44.14 & 2.72 & 0.10 & 0.08 & 0.00 \\
GPT-2 & L6 / query & 0.27 & 0.28 & 98.79 & 0.40 & 0.11 & 0.12 & 0.03 \\
GPT-2 & L6 / pred & 16.68 & 15.92 & 19.83 & 40.15 & 2.06 & 2.19 & 3.17 \\
GPT-2 & L10 / query & 4.70 & 4.99 & 68.97 & 18.99 & 0.57 & 0.57 & 1.21 \\
GPT-2 & L10 / pred & 38.62 & 37.70 & 3.71 & 18.83 & 0.28 & 0.39 & 0.46 \\
Qwen2.5 & L5 / query & 3.10 & 3.16 & 91.54 & 1.59 & 0.25 & 0.27 & 0.08 \\
Qwen2.5 & L5 / pred & 32.49 & 32.86 & 28.51 & 5.79 & 0.16 & 0.15 & 0.04 \\
Qwen2.5 & L14 / query & 5.41 & 5.24 & 54.39 & 29.31 & 0.72 & 0.68 & 4.25 \\
Qwen2.5 & L14 / pred & 1.48 & 1.49 & 89.77 & 5.88 & 0.28 & 0.26 & 0.85 \\
Qwen2.5 & L23 / query & 7.08 & 7.01 & 41.45 & 36.67 & 1.08 & 1.04 & 5.67 \\
Qwen2.5 & L23 / pred & 35.99 & 33.78 & 11.42 & 9.45 & 3.77 & 3.45 & 2.14 \\
Llama3.2 & L3 / query & 0.96 & 0.95 & 97.66 & 0.22 & 0.10 & 0.11 & 0.00 \\
Llama3.2 & L3 / pred & 26.20 & 26.84 & 30.03 & 16.50 & 0.17 & 0.19 & 0.07 \\
Llama3.2 & L8 / query & 1.83 & 1.89 & 80.02 & 12.30 & 0.66 & 0.68 & 2.62 \\
Llama3.2 & L8 / pred & 3.56 & 3.62 & 46.31 & 44.16 & 0.29 & 0.29 & 1.78 \\
Llama3.2 & L13 / query & 5.13 & 5.55 & 73.95 & 13.80 & 0.37 & 0.35 & 0.85 \\
Llama3.2 & L13 / pred & 30.87 & 31.71 & 12.66 & 14.83 & 4.35 & 4.27 & 1.30 \\
\addlinespace\multicolumn{9}{l}{\emph{B. Middle-query relation fraction, role-swap cosine and held-out-name coverage}}\\
Model & \multicolumn{2}{c}{$AB/(AB+ABR)$ (\%)} & \multicolumn{2}{c}{Role-swap cosine} & \multicolumn{2}{c}{Rank 1 (\%)} & \multicolumn{2}{c}{Rank 16 (\%)}\\
GPT-2 & \multicolumn{2}{c}{93.64 [93.07, 94.19]} & \multicolumn{2}{c}{0.880 [0.869, 0.891]} & \multicolumn{2}{c}{84.88 [83.47, 86.12]} & \multicolumn{2}{c}{94.26 [93.71, 94.73]} \\
Qwen2.5 & \multicolumn{2}{c}{87.27 [86.65, 87.89]} & \multicolumn{2}{c}{0.769 [0.759, 0.779]} & \multicolumn{2}{c}{79.86 [78.78, 80.93]} & \multicolumn{2}{c}{94.04 [93.57, 94.52]} \\
Llama3.2 & \multicolumn{2}{c}{82.16 [81.43, 82.90]} & \multicolumn{2}{c}{0.653 [0.639, 0.667]} & \multicolumn{2}{c}{89.03 [88.39, 89.58]} & \multicolumn{2}{c}{96.97 [96.69, 97.22]} \\
\end{longtable}
\endgroup

\FloatBarrier
\subsection{Domain, word mapping and group-supported relation transfer}
The full 64-condition color--shape design crossed four binary fact properties with queried entity and queried domain. For each source group, the analysis averaged symmetric query-role/domain slices and factor pairs, then compared queried same-domain, unqueried same-domain and cross-entity cross-domain interactions. Direct attribute answers, domain-specific codes and a shared binary answer code were all evaluated at the middle query state (32 test source groups). These states include the processed query.

\begin{figure}[!htbp]\centering
\includegraphics[width=\linewidth,height=.77\textheight,keepaspectratio]{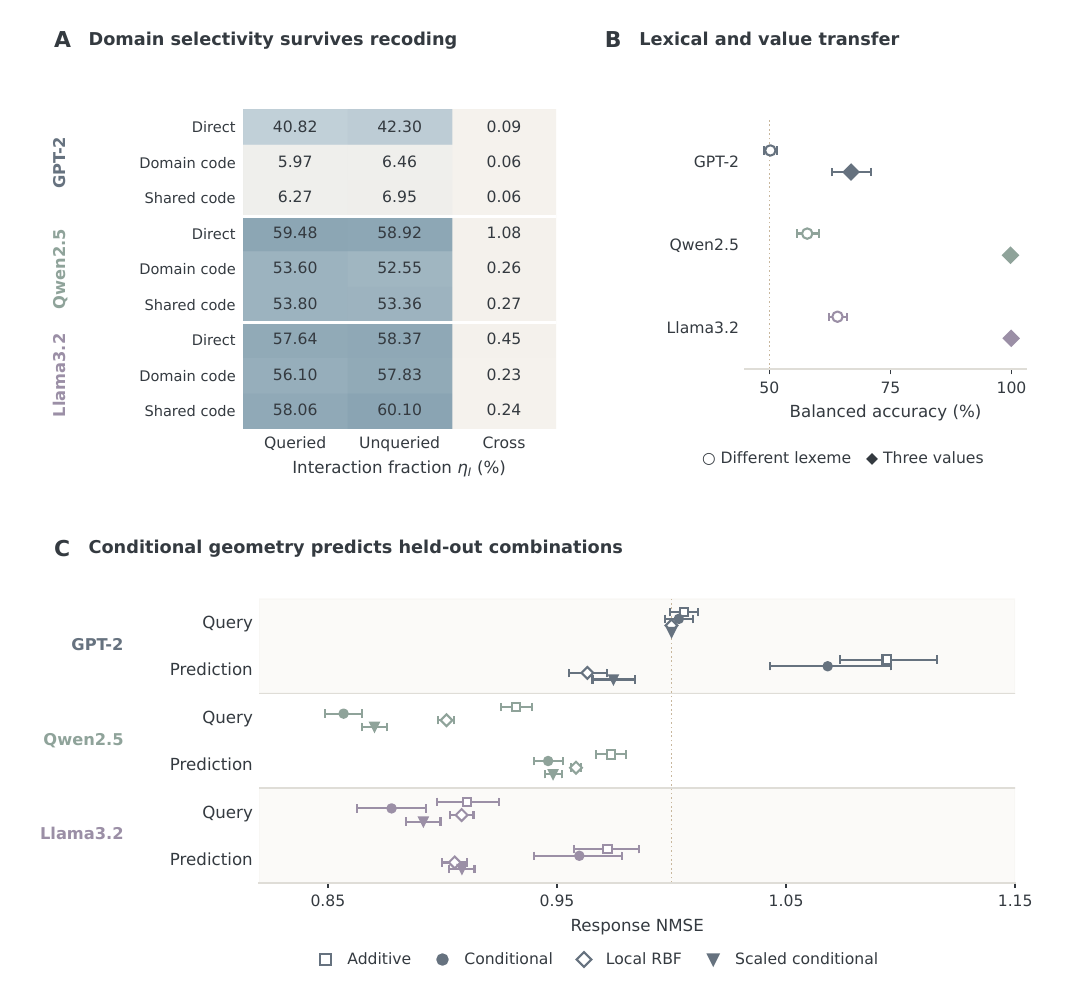}
\caption{Domain selectivity, lexical/value transfer and fixed-basis prediction. A, queried and unqueried same-domain and cross-domain interaction under direct answers, domain-specific answer codes and a shared binary code. B, complete-group-centered classification with the original frozen binary relation axis in length-matched different-lexeme and three-value materials. C, query and prediction receiver NMSE for held-out combinations, comparing additive, conditional-linear, local-RBF and development-scaled conditional models. Whiskers are the saved 95\% source-group bootstrap intervals. The 50\% and NMSE-1 reference lines correspond to chance classification and zero-response prediction, respectively.}\label{fig:supp-v7-generalization}
\end{figure}

The length-matched lexical controls compared repeated and different surface forms under explicit category mappings. Relation classification used the frozen original binary interaction axis $I$, oriented on training data. Each complete counterfactual group was centered by its unlabeled group mean before projection, and same/different outcomes were evaluated using balanced accuracy. The three-value test used the complete $3\times3\times R$ design with the same centering procedure. These are group-supported relation-axis diagnostics: the reference mean is computed from the full condition group. They complement the separate eight-condition $AB$ transfer test across new names. Both lexical and three-value tests used 64 source groups; reported intervals used 2,000 source-group bootstrap resamples (Fig.~\ref{fig:supp-v7-generalization}A--B; Table~\ref{tab:supp-v7-generalization}).
\begin{table}[!htbp]
\centering\footnotesize\setlength{\tabcolsep}{3.5pt}
\caption{Middle-query domain, word-mapping and value controls. Energy entries are interaction fractions (\%), cosine entries are dimensionless, and accuracy entries are balanced percentages after complete-group unlabeled mean centering. Brackets are 95\% source-group intervals. Domain controls use 32 source groups; lexical and three-value controls use 64.}\label{tab:supp-v7-generalization}
\begin{tabular}{lrrr}
\toprule
Condition / measurement & GPT-2 & Qwen2.5 & Llama3.2 \\
\midrule

\multicolumn{4}{l}{\emph{Direct attribute answers}}\\
Queried same-domain $\eta_I$ & 40.82 [39.49, 42.24] & 59.48 [58.57, 60.35] & 57.64 [57.05, 58.22] \\
Unqueried same-domain $\eta_I$ & 42.30 [41.08, 43.51] & 58.92 [57.92, 59.86] & 58.37 [57.61, 59.10] \\
Cross-domain $\eta_I$ & 0.090 [0.085, 0.095] & 1.076 [1.050, 1.104] & 0.449 [0.433, 0.468] \\
\multicolumn{4}{l}{\emph{Domain-specific answer codes}}\\
Queried same-domain $\eta_I$ & 5.97 [5.68, 6.26] & 53.60 [52.41, 54.80] & 56.10 [54.56, 57.60] \\
Unqueried same-domain $\eta_I$ & 6.46 [6.19, 6.74] & 52.55 [51.18, 54.01] & 57.83 [56.28, 59.31] \\
Cross-domain $\eta_I$ & 0.056 [0.051, 0.062] & 0.261 [0.248, 0.275] & 0.234 [0.225, 0.244] \\
\multicolumn{4}{l}{\emph{Shared binary answer code}}\\
Queried same-domain $\eta_I$ & 6.27 [5.99, 6.58] & 53.80 [52.63, 55.00] & 58.06 [56.74, 59.32] \\
Unqueried same-domain $\eta_I$ & 6.95 [6.67, 7.24] & 53.36 [51.97, 54.87] & 60.10 [58.76, 61.39] \\
Cross-domain $\eta_I$ & 0.058 [0.052, 0.064] & 0.268 [0.254, 0.281] & 0.239 [0.229, 0.248] \\
\addlinespace\multicolumn{4}{l}{\emph{Length-matched mappings and relation-axis transfer}}\\
Repeated-lexeme $\eta_I$ & 5.45 [5.01, 5.87] & 47.99 [47.19, 48.76] & 56.16 [55.41, 56.87] \\
Different-lexeme $\eta_I$ & 0.45 [0.43, 0.47] & 6.00 [5.81, 6.19] & 7.94 [7.58, 8.32] \\
Different-lexeme axis cosine & -0.078 [-0.091, -0.066] & 0.152 [0.143, 0.160] & 0.350 [0.342, 0.357] \\
Different-lexeme accuracy & 50.20 [48.83, 51.56] & 57.81 [55.66, 60.16] & 64.06 [62.30, 66.02] \\
Three-value accuracy & 66.86 [62.89, 71.03] & 99.80 [99.48, 100.00] & 99.93 [99.80, 100.00] \\
\bottomrule
\end{tabular}
\end{table}

\FloatBarrier
\subsection{Conditional prediction with held-out combinations}
The response-prediction task withheld combinations $(0,2)$, $(1,0)$ and $(2,1)$ and the test vocabulary. All predictors received the same condition features and the same within-group $(0,0,R0)$ reference state. Predictions were displacements from that state in a common rank-16 output basis fitted using only permitted training combinations. The additive, conditional-linear and local-RBF models each had nine development candidates. Conditional and RBF models used the same condition inputs; comparison therefore tested their prediction rules within the specified output coordinates.

An additional scalar-amplitude diagnostic multiplied each conditional-linear prediction by one gain fitted only on development groups. It preserved the original frozen model comparisons. At the prediction receiver, the development gain reduced GPT-2 and Llama3.2 conditional errors, while scaled-conditional minus RBF NMSE remained positive: $+0.011389$ [0.009903, 0.012918] and $+0.003153$ [0.002166, 0.004114]. Table~\ref{tab:supp-v7-prediction} reports all receiver/model estimates, gains and paired prediction contrasts. The zero-response reference is 1 in each case.
\begingroup\footnotesize\setlength{\tabcolsep}{3.5pt}\setlength{\LTpre}{6pt}
\begin{longtable}{llll}
\caption{Response prediction within the common rank-16 basis. NMSE entries and paired contrasts have 95\% source-group bootstrap intervals ($n=64$). A development gain is fitted once per model and receiver; it is an amplitude diagnostic rather than a new output basis.}\label{tab:supp-v7-prediction}\\
\toprule
Model / receiver & Predictor / comparison & NMSE / paired difference & Dev. gain \\
\midrule\endfirsthead
\multicolumn{4}{l}{\textit{Continued}}\\\toprule
Model / receiver & Predictor / comparison & NMSE / paired difference & Dev. gain \\
\midrule\endhead
\bottomrule\endfoot
GPT-2 / query & Additive & 1.00546 [0.99923, 1.01170] & -- \\
GPT-2 / query & Conditional linear & 1.00318 [0.99718, 1.00919] & -- \\
GPT-2 / query & Local RBF & 0.99999 [0.99998, 0.99999] & -- \\
GPT-2 / query & Scaled conditional & 1.00000 [1.00000, 1.00000] & 0.0000 \\
GPT-2 / query & Zero response & 1.00000 [1.00000, 1.00000] & -- \\
GPT-2 / query & Scaled $-$ RBF & 0.00001 [0.00001, 0.00002] & -- \\
GPT-2 / query & Conditional $-$ RBF & 0.00319 [-0.00280, 0.00920] & -- \\
\addlinespace
GPT-2 / pred & Additive & 1.09389 [1.07335, 1.11569] & -- \\
GPT-2 / pred & Conditional linear & 1.06815 [1.04307, 1.09566] & -- \\
GPT-2 / pred & Local RBF & 0.96323 [0.95511, 0.97178] & -- \\
GPT-2 / pred & Scaled conditional & 0.97462 [0.96549, 0.98414] & 0.4043 \\
GPT-2 / pred & Zero response & 1.00000 [1.00000, 1.00000] & -- \\
GPT-2 / pred & Scaled $-$ RBF & 0.01139 [0.00990, 0.01292] & -- \\
GPT-2 / pred & Conditional $-$ RBF & 0.10492 [0.08618, 0.12509] & -- \\
\addlinespace
Qwen2.5 / query & Additive & 0.93218 [0.92568, 0.93926] & -- \\
Qwen2.5 / query & Conditional linear & 0.85685 [0.84891, 0.86498] & -- \\
Qwen2.5 / query & Local RBF & 0.90176 [0.89815, 0.90504] & -- \\
Qwen2.5 / query & Scaled conditional & 0.87033 [0.86485, 0.87562] & 0.6244 \\
Qwen2.5 / query & Zero response & 1.00000 [1.00000, 1.00000] & -- \\
Qwen2.5 / query & Scaled $-$ RBF & -0.03142 [-0.03368, -0.02904] & -- \\
Qwen2.5 / query & Conditional $-$ RBF & -0.04491 [-0.05028, -0.03901] & -- \\
\addlinespace
Qwen2.5 / pred & Additive & 0.97359 [0.96713, 0.98025] & -- \\
Qwen2.5 / pred & Conditional linear & 0.94612 [0.93988, 0.95244] & -- \\
Qwen2.5 / pred & Local RBF & 0.95828 [0.95621, 0.96042] & -- \\
Qwen2.5 / pred & Scaled conditional & 0.94826 [0.94469, 0.95206] & 0.5656 \\
Qwen2.5 / pred & Zero response & 1.00000 [1.00000, 1.00000] & -- \\
Qwen2.5 / pred & Scaled $-$ RBF & -0.01003 [-0.01174, -0.00825] & -- \\
Qwen2.5 / pred & Conditional $-$ RBF & -0.01217 [-0.01655, -0.00758] & -- \\
\addlinespace
Llama3.2 / query & Additive & 0.91081 [0.89767, 0.92457] & -- \\
Llama3.2 / query & Conditional linear & 0.87779 [0.86278, 0.89284] & -- \\
Llama3.2 / query & Local RBF & 0.90837 [0.90314, 0.91356] & -- \\
Llama3.2 / query & Scaled conditional & 0.89165 [0.88396, 0.89912] & 0.4795 \\
Llama3.2 / query & Zero response & 1.00000 [1.00000, 1.00000] & -- \\
Llama3.2 / query & Scaled $-$ RBF & -0.01672 [-0.01919, -0.01414] & -- \\
Llama3.2 / query & Conditional $-$ RBF & -0.03058 [-0.04083, -0.01992] & -- \\
\addlinespace
Llama3.2 / pred & Additive & 0.97205 [0.95742, 0.98563] & -- \\
Llama3.2 / pred & Conditional linear & 0.95974 [0.93996, 0.97823] & -- \\
Llama3.2 / pred & Local RBF & 0.90535 [0.89991, 0.91067] & -- \\
Llama3.2 / pred & Scaled conditional & 0.90850 [0.90286, 0.91398] & 0.3731 \\
Llama3.2 / pred & Zero response & 1.00000 [1.00000, 1.00000] & -- \\
Llama3.2 / pred & Scaled $-$ RBF & 0.00315 [0.00217, 0.00411] & -- \\
Llama3.2 / pred & Conditional $-$ RBF & 0.05439 [0.03899, 0.06882] & -- \\
\addlinespace
\end{longtable}
\endgroup
\FloatBarrier

\FloatBarrier
\section{Conditional signed transfer and receiver roles}
\label{sup:s4-transfer}
This section connects receiver energy and orientation, cross-group donor recovery, conditional directional gain, and residual transmission under clean-baseline path controls.

\subsection{Receiver energy and signed output}
In the 16-token four-corner binding materials, deep query receivers retained interaction fractions of $59.39\%$, $68.66\%$ and $42.25\%$ in GPT-2, Qwen2.5 and Llama3.2, whereas the adjacent prediction receivers retained $18.50\%$, $15.03\%$ and $15.69\%$. The prediction receiver's own middle-to-deep decreases were $39.08$ $[36.59,41.52]$, $61.51$ $[59.65,63.42]$ and $64.71$ $[61.14,67.89]$ percentage points. Absolute interaction energy increased in GPT-2 and Qwen2.5, while stronger main-effect growth reduced its relative share.

The separate 15-token eight-condition design defines total attribute energy $T=\|A\|^2+\|B\|^2$ and fixed-entity asymmetry $S_{\mathrm{entity}}=(\|A\|^2-\|B\|^2)/T$. For role alignment, $A_r=A+rI_{AR}$ and $B_r=B+rI_{BR}$ supply target and distractor contributions according to the queried entity. Their squared-energy difference divided by their sum gives $S_{\mathrm{target}}$. Table~\ref{tab:supp-role-correlations} reports depth-specific fixed-coordinate correlations and the deep role-aligned comparison. GPT-2 retained a positive $S_{\mathrm{target}}$--$G$ correlation; Qwen2.5 and Llama3.2 had intervals spanning zero.

\begin{table}[!htbp]
\centering\footnotesize
\setlength{\tabcolsep}{4pt}
\caption{Prediction-receiver attention-output Spearman correlations with symmetric role-selection margin $G$. Each row uses 64 source groups; brackets are exploratory 95\% source-group bootstrap intervals. Role alignment is shown at the deep anchor.}
\label{tab:supp-role-correlations}
\begin{tabular}{llrrr}
\toprule
Model & Receiver & $T$ versus $G$ & $S_{\rm entity}$ versus $G$ & $S_{\rm target}$ versus $G$ \\
\midrule
GPT-2 & L2/pred & \shortstack{$-0.097$\\$[-0.347,+0.184]$} & \shortstack{$-0.250$\\$[-0.459,-0.032]$} & --- \\
GPT-2 & L6/pred & \shortstack{$-0.490$\\$[-0.680,-0.245]$} & \shortstack{$-0.219$\\$[-0.450,+0.033]$} & --- \\
GPT-2 & L10/pred & \shortstack{$+0.195$\\$[-0.053,+0.436]$} & \shortstack{$-0.227$\\$[-0.455,+0.046]$} & \shortstack{$+0.796$\\$[+0.668,+0.875]$} \\
Qwen2.5 & L5/pred & \shortstack{$+0.001$\\$[-0.271,+0.263]$} & \shortstack{$+0.022$\\$[-0.202,+0.232]$} & --- \\
Qwen2.5 & L14/pred & \shortstack{$+0.144$\\$[-0.101,+0.363]$} & \shortstack{$-0.018$\\$[-0.241,+0.225]$} & --- \\
Qwen2.5 & L23/pred & \shortstack{$-0.437$\\$[-0.630,-0.197]$} & \shortstack{$-0.057$\\$[-0.299,+0.188]$} & \shortstack{$+0.011$\\$[-0.265,+0.264]$} \\
Llama3.2 & L3/pred & \shortstack{$+0.125$\\$[-0.142,+0.364]$} & \shortstack{$-0.127$\\$[-0.370,+0.120]$} & --- \\
Llama3.2 & L8/pred & \shortstack{$-0.115$\\$[-0.329,+0.125]$} & \shortstack{$-0.074$\\$[-0.341,+0.208]$} & --- \\
Llama3.2 & L13/pred & \shortstack{$+0.126$\\$[-0.134,+0.375]$} & \shortstack{$-0.144$\\$[-0.342,+0.092]$} & \shortstack{$+0.007$\\$[-0.270,+0.281]$} \\
\bottomrule
\end{tabular}
\end{table}

\subsection{Context and absolute orientation}
The factor-direction intervention uses the 16-token four-corner materials, with equal-norm dose $0.25$. The matching-context arm injects the factor direction for the recipient context; the alternative arm uses the other factor context. A common orthogonal rotation preserves the internal angles of the injected vectors while changing their orientation relative to the downstream network. Table~\ref{tab:supp-dose-proportionality} retains the original frozen source-group intervals for both contrasts.

\begin{figure}[!htbp]
\centering
\includegraphics[width=\linewidth,height=.78\textheight,keepaspectratio]{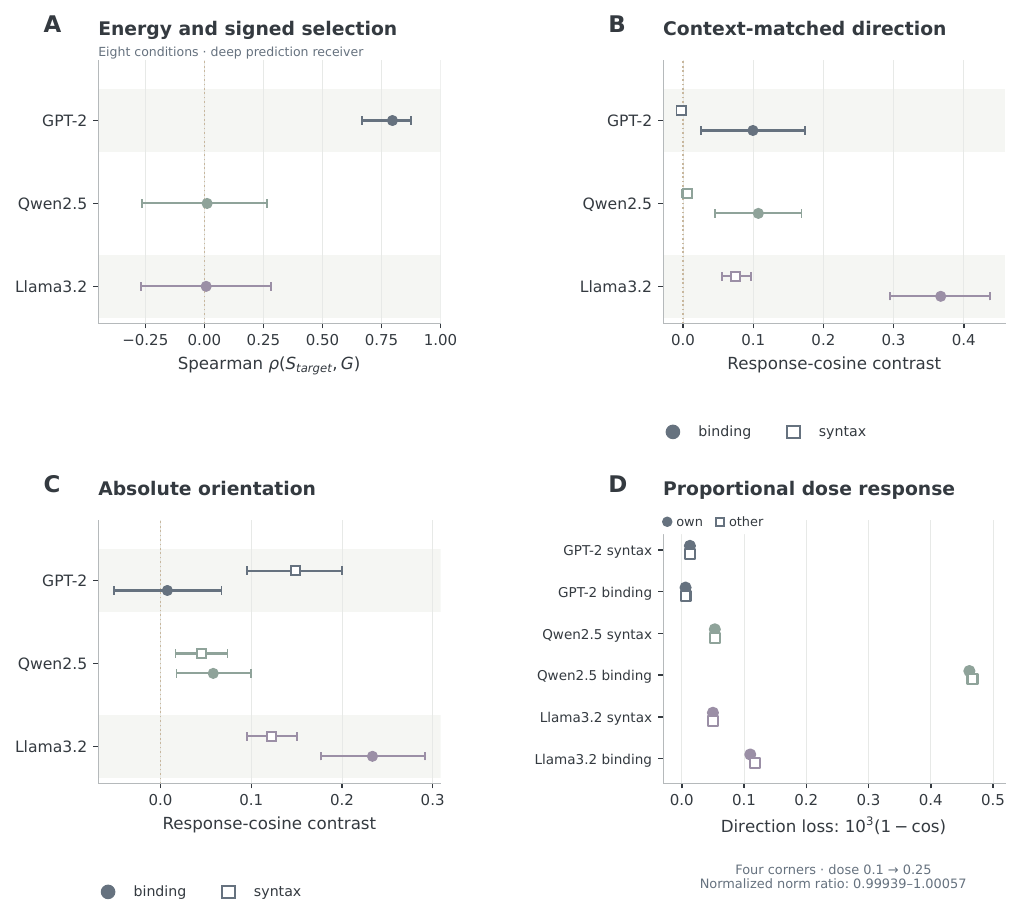}
\caption{Receiver energy and downstream reading geometry. (A) Deep prediction-receiver role-aligned energy correlation with $G$ in the 15-token eight-condition design. (B,C) Matching-context and rotation contrasts for binding (filled circles) and syntax (open squares). (D) Dose-scaling direction loss for matching (circles) and alternative (squares) contexts. Intervals in A--C resample source groups; D shows reported means. Panels B--D use context-conditioned factor directions in the 16-token four-corner materials; dose norm ratios are normalized by the expected factor of 2.5.}
\label{fig:supp-v7-reading}
\end{figure}

For each group and direction arm we also compare $\delta y_{0.1}$ with $\delta y_{0.25}$, their norm ratio $\|\delta y_{0.25}\|/(2.5\|\delta y_{0.1}\|)$, and the relative error $\|\delta y_{0.25}-2.5\delta y_{0.1}\|/\|\delta y_{0.25}\|$. All six model--task pairs retain response cosine $\geq0.99953$ for both context arms. These measurements characterize the local dose range for factor directions; the relational projection edit below has its own derivative and finite-difference checks (Fig.~\ref{fig:supp-v7-reading}).

\begin{table}[!htbp]
\centering\footnotesize
\setlength{\tabcolsep}{4pt}
\caption{Context-conditioned factor directions: functional contrasts and dose proportionality. The top block uses response-cosine differences; the bottom block records 64-group means in their stated units.}
\label{tab:supp-dose-proportionality}
\begin{tabular}{lrrr}
\toprule
Model/task & Own $-$ other & Own $-$ rotated & $n$ \\
\midrule
\addlinespace
\multicolumn{4}{l}{\emph{Context contrast at dose 0.25: estimates [95\% CI]}}\\
GPT-2/syntax & \shortstack{$-0.0029$\\$[-0.0064,+0.0005]$} & \shortstack{$+0.1484$\\$[+0.0957,+0.2003]$} & 64 \\
GPT-2/binding & \shortstack{$+0.0994$\\$[+0.0254,+0.1735]$} & \shortstack{$+0.0075$\\$[-0.0515,+0.0672]$} & 64 \\
Qwen2.5/syntax & \shortstack{$+0.0057$\\$[-0.0023,+0.0134]$} & \shortstack{$+0.0451$\\$[+0.0165,+0.0738]$} & 64 \\
Qwen2.5/binding & \shortstack{$+0.1071$\\$[+0.0450,+0.1687]$} & \shortstack{$+0.0581$\\$[+0.0176,+0.1000]$} & 64 \\
Llama3.2/syntax & \shortstack{$+0.0745$\\$[+0.0550,+0.0963]$} & \shortstack{$+0.1224$\\$[+0.0955,+0.1502]$} & 64 \\
Llama3.2/binding & \shortstack{$+0.3671$\\$[+0.2950,+0.4372]$} & \shortstack{$+0.2336$\\$[+0.1767,+0.2915]$} & 64 \\
\addlinespace
\multicolumn{4}{l}{\emph{Dose 0.1 to 0.25: each cell gives matching / alternative context}}\\
Model/task & Response cosine & Relative norm ratio & Relative error (\%) \\
GPT-2/syntax & 0.999987 / 0.999987 & 0.999789 / 0.999784 & 0.44 / 0.44 \\
GPT-2/binding & 0.999994 / 0.999994 & 1.000019 / 0.999926 & 0.35 / 0.35 \\
Qwen2.5/syntax & 0.999947 / 0.999946 & 0.999609 / 0.999700 & 1.00 / 1.00 \\
Qwen2.5/binding & 0.999538 / 0.999533 & 1.000216 / 0.999948 & 3.00 / 3.01 \\
Llama3.2/syntax & 0.999950 / 0.999950 & 0.999946 / 1.000017 & 0.99 / 1.00 \\
Llama3.2/binding & 0.999890 / 0.999882 & 0.999390 / 1.000566 & 1.40 / 1.43 \\
\bottomrule
\end{tabular}
\end{table}

\subsection{Cross-group recovery and conditional gain}
The independent training donor was estimated from 128 training name groups. We deleted each held-out group's rank-16 $AB$ projection and refilled it with that donor, its sign reversal, or an equal-norm random direction. Recovery is the reduction in absolute distance from the group's native $G$. The training donor outperformed both controls in all three models (Table~\ref{tab:supp-donor-restore}); this criterion measures return toward the original response.

For the balanced eight conditions, $m(a,b,r)$ is candidate-1 minus candidate-0 logit. The fixed-direction edit is $h_{abr}(\epsilon)=h_{abr}+ab\epsilon v$, with $v$ constant across conditions within a group. In the projection experiment, $v$ is the group's $AB$ coefficient projected onto the frozen training rank-16 basis. Defining $\kappa(a,b,r)=\nabla_hm(a,b,r)^{\mathsf T}v$ and $\kappa_X=\mathbb E[\chi_X\kappa]$ gives
\begin{equation}
\beta_G=\mathbb E[(ar-br)ab\kappa]=\kappa_{BR}-\kappa_{AR},\qquad
\beta_C=\mathbb E[(a+b)ab\kappa]=\kappa_A+\kappa_B.
\label{eq:supp-conditional-gain}
\end{equation}
The first 32 test groups supply analytic derivatives; all 64 supply finite-dose responses (Table~\ref{tab:supp-d2-protocols}). The late-clamp extension reuses these same 64 groups, extending the derivative calculation and introducing component interventions. Each protocol retains its sample size and editing direction.

\begin{table}[!htbp]
\centering\footnotesize
\setlength{\tabcolsep}{5pt}
\caption{Component intervention outcomes, with analytic and finite-dose protocols separated. All displayed numbers are multiplied by $10^3$; brackets are 95\% source-group bootstrap intervals. The finite intervention subtracts half of the rank-16 $AB$ projection.}
\label{tab:supp-d2-protocols}
\begin{tabular}{llll}
\toprule
Model & $\beta_G$, analytic ($n=32$) & $\Delta G$, dose $-0.5$ ($n=64$) & $\Delta C$, dose $-0.5$ ($n=64$) \\
\midrule
GPT-2 & \shortstack{$-0.0175$\\$[-0.1409,+0.1051]$} & \shortstack{$+0.021$\\$[-0.018,+0.058]$} & \shortstack{$-0.432$\\$[-0.540,-0.321]$} \\
Qwen2.5 & \shortstack{$+6.505$\\$[+3.898,+9.111]$} & \shortstack{$-1.887$\\$[-2.919,-0.867]$} & \shortstack{$+5.546$\\$[+2.988,+8.114]$} \\
Llama3.2 & \shortstack{$-7.273$\\$[-9.393,-5.086]$} & \shortstack{$+3.678$\\$[+2.892,+4.372]$} & \shortstack{$+7.092$\\$[+6.280,+7.928]$} \\
\bottomrule
\end{tabular}
\end{table}

\subsection{A fixed direction across material working points}
\label{sup:s4-material-protocol}
\begin{figure}[!htbp]
\centering
\includegraphics[width=\linewidth,height=.78\textheight,keepaspectratio]{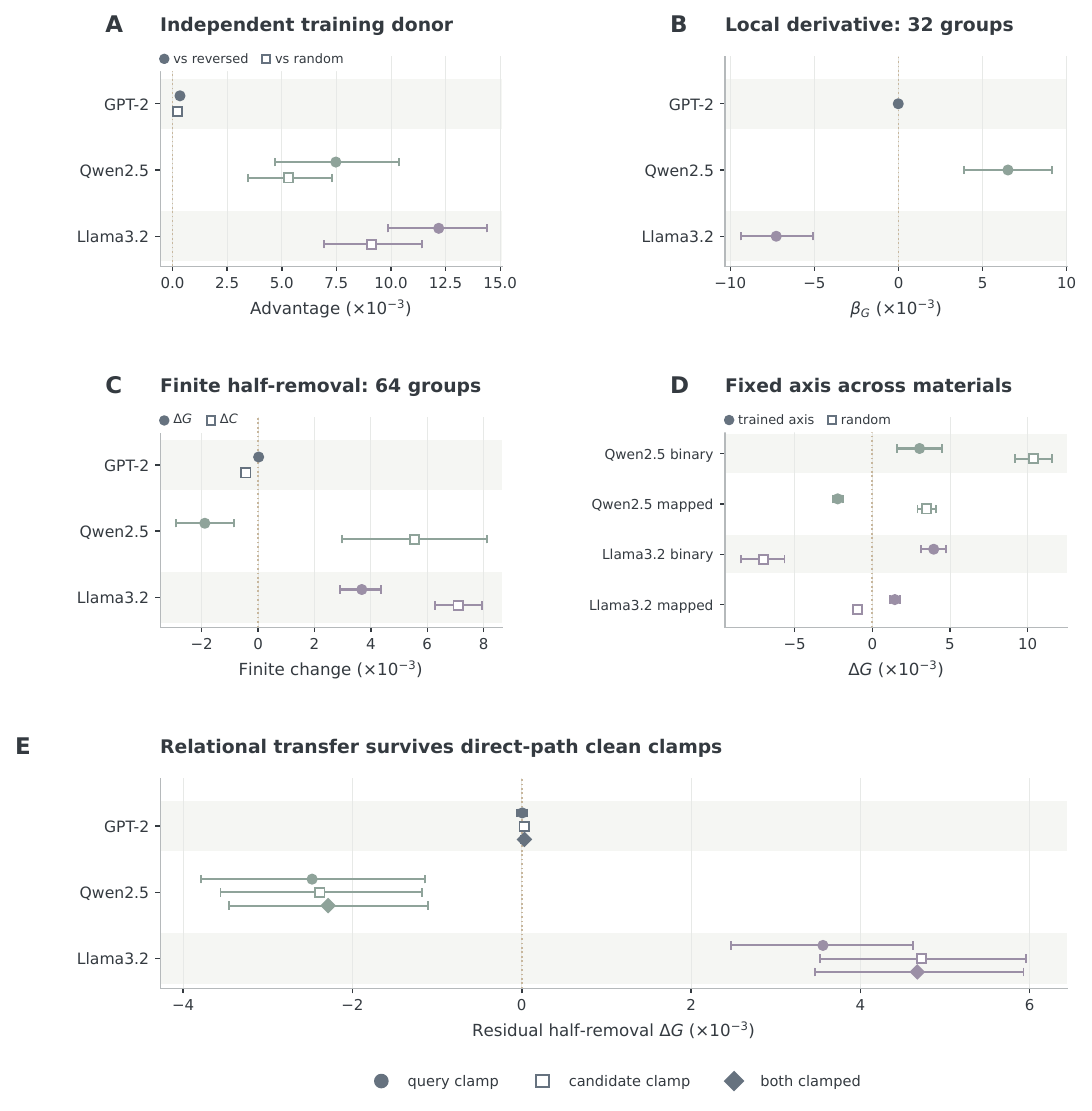}
\caption{Relational transfer under distinct protocols. (A) Donor recovery against reversed (filled) and random (open) refills. (B) Analytic derivatives in 32 groups. (C) Half-projection removal in 64 groups: $\Delta G$ (filled) and $\Delta C$ (open). (D) Fixed training-mean axis (filled) and equal-norm random (open) edits across binary and mapped materials, using orange/purple and red/blue output pairs with the same category and query-role coding. (E) Residual half-removal effects under direct query, candidate or joint clean clamps in 32 groups. Error bars retain the original 95\% source-group intervals.}
\label{fig:supp-v7-ab-transfer}
\end{figure}

A separate experiment holds the checkpoint, training-mean $AB$ vector and dose $+0.5abv$ fixed across two material regimes. The vector is the mean middle-query $AB$ coefficient over 128 independent training-name groups, injected at the query entity's last token in Attention~6 of GPT-2, Attention~14 of Qwen2.5 or Attention~8 of Llama3.2. Each regime contains the same 64 test name-pair groups and all eight attribute--query conditions. Qwen2.5 changes sign; GPT-2 and Llama3.2 change magnitude while remaining positive (Table~\ref{tab:supp-donor-restore}; the Qwen2.5/Llama3.2 comparison is shown in Fig.~\ref{fig:supp-v7-ab-transfer}D). GPT-2's edits and unedited baseline were both evaluated on CPU in FP32 with batch size 16; the saved baseline audit pairs each edit with this recomputed baseline.

The following complete prompts are the saved test group 000 at category indices $(0,1)$, querying the first entity. A leading space is part of every candidate token.
\begin{center}\small
\setlength{\tabcolsep}{5pt}
\begin{tabularx}{\linewidth}{@{}XX@{}}
\toprule
\textbf{Binary material} & \textbf{Mapped-lexeme material}\\
\midrule
Rachel is orange. Simon is purple. The color of Rachel is &
In this code, crimson and cherry mean red; navy and denim mean blue. Note: quiet calm. Rachel is crimson. Simon is denim. The color category of Rachel is\\[4pt]
Candidate 0: \texttt{\textvisiblespace orange}; candidate 1: \texttt{\textvisiblespace purple} &
Candidate 0: \texttt{\textvisiblespace red}; candidate 1: \texttt{\textvisiblespace blue}\\[4pt]
GPT-2 IDs: 10912 / 14032 & GPT-2 IDs: 2266 / 4171\\
Qwen2.5 IDs: 18575 / 24932 & Qwen2.5 IDs: 2518 / 6303\\
Llama3.2 IDs: 19087 / 25977 & Llama3.2 IDs: 2579 / 6437\\
\bottomrule
\end{tabularx}
\end{center}
In both regimes, category indices $i_A,i_B\in\{0,1\}$ give $a=2i_A-1$ and $b=2i_B-1$, while $r=+1$ queries entity $A$ and $r=-1$ queries entity $B$. The margin is always candidate-1 minus candidate-0 logit, and $G=\mathbb E[(ar-br)m]$ equals mean target-minus-distractor margin in the four conflict conditions. Thus the categorical coding is fixed while the numerical output axis changes from purple-minus-orange to blue-minus-red. The binary prompts have 13--15 tokens with query positions 11--13; mapped prompts have 35 tokens with query position 33 in all three models. The mapped same-lexeme and different-lexeme controls share these lengths and positions. Their length matching is within the mapping design. The fixed-axis experiment measures transfer across the full prompt and output-vocabulary change.

The exported \texttt{fixed\_axis\_protocol.json} records complete examples, token IDs, query positions, direction hashes and the saved-score identity for $G$; \texttt{transfer\_metadata/} contains all test conditions (Supplementary Data, \texttt{materials/}). The source package also includes the GPT-2 protocol and matched-backend records. Equal-norm random directions retain their measured signed effects. The fixed mean axis measures transfer across material regimes, whereas the group-projection edit measures each test group's relational component.

\begin{table}[!htbp]
\centering\footnotesize
\setlength{\tabcolsep}{4pt}
\caption{Cross-group donor recovery and fixed-axis material transfer. All numerical effects are in $10^{-3}$ units with 95\% source-group bootstrap intervals. Both blocks use 64 source groups. Fixed-axis transfer holds the injection vector and dose constant across binary orange/purple and mapped red/blue output pairs, with common category and query-role coding.}
\label{tab:supp-donor-restore}
\begin{tabular}{llrr}
\toprule
Model & Protocol & Donor $-$ reversed & Donor $-$ random \\
\midrule
\addlinespace
\multicolumn{4}{l}{\emph{Donor recovery advantage; $n=64$}}\\
GPT-2 & recovery & \shortstack{$+0.338$\\$[+0.257,+0.427]$} & \shortstack{$+0.209$\\$[+0.143,+0.281]$} \\
Qwen2.5 & recovery & \shortstack{$+7.473$\\$[+4.684,+10.356]$} & \shortstack{$+5.319$\\$[+3.468,+7.316]$} \\
Llama3.2 & recovery & \shortstack{$+12.181$\\$[+9.845,+14.388]$} & \shortstack{$+9.099$\\$[+6.946,+11.399]$} \\
\addlinespace
\multicolumn{4}{l}{\emph{Fixed mean-axis edit, dose $+0.5$: trained-axis and random-direction $\Delta G$}}\\
Model & Material & Trained $AB$ axis & Equal-norm random \\
GPT-2 & binary & \shortstack{$+0.039$\\$[+0.021,+0.057]$} & \shortstack{$-0.081$\\$[-0.097,-0.066]$} \\
GPT-2 & mapped & \shortstack{$+0.065$\\$[+0.051,+0.079]$} & \shortstack{$-0.005$\\$[-0.016,+0.007]$} \\
Qwen2.5 & binary & \shortstack{$+3.040$\\$[+1.560,+4.500]$} & \shortstack{$+10.380$\\$[+9.200,+11.550]$} \\
Qwen2.5 & mapped & \shortstack{$-2.230$\\$[-2.550,-1.910]$} & \shortstack{$+3.470$\\$[+2.910,+4.080]$} \\
Llama3.2 & binary & \shortstack{$+3.950$\\$[+3.160,+4.750]$} & \shortstack{$-7.000$\\$[-8.480,-5.650]$} \\
Llama3.2 & mapped & \shortstack{$+1.450$\\$[+1.120,+1.770]$} & \shortstack{$-0.960$\\$[-1.220,-0.720]$} \\
\bottomrule
\end{tabular}
\end{table}

\subsection{Residual transfer under direct-contribution clamps}
We crossed half-removal of the rank-16 $AB$ projection with clean-baseline clamps on the designated deep-attention query contribution, candidate contribution, or both. Each clamp replaces the selected post-softmax value contribution with its clean value for the same condition, retaining the other attention probabilities. Zero-edit clean clamps reproduce the baseline. With the edit present, Qwen2.5 retains negative and Llama3.2 positive residual $\Delta G$ under all three clamp configurations (Table~\ref{tab:supp-d2-path-residual}). This residual establishes transmission beyond the specified direct contributions and motivates the late-component decomposition in the next section.

\begin{table}[!htbp]
\centering\footnotesize
\setlength{\tabcolsep}{5pt}
\caption{Residual $AB$ half-removal effect after direct contributions are clamped to the clean baseline ($\times10^3$; $n=32$ paired source groups). These are finite edited-minus-clean responses, not derivatives. Brackets are 95\% source-group bootstrap intervals.}
\label{tab:supp-d2-path-residual}
\begin{tabular}{llll}
\toprule
Model & Contribution clamped & Residual $\Delta G$ & Residual $\Delta C$ \\
\midrule
GPT-2 & Query & \shortstack{$+0.006$\\$[-0.053,+0.064]$} & \shortstack{$-0.257$\\$[-0.427,-0.097]$} \\
GPT-2 & Candidate & \shortstack{$+0.027$\\$[-0.008,+0.064]$} & \shortstack{$-0.377$\\$[-0.509,-0.249]$} \\
GPT-2 & Both & \shortstack{$+0.031$\\$[-0.005,+0.067]$} & \shortstack{$-0.292$\\$[-0.430,-0.161]$} \\
Qwen2.5 & Query & \shortstack{$-2.477$\\$[-3.785,-1.145]$} & \shortstack{$+5.431$\\$[+2.170,+8.711]$} \\
Qwen2.5 & Candidate & \shortstack{$-2.385$\\$[-3.559,-1.182]$} & \shortstack{$+5.384$\\$[+2.529,+8.424]$} \\
Qwen2.5 & Both & \shortstack{$-2.288$\\$[-3.460,-1.109]$} & \shortstack{$+4.967$\\$[+2.108,+7.936]$} \\
Llama3.2 & Query & \shortstack{$+3.558$\\$[+2.470,+4.620]$} & \shortstack{$+6.960$\\$[+5.943,+7.961]$} \\
Llama3.2 & Candidate & \shortstack{$+4.723$\\$[+3.526,+5.958]$} & \shortstack{$+8.235$\\$[+7.155,+9.357]$} \\
Llama3.2 & Both & \shortstack{$+4.673$\\$[+3.461,+5.927]$} & \shortstack{$+8.222$\\$[+7.187,+9.308]$} \\
\bottomrule
\end{tabular}
\end{table}

\FloatBarrier
\section{Signed transfer through fixed late-layer components}
\label{sup:s5-clamp}
This section moves from prespecified component clamps to additive tangent-path totals, then tests direction dependence, static/dynamic differences and numerical stability.

\subsection{Shared material and zero-edit state}
The late-clamp extension reuses the same 64 held-out eight-condition groups and training rank-16 basis as the earlier projection experiment. The previous derivative subset comprised the first 32 groups. The injected vector in each group is exactly its $AB$ coefficient projected onto that basis, with $\epsilon abv$ added at the middle query output (Qwen2.5 Attention~14; Llama3.2 Attention~8). Query and prediction positions are 13 and 14 in the 15-token sequence. Each clean clamp replaces only the selected component's prediction-position write with the corresponding same-condition unedited write, so all clamp graphs share the zero-edit state.

The fixed attention sets are layers $\{21,23,25,27\}$ in Qwen2.5 and $\{12,13,15,16\}$ in Llama3.2; the MLP sets are $\{24,28\}$ and $\{15,16\}$. These components were selected using preceding static attribution. Pre-injection attention controls are Qwen2.5 layers 5 and 13 and Llama3.2 layers 3 and 7. Table~\ref{tab:supp-clamp-all} contains all single, control and joint operations.

\subsection{Single-component shifts and inference}
We record $\Delta\beta_G=\beta_{\mathrm{clamp}}-\beta_0$, the negative of main-text net transmission $\tau_k=\beta_0-\beta_{\mathrm{clamp}(k)}$. A positive clamp shift therefore removes net negative native transmission. Source-group intervals use 10,000 paired bootstrap resamples; two-sided sign-flip tests use 50,000 draws. Holm adjustment applies within the 12 single-component comparisons, with separate families for 12 signed random-reference comparisons and two final-denominator comparisons. The magnitude controls below are reported with their own exploratory paired intervals.

\begingroup\footnotesize
\setlength{\tabcolsep}{4pt}
\begin{longtable}{lrrrr}
\caption{All fixed operations in the 64-group late-clamp extension. Derivatives and shifts are in $10^{-3}$ units; brackets are 95\% source-group bootstrap intervals. Holm values apply to the single-node family. Pre-injection controls and the native row have exactly zero shifts.}\label{tab:supp-clamp-all}\label{tab:sup-clamp-single}\label{tab:sup-clamp-nodes}\\
\toprule
Clamped operation & $\beta_G$ & $\Delta\beta_G$ & $\Delta\beta_C$ & $p_{\rm Holm}$ \\
\midrule
\endfirsthead
\multicolumn{5}{l}{\emph{Table \thetable\ continued}}\\
\toprule
Clamped operation & $\beta_G$ & $\Delta\beta_G$ & $\Delta\beta_C$ & $p_{\rm Holm}$ \\
\midrule
\endhead
\bottomrule\endfoot
\addlinespace
\multicolumn{5}{l}{\emph{Qwen2.5: local derivatives at the common clean state}}\\
none & \shortstack{$+5.193$\\$[+3.119,+7.281]$} & \shortstack{$+0.000$\\$[+0.000,+0.000]$} & \shortstack{$+0.000$\\$[+0.000,+0.000]$} & --- \\
attn5 & \shortstack{$+5.193$\\$[+3.119,+7.281]$} & \shortstack{$+0.000$\\$[+0.000,+0.000]$} & \shortstack{$+0.000$\\$[+0.000,+0.000]$} & --- \\
attn13 & \shortstack{$+5.193$\\$[+3.119,+7.281]$} & \shortstack{$+0.000$\\$[+0.000,+0.000]$} & \shortstack{$+0.000$\\$[+0.000,+0.000]$} & --- \\
attn21 & \shortstack{$+2.104$\\$[+0.396,+3.815]$} & \shortstack{$-3.089$\\$[-3.907,-2.289]$} & \shortstack{$-2.062$\\$[-2.717,-1.388]$} & 0.00024 \\
attn23 & \shortstack{$+5.173$\\$[+3.269,+7.112]$} & \shortstack{$-0.019$\\$[-0.412,+0.336]$} & \shortstack{$+1.428$\\$[+0.710,+2.207]$} & 0.92256 \\
attn25 & \shortstack{$+6.704$\\$[+4.648,+8.889]$} & \shortstack{$+1.511$\\$[+1.053,+2.010]$} & \shortstack{$+1.338$\\$[+0.489,+2.231]$} & 0.00024 \\
attn27 & \shortstack{$+3.693$\\$[+2.076,+5.316]$} & \shortstack{$-1.500$\\$[-2.194,-0.859]$} & \shortstack{$+4.293$\\$[+2.381,+6.142]$} & 0.00024 \\
mlp24 & \shortstack{$+4.392$\\$[+2.399,+6.428]$} & \shortstack{$-0.800$\\$[-1.132,-0.482]$} & \shortstack{$+1.322$\\$[+0.587,+2.086]$} & 0.00024 \\
mlp28 & \shortstack{$+10.499$\\$[+6.982,+14.169]$} & \shortstack{$+5.307$\\$[+3.483,+7.257]$} & \shortstack{$-9.075$\\$[-13.418,-5.287]$} & 0.00024 \\
attention\_all & \shortstack{$+1.625$\\$[+0.773,+2.481]$} & \shortstack{$-3.567$\\$[-4.975,-2.182]$} & \shortstack{$+6.147$\\$[+2.722,+9.565]$} & --- \\
mlp\_all & \shortstack{$+8.365$\\$[+5.161,+11.712]$} & \shortstack{$+3.173$\\$[+1.601,+4.839]$} & \shortstack{$-5.617$\\$[-8.884,-2.772]$} & --- \\
all\_late & \shortstack{$+4.165$\\$[+2.809,+5.581]$} & \shortstack{$-1.027$\\$[-2.367,+0.334]$} & \shortstack{$+4.449$\\$[+1.132,+7.732]$} & --- \\
\addlinespace
\multicolumn{5}{l}{\emph{Llama3.2: local derivatives at the common clean state}}\\
none & \shortstack{$-7.425$\\$[-8.930,-5.902]$} & \shortstack{$+0.000$\\$[+0.000,+0.000]$} & \shortstack{$+0.000$\\$[+0.000,+0.000]$} & --- \\
attn3 & \shortstack{$-7.425$\\$[-8.930,-5.902]$} & \shortstack{$+0.000$\\$[+0.000,+0.000]$} & \shortstack{$+0.000$\\$[+0.000,+0.000]$} & --- \\
attn7 & \shortstack{$-7.425$\\$[-8.930,-5.902]$} & \shortstack{$+0.000$\\$[+0.000,+0.000]$} & \shortstack{$+0.000$\\$[+0.000,+0.000]$} & --- \\
attn12 & \shortstack{$-4.000$\\$[-5.040,-2.955]$} & \shortstack{$+3.425$\\$[+2.836,+4.008]$} & \shortstack{$+3.778$\\$[+3.265,+4.304]$} & 0.00024 \\
attn13 & \shortstack{$-9.799$\\$[-11.531,-8.038]$} & \shortstack{$-2.373$\\$[-2.820,-1.925]$} & \shortstack{$-2.173$\\$[-2.960,-1.439]$} & 0.00024 \\
attn15 & \shortstack{$-5.741$\\$[-6.940,-4.531]$} & \shortstack{$+1.684$\\$[+1.227,+2.139]$} & \shortstack{$+3.570$\\$[+2.895,+4.262]$} & 0.00024 \\
attn16 & \shortstack{$-8.003$\\$[-9.438,-6.500]$} & \shortstack{$-0.578$\\$[-0.898,-0.262]$} & \shortstack{$+1.862$\\$[+0.902,+2.831]$} & 0.00108 \\
mlp15 & \shortstack{$-10.091$\\$[-11.810,-8.294]$} & \shortstack{$-2.665$\\$[-3.304,-2.047]$} & \shortstack{$-3.077$\\$[-3.390,-2.774]$} & 0.00024 \\
mlp16 & \shortstack{$-10.316$\\$[-12.693,-8.108]$} & \shortstack{$-2.890$\\$[-3.992,-1.918]$} & \shortstack{$-1.918$\\$[-2.858,-1.050]$} & 0.00024 \\
attention\_all & \shortstack{$-0.750$\\$[-1.033,-0.486]$} & \shortstack{$+6.675$\\$[+5.284,+8.006]$} & \shortstack{$+13.295$\\$[+11.867,+14.717]$} & --- \\
mlp\_all & \shortstack{$-13.369$\\$[-15.727,-11.035]$} & \shortstack{$-5.943$\\$[-6.889,-5.039]$} & \shortstack{$-4.300$\\$[-4.893,-3.718]$} & --- \\
all\_late & \shortstack{$-0.685$\\$[-0.871,-0.509]$} & \shortstack{$+6.740$\\$[+5.332,+8.107]$} & \shortstack{$+13.226$\\$[+11.586,+14.876]$} & --- \\
\end{longtable}
\endgroup

\subsection{Joint graphs and the four tangent-path categories}
\begin{figure}[!htbp]
\centering
\includegraphics[width=\linewidth,height=.78\textheight,keepaspectratio]{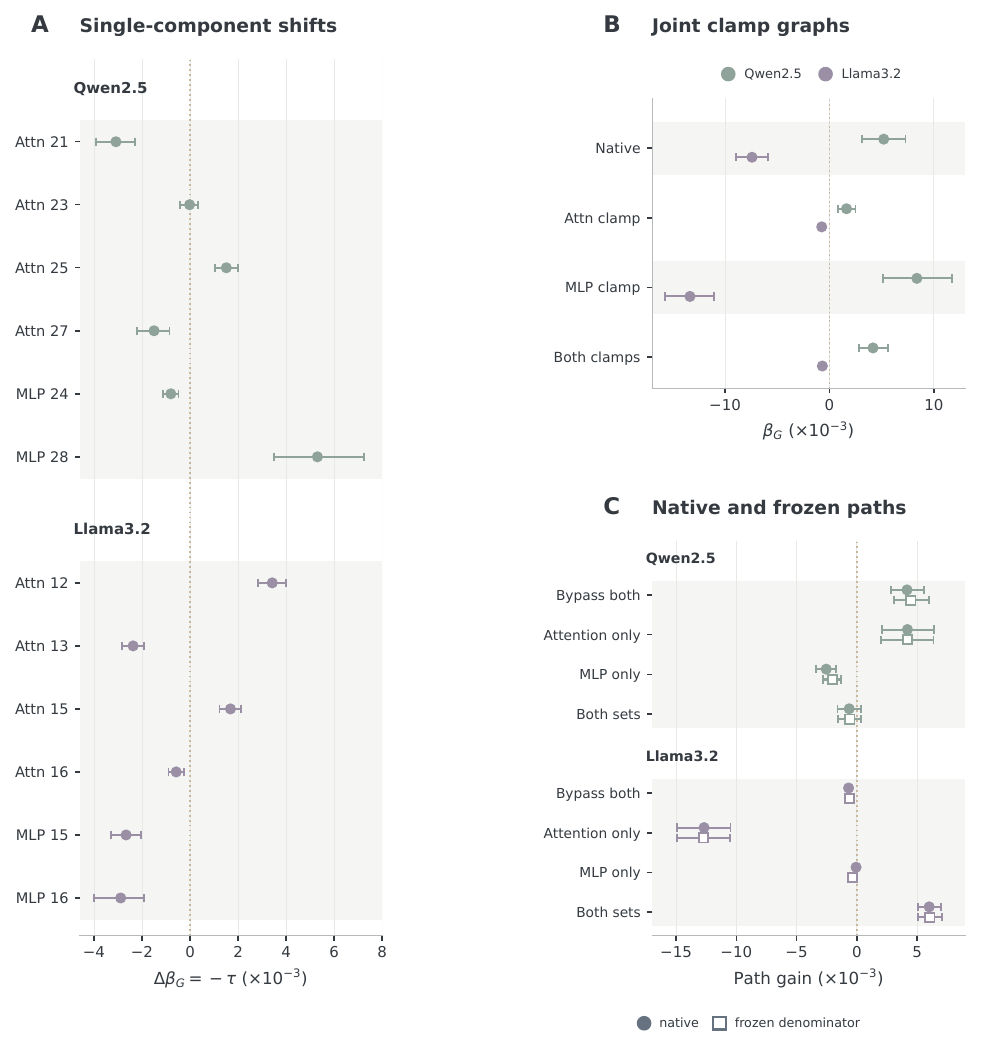}
\caption{Late-component transfer. (A) Twelve single-node clamp shifts on one scale, $\Delta\beta_G=\beta_{\rm clamp}-\beta_0=-\tau$. (B) Native and jointly clamped graph gains. (C) Four additive tangent-path totals under native (filled circles) and frozen-final-denominator (open squares) readout. Points and 95\% source-group intervals use all 64 groups per model; the selected attention and MLP sets define the path categories.}
\label{fig:supp-v7-clamp}
\end{figure}

Let $\beta_A$, $\beta_M$ and $\beta_{AM}$ denote gains after clamping the attention, MLP or both selected sets. The common clean state permits the additive partition
\begin{align}
b_{\varnothing}&=\beta_{AM}, & b_{A\bar M}&=\beta_M-\beta_{AM},\\
b_{\bar A M}&=\beta_A-\beta_{AM}, & b_{AM}&=\beta_0-\beta_A-\beta_M+\beta_{AM}.
\end{align}
Each term sums tangent paths classified by traversal of the selected sets; their sum is exactly $\beta_0$ within every group. Llama3.2's positive compensation is concentrated in paths traversing both sets, against a stronger negative attention-only total. Qwen2.5 instead combines positive bypass and attention-only totals with a negative MLP-only total (Table~\ref{tab:supp-clamp-paths}; Fig.~\ref{fig:supp-v7-clamp}).

\begin{table}[!htbp]
\centering\footnotesize
\setlength{\tabcolsep}{4pt}
\caption{Joint graph derivatives and their additive tangent-path re-expression ($10^{-3}$ units; $n=64$ per model). Brackets are paired source-group 95\% intervals. Native and frozen-final-denominator path totals use the same selected sets and source groups.}
\label{tab:supp-clamp-paths}
\label{tab:sup-clamp-joint}
\label{tab:sup-clamp-partition}
\begin{tabular}{lrr}
\toprule
Graph / path class & Qwen2.5 & Llama3.2 \\
\midrule
\addlinespace
\multicolumn{3}{l}{\emph{Forward-graph gains}}\\
Native & \shortstack{$+5.193$\\$[+3.119,+7.281]$} & \shortstack{$-7.425$\\$[-8.930,-5.902]$} \\
Attention clamp & \shortstack{$+1.625$\\$[+0.773,+2.481]$} & \shortstack{$-0.750$\\$[-1.033,-0.486]$} \\
MLP clamp & \shortstack{$+8.365$\\$[+5.161,+11.712]$} & \shortstack{$-13.369$\\$[-15.727,-11.035]$} \\
Joint clamp & \shortstack{$+4.165$\\$[+2.809,+5.581]$} & \shortstack{$-0.685$\\$[-0.871,-0.509]$} \\
\addlinespace
\multicolumn{3}{l}{\emph{Additive tangent-path totals}}\\
Bypass both & \shortstack{$+4.165$\\$[+2.809,+5.581]$} & \shortstack{$-0.685$\\$[-0.871,-0.509]$} \\
Attention only & \shortstack{$+4.200$\\$[+2.061,+6.379]$} & \shortstack{$-12.683$\\$[-14.915,-10.488]$} \\
MLP only & \shortstack{$-2.540$\\$[-3.384,-1.731]$} & \shortstack{$-0.065$\\$[-0.231,+0.101]$} \\
Both sets & \shortstack{$-0.633$\\$[-1.610,+0.326]$} & \shortstack{$+6.008$\\$[+5.074,+7.003]$} \\
\addlinespace
\multicolumn{3}{l}{\emph{Frozen-final-denominator tangent-path totals}}\\
% BEGIN GENERATED ROWS: frozen_path_rows
Bypass both & \shortstack{$+4.493$\\$[+3.092,\,+5.963]$} & \shortstack{$-0.639$\\$[-0.813,\,-0.472]$} \\
Attention only & \shortstack{$+4.185$\\$[+2.039,\,+6.369]$} & \shortstack{$-12.713$\\$[-14.947,\,-10.516]$} \\
MLP only & \shortstack{$-2.034$\\$[-2.788,\,-1.314]$} & \shortstack{$-0.367$\\$[-0.531,\,-0.205]$} \\
Both sets & \shortstack{$-0.579$\\$[-1.556,\,+0.375]$} & \shortstack{$+6.063$\\$[+5.108,\,+7.078]$} \\
% END GENERATED ROWS: frozen_path_rows
\bottomrule
\end{tabular}
\end{table}

\subsection{Static writes, conditional cancellation and random directions}
Static attribution holds each complete sample's final normalization coefficients fixed and decomposes its baseline logit contrast into residual writes. Dynamic net transmission measures a derivative difference between clean-clamped forward graphs. Qwen2.5 MLP~28 and Llama3.2 Attention~13 have opposite signs in these two coordinates. Qwen2.5 Attention~23 has substantial early and late effects with opposite signs that nearly cancel in the pooled mean (Table~\ref{tab:supp-clamp-controls}).

\begin{figure}[!htbp]
\centering
\includegraphics[width=\linewidth,height=.78\textheight,keepaspectratio]{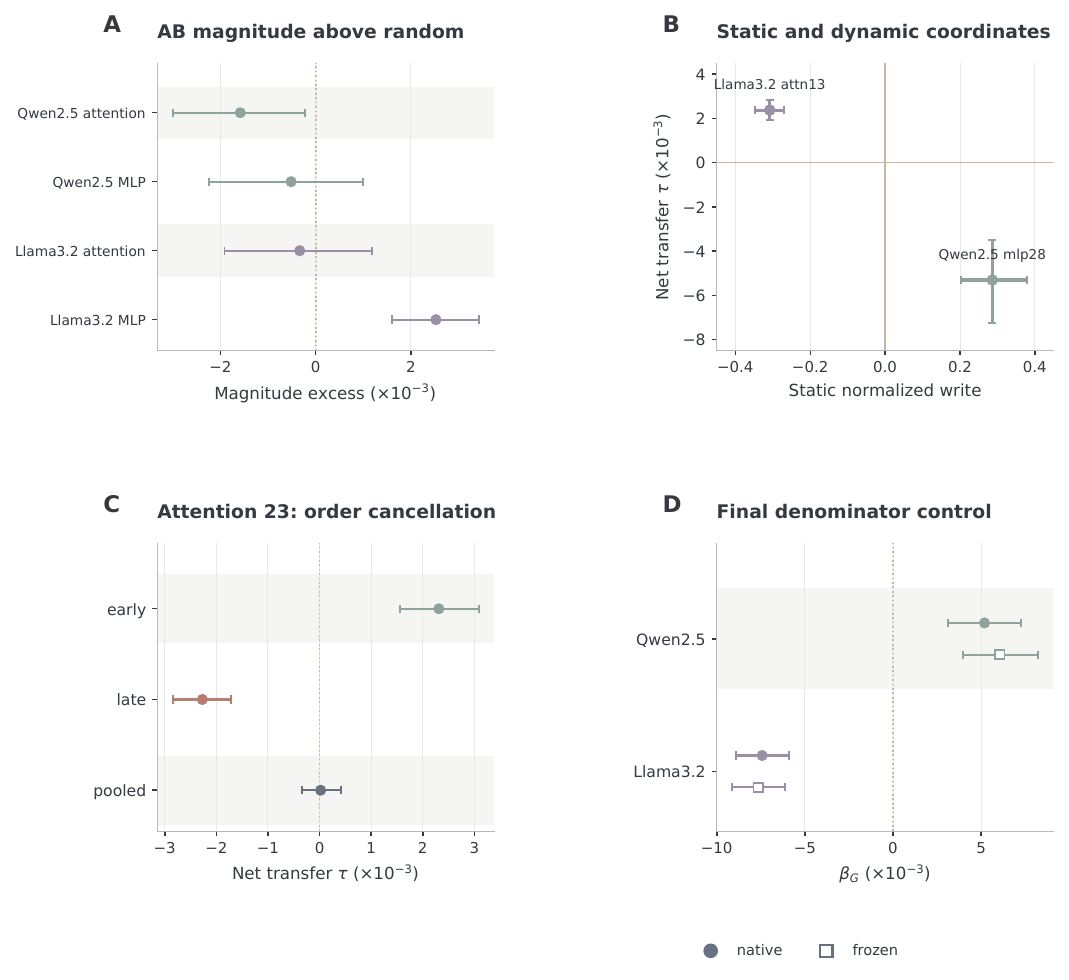}
\caption{Direction, attribution and scale controls. (A) Absolute $AB$ clamp shift minus the within-group random magnitude reference. (B) Static write versus net dynamic transmission $\tau$ for two components. (C) Qwen2.5 Attention~23 early/late cancellation. (D) Native and frozen-final-denominator $\beta_G$. Error bars are 95\% source-group bootstrap intervals. Static writes and dynamic derivatives use their stated coordinates.}
\label{fig:supp-v7-controls}
\end{figure}

Two equal-norm Gaussian directions per group probe the same selected sets. Random magnitudes are averaged within group before inference. The attention sets carry these directions as well as the trained $AB$ component; Llama3.2's MLP set has an absolute $AB$ shift exceeding the random reference by $2.527$ $[1.596,3.431]\times10^{-3}$. Thus general downstream channels have direction-dependent signed use (Fig.~\ref{fig:supp-v7-controls}).

\begin{table}[!htbp]
\centering\footnotesize
\setlength{\tabcolsep}{4pt}
\caption{Static/dynamic and direction controls ($n=64$). Units are stated separately for each block. Each interval resamples complete source groups.}
\label{tab:supp-clamp-controls}
\label{tab:sup-clamp-random}
\begin{tabular}{lrr}
\toprule
Component & Static write & Net transmission \\
\midrule
\addlinespace
\multicolumn{3}{l}{\emph{Static write (raw margin) and net dynamic transmission ($10^{-3}$)}}\\
Qwen2.5 mlp28 & \shortstack{$+0.286$\\$[+0.204,+0.378]$} & \shortstack{$-5.307$\\$[-7.257,-3.483]$} \\
Llama3.2 attn13 & \shortstack{$-0.308$\\$[-0.349,-0.271]$} & \shortstack{$+2.373$\\$[+1.925,+2.820]$} \\
\addlinespace
\multicolumn{3}{l}{\emph{Qwen2.5 Attention 23 net transmission ($10^{-3}$)}}\\
Stratum & Net transmission &  \\
early & \shortstack{$+2.310$\\$[+1.561,+3.088]$} &  \\
late & \shortstack{$-2.272$\\$[-2.849,-1.717]$} &  \\
pooled & \shortstack{$+0.019$\\$[-0.336,+0.412]$} &  \\
\addlinespace
\multicolumn{3}{l}{\emph{Random magnitude and $AB$ magnitude excess (both $10^{-3}$)}}\\
Model / set & Mean $|\Delta\beta_{random}|$ & $|\Delta\beta_{AB}|-$ random \\
Qwen2.5 Attention & \shortstack{$+6.921$\\$[+5.802,+8.096]$} & \shortstack{$-1.585$\\$[-3.009,-0.225]$} \\
Qwen2.5 MLP & \shortstack{$+5.890$\\$[+4.375,+7.561]$} & \shortstack{$-0.517$\\$[-2.251,+0.999]$} \\
Llama3.2 Attention & \shortstack{$+7.800$\\$[+6.660,+9.063]$} & \shortstack{$-0.335$\\$[-1.916,+1.184]$} \\
Llama3.2 MLP & \shortstack{$+3.416$\\$[+2.767,+4.148]$} & \shortstack{$+2.527$\\$[+1.596,+3.431]$} \\
\bottomrule
\end{tabular}
\end{table}

\subsection{Final normalization and numerical validation}
Freezing each condition's final RMSNorm denominator preserves the positive Qwen2.5 and negative Llama3.2 gain signs. Internal normalization remains native. Network activations and forward computation use FP32; candidate-head dot products, output contrasts and derivative contractions use FP64 accumulation. The final-denominator comparison therefore tests a specific downstream scale while the numerical audit checks consistency of the computed local derivatives. Clean-clamp margin differences and pre-injection derivative shifts are exactly zero. Central differences at $\epsilon=\pm0.1$ provide an independent local check (Table~\ref{tab:supp-clamp-numerics}).

\begin{table}[!htbp]
\centering\footnotesize
\setlength{\tabcolsep}{4pt}
\caption{Final-denominator and numerical checks over all 64 source groups. The derivative block gives estimates and 95\% intervals; the audit block gives maximum absolute discrepancies.}
\label{tab:supp-clamp-numerics}
\label{tab:sup-clamp-norm}
\label{tab:sup-clamp-numerical}
\begin{tabular}{lrr}
\toprule
Quantity & Qwen2.5 & Llama3.2 \\
\midrule
\addlinespace
\multicolumn{3}{l}{\emph{Derivative estimates and intervals ($10^{-3}$)}}\\
Native $\beta_G$ & \shortstack{$+5.193$\\$[+3.119,+7.281]$} & \shortstack{$-7.425$\\$[-8.930,-5.902]$} \\
Frozen-denominator $\beta_G$ & \shortstack{$+6.065$\\$[+3.948,+8.246]$} & \shortstack{$-7.655$\\$[-9.138,-6.141]$} \\
Native $\beta_C$ & \shortstack{$-10.371$\\$[-15.498,-5.346]$} & \shortstack{$-14.262$\\$[-15.875,-12.637]$} \\
\addlinespace
\multicolumn{3}{l}{\emph{Maximum absolute numerical differences (unscaled original units)}}\\
Clean-clamp margin & $0$ & $0$ \\
Pre-injection derivative shift & $0$ & $0$ \\
Native/frozen zero-dose readout & $3.51\times10^{-7}$ & $4.36\times10^{-7}$ \\
Analytic/central margin derivative & $8.74\times10^{-5}$ & $5.08\times10^{-5}$ \\
Independent contrast recomputation & $2.78\times10^{-17}$ & $1.39\times10^{-17}$ \\
\bottomrule
\end{tabular}
\end{table}

\FloatBarrier
\section{Shared receivers and fact-order competition}
\label{sup:s6-receiving}
This section tests non-additive receiving-point combination, separates candidate and query contributions, and resolves their order-conditioned effects against the baseline write balance.

\subsection{Matched shared and separated receivers}
\begin{figure}[!htbp]
\centering
\includegraphics[width=\linewidth,height=.78\textheight,keepaspectratio]{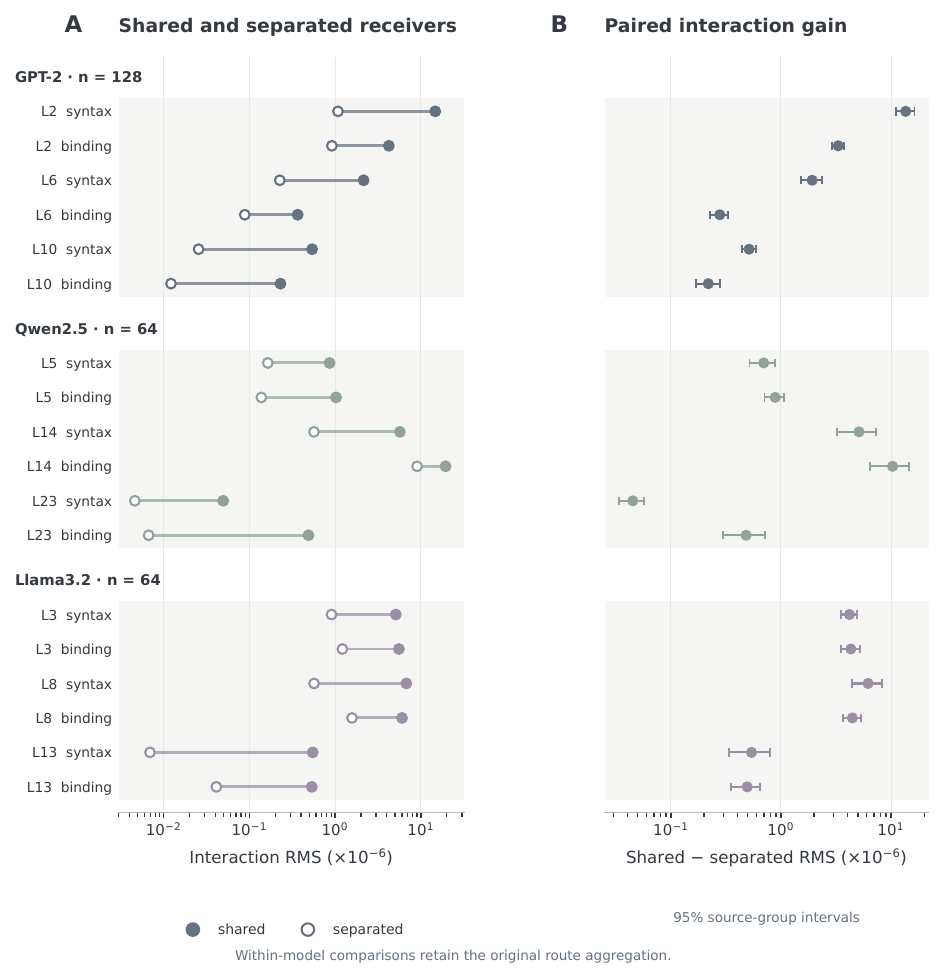}
\caption{Complete receiving-point comparisons. (A) Shared (filled) and separated (open) mean interaction RMS for all 18 model--task--depth comparisons. (B) Row-matched shared-minus-separated differences and 95\% source-group intervals; every lower endpoint is positive. Both axes are logarithmic. GPT-2 uses 128 groups per task with route-wise RMS averaging; Qwen2.5 and Llama3.2 use 64 groups per task with pooled-route RMS. The comparisons are paired within each model.}
\label{fig:supp-v7-receiving}
\end{figure}

For each route, the four intervention masks define the descendant-margin interaction $I_m=m_{11}-m_{10}-m_{01}+m_{00}$. The contrast is the RMS interaction at shared receivers minus that at separated receivers. GPT-2 averages route-wise RMS values; Qwen2.5 and Llama3.2 pool routes within receiver class before computing RMS. Each contrast retains its original within-model estimator. GPT-2 uses 128 source groups per task and total dose $0.25$; Qwen2.5 and Llama3.2 use 64 groups per task under their matched source-direction route protocol. All 18 model--task--depth intervals are positive (Table~\ref{tab:sup-c6-values}; Fig.~\ref{fig:supp-v7-receiving}).

\begin{table}[!htbp]
\centering\footnotesize
\setlength{\tabcolsep}{4pt}
\caption{Shared-versus-separated receiver interaction. Values are in $10^{-6}$ margin units; contrast intervals use source-group bootstrap. GPT-2 uses 128 source groups per task and averages route-wise RMS; Qwen2.5/Llama3.2 use 64 groups per task and pool routes before computing RMS. Qwen2.5/Llama3.2 contrast intervals use the saved geometry/function reanalysis; shared and separate means retain the route summaries.}
\label{tab:sup-c6-values}
\begin{tabular}{lllrrr}
\toprule
Model & Layer & Task & Shared RMS & Separate RMS & Difference [95\% CI] \\
\midrule
GPT-2 & 2 & syntax & 14.6733 & 1.0790 & \shortstack{$+13.5943$\\$[+11.0088,+16.3434]$} \\
GPT-2 & 2 & binding & 4.2288 & 0.9165 & \shortstack{$+3.3123$\\$[+2.8922,+3.7573]$} \\
GPT-2 & 6 & syntax & 2.1464 & 0.2265 & \shortstack{$+1.9200$\\$[+1.5096,+2.3765]$} \\
GPT-2 & 6 & binding & 0.3662 & 0.0885 & \shortstack{$+0.2777$\\$[+0.2281,+0.3295]$} \\
GPT-2 & 10 & syntax & 0.5386 & 0.0256 & \shortstack{$+0.5130$\\$[+0.4416,+0.5958]$} \\
GPT-2 & 10 & binding & 0.2307 & 0.0122 & \shortstack{$+0.2185$\\$[+0.1691,+0.2797]$} \\
Qwen2.5 & 5 & syntax & 0.8613 & 0.1640 & \shortstack{$+0.6973$\\$[+0.5179,+0.8907]$} \\
Qwen2.5 & 5 & binding & 1.0240 & 0.1381 & \shortstack{$+0.8859$\\$[+0.7091,+1.0716]$} \\
Qwen2.5 & 14 & syntax & 5.6851 & 0.5666 & \shortstack{$+5.1186$\\$[+3.2277,+7.2777]$} \\
Qwen2.5 & 14 & binding & 19.3611 & 9.0249 & \shortstack{$+10.3362$\\$[+6.4184,+14.6379]$} \\
Qwen2.5 & 23 & syntax & 0.0496 & 0.0046 & \shortstack{$+0.0450$\\$[+0.0337,+0.0574]$} \\
Qwen2.5 & 23 & binding & 0.4885 & 0.0067 & \shortstack{$+0.4818$\\$[+0.2988,+0.7099]$} \\
Llama3.2 & 3 & syntax & 5.0981 & 0.9073 & \shortstack{$+4.1908$\\$[+3.4973,+4.9276]$} \\
Llama3.2 & 3 & binding & 5.5358 & 1.2162 & \shortstack{$+4.3196$\\$[+3.5308,+5.1962]$} \\
Llama3.2 & 8 & syntax & 6.7593 & 0.5680 & \shortstack{$+6.1913$\\$[+4.3821,+8.2654]$} \\
Llama3.2 & 8 & binding & 6.0271 & 1.5731 & \shortstack{$+4.4540$\\$[+3.6387,+5.3265]$} \\
Llama3.2 & 13 & syntax & 0.5471 & 0.0069 & \shortstack{$+0.5401$\\$[+0.3395,+0.7914]$} \\
Llama3.2 & 13 & binding & 0.5344 & 0.0413 & \shortstack{$+0.4931$\\$[+0.3523,+0.6442]$} \\
\bottomrule
\end{tabular}
\end{table}

\subsection{Source contributions and paired fact order}
In the separate 15-token eight-condition design, candidate, query and joint removal set the designated post-softmax source-value contributions to zero while retaining the other attention probabilities. Their edited-minus-clean $\Delta G$ and $\Delta C$ can differ in sign (Table~\ref{tab:supp-source-order}).

Early and late identify whether the queried entity's fact occurs first or second within the same source group. Pairing the two roles holds the group fixed. Candidate removal has positive early means and negative late means in all three models; every paired late-minus-early interval is negative. Qwen2.5's individual late interval spans zero, while its paired difference remains negative. This identifies relative support for the later fact by the candidate contribution.

\begin{figure}[H]
\centering
\includegraphics[width=\linewidth,height=.78\textheight,keepaspectratio]{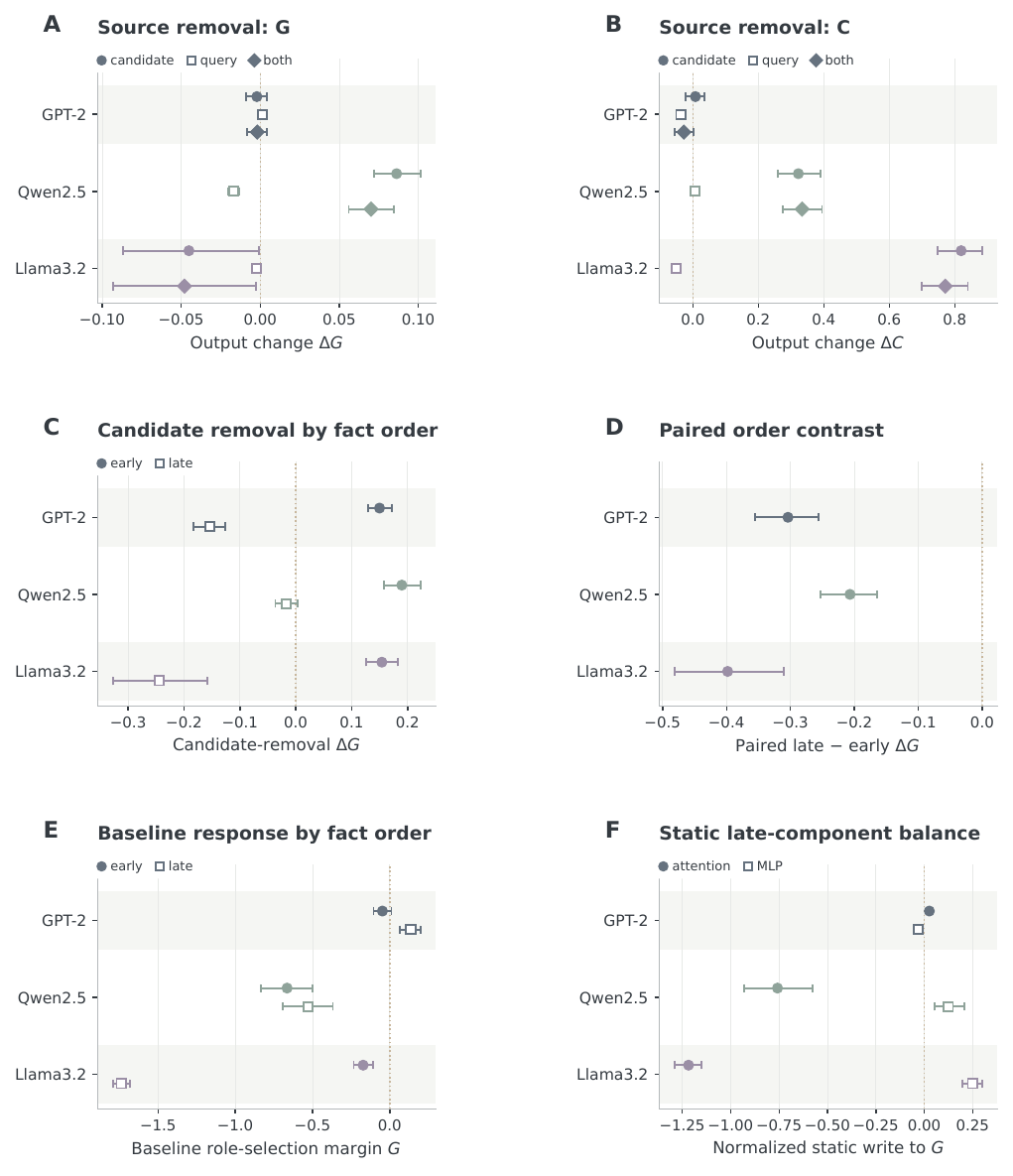}
\caption{Source and order effects. (A,B) Candidate (filled circles), query (open squares) and joint (diamonds) removal in $G$ and $C$. (C) Candidate removal for early (filled) and late (open) facts. (D) Paired late-minus-early removal effect. (E) Early/late unedited $G$. (F) Static late attention (filled) and MLP (open) sums. All error bars are 95\% source-group intervals. The source, order and static comparisons retain their separate operations.}
\label{fig:supp-v7-order}
\end{figure}

Baseline and normalized residual-write contrasts locate the surrounding competition. Llama3.2's baseline strongly favors early facts, while its candidate contribution provides relative late-fact support. The late attention sum contributes negatively to $G$ and its order contrast, opposed by positive MLP writes. Qwen2.5 has a similar static sign pattern with a different pooled removal effect. Static writes describe baseline composition; source removal measures a specified forward intervention (Fig.~\ref{fig:supp-v7-order}).

\begingroup\footnotesize
\setlength{\tabcolsep}{4pt}
\begin{longtable}{lrrr}
\caption{Source-position interventions, fact-order contrasts and baseline write balance. All estimates use 64 paired source groups, with 95\% bootstrap intervals. Single-source removal intervals retain the original projection study; joint-removal, order and write intervals use the saved follow-up analysis of the same source groups.}\label{tab:supp-source-order}\label{tab:sup-source-controls}\label{tab:sup-order}\\
\toprule
Model / contrast & First quantity & Second quantity & Paired contrast \\
\midrule
\endfirsthead
\multicolumn{4}{l}{\emph{Table \thetable\ continued}}\\
\toprule
Model / contrast & First quantity & Second quantity & Paired contrast \\
\midrule
\endhead
\bottomrule\endfoot
\addlinespace
\multicolumn{4}{l}{\emph{Source removal: edited minus clean output (raw margin units)}}\\
Model / operation & $\Delta G$ & $\Delta C$ &  \\
GPT-2 / candidate\_zero & \shortstack{$-0.00218$\\$[-0.00888,+0.00426]$} & \shortstack{$+0.00749$\\$[-0.02166,+0.03550]$} &  \\
GPT-2 / query\_zero & \shortstack{$+0.00135$\\$[+0.00047,+0.00223]$} & \shortstack{$-0.03767$\\$[-0.04024,-0.03515]$} &  \\
Qwen2.5 / candidate\_zero & \shortstack{$+0.08621$\\$[+0.07220,+0.10134]$} & \shortstack{$+0.32187$\\$[+0.25856,+0.39050]$} &  \\
Qwen2.5 / query\_zero & \shortstack{$-0.01671$\\$[-0.02053,-0.01307]$} & \shortstack{$+0.00516$\\$[-0.00092,+0.01111]$} &  \\
Llama3.2 / candidate\_zero & \shortstack{$-0.04516$\\$[-0.08708,-0.00076]$} & \shortstack{$+0.81916$\\$[+0.74770,+0.88346]$} &  \\
Llama3.2 / query\_zero & \shortstack{$-0.00258$\\$[-0.00416,-0.00100]$} & \shortstack{$-0.05142$\\$[-0.05398,-0.04889]$} &  \\
\addlinespace[2pt]
% BEGIN GENERATED ROWS: both_source_rows
GPT-2 / both\_zero & \shortstack{$-0.00190$\\$[-0.00837,\,+0.00409]$} & \shortstack{$-0.02790$\\$[-0.05616,\,+0.00094]$} & \\
Qwen2.5 / both\_zero & \shortstack{$+0.07006$\\$[+0.05596,\,+0.08474]$} & \shortstack{$+0.33322$\\$[+0.27395,\,+0.39409]$} & \\
Llama3.2 / both\_zero & \shortstack{$-0.04796$\\$[-0.09326,\,-0.00270]$} & \shortstack{$+0.77075$\\$[+0.69756,\,+0.83954]$} & \\
% END GENERATED ROWS: both_source_rows
\addlinespace
\multicolumn{4}{l}{\emph{Candidate removal paired by fact order (raw margin units)}}\\
Model & Early $\Delta G$ & Late $\Delta G$ & Late $-$ early \\
GPT-2 & \shortstack{$+0.14977$\\$[+0.12891,+0.17235]$} & \shortstack{$-0.15412$\\$[-0.18323,-0.12664]$} & \shortstack{$-0.30389$\\$[-0.35474,-0.25643]$} \\
Qwen2.5 & \shortstack{$+0.18962$\\$[+0.15833,+0.22310]$} & \shortstack{$-0.01721$\\$[-0.03676,+0.00362]$} & \shortstack{$-0.20683$\\$[-0.25314,-0.16410]$} \\
Llama3.2 & \shortstack{$+0.15397$\\$[+0.12550,+0.18206]$} & \shortstack{$-0.24429$\\$[-0.32741,-0.15891]$} & \shortstack{$-0.39826$\\$[-0.48122,-0.31030]$} \\
\addlinespace
\multicolumn{4}{l}{\emph{Unedited baseline paired by fact order (raw margin units)}}\\
Model & Early $G$ & Late $G$ & Late $-$ early \\
GPT-2 & \shortstack{$-0.04786$\\$[-0.10531,+0.01026]$} & \shortstack{$+0.13580$\\$[+0.06742,+0.19883]$} & \shortstack{$+0.18366$\\$[+0.05778,+0.30151]$} \\
Qwen2.5 & \shortstack{$-0.66513$\\$[-0.83658,-0.50246]$} & \shortstack{$-0.53143$\\$[-0.69257,-0.36840]$} & \shortstack{$+0.13370$\\$[-0.03574,+0.30329]$} \\
Llama3.2 & \shortstack{$-0.17249$\\$[-0.23390,-0.10572]$} & \shortstack{$-1.73944$\\$[-1.79293,-1.68310]$} & \shortstack{$-1.56696$\\$[-1.66287,-1.47040]$} \\
\addlinespace
\multicolumn{4}{l}{\emph{Normalized late-component write sums (raw margin units)}}\\
Model / late range & Component & Write to $G$ & Late $-$ early write \\
GPT-2 / L8--12 & Attention & \shortstack{$+0.02620$\\$[+0.01392,+0.03864]$} & \shortstack{$+0.22150$\\$[+0.12958,+0.31373]$} \\
GPT-2 / L8--12 & MLP & \shortstack{$-0.03141$\\$[-0.03748,-0.02523]$} & \shortstack{$-0.05836$\\$[-0.09441,-0.02479]$} \\
Qwen2.5 / L21--28 & Attention & \shortstack{$-0.75834$\\$[-0.93063,-0.57730]$} & \shortstack{$-0.23821$\\$[-0.46336,-0.02050]$} \\
Qwen2.5 / L21--28 & MLP & \shortstack{$+0.12471$\\$[+0.05181,+0.20861]$} & \shortstack{$+0.25830$\\$[+0.17419,+0.34844]$} \\
Llama3.2 / L12--16 & Attention & \shortstack{$-1.21788$\\$[-1.28985,-1.15004]$} & \shortstack{$-1.94321$\\$[-2.07797,-1.81123]$} \\
Llama3.2 / L12--16 & MLP & \shortstack{$+0.24844$\\$[+0.19870,+0.29935]$} & \shortstack{$+0.43123$\\$[+0.33778,+0.52833]$} \\
\end{longtable}
\endgroup

\FloatBarrier
\section{Source-coordinate transport and hidden-receiver layout}
\label{sup:s7-source}

\subsection{Future components and statistical source projections}
\begin{figure}[!htbp]
\centering
\includegraphics[width=\linewidth,height=.79\textheight,keepaspectratio]{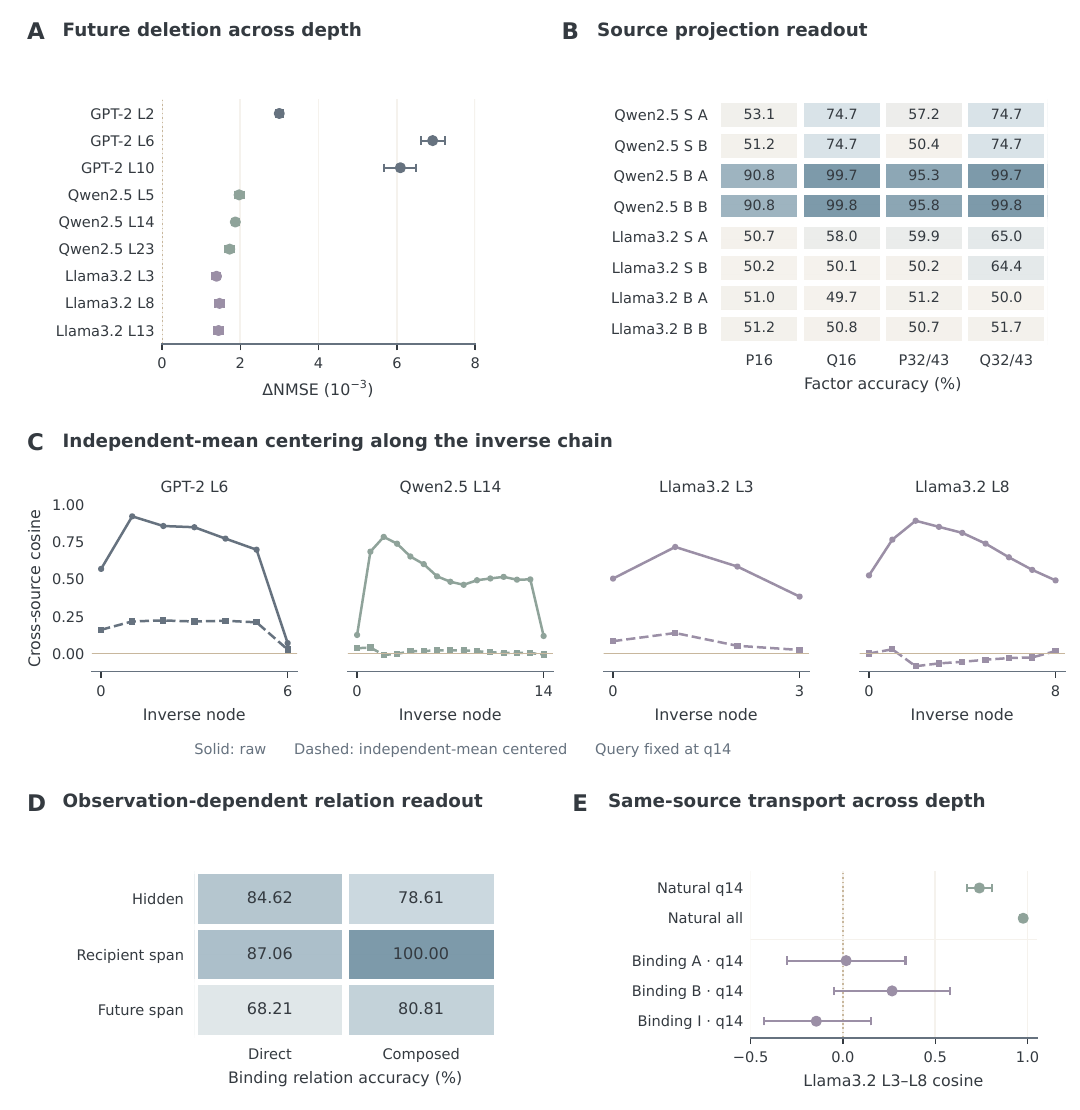}
\caption{Source-coordinate carrying and transport. \textbf{A}, Deletion of q7's estimated source coordinates at $s>7$ from the fixed full-query inverse sum, across three depths per model; intervals are document-bootstrap 95\% intervals. \textbf{B}, Factor accuracies in prefix source spans, using $P_{16}/Q_{16}$ and model-specific higher ranks (32 in Qwen2.5; 43 in Llama3.2). S/B denote syntax/binding. \textbf{C}, Raw and independently centered cross-source cosine at every inverse node, with the query fixed at q14. Node 0 is the input; the largest node index is the selected hidden endpoint. \textbf{D}, Direct and evaluator-composed relation accuracy in matched GPT-2 binding groups. \textbf{E}, Llama3.2 L3--L8 same-window centered cosine, followed by controlled binding-factor difference cosines at q14. Natural vectors and controlled differences use distinct paired comparisons. All intervals are inherited from the corresponding frozen summaries.}
\label{fig:sup-source-transport}
\end{figure}

The source map expresses a selected hidden feature in input coordinates using the instance Jacobian and calibrated data statistics. At non-terminal q7, the complete future component comprises all estimated source coordinates at $s>7$. We subtract this single-query component from the same source's fixed sum of inverses over all queries, then compare NMSE against the same input reference. All other contributions remain fixed. The support distinction between the selected native feature and this source estimate is derived in Supplementary Section~\ref{sup:s2-support}. We averaged repeated calibration-seed measurements within each source before document-bootstrap inference. All nine model-by-depth contrasts were positive (Fig.~\ref{fig:sup-source-transport}A; Table~\ref{tab:sup-source-readouts}). At the main middle-layer budgets, the effects were $+0.006913$, $+0.001873$ and $+0.001470$ for GPT-2, Qwen2.5 and Llama3.2.

The principal projection $P_r$ and complement $Q_r=I-P_r$ were fitted in the source representation's channel space. Their readouts used the original controlled training/development protocol and the complete confirmation set. Known source spans selected the factor-bearing token positions. Figure~\ref{fig:sup-source-transport}B compares ranks 16 and 32 for Qwen2.5, and ranks 16 and 43 for Llama3.2. Qwen2.5 binding remained highly readable in both principal and complementary coordinates. Llama3.2 binding was near chance in these projected readouts, whereas its syntax complement at rank 43 supported $65.03\%/64.38\%$ factor accuracy. Readout depends jointly on task, projection and training coordinates.

\begingroup\footnotesize
\setlength{\tabcolsep}{4pt}
\begin{longtable}{lrr}
\caption{Future-deletion distances and factor readouts in statistical source projections. Deletion subtracts the complete q7 estimate at later source positions from the fixed full-query inverse sum, using the main calibration budgets; projected accuracies use prefix source spans, the original controlled training/development split and the full confirmation set. Brackets are 95\% source-group or document-bootstrap intervals.}\label{tab:sup-source-readouts}\\
\toprule
Model / depth or task & Measurement & Estimate [95\% interval]\\
\midrule
\endfirsthead
\multicolumn{3}{l}{\tablename\ \thetable\ (continued)}\\
\toprule
Model / depth or task & Measurement & Estimate [95\% interval]\\
\midrule
\endhead
\bottomrule\endfoot
GPT-2 L2 & Future deletion & \shortstack{$0.002995$\\$[0.002907,0.003086]$}\\
GPT-2 L6 & Future deletion & \shortstack{$0.006913$\\$[0.006611,0.007233]$}\\
GPT-2 L10 & Future deletion & \shortstack{$0.006089$\\$[0.005672,0.006495]$}\\
Qwen2.5 L5 & Future deletion & \shortstack{$0.001971$\\$[0.001853,0.002097]$}\\
Qwen2.5 L14 & Future deletion & \shortstack{$0.001873$\\$[0.001782,0.001970]$}\\
Qwen2.5 L23 & Future deletion & \shortstack{$0.001728$\\$[0.001622,0.001845]$}\\
Llama3.2 L3 & Future deletion & \shortstack{$0.001393$\\$[0.001283,0.001492]$}\\
Llama3.2 L8 & Future deletion & \shortstack{$0.001470$\\$[0.001351,0.001583]$}\\
Llama3.2 L13 & Future deletion & \shortstack{$0.001444$\\$[0.001335,0.001550]$}\\
Qwen2.5 Syntax & P16, factor A (\%) & \shortstack{$53.11$\\$[50.60,55.36]$}\\
Qwen2.5 Syntax & P16, factor B (\%) & \shortstack{$51.20$\\$[48.45,53.86]$}\\
Qwen2.5 Syntax & Q16, factor A (\%) & \shortstack{$74.75$\\$[72.55,77.05]$}\\
Qwen2.5 Syntax & Q16, factor B (\%) & \shortstack{$74.75$\\$[72.55,77.05]$}\\
Qwen2.5 Syntax & P32, factor A (\%) & \shortstack{$57.16$\\$[54.81,59.32]$}\\
Qwen2.5 Syntax & P32, factor B (\%) & \shortstack{$50.40$\\$[46.94,53.76]$}\\
Qwen2.5 Syntax & Q32, factor A (\%) & \shortstack{$74.75$\\$[72.55,77.05]$}\\
Qwen2.5 Syntax & Q32, factor B (\%) & \shortstack{$74.75$\\$[72.55,77.05]$}\\
Qwen2.5 Binding & P16, factor A (\%) & \shortstack{$90.77$\\$[89.11,92.33]$}\\
Qwen2.5 Binding & P16, factor B (\%) & \shortstack{$90.82$\\$[89.21,92.38]$}\\
Qwen2.5 Binding & Q16, factor A (\%) & \shortstack{$99.71$\\$[99.22,100.00]$}\\
Qwen2.5 Binding & Q16, factor B (\%) & \shortstack{$99.76$\\$[99.32,100.00]$}\\
Qwen2.5 Binding & P32, factor A (\%) & \shortstack{$95.31$\\$[94.09,96.48]$}\\
Qwen2.5 Binding & P32, factor B (\%) & \shortstack{$95.85$\\$[94.63,96.97]$}\\
Qwen2.5 Binding & Q32, factor A (\%) & \shortstack{$99.71$\\$[99.22,100.00]$}\\
Qwen2.5 Binding & Q32, factor B (\%) & \shortstack{$99.76$\\$[99.32,100.00]$}\\
Llama3.2 Syntax & P16, factor A (\%) & \shortstack{$50.65$\\$[50.05,51.30]$}\\
Llama3.2 Syntax & P16, factor B (\%) & \shortstack{$50.15$\\$[49.95,50.45]$}\\
Llama3.2 Syntax & Q16, factor A (\%) & \shortstack{$57.97$\\$[56.41,59.47]$}\\
Llama3.2 Syntax & Q16, factor B (\%) & \shortstack{$50.10$\\$[50.00,50.25]$}\\
Llama3.2 Syntax & P43, factor A (\%) & \shortstack{$59.92$\\$[58.22,61.62]$}\\
Llama3.2 Syntax & P43, factor B (\%) & \shortstack{$50.20$\\$[49.95,50.50]$}\\
Llama3.2 Syntax & Q43, factor A (\%) & \shortstack{$65.03$\\$[63.13,66.93]$}\\
Llama3.2 Syntax & Q43, factor B (\%) & \shortstack{$64.38$\\$[62.63,66.18]$}\\
Llama3.2 Binding & P16, factor A (\%) & \shortstack{$50.98$\\$[50.29,51.76]$}\\
Llama3.2 Binding & P16, factor B (\%) & \shortstack{$51.17$\\$[50.49,51.90]$}\\
Llama3.2 Binding & Q16, factor A (\%) & \shortstack{$49.66$\\$[48.58,50.63]$}\\
Llama3.2 Binding & Q16, factor B (\%) & \shortstack{$50.83$\\$[49.80,51.81]$}\\
Llama3.2 Binding & P43, factor A (\%) & \shortstack{$51.22$\\$[50.54,51.95]$}\\
Llama3.2 Binding & P43, factor B (\%) & \shortstack{$50.68$\\$[50.00,51.42]$}\\
Llama3.2 Binding & Q43, factor A (\%) & \shortstack{$50.00$\\$[49.22,50.78]$}\\
Llama3.2 Binding & Q43, factor B (\%) & \shortstack{$51.71$\\$[50.78,52.64]$}\\
\end{longtable}
\endgroup

\subsection{Independent natural means and controlled differences along the inverse chain}
We followed every adjacent inverse edge with the terminal query and calibrated $\WienerG/D$ maps held fixed. Natural reference documents, documents for inverse-state mean estimation and evaluation documents were mutually exclusive. A separate mean was estimated at each node for the same query. Natural rank-16 PCA in this transport analysis was fitted separately at each node and position; its direct sum over 16 positions can have rank 256. It therefore describes a different projection from the shared channel-space $P_{16}$ above.

At fixed q14 and input node 0, independent mean subtraction reduced cross-source cosine in all four model/depth configurations (Table~\ref{tab:sup-d5-input}; Fig.~\ref{fig:sup-source-transport}C). GPT-2 retained a positive centered mean of $0.1610$ $[0.0874,0.2367]$; the centered intervals of Qwen2.5 L14 and Llama3.2 L3/L8 included zero. This comparison used fixed cyclic mismatches and source-document resampling. It isolates the measured shared background under that pairing.

Same-source transport across depth supplied a separate comparison. For 64 matched Llama3.2 windows, input cosine across L3 and L8 after independent centering was $0.7391$ $[0.6697,0.8063]$ at q14 and $0.9764$ $[0.9603,0.9869]$ for all queries. Controlled binding $A/B/I$ differences at q14 instead had signed cross-depth cosines $0.0175/0.2663/-0.1442$, each with an interval spanning zero, and mean absolute cosines $0.8932/0.9433/0.7831$. In the all-query view their signed cosines were $0.3683/0.5227/0.2864$. Preserved natural-sample directions therefore coexist with factor differences that change sign across groups (Fig.~\ref{fig:sup-source-transport}E).

Local edge controls held the entering inverse state fixed, then either removed that edge's $D$ correction or substituted a same-layout donor Jacobian while retaining $\WienerG/D$. Subsequent edges used the original geometry. At the first inverse edge in the all-query binding view, removing $D$ retained raw-vector/$A$-difference cosines of $0.9824/0.9460$ in GPT-2 and $0.9274/0.9108$ in Qwen2.5; mismatching $J$ gave $0.9496/0.7256$ and $0.8289/0.6891$. This identifies distinct effects of statistical correction and instance geometry on the transported differences. Complete node trajectories and local controls are provided with the figure source data.

A fixed-readout control further isolated a group-level offset. Keeping the original coefficients, global RMS and vectors fixed, we subtracted the unlabeled four-corner group mean from confirmation scores. Llama3.2 L3 binding $A/B$ accuracy rose from $50.78/50.78\%$ to $67.97/66.41\%$, with paired gains of $17.19$ $[6.25,28.13]$ and $15.63$ $[4.69,25.78]$ percentage points. This is a group-supported score-centering operation, with every corner available for its group mean.

\begingroup\footnotesize
\setlength{\tabcolsep}{4pt}
\begin{longtable}{lrrr}
\caption{Natural background across the complete adjacent inverse chain. Raw and centered cross-source cosines use fixed mismatched windows resampled by source document; centering uses independent documents and the same query/node. The last two rows instead pair the same windows across depth, explicitly separating that comparison from cross-source alignment.}\label{tab:sup-d5-input}\\
\toprule
Model / depth & Query / node & Raw cosine & Centered cosine\\
\midrule
\endfirsthead
\multicolumn{4}{l}{\tablename\ \thetable\ (continued)}\\
\toprule
Model / depth & Query / node & Raw cosine & Centered cosine\\
\midrule
\endhead
\bottomrule\endfoot
GPT-2 L6 & all, node 0 & \shortstack{$0.5372$\\$[0.5264,0.5482]$} & \shortstack{$0.0364$\\$[0.0158,0.0577]$}\\
GPT-2 L6 & all, node 1 & \shortstack{$0.9327$\\$[0.9290,0.9364]$} & \shortstack{$0.0465$\\$[-0.0150,0.1055]$}\\
GPT-2 L6 & all, node 2 & \shortstack{$0.9603$\\$[0.9566,0.9638]$} & \shortstack{$0.0606$\\$[-0.0288,0.1434]$}\\
GPT-2 L6 & all, node 3 & \shortstack{$0.9810$\\$[0.9790,0.9828]$} & \shortstack{$0.0771$\\$[-0.0422,0.1945]$}\\
GPT-2 L6 & all, node 4 & \shortstack{$0.9695$\\$[0.9662,0.9727]$} & \shortstack{$0.0683$\\$[-0.0338,0.1692]$}\\
GPT-2 L6 & all, node 5 & \shortstack{$0.9535$\\$[0.9478,0.9588]$} & \shortstack{$0.0721$\\$[-0.0202,0.1642]$}\\
GPT-2 L6 & all, node 6 & \shortstack{$0.0931$\\$[0.0831,0.1033]$} & \shortstack{$0.0264$\\$[0.0159,0.0366]$}\\
GPT-2 L6 & q0, node 0 & \shortstack{$0.2410$\\$[0.1929,0.2903]$} & \shortstack{$0.0847$\\$[0.0327,0.1388]$}\\
GPT-2 L6 & q0, node 1 & \shortstack{$0.4230$\\$[0.3400,0.5039]$} & \shortstack{$0.0680$\\$[-0.0302,0.1652]$}\\
GPT-2 L6 & q0, node 2 & \shortstack{$0.3370$\\$[0.2695,0.4038]$} & \shortstack{$0.0560$\\$[-0.0167,0.1321]$}\\
GPT-2 L6 & q0, node 3 & \shortstack{$0.4794$\\$[0.4127,0.5454]$} & \shortstack{$0.0396$\\$[-0.0532,0.1337]$}\\
GPT-2 L6 & q0, node 4 & \shortstack{$0.5531$\\$[0.4965,0.6061]$} & \shortstack{$0.0526$\\$[-0.0378,0.1405]$}\\
GPT-2 L6 & q0, node 5 & \shortstack{$0.6736$\\$[0.6134,0.7325]$} & \shortstack{$0.0949$\\$[-0.0265,0.2160]$}\\
GPT-2 L6 & q0, node 6 & \shortstack{$0.9952$\\$[0.9948,0.9955]$} & \shortstack{$0.0237$\\$[-0.0098,0.0565]$}\\
GPT-2 L6 & q1, node 0 & \shortstack{$0.3010$\\$[0.2128,0.3893]$} & \shortstack{$0.0861$\\$[0.0104,0.1587]$}\\
GPT-2 L6 & q1, node 1 & \shortstack{$0.5058$\\$[0.3623,0.6492]$} & \shortstack{$0.0605$\\$[-0.0782,0.1927]$}\\
GPT-2 L6 & q1, node 2 & \shortstack{$0.4649$\\$[0.3356,0.5980]$} & \shortstack{$0.0652$\\$[-0.0564,0.1770]$}\\
GPT-2 L6 & q1, node 3 & \shortstack{$0.4716$\\$[0.3328,0.6106]$} & \shortstack{$0.0491$\\$[-0.0791,0.1671]$}\\
GPT-2 L6 & q1, node 4 & \shortstack{$0.4338$\\$[0.3047,0.5657]$} & \shortstack{$0.0538$\\$[-0.0587,0.1596]$}\\
GPT-2 L6 & q1, node 5 & \shortstack{$0.4097$\\$[0.2829,0.5428]$} & \shortstack{$0.0420$\\$[-0.0761,0.1516]$}\\
GPT-2 L6 & q1, node 6 & \shortstack{$0.7247$\\$[0.6900,0.7547]$} & \shortstack{$0.0319$\\$[0.0044,0.0607]$}\\
GPT-2 L6 & q14, node 0 & \shortstack{$0.5705$\\$[0.5593,0.5815]$} & \shortstack{$0.1610$\\$[0.0874,0.2367]$}\\
GPT-2 L6 & q14, node 1 & \shortstack{$0.9241$\\$[0.9153,0.9323]$} & \shortstack{$0.2174$\\$[0.0573,0.3777]$}\\
GPT-2 L6 & q14, node 2 & \shortstack{$0.8591$\\$[0.8410,0.8771]$} & \shortstack{$0.2238$\\$[0.0816,0.3679]$}\\
GPT-2 L6 & q14, node 3 & \shortstack{$0.8508$\\$[0.8244,0.8733]$} & \shortstack{$0.2167$\\$[0.0698,0.3659]$}\\
GPT-2 L6 & q14, node 4 & \shortstack{$0.7741$\\$[0.7368,0.8041]$} & \shortstack{$0.2217$\\$[0.0874,0.3550]$}\\
GPT-2 L6 & q14, node 5 & \shortstack{$0.6996$\\$[0.6469,0.7442]$} & \shortstack{$0.2111$\\$[0.0803,0.3409]$}\\
GPT-2 L6 & q14, node 6 & \shortstack{$0.0710$\\$[0.0471,0.0958]$} & \shortstack{$0.0279$\\$[0.0041,0.0536]$}\\
GPT-2 L6 & q15, node 0 & \shortstack{$0.5778$\\$[0.5560,0.5971]$} & \shortstack{$0.0338$\\$[-0.0155,0.0810]$}\\
GPT-2 L6 & q15, node 1 & \shortstack{$0.9217$\\$[0.8880,0.9466]$} & \shortstack{$0.0086$\\$[-0.1238,0.1348]$}\\
GPT-2 L6 & q15, node 2 & \shortstack{$0.8755$\\$[0.8300,0.9093]$} & \shortstack{$0.0229$\\$[-0.0904,0.1318]$}\\
GPT-2 L6 & q15, node 3 & \shortstack{$0.8814$\\$[0.8337,0.9184]$} & \shortstack{$0.0336$\\$[-0.0898,0.1499]$}\\
GPT-2 L6 & q15, node 4 & \shortstack{$0.8355$\\$[0.7838,0.8776]$} & \shortstack{$0.0510$\\$[-0.0620,0.1600]$}\\
GPT-2 L6 & q15, node 5 & \shortstack{$0.7991$\\$[0.7411,0.8488]$} & \shortstack{$0.0686$\\$[-0.0556,0.1899]$}\\
GPT-2 L6 & q15, node 6 & \shortstack{$0.0781$\\$[0.0538,0.1036]$} & \shortstack{$0.0423$\\$[0.0141,0.0705]$}\\
GPT-2 L6 & q3, node 0 & \shortstack{$0.5411$\\$[0.5257,0.5556]$} & \shortstack{$0.0018$\\$[-0.0697,0.0817]$}\\
GPT-2 L6 & q3, node 1 & \shortstack{$0.8895$\\$[0.8681,0.9081]$} & \shortstack{$-0.1106$\\$[-0.2640,0.0485]$}\\
GPT-2 L6 & q3, node 2 & \shortstack{$0.8183$\\$[0.7794,0.8512]$} & \shortstack{$-0.0988$\\$[-0.2389,0.0462]$}\\
GPT-2 L6 & q3, node 3 & \shortstack{$0.8268$\\$[0.7859,0.8623]$} & \shortstack{$-0.0862$\\$[-0.2255,0.0603]$}\\
GPT-2 L6 & q3, node 4 & \shortstack{$0.7535$\\$[0.7045,0.7950]$} & \shortstack{$-0.0702$\\$[-0.1942,0.0615]$}\\
GPT-2 L6 & q3, node 5 & \shortstack{$0.6989$\\$[0.6417,0.7477]$} & \shortstack{$-0.0546$\\$[-0.1752,0.0733]$}\\
GPT-2 L6 & q3, node 6 & \shortstack{$0.1199$\\$[0.0926,0.1487]$} & \shortstack{$0.0242$\\$[0.0028,0.0463]$}\\
GPT-2 L6 & q4, node 0 & \shortstack{$0.4878$\\$[0.4689,0.5067]$} & \shortstack{$0.1529$\\$[0.0866,0.2207]$}\\
GPT-2 L6 & q4, node 1 & \shortstack{$0.8450$\\$[0.8218,0.8668]$} & \shortstack{$0.2550$\\$[0.1152,0.3879]$}\\
GPT-2 L6 & q4, node 2 & \shortstack{$0.7443$\\$[0.7040,0.7830]$} & \shortstack{$0.2239$\\$[0.0987,0.3399]$}\\
GPT-2 L6 & q4, node 3 & \shortstack{$0.7472$\\$[0.6971,0.7914]$} & \shortstack{$0.2482$\\$[0.1213,0.3689]$}\\
GPT-2 L6 & q4, node 4 & \shortstack{$0.6392$\\$[0.5768,0.6981]$} & \shortstack{$0.2269$\\$[0.1144,0.3391]$}\\
GPT-2 L6 & q4, node 5 & \shortstack{$0.5486$\\$[0.4729,0.6227]$} & \shortstack{$0.2335$\\$[0.1275,0.3480]$}\\
GPT-2 L6 & q4, node 6 & \shortstack{$0.1062$\\$[0.0773,0.1346]$} & \shortstack{$0.0627$\\$[0.0346,0.0907]$}\\
GPT-2 L6 & q5, node 0 & \shortstack{$0.4426$\\$[0.4180,0.4655]$} & \shortstack{$0.0569$\\$[0.0046,0.1144]$}\\
GPT-2 L6 & q5, node 1 & \shortstack{$0.7897$\\$[0.7580,0.8185]$} & \shortstack{$0.0472$\\$[-0.0527,0.1490]$}\\
GPT-2 L6 & q5, node 2 & \shortstack{$0.6710$\\$[0.6284,0.7100]$} & \shortstack{$0.0394$\\$[-0.0438,0.1255]$}\\
GPT-2 L6 & q5, node 3 & \shortstack{$0.6919$\\$[0.6400,0.7382]$} & \shortstack{$0.0355$\\$[-0.0607,0.1388]$}\\
GPT-2 L6 & q5, node 4 & \shortstack{$0.5959$\\$[0.5385,0.6460]$} & \shortstack{$0.0486$\\$[-0.0399,0.1398]$}\\
GPT-2 L6 & q5, node 5 & \shortstack{$0.5358$\\$[0.4690,0.5957]$} & \shortstack{$0.0586$\\$[-0.0358,0.1533]$}\\
GPT-2 L6 & q5, node 6 & \shortstack{$0.0742$\\$[0.0511,0.0994]$} & \shortstack{$0.0210$\\$[-0.0035,0.0465]$}\\
GPT-2 L6 & q6, node 0 & \shortstack{$0.4219$\\$[0.3984,0.4439]$} & \shortstack{$0.0741$\\$[0.0236,0.1250]$}\\
GPT-2 L6 & q6, node 1 & \shortstack{$0.7501$\\$[0.7197,0.7774]$} & \shortstack{$0.0915$\\$[-0.0150,0.1927]$}\\
GPT-2 L6 & q6, node 2 & \shortstack{$0.5897$\\$[0.5484,0.6296]$} & \shortstack{$0.0735$\\$[-0.0082,0.1553]$}\\
GPT-2 L6 & q6, node 3 & \shortstack{$0.6002$\\$[0.5512,0.6474]$} & \shortstack{$0.0852$\\$[-0.0023,0.1727]$}\\
GPT-2 L6 & q6, node 4 & \shortstack{$0.4799$\\$[0.4199,0.5357]$} & \shortstack{$0.0695$\\$[-0.0041,0.1501]$}\\
GPT-2 L6 & q6, node 5 & \shortstack{$0.3968$\\$[0.3174,0.4731]$} & \shortstack{$0.0582$\\$[-0.0241,0.1467]$}\\
GPT-2 L6 & q6, node 6 & \shortstack{$0.0614$\\$[0.0360,0.0874]$} & \shortstack{$0.0139$\\$[-0.0141,0.0421]$}\\
GPT-2 L6 & q7, node 0 & \shortstack{$0.4069$\\$[0.3795,0.4342]$} & \shortstack{$0.0322$\\$[-0.0037,0.0676]$}\\
GPT-2 L6 & q7, node 1 & \shortstack{$0.7214$\\$[0.6830,0.7570]$} & \shortstack{$0.0514$\\$[-0.0323,0.1357]$}\\
GPT-2 L6 & q7, node 2 & \shortstack{$0.5543$\\$[0.5085,0.6041]$} & \shortstack{$0.0428$\\$[-0.0246,0.1119]$}\\
GPT-2 L6 & q7, node 3 & \shortstack{$0.5700$\\$[0.5230,0.6190]$} & \shortstack{$0.0450$\\$[-0.0309,0.1226]$}\\
GPT-2 L6 & q7, node 4 & \shortstack{$0.4683$\\$[0.4175,0.5224]$} & \shortstack{$0.0518$\\$[-0.0233,0.1256]$}\\
GPT-2 L6 & q7, node 5 & \shortstack{$0.3969$\\$[0.3337,0.4629]$} & \shortstack{$0.0462$\\$[-0.0413,0.1335]$}\\
GPT-2 L6 & q7, node 6 & \shortstack{$0.0617$\\$[0.0384,0.0849]$} & \shortstack{$0.0119$\\$[-0.0112,0.0354]$}\\
Qwen2.5 L14 & all, node 0 & \shortstack{$0.1261$\\$[0.0904,0.1631]$} & \shortstack{$0.0281$\\$[-0.0138,0.0714]$}\\
Qwen2.5 L14 & all, node 1 & \shortstack{$0.6865$\\$[0.6609,0.7117]$} & \shortstack{$0.0225$\\$[-0.0116,0.0582]$}\\
Qwen2.5 L14 & all, node 2 & \shortstack{$0.8134$\\$[0.7198,0.9061]$} & \shortstack{$0.2847$\\$[0.1419,0.4166]$}\\
Qwen2.5 L14 & all, node 3 & \shortstack{$0.8140$\\$[0.7204,0.9060]$} & \shortstack{$0.2523$\\$[0.1295,0.3661]$}\\
Qwen2.5 L14 & all, node 4 & \shortstack{$0.8139$\\$[0.7203,0.9055]$} & \shortstack{$0.2083$\\$[0.0991,0.3132]$}\\
Qwen2.5 L14 & all, node 5 & \shortstack{$0.8143$\\$[0.7207,0.9051]$} & \shortstack{$0.1735$\\$[0.0698,0.2721]$}\\
Qwen2.5 L14 & all, node 6 & \shortstack{$0.8153$\\$[0.7222,0.9048]$} & \shortstack{$0.1358$\\$[0.0405,0.2285]$}\\
Qwen2.5 L14 & all, node 7 & \shortstack{$0.8150$\\$[0.7221,0.9043]$} & \shortstack{$0.1198$\\$[0.0302,0.2083]$}\\
Qwen2.5 L14 & all, node 8 & \shortstack{$0.8150$\\$[0.7225,0.9042]$} & \shortstack{$0.1125$\\$[0.0243,0.1981]$}\\
Qwen2.5 L14 & all, node 9 & \shortstack{$0.8153$\\$[0.7227,0.9045]$} & \shortstack{$0.1182$\\$[0.0264,0.2079]$}\\
Qwen2.5 L14 & all, node 10 & \shortstack{$0.8156$\\$[0.7231,0.9046]$} & \shortstack{$0.1186$\\$[0.0256,0.2098]$}\\
Qwen2.5 L14 & all, node 11 & \shortstack{$0.8156$\\$[0.7231,0.9047]$} & \shortstack{$0.1203$\\$[0.0266,0.2126]$}\\
Qwen2.5 L14 & all, node 12 & \shortstack{$0.8154$\\$[0.7229,0.9044]$} & \shortstack{$0.1136$\\$[0.0221,0.2046]$}\\
Qwen2.5 L14 & all, node 13 & \shortstack{$0.8154$\\$[0.7229,0.9044]$} & \shortstack{$0.1131$\\$[0.0209,0.2048]$}\\
Qwen2.5 L14 & all, node 14 & \shortstack{$0.2912$\\$[0.2741,0.3079]$} & \shortstack{$0.0154$\\$[-0.0053,0.0376]$}\\
Qwen2.5 L14 & q0, node 0 & \shortstack{$0.0971$\\$[0.0555,0.1386]$} & \shortstack{$0.6845$\\$[0.5551,0.8023]$}\\
Qwen2.5 L14 & q0, node 1 & \shortstack{$0.5270$\\$[0.4121,0.6312]$} & \shortstack{$0.6838$\\$[0.5436,0.8087]$}\\
Qwen2.5 L14 & q0, node 2 & \shortstack{$0.8264$\\$[0.7342,0.9088]$} & \shortstack{$0.7928$\\$[0.6885,0.8885]$}\\
Qwen2.5 L14 & q0, node 3 & \shortstack{$0.8187$\\$[0.7390,0.8906]$} & \shortstack{$0.7762$\\$[0.6751,0.8703]$}\\
Qwen2.5 L14 & q0, node 4 & \shortstack{$0.7913$\\$[0.7136,0.8622]$} & \shortstack{$0.7606$\\$[0.6603,0.8550]$}\\
Qwen2.5 L14 & q0, node 5 & \shortstack{$0.7790$\\$[0.7021,0.8503]$} & \shortstack{$0.7508$\\$[0.6494,0.8452]$}\\
Qwen2.5 L14 & q0, node 6 & \shortstack{$0.7462$\\$[0.6736,0.8131]$} & \shortstack{$0.7267$\\$[0.6269,0.8192]$}\\
Qwen2.5 L14 & q0, node 7 & \shortstack{$0.7313$\\$[0.6600,0.7976]$} & \shortstack{$0.7151$\\$[0.6145,0.8075]$}\\
Qwen2.5 L14 & q0, node 8 & \shortstack{$0.7279$\\$[0.6549,0.7943]$} & \shortstack{$0.7102$\\$[0.6091,0.8031]$}\\
Qwen2.5 L14 & q0, node 9 & \shortstack{$0.7674$\\$[0.6907,0.8371]$} & \shortstack{$0.7335$\\$[0.6281,0.8288]$}\\
Qwen2.5 L14 & q0, node 10 & \shortstack{$0.7857$\\$[0.7070,0.8577]$} & \shortstack{$0.7439$\\$[0.6373,0.8413]$}\\
Qwen2.5 L14 & q0, node 11 & \shortstack{$0.7987$\\$[0.7189,0.8732]$} & \shortstack{$0.7517$\\$[0.6449,0.8510]$}\\
Qwen2.5 L14 & q0, node 12 & \shortstack{$0.7945$\\$[0.7145,0.8692]$} & \shortstack{$0.7477$\\$[0.6404,0.8467]$}\\
Qwen2.5 L14 & q0, node 13 & \shortstack{$0.8023$\\$[0.7222,0.8780]$} & \shortstack{$0.7516$\\$[0.6432,0.8511]$}\\
Qwen2.5 L14 & q0, node 14 & \shortstack{$0.8843$\\$[0.8166,0.9435]$} & \shortstack{$0.4515$\\$[0.3071,0.5881]$}\\
Qwen2.5 L14 & q1, node 0 & \shortstack{$0.1061$\\$[0.0673,0.1445]$} & \shortstack{$0.0314$\\$[-0.0104,0.0734]$}\\
Qwen2.5 L14 & q1, node 1 & \shortstack{$0.6126$\\$[0.5270,0.6844]$} & \shortstack{$0.0100$\\$[-0.0465,0.0679]$}\\
Qwen2.5 L14 & q1, node 2 & \shortstack{$0.7977$\\$[0.6967,0.8905]$} & \shortstack{$0.1188$\\$[-0.0765,0.2983]$}\\
Qwen2.5 L14 & q1, node 3 & \shortstack{$0.7749$\\$[0.6647,0.8718]$} & \shortstack{$0.1074$\\$[-0.0532,0.2567]$}\\
Qwen2.5 L14 & q1, node 4 & \shortstack{$0.7617$\\$[0.6553,0.8553]$} & \shortstack{$0.0932$\\$[-0.0434,0.2252]$}\\
Qwen2.5 L14 & q1, node 5 & \shortstack{$0.7472$\\$[0.6420,0.8405]$} & \shortstack{$0.0884$\\$[-0.0456,0.2159]$}\\
Qwen2.5 L14 & q1, node 6 & \shortstack{$0.7238$\\$[0.6198,0.8177]$} & \shortstack{$0.0780$\\$[-0.0430,0.1910]$}\\
Qwen2.5 L14 & q1, node 7 & \shortstack{$0.7122$\\$[0.6096,0.8045]$} & \shortstack{$0.0724$\\$[-0.0418,0.1817]$}\\
Qwen2.5 L14 & q1, node 8 & \shortstack{$0.7060$\\$[0.6027,0.7983]$} & \shortstack{$0.0764$\\$[-0.0373,0.1838]$}\\
Qwen2.5 L14 & q1, node 9 & \shortstack{$0.7131$\\$[0.6085,0.8072]$} & \shortstack{$0.0838$\\$[-0.0406,0.2000]$}\\
Qwen2.5 L14 & q1, node 10 & \shortstack{$0.7156$\\$[0.6100,0.8113]$} & \shortstack{$0.0866$\\$[-0.0441,0.2081]$}\\
Qwen2.5 L14 & q1, node 11 & \shortstack{$0.7176$\\$[0.6098,0.8144]$} & \shortstack{$0.0851$\\$[-0.0495,0.2093]$}\\
Qwen2.5 L14 & q1, node 12 & \shortstack{$0.7127$\\$[0.6053,0.8086]$} & \shortstack{$0.0849$\\$[-0.0480,0.2077]$}\\
Qwen2.5 L14 & q1, node 13 & \shortstack{$0.7116$\\$[0.6040,0.8079]$} & \shortstack{$0.0852$\\$[-0.0493,0.2096]$}\\
Qwen2.5 L14 & q1, node 14 & \shortstack{$0.8395$\\$[0.7782,0.8934]$} & \shortstack{$0.0301$\\$[-0.0378,0.0985]$}\\
Qwen2.5 L14 & q14, node 0 & \shortstack{$0.1261$\\$[0.0904,0.1631]$} & \shortstack{$0.0368$\\$[-0.0064,0.0806]$}\\
Qwen2.5 L14 & q14, node 1 & \shortstack{$0.6866$\\$[0.6611,0.7118]$} & \shortstack{$0.0416$\\$[-0.0209,0.1033]$}\\
Qwen2.5 L14 & q14, node 2 & \shortstack{$0.7859$\\$[0.7102,0.8537]$} & \shortstack{$-0.0082$\\$[-0.1346,0.1254]$}\\
Qwen2.5 L14 & q14, node 3 & \shortstack{$0.7400$\\$[0.6765,0.7986]$} & \shortstack{$-0.0009$\\$[-0.0992,0.0977]$}\\
Qwen2.5 L14 & q14, node 4 & \shortstack{$0.6540$\\$[0.5945,0.7114]$} & \shortstack{$0.0160$\\$[-0.0593,0.0911]$}\\
Qwen2.5 L14 & q14, node 5 & \shortstack{$0.6026$\\$[0.5466,0.6574]$} & \shortstack{$0.0178$\\$[-0.0511,0.0859]$}\\
Qwen2.5 L14 & q14, node 6 & \shortstack{$0.5207$\\$[0.4666,0.5743]$} & \shortstack{$0.0231$\\$[-0.0349,0.0821]$}\\
Qwen2.5 L14 & q14, node 7 & \shortstack{$0.4840$\\$[0.4287,0.5382]$} & \shortstack{$0.0229$\\$[-0.0347,0.0825]$}\\
Qwen2.5 L14 & q14, node 8 & \shortstack{$0.4636$\\$[0.4059,0.5199]$} & \shortstack{$0.0231$\\$[-0.0361,0.0842]$}\\
Qwen2.5 L14 & q14, node 9 & \shortstack{$0.4942$\\$[0.4304,0.5544]$} & \shortstack{$0.0162$\\$[-0.0567,0.0917]$}\\
Qwen2.5 L14 & q14, node 10 & \shortstack{$0.5062$\\$[0.4395,0.5692]$} & \shortstack{$0.0109$\\$[-0.0707,0.0942]$}\\
Qwen2.5 L14 & q14, node 11 & \shortstack{$0.5169$\\$[0.4492,0.5822]$} & \shortstack{$0.0047$\\$[-0.0851,0.0958]$}\\
Qwen2.5 L14 & q14, node 12 & \shortstack{$0.4979$\\$[0.4282,0.5650]$} & \shortstack{$0.0063$\\$[-0.0846,0.0977]$}\\
Qwen2.5 L14 & q14, node 13 & \shortstack{$0.5000$\\$[0.4286,0.5684]$} & \shortstack{$0.0044$\\$[-0.0909,0.0995]$}\\
Qwen2.5 L14 & q14, node 14 & \shortstack{$0.1193$\\$[0.0905,0.1477]$} & \shortstack{$-0.0051$\\$[-0.0330,0.0226]$}\\
Qwen2.5 L14 & q15, node 0 & \shortstack{$0.0951$\\$[0.0549,0.1356]$} & \shortstack{$0.0420$\\$[-0.0021,0.0893]$}\\
Qwen2.5 L14 & q15, node 1 & \shortstack{$0.6034$\\$[0.5115,0.6787]$} & \shortstack{$0.0464$\\$[-0.0289,0.1219]$}\\
Qwen2.5 L14 & q15, node 2 & \shortstack{$0.6940$\\$[0.5739,0.8052]$} & \shortstack{$0.1072$\\$[-0.0444,0.2712]$}\\
Qwen2.5 L14 & q15, node 3 & \shortstack{$0.5978$\\$[0.4842,0.7072]$} & \shortstack{$0.0480$\\$[-0.0470,0.1467]$}\\
Qwen2.5 L14 & q15, node 4 & \shortstack{$0.5376$\\$[0.4402,0.6330]$} & \shortstack{$0.0587$\\$[-0.0153,0.1333]$}\\
Qwen2.5 L14 & q15, node 5 & \shortstack{$0.5019$\\$[0.4040,0.5977]$} & \shortstack{$0.0548$\\$[-0.0178,0.1286]$}\\
Qwen2.5 L14 & q15, node 6 & \shortstack{$0.4636$\\$[0.3745,0.5486]$} & \shortstack{$0.0586$\\$[-0.0030,0.1231]$}\\
Qwen2.5 L14 & q15, node 7 & \shortstack{$0.4600$\\$[0.3748,0.5407]$} & \shortstack{$0.0501$\\$[-0.0089,0.1117]$}\\
Qwen2.5 L14 & q15, node 8 & \shortstack{$0.4773$\\$[0.3961,0.5564]$} & \shortstack{$0.0521$\\$[-0.0085,0.1170]$}\\
Qwen2.5 L14 & q15, node 9 & \shortstack{$0.5545$\\$[0.4661,0.6399]$} & \shortstack{$0.0478$\\$[-0.0301,0.1289]$}\\
Qwen2.5 L14 & q15, node 10 & \shortstack{$0.5918$\\$[0.5002,0.6805]$} & \shortstack{$0.0531$\\$[-0.0336,0.1433]$}\\
Qwen2.5 L14 & q15, node 11 & \shortstack{$0.6200$\\$[0.5250,0.7124]$} & \shortstack{$0.0534$\\$[-0.0481,0.1589]$}\\
Qwen2.5 L14 & q15, node 12 & \shortstack{$0.6236$\\$[0.5317,0.7133]$} & \shortstack{$0.0529$\\$[-0.0487,0.1581]$}\\
Qwen2.5 L14 & q15, node 13 & \shortstack{$0.6429$\\$[0.5507,0.7328]$} & \shortstack{$0.0540$\\$[-0.0607,0.1708]$}\\
Qwen2.5 L14 & q15, node 14 & \shortstack{$0.1400$\\$[0.1132,0.1667]$} & \shortstack{$0.0181$\\$[-0.0107,0.0470]$}\\
Qwen2.5 L14 & q3, node 0 & \shortstack{$0.0957$\\$[0.0556,0.1330]$} & \shortstack{$0.0291$\\$[-0.0120,0.0655]$}\\
Qwen2.5 L14 & q3, node 1 & \shortstack{$0.6218$\\$[0.5396,0.6875]$} & \shortstack{$0.0281$\\$[-0.0194,0.0741]$}\\
Qwen2.5 L14 & q3, node 2 & \shortstack{$0.7885$\\$[0.6918,0.8751]$} & \shortstack{$-0.0148$\\$[-0.1461,0.1223]$}\\
Qwen2.5 L14 & q3, node 3 & \shortstack{$0.7710$\\$[0.6796,0.8520]$} & \shortstack{$-0.0051$\\$[-0.0991,0.0899]$}\\
Qwen2.5 L14 & q3, node 4 & \shortstack{$0.7271$\\$[0.6401,0.8060]$} & \shortstack{$0.0015$\\$[-0.0726,0.0749]$}\\
Qwen2.5 L14 & q3, node 5 & \shortstack{$0.6909$\\$[0.6059,0.7669]$} & \shortstack{$-0.0008$\\$[-0.0717,0.0674]$}\\
Qwen2.5 L14 & q3, node 6 & \shortstack{$0.6374$\\$[0.5587,0.7089]$} & \shortstack{$0.0022$\\$[-0.0601,0.0613]$}\\
Qwen2.5 L14 & q3, node 7 & \shortstack{$0.6042$\\$[0.5280,0.6738]$} & \shortstack{$0.0017$\\$[-0.0594,0.0576]$}\\
Qwen2.5 L14 & q3, node 8 & \shortstack{$0.5867$\\$[0.5100,0.6561]$} & \shortstack{$-0.0019$\\$[-0.0651,0.0562]$}\\
Qwen2.5 L14 & q3, node 9 & \shortstack{$0.5986$\\$[0.5199,0.6699]$} & \shortstack{$-0.0114$\\$[-0.0853,0.0542]$}\\
Qwen2.5 L14 & q3, node 10 & \shortstack{$0.6045$\\$[0.5250,0.6768]$} & \shortstack{$-0.0160$\\$[-0.0947,0.0547]$}\\
Qwen2.5 L14 & q3, node 11 & \shortstack{$0.6083$\\$[0.5280,0.6818]$} & \shortstack{$-0.0200$\\$[-0.1014,0.0535]$}\\
Qwen2.5 L14 & q3, node 12 & \shortstack{$0.5921$\\$[0.5121,0.6652]$} & \shortstack{$-0.0215$\\$[-0.1025,0.0524]$}\\
Qwen2.5 L14 & q3, node 13 & \shortstack{$0.5893$\\$[0.5091,0.6624]$} & \shortstack{$-0.0225$\\$[-0.1052,0.0533]$}\\
Qwen2.5 L14 & q3, node 14 & \shortstack{$0.3393$\\$[0.3071,0.3743]$} & \shortstack{$0.0257$\\$[-0.0222,0.0742]$}\\
Qwen2.5 L14 & q4, node 0 & \shortstack{$0.1001$\\$[0.0604,0.1402]$} & \shortstack{$0.0353$\\$[-0.0087,0.0799]$}\\
Qwen2.5 L14 & q4, node 1 & \shortstack{$0.6518$\\$[0.5875,0.7006]$} & \shortstack{$0.0632$\\$[0.0201,0.1113]$}\\
Qwen2.5 L14 & q4, node 2 & \shortstack{$0.7634$\\$[0.6764,0.8425]$} & \shortstack{$0.2077$\\$[0.1045,0.3088]$}\\
Qwen2.5 L14 & q4, node 3 & \shortstack{$0.7171$\\$[0.6371,0.7911]$} & \shortstack{$0.1416$\\$[0.0692,0.2171]$}\\
Qwen2.5 L14 & q4, node 4 & \shortstack{$0.6414$\\$[0.5691,0.7116]$} & \shortstack{$0.1211$\\$[0.0662,0.1797]$}\\
Qwen2.5 L14 & q4, node 5 & \shortstack{$0.5814$\\$[0.5147,0.6467]$} & \shortstack{$0.1160$\\$[0.0648,0.1697]$}\\
Qwen2.5 L14 & q4, node 6 & \shortstack{$0.5042$\\$[0.4402,0.5675]$} & \shortstack{$0.1104$\\$[0.0633,0.1608]$}\\
Qwen2.5 L14 & q4, node 7 & \shortstack{$0.4637$\\$[0.4017,0.5258]$} & \shortstack{$0.1086$\\$[0.0618,0.1600]$}\\
Qwen2.5 L14 & q4, node 8 & \shortstack{$0.4440$\\$[0.3816,0.5058]$} & \shortstack{$0.1066$\\$[0.0565,0.1624]$}\\
Qwen2.5 L14 & q4, node 9 & \shortstack{$0.4626$\\$[0.3967,0.5280]$} & \shortstack{$0.1177$\\$[0.0572,0.1833]$}\\
Qwen2.5 L14 & q4, node 10 & \shortstack{$0.4725$\\$[0.4046,0.5397]$} & \shortstack{$0.1262$\\$[0.0604,0.1974]$}\\
Qwen2.5 L14 & q4, node 11 & \shortstack{$0.4776$\\$[0.4086,0.5457]$} & \shortstack{$0.1284$\\$[0.0584,0.2031]$}\\
Qwen2.5 L14 & q4, node 12 & \shortstack{$0.4609$\\$[0.3921,0.5289]$} & \shortstack{$0.1285$\\$[0.0578,0.2031]$}\\
Qwen2.5 L14 & q4, node 13 & \shortstack{$0.4590$\\$[0.3894,0.5275]$} & \shortstack{$0.1309$\\$[0.0579,0.2073]$}\\
Qwen2.5 L14 & q4, node 14 & \shortstack{$0.2914$\\$[0.2644,0.3209]$} & \shortstack{$0.0329$\\$[-0.0035,0.0703]$}\\
Qwen2.5 L14 & q5, node 0 & \shortstack{$0.1261$\\$[0.0905,0.1631]$} & \shortstack{$0.0222$\\$[-0.0225,0.0677]$}\\
Qwen2.5 L14 & q5, node 1 & \shortstack{$0.6865$\\$[0.6609,0.7118]$} & \shortstack{$0.0362$\\$[-0.0125,0.0870]$}\\
Qwen2.5 L14 & q5, node 2 & \shortstack{$0.7667$\\$[0.6830,0.8422]$} & \shortstack{$0.1296$\\$[0.0306,0.2360]$}\\
Qwen2.5 L14 & q5, node 3 & \shortstack{$0.6999$\\$[0.6221,0.7704]$} & \shortstack{$0.0768$\\$[0.0121,0.1448]$}\\
Qwen2.5 L14 & q5, node 4 & \shortstack{$0.6058$\\$[0.5385,0.6691]$} & \shortstack{$0.0549$\\$[0.0070,0.1050]$}\\
Qwen2.5 L14 & q5, node 5 & \shortstack{$0.5358$\\$[0.4761,0.5957]$} & \shortstack{$0.0556$\\$[0.0068,0.1066]$}\\
Qwen2.5 L14 & q5, node 6 & \shortstack{$0.4510$\\$[0.3972,0.5060]$} & \shortstack{$0.0569$\\$[0.0070,0.1088]$}\\
Qwen2.5 L14 & q5, node 7 & \shortstack{$0.4102$\\$[0.3588,0.4654]$} & \shortstack{$0.0574$\\$[0.0054,0.1116]$}\\
Qwen2.5 L14 & q5, node 8 & \shortstack{$0.3924$\\$[0.3394,0.4495]$} & \shortstack{$0.0563$\\$[-0.0010,0.1158]$}\\
Qwen2.5 L14 & q5, node 9 & \shortstack{$0.4191$\\$[0.3605,0.4809]$} & \shortstack{$0.0661$\\$[-0.0042,0.1381]$}\\
Qwen2.5 L14 & q5, node 10 & \shortstack{$0.4304$\\$[0.3683,0.4962]$} & \shortstack{$0.0701$\\$[-0.0071,0.1486]$}\\
Qwen2.5 L14 & q5, node 11 & \shortstack{$0.4376$\\$[0.3735,0.5061]$} & \shortstack{$0.0734$\\$[-0.0080,0.1568]$}\\
Qwen2.5 L14 & q5, node 12 & \shortstack{$0.4193$\\$[0.3538,0.4886]$} & \shortstack{$0.0731$\\$[-0.0092,0.1575]$}\\
Qwen2.5 L14 & q5, node 13 & \shortstack{$0.4195$\\$[0.3517,0.4898]$} & \shortstack{$0.0754$\\$[-0.0094,0.1635]$}\\
Qwen2.5 L14 & q5, node 14 & \shortstack{$0.2613$\\$[0.2334,0.2927]$} & \shortstack{$0.0302$\\$[-0.0064,0.0687]$}\\
Qwen2.5 L14 & q6, node 0 & \shortstack{$0.1260$\\$[0.0911,0.1632]$} & \shortstack{$0.0465$\\$[0.0002,0.0949]$}\\
Qwen2.5 L14 & q6, node 1 & \shortstack{$0.6509$\\$[0.5871,0.7047]$} & \shortstack{$0.0190$\\$[-0.0343,0.0761]$}\\
Qwen2.5 L14 & q6, node 2 & \shortstack{$0.6901$\\$[0.5919,0.7833]$} & \shortstack{$0.0517$\\$[-0.0338,0.1344]$}\\
Qwen2.5 L14 & q6, node 3 & \shortstack{$0.6363$\\$[0.5513,0.7195]$} & \shortstack{$0.0212$\\$[-0.0291,0.0727]$}\\
Qwen2.5 L14 & q6, node 4 & \shortstack{$0.5385$\\$[0.4674,0.6102]$} & \shortstack{$0.0198$\\$[-0.0155,0.0552]$}\\
Qwen2.5 L14 & q6, node 5 & \shortstack{$0.4669$\\$[0.4057,0.5305]$} & \shortstack{$0.0091$\\$[-0.0283,0.0471]$}\\
Qwen2.5 L14 & q6, node 6 & \shortstack{$0.3735$\\$[0.3201,0.4275]$} & \shortstack{$0.0022$\\$[-0.0368,0.0414]$}\\
Qwen2.5 L14 & q6, node 7 & \shortstack{$0.3301$\\$[0.2804,0.3826]$} & \shortstack{$-0.0014$\\$[-0.0409,0.0400]$}\\
Qwen2.5 L14 & q6, node 8 & \shortstack{$0.3107$\\$[0.2604,0.3619]$} & \shortstack{$-0.0053$\\$[-0.0489,0.0392]$}\\
Qwen2.5 L14 & q6, node 9 & \shortstack{$0.3260$\\$[0.2695,0.3821]$} & \shortstack{$-0.0160$\\$[-0.0706,0.0380]$}\\
Qwen2.5 L14 & q6, node 10 & \shortstack{$0.3332$\\$[0.2732,0.3919]$} & \shortstack{$-0.0206$\\$[-0.0821,0.0396]$}\\
Qwen2.5 L14 & q6, node 11 & \shortstack{$0.3376$\\$[0.2765,0.3980]$} & \shortstack{$-0.0235$\\$[-0.0889,0.0400]$}\\
Qwen2.5 L14 & q6, node 12 & \shortstack{$0.3171$\\$[0.2558,0.3779]$} & \shortstack{$-0.0250$\\$[-0.0917,0.0397]$}\\
Qwen2.5 L14 & q6, node 13 & \shortstack{$0.3143$\\$[0.2518,0.3762]$} & \shortstack{$-0.0273$\\$[-0.0967,0.0395]$}\\
Qwen2.5 L14 & q6, node 14 & \shortstack{$0.2067$\\$[0.1770,0.2370]$} & \shortstack{$0.0079$\\$[-0.0255,0.0424]$}\\
Qwen2.5 L14 & q7, node 0 & \shortstack{$0.1262$\\$[0.0905,0.1631]$} & \shortstack{$0.0395$\\$[-0.0054,0.0857]$}\\
Qwen2.5 L14 & q7, node 1 & \shortstack{$0.6867$\\$[0.6611,0.7119]$} & \shortstack{$0.0021$\\$[-0.0522,0.0531]$}\\
Qwen2.5 L14 & q7, node 2 & \shortstack{$0.7258$\\$[0.6475,0.7986]$} & \shortstack{$-0.0569$\\$[-0.1531,0.0355]$}\\
Qwen2.5 L14 & q7, node 3 & \shortstack{$0.6293$\\$[0.5579,0.6950]$} & \shortstack{$-0.0413$\\$[-0.0995,0.0163]$}\\
Qwen2.5 L14 & q7, node 4 & \shortstack{$0.5170$\\$[0.4568,0.5734]$} & \shortstack{$-0.0079$\\$[-0.0501,0.0326]$}\\
Qwen2.5 L14 & q7, node 5 & \shortstack{$0.4532$\\$[0.3997,0.5034]$} & \shortstack{$0.0120$\\$[-0.0304,0.0530]$}\\
Qwen2.5 L14 & q7, node 6 & \shortstack{$0.3794$\\$[0.3309,0.4218]$} & \shortstack{$0.0273$\\$[-0.0157,0.0692]$}\\
Qwen2.5 L14 & q7, node 7 & \shortstack{$0.3425$\\$[0.2955,0.3864]$} & \shortstack{$0.0315$\\$[-0.0131,0.0761]$}\\
Qwen2.5 L14 & q7, node 8 & \shortstack{$0.3331$\\$[0.2855,0.3783]$} & \shortstack{$0.0354$\\$[-0.0134,0.0841]$}\\
Qwen2.5 L14 & q7, node 9 & \shortstack{$0.3577$\\$[0.3048,0.4076]$} & \shortstack{$0.0404$\\$[-0.0187,0.0998]$}\\
Qwen2.5 L14 & q7, node 10 & \shortstack{$0.3696$\\$[0.3134,0.4222]$} & \shortstack{$0.0456$\\$[-0.0191,0.1121]$}\\
Qwen2.5 L14 & q7, node 11 & \shortstack{$0.3782$\\$[0.3202,0.4332]$} & \shortstack{$0.0490$\\$[-0.0196,0.1194]$}\\
Qwen2.5 L14 & q7, node 12 & \shortstack{$0.3617$\\$[0.3029,0.4170]$} & \shortstack{$0.0505$\\$[-0.0193,0.1216]$}\\
Qwen2.5 L14 & q7, node 13 & \shortstack{$0.3622$\\$[0.3012,0.4192]$} & \shortstack{$0.0518$\\$[-0.0202,0.1258]$}\\
Qwen2.5 L14 & q7, node 14 & \shortstack{$0.1977$\\$[0.1676,0.2306]$} & \shortstack{$0.0179$\\$[-0.0164,0.0543]$}\\
Llama3.2 L3 & all, node 0 & \shortstack{$0.5641$\\$[0.5331,0.5936]$} & \shortstack{$0.0022$\\$[-0.0549,0.0590]$}\\
Llama3.2 L3 & all, node 1 & \shortstack{$0.8234$\\$[0.7915,0.8516]$} & \shortstack{$-0.0112$\\$[-0.1167,0.1006]$}\\
Llama3.2 L3 & all, node 2 & \shortstack{$0.9668$\\$[0.9600,0.9730]$} & \shortstack{$-0.0796$\\$[-0.1840,0.0210]$}\\
Llama3.2 L3 & all, node 3 & \shortstack{$0.3915$\\$[0.3781,0.4057]$} & \shortstack{$0.0184$\\$[0.0089,0.0287]$}\\
Llama3.2 L3 & q0, node 0 & \shortstack{$0.5640$\\$[0.5327,0.5933]$} & \shortstack{$0.0443$\\$[-0.0265,0.1211]$}\\
Llama3.2 L3 & q0, node 1 & \shortstack{$0.8228$\\$[0.7909,0.8511]$} & \shortstack{$0.1001$\\$[-0.0209,0.2253]$}\\
Llama3.2 L3 & q0, node 2 & \shortstack{$0.8396$\\$[0.8052,0.8706]$} & \shortstack{$0.0761$\\$[-0.0438,0.1955]$}\\
Llama3.2 L3 & q0, node 3 & \shortstack{$0.9999$\\$[0.9998,0.9999]$} & \shortstack{$0.0239$\\$[-0.0116,0.0616]$}\\
Llama3.2 L3 & q1, node 0 & \shortstack{$0.4922$\\$[0.4091,0.5587]$} & \shortstack{$0.1249$\\$[0.0346,0.2188]$}\\
Llama3.2 L3 & q1, node 1 & \shortstack{$0.7236$\\$[0.6112,0.8139]$} & \shortstack{$0.1764$\\$[0.0436,0.3080]$}\\
Llama3.2 L3 & q1, node 2 & \shortstack{$0.5544$\\$[0.4596,0.6425]$} & \shortstack{$0.0623$\\$[-0.0779,0.2008]$}\\
Llama3.2 L3 & q1, node 3 & \shortstack{$0.7141$\\$[0.6894,0.7386]$} & \shortstack{$0.0743$\\$[0.0359,0.1141]$}\\
Llama3.2 L3 & q14, node 0 & \shortstack{$0.5055$\\$[0.4348,0.5663]$} & \shortstack{$0.0843$\\$[-0.0021,0.1728]$}\\
Llama3.2 L3 & q14, node 1 & \shortstack{$0.7181$\\$[0.6081,0.8113]$} & \shortstack{$0.1392$\\$[-0.0079,0.2816]$}\\
Llama3.2 L3 & q14, node 2 & \shortstack{$0.5863$\\$[0.5047,0.6605]$} & \shortstack{$0.0533$\\$[-0.0678,0.1657]$}\\
Llama3.2 L3 & q14, node 3 & \shortstack{$0.3843$\\$[0.3491,0.4215]$} & \shortstack{$0.0253$\\$[-0.0006,0.0504]$}\\
Llama3.2 L3 & q15, node 0 & \shortstack{$0.5639$\\$[0.5328,0.5934]$} & \shortstack{$0.0001$\\$[-0.0637,0.0654]$}\\
Llama3.2 L3 & q15, node 1 & \shortstack{$0.8233$\\$[0.7914,0.8516]$} & \shortstack{$-0.0793$\\$[-0.1877,0.0359]$}\\
Llama3.2 L3 & q15, node 2 & \shortstack{$0.9307$\\$[0.9193,0.9417]$} & \shortstack{$-0.0033$\\$[-0.1373,0.1332]$}\\
Llama3.2 L3 & q15, node 3 & \shortstack{$0.3914$\\$[0.3551,0.4261]$} & \shortstack{$-0.0069$\\$[-0.0302,0.0189]$}\\
Llama3.2 L3 & q3, node 0 & \shortstack{$0.5644$\\$[0.5333,0.5938]$} & \shortstack{$0.0148$\\$[-0.0542,0.0885]$}\\
Llama3.2 L3 & q3, node 1 & \shortstack{$0.8234$\\$[0.7917,0.8516]$} & \shortstack{$0.0256$\\$[-0.0968,0.1506]$}\\
Llama3.2 L3 & q3, node 2 & \shortstack{$0.6367$\\$[0.5691,0.7012]$} & \shortstack{$0.0347$\\$[-0.0793,0.1485]$}\\
Llama3.2 L3 & q3, node 3 & \shortstack{$0.4221$\\$[0.3882,0.4600]$} & \shortstack{$0.0343$\\$[0.0024,0.0730]$}\\
Llama3.2 L3 & q4, node 0 & \shortstack{$0.5643$\\$[0.5332,0.5937]$} & \shortstack{$-0.0262$\\$[-0.0869,0.0331]$}\\
Llama3.2 L3 & q4, node 1 & \shortstack{$0.8232$\\$[0.7912,0.8515]$} & \shortstack{$-0.0623$\\$[-0.1750,0.0566]$}\\
Llama3.2 L3 & q4, node 2 & \shortstack{$0.5362$\\$[0.4775,0.5932]$} & \shortstack{$-0.0131$\\$[-0.0932,0.0686]$}\\
Llama3.2 L3 & q4, node 3 & \shortstack{$0.4227$\\$[0.3887,0.4565]$} & \shortstack{$0.0174$\\$[-0.0120,0.0508]$}\\
Llama3.2 L3 & q5, node 0 & \shortstack{$0.5642$\\$[0.5332,0.5936]$} & \shortstack{$-0.0123$\\$[-0.0837,0.0617]$}\\
Llama3.2 L3 & q5, node 1 & \shortstack{$0.8233$\\$[0.7915,0.8516]$} & \shortstack{$-0.0172$\\$[-0.1355,0.0979]$}\\
Llama3.2 L3 & q5, node 2 & \shortstack{$0.5054$\\$[0.4440,0.5664]$} & \shortstack{$-0.0194$\\$[-0.1084,0.0812]$}\\
Llama3.2 L3 & q5, node 3 & \shortstack{$0.3907$\\$[0.3492,0.4276]$} & \shortstack{$0.0124$\\$[-0.0140,0.0401]$}\\
Llama3.2 L3 & q6, node 0 & \shortstack{$0.5639$\\$[0.5328,0.5932]$} & \shortstack{$0.0412$\\$[-0.0346,0.1138]$}\\
Llama3.2 L3 & q6, node 1 & \shortstack{$0.8231$\\$[0.7911,0.8513]$} & \shortstack{$0.0994$\\$[-0.0336,0.2210]$}\\
Llama3.2 L3 & q6, node 2 & \shortstack{$0.3871$\\$[0.3092,0.4592]$} & \shortstack{$0.0168$\\$[-0.0864,0.1109]$}\\
Llama3.2 L3 & q6, node 3 & \shortstack{$0.4019$\\$[0.3636,0.4388]$} & \shortstack{$0.0055$\\$[-0.0223,0.0343]$}\\
Llama3.2 L3 & q7, node 0 & \shortstack{$0.5263$\\$[0.4638,0.5803]$} & \shortstack{$0.0220$\\$[-0.0435,0.0910]$}\\
Llama3.2 L3 & q7, node 1 & \shortstack{$0.7748$\\$[0.6945,0.8402]$} & \shortstack{$0.1149$\\$[-0.0074,0.2261]$}\\
Llama3.2 L3 & q7, node 2 & \shortstack{$0.3945$\\$[0.3042,0.4832]$} & \shortstack{$0.1225$\\$[0.0228,0.2214]$}\\
Llama3.2 L3 & q7, node 3 & \shortstack{$0.4127$\\$[0.3745,0.4498]$} & \shortstack{$0.0177$\\$[-0.0036,0.0403]$}\\
Llama3.2 L8 & all, node 0 & \shortstack{$0.5642$\\$[0.5331,0.5936]$} & \shortstack{$-0.0031$\\$[-0.0568,0.0509]$}\\
Llama3.2 L8 & all, node 1 & \shortstack{$0.8234$\\$[0.7914,0.8517]$} & \shortstack{$-0.0134$\\$[-0.1145,0.0889]$}\\
Llama3.2 L8 & all, node 2 & \shortstack{$0.9999$\\$[0.9998,0.9999]$} & \shortstack{$0.0933$\\$[-0.1102,0.2937]$}\\
Llama3.2 L8 & all, node 3 & \shortstack{$0.9997$\\$[0.9997,0.9997]$} & \shortstack{$0.0773$\\$[-0.1155,0.2605]$}\\
Llama3.2 L8 & all, node 4 & \shortstack{$0.9995$\\$[0.9994,0.9995]$} & \shortstack{$0.0632$\\$[-0.1159,0.2324]$}\\
Llama3.2 L8 & all, node 5 & \shortstack{$0.9989$\\$[0.9987,0.9991]$} & \shortstack{$0.0471$\\$[-0.1087,0.2028]$}\\
Llama3.2 L8 & all, node 6 & \shortstack{$0.9979$\\$[0.9975,0.9982]$} & \shortstack{$0.0380$\\$[-0.0934,0.1733]$}\\
Llama3.2 L8 & all, node 7 & \shortstack{$0.9962$\\$[0.9956,0.9968]$} & \shortstack{$0.0307$\\$[-0.0821,0.1477]$}\\
Llama3.2 L8 & all, node 8 & \shortstack{$0.5773$\\$[0.5659,0.5889]$} & \shortstack{$0.0111$\\$[-0.0027,0.0253]$}\\
Llama3.2 L8 & q0, node 0 & \shortstack{$0.1906$\\$[0.0510,0.3220]$} & \shortstack{$-0.0603$\\$[-0.1583,0.0410]$}\\
Llama3.2 L8 & q0, node 1 & \shortstack{$0.2332$\\$[0.0336,0.4193]$} & \shortstack{$-0.1577$\\$[-0.3086,0.0018]$}\\
Llama3.2 L8 & q0, node 2 & \shortstack{$0.2833$\\$[0.0898,0.4697]$} & \shortstack{$-0.1067$\\$[-0.2999,0.0860]$}\\
Llama3.2 L8 & q0, node 3 & \shortstack{$0.2663$\\$[0.0885,0.4365]$} & \shortstack{$-0.1016$\\$[-0.2727,0.0653]$}\\
Llama3.2 L8 & q0, node 4 & \shortstack{$0.2461$\\$[0.0796,0.4019]$} & \shortstack{$-0.0960$\\$[-0.2544,0.0504]$}\\
Llama3.2 L8 & q0, node 5 & \shortstack{$0.2158$\\$[0.0688,0.3518]$} & \shortstack{$-0.0759$\\$[-0.2098,0.0514]$}\\
Llama3.2 L8 & q0, node 6 & \shortstack{$0.1852$\\$[0.0600,0.3023]$} & \shortstack{$-0.0520$\\$[-0.1638,0.0555]$}\\
Llama3.2 L8 & q0, node 7 & \shortstack{$0.1627$\\$[0.0501,0.2661]$} & \shortstack{$-0.0294$\\$[-0.1290,0.0648]$}\\
Llama3.2 L8 & q0, node 8 & \shortstack{$0.9999$\\$[0.9999,0.9999]$} & \shortstack{$0.0371$\\$[-0.0125,0.0898]$}\\
Llama3.2 L8 & q1, node 0 & \shortstack{$0.2832$\\$[0.1610,0.4027]$} & \shortstack{$-0.0461$\\$[-0.1374,0.0488]$}\\
Llama3.2 L8 & q1, node 1 & \shortstack{$0.4032$\\$[0.2217,0.5790]$} & \shortstack{$-0.1251$\\$[-0.2631,0.0123]$}\\
Llama3.2 L8 & q1, node 2 & \shortstack{$0.4431$\\$[0.2646,0.6138]$} & \shortstack{$-0.2039$\\$[-0.3853,-0.0156]$}\\
Llama3.2 L8 & q1, node 3 & \shortstack{$0.4126$\\$[0.2498,0.5706]$} & \shortstack{$-0.2006$\\$[-0.3632,-0.0295]$}\\
Llama3.2 L8 & q1, node 4 & \shortstack{$0.3838$\\$[0.2354,0.5299]$} & \shortstack{$-0.1966$\\$[-0.3429,-0.0415]$}\\
Llama3.2 L8 & q1, node 5 & \shortstack{$0.3407$\\$[0.2128,0.4725]$} & \shortstack{$-0.1786$\\$[-0.3068,-0.0513]$}\\
Llama3.2 L8 & q1, node 6 & \shortstack{$0.2956$\\$[0.1833,0.4099]$} & \shortstack{$-0.1556$\\$[-0.2621,-0.0514]$}\\
Llama3.2 L8 & q1, node 7 & \shortstack{$0.2505$\\$[0.1512,0.3509]$} & \shortstack{$-0.1317$\\$[-0.2217,-0.0418]$}\\
Llama3.2 L8 & q1, node 8 & \shortstack{$0.8626$\\$[0.8522,0.8718]$} & \shortstack{$-0.0041$\\$[-0.0393,0.0295]$}\\
Llama3.2 L8 & q14, node 0 & \shortstack{$0.5266$\\$[0.4627,0.5774]$} & \shortstack{$0.0013$\\$[-0.0635,0.0713]$}\\
Llama3.2 L8 & q14, node 1 & \shortstack{$0.7674$\\$[0.6778,0.8350]$} & \shortstack{$0.0295$\\$[-0.0789,0.1415]$}\\
Llama3.2 L8 & q14, node 2 & \shortstack{$0.8946$\\$[0.8084,0.9568]$} & \shortstack{$-0.0842$\\$[-0.2369,0.0821]$}\\
Llama3.2 L8 & q14, node 3 & \shortstack{$0.8533$\\$[0.7700,0.9158]$} & \shortstack{$-0.0651$\\$[-0.1972,0.0792]$}\\
Llama3.2 L8 & q14, node 4 & \shortstack{$0.8130$\\$[0.7313,0.8759]$} & \shortstack{$-0.0548$\\$[-0.1767,0.0802]$}\\
Llama3.2 L8 & q14, node 5 & \shortstack{$0.7405$\\$[0.6610,0.8073]$} & \shortstack{$-0.0405$\\$[-0.1522,0.0842]$}\\
Llama3.2 L8 & q14, node 6 & \shortstack{$0.6486$\\$[0.5710,0.7214]$} & \shortstack{$-0.0294$\\$[-0.1322,0.0876]$}\\
Llama3.2 L8 & q14, node 7 & \shortstack{$0.5641$\\$[0.4832,0.6460]$} & \shortstack{$-0.0250$\\$[-0.1284,0.0918]$}\\
Llama3.2 L8 & q14, node 8 & \shortstack{$0.4932$\\$[0.4727,0.5132]$} & \shortstack{$0.0175$\\$[-0.0002,0.0345]$}\\
Llama3.2 L8 & q15, node 0 & \shortstack{$0.5642$\\$[0.5331,0.5936]$} & \shortstack{$0.0027$\\$[-0.0509,0.0543]$}\\
Llama3.2 L8 & q15, node 1 & \shortstack{$0.8235$\\$[0.7914,0.8517]$} & \shortstack{$0.0067$\\$[-0.0943,0.1136]$}\\
Llama3.2 L8 & q15, node 2 & \shortstack{$0.9948$\\$[0.9937,0.9959]$} & \shortstack{$-0.1013$\\$[-0.2818,0.0744]$}\\
Llama3.2 L8 & q15, node 3 & \shortstack{$0.9882$\\$[0.9857,0.9906]$} & \shortstack{$-0.0775$\\$[-0.2342,0.0780]$}\\
Llama3.2 L8 & q15, node 4 & \shortstack{$0.9800$\\$[0.9755,0.9841]$} & \shortstack{$-0.0559$\\$[-0.1978,0.0867]$}\\
Llama3.2 L8 & q15, node 5 & \shortstack{$0.9600$\\$[0.9512,0.9682]$} & \shortstack{$-0.0313$\\$[-0.1515,0.0936]$}\\
Llama3.2 L8 & q15, node 6 & \shortstack{$0.9252$\\$[0.9092,0.9401]$} & \shortstack{$-0.0129$\\$[-0.1191,0.0968]$}\\
Llama3.2 L8 & q15, node 7 & \shortstack{$0.8804$\\$[0.8539,0.9045]$} & \shortstack{$-0.0039$\\$[-0.1106,0.1077]$}\\
Llama3.2 L8 & q15, node 8 & \shortstack{$0.4865$\\$[0.4671,0.5051]$} & \shortstack{$0.0016$\\$[-0.0185,0.0195]$}\\
Llama3.2 L8 & q3, node 0 & \shortstack{$0.5643$\\$[0.5331,0.5937]$} & \shortstack{$0.0178$\\$[-0.0438,0.0837]$}\\
Llama3.2 L8 & q3, node 1 & \shortstack{$0.8235$\\$[0.7914,0.8518]$} & \shortstack{$-0.0111$\\$[-0.1174,0.0930]$}\\
Llama3.2 L8 & q3, node 2 & \shortstack{$0.9538$\\$[0.9258,0.9772]$} & \shortstack{$-0.0571$\\$[-0.1988,0.0832]$}\\
Llama3.2 L8 & q3, node 3 & \shortstack{$0.9228$\\$[0.8852,0.9537]$} & \shortstack{$-0.0293$\\$[-0.1470,0.0888]$}\\
Llama3.2 L8 & q3, node 4 & \shortstack{$0.8934$\\$[0.8503,0.9281]$} & \shortstack{$-0.0127$\\$[-0.1201,0.0897]$}\\
Llama3.2 L8 & q3, node 5 & \shortstack{$0.8359$\\$[0.7876,0.8767]$} & \shortstack{$0.0048$\\$[-0.0907,0.0964]$}\\
Llama3.2 L8 & q3, node 6 & \shortstack{$0.7591$\\$[0.7066,0.8051]$} & \shortstack{$0.0176$\\$[-0.0668,0.0995]$}\\
Llama3.2 L8 & q3, node 7 & \shortstack{$0.6704$\\$[0.6128,0.7234]$} & \shortstack{$0.0258$\\$[-0.0544,0.1027]$}\\
Llama3.2 L8 & q3, node 8 & \shortstack{$0.6730$\\$[0.6552,0.6900]$} & \shortstack{$-0.0011$\\$[-0.0289,0.0292]$}\\
Llama3.2 L8 & q4, node 0 & \shortstack{$0.5642$\\$[0.5331,0.5937]$} & \shortstack{$0.0079$\\$[-0.0563,0.0745]$}\\
Llama3.2 L8 & q4, node 1 & \shortstack{$0.8236$\\$[0.7915,0.8518]$} & \shortstack{$0.0229$\\$[-0.0812,0.1289]$}\\
Llama3.2 L8 & q4, node 2 & \shortstack{$0.9639$\\$[0.9537,0.9731]$} & \shortstack{$0.0782$\\$[-0.0569,0.2282]$}\\
Llama3.2 L8 & q4, node 3 & \shortstack{$0.9206$\\$[0.8997,0.9393]$} & \shortstack{$0.0698$\\$[-0.0368,0.1823]$}\\
Llama3.2 L8 & q4, node 4 & \shortstack{$0.8757$\\$[0.8446,0.9037]$} & \shortstack{$0.0605$\\$[-0.0290,0.1568]$}\\
Llama3.2 L8 & q4, node 5 & \shortstack{$0.7933$\\$[0.7474,0.8341]$} & \shortstack{$0.0500$\\$[-0.0230,0.1311]$}\\
Llama3.2 L8 & q4, node 6 & \shortstack{$0.6919$\\$[0.6347,0.7441]$} & \shortstack{$0.0416$\\$[-0.0309,0.1167]$}\\
Llama3.2 L8 & q4, node 7 & \shortstack{$0.5876$\\$[0.5220,0.6495]$} & \shortstack{$0.0374$\\$[-0.0382,0.1140]$}\\
Llama3.2 L8 & q4, node 8 & \shortstack{$0.6338$\\$[0.6143,0.6529]$} & \shortstack{$-0.0021$\\$[-0.0296,0.0267]$}\\
Llama3.2 L8 & q5, node 0 & \shortstack{$0.5305$\\$[0.4713,0.5818]$} & \shortstack{$-0.0075$\\$[-0.0639,0.0500]$}\\
Llama3.2 L8 & q5, node 1 & \shortstack{$0.7741$\\$[0.6908,0.8379]$} & \shortstack{$0.0028$\\$[-0.1021,0.1093]$}\\
Llama3.2 L8 & q5, node 2 & \shortstack{$0.9151$\\$[0.8341,0.9706]$} & \shortstack{$0.0365$\\$[-0.1033,0.1743]$}\\
Llama3.2 L8 & q5, node 3 & \shortstack{$0.8797$\\$[0.8074,0.9319]$} & \shortstack{$0.0357$\\$[-0.0708,0.1442]$}\\
Llama3.2 L8 & q5, node 4 & \shortstack{$0.8402$\\$[0.7738,0.8895]$} & \shortstack{$0.0409$\\$[-0.0447,0.1319]$}\\
Llama3.2 L8 & q5, node 5 & \shortstack{$0.7606$\\$[0.6991,0.8089]$} & \shortstack{$0.0459$\\$[-0.0241,0.1213]$}\\
Llama3.2 L8 & q5, node 6 & \shortstack{$0.6550$\\$[0.5949,0.7061]$} & \shortstack{$0.0487$\\$[-0.0159,0.1189]$}\\
Llama3.2 L8 & q5, node 7 & \shortstack{$0.5498$\\$[0.4878,0.6038]$} & \shortstack{$0.0519$\\$[-0.0166,0.1224]$}\\
Llama3.2 L8 & q5, node 8 & \shortstack{$0.6047$\\$[0.5881,0.6214]$} & \shortstack{$0.0097$\\$[-0.0164,0.0339]$}\\
Llama3.2 L8 & q6, node 0 & \shortstack{$0.5641$\\$[0.5331,0.5936]$} & \shortstack{$-0.0085$\\$[-0.0664,0.0492]$}\\
Llama3.2 L8 & q6, node 1 & \shortstack{$0.8233$\\$[0.7914,0.8515]$} & \shortstack{$-0.0314$\\$[-0.1364,0.0780]$}\\
Llama3.2 L8 & q6, node 2 & \shortstack{$0.9409$\\$[0.9278,0.9523]$} & \shortstack{$-0.0496$\\$[-0.1753,0.0749]$}\\
Llama3.2 L8 & q6, node 3 & \shortstack{$0.8678$\\$[0.8383,0.8932]$} & \shortstack{$-0.0294$\\$[-0.1268,0.0678]$}\\
Llama3.2 L8 & q6, node 4 & \shortstack{$0.8025$\\$[0.7662,0.8353]$} & \shortstack{$-0.0299$\\$[-0.1177,0.0559]$}\\
Llama3.2 L8 & q6, node 5 & \shortstack{$0.6819$\\$[0.6307,0.7286]$} & \shortstack{$-0.0212$\\$[-0.1029,0.0598]$}\\
Llama3.2 L8 & q6, node 6 & \shortstack{$0.5421$\\$[0.4800,0.5994]$} & \shortstack{$-0.0170$\\$[-0.0977,0.0633]$}\\
Llama3.2 L8 & q6, node 7 & \shortstack{$0.4203$\\$[0.3494,0.4875]$} & \shortstack{$-0.0171$\\$[-0.1037,0.0704]$}\\
Llama3.2 L8 & q6, node 8 & \shortstack{$0.5843$\\$[0.5675,0.6014]$} & \shortstack{$0.0061$\\$[-0.0171,0.0308]$}\\
Llama3.2 L8 & q7, node 0 & \shortstack{$0.5641$\\$[0.5331,0.5936]$} & \shortstack{$0.0010$\\$[-0.0610,0.0621]$}\\
Llama3.2 L8 & q7, node 1 & \shortstack{$0.8234$\\$[0.7913,0.8517]$} & \shortstack{$0.0006$\\$[-0.1004,0.0967]$}\\
Llama3.2 L8 & q7, node 2 & \shortstack{$0.9314$\\$[0.9106,0.9492]$} & \shortstack{$0.0900$\\$[-0.0322,0.2072]$}\\
Llama3.2 L8 & q7, node 3 & \shortstack{$0.8732$\\$[0.8438,0.8986]$} & \shortstack{$0.0864$\\$[-0.0090,0.1779]$}\\
Llama3.2 L8 & q7, node 4 & \shortstack{$0.8102$\\$[0.7734,0.8434]$} & \shortstack{$0.0868$\\$[0.0055,0.1657]$}\\
Llama3.2 L8 & q7, node 5 & \shortstack{$0.7017$\\$[0.6554,0.7443]$} & \shortstack{$0.0816$\\$[0.0097,0.1512]$}\\
Llama3.2 L8 & q7, node 6 & \shortstack{$0.5744$\\$[0.5194,0.6269]$} & \shortstack{$0.0757$\\$[0.0043,0.1418]$}\\
Llama3.2 L8 & q7, node 7 & \shortstack{$0.4641$\\$[0.4007,0.5219]$} & \shortstack{$0.0721$\\$[-0.0072,0.1447]$}\\
Llama3.2 L8 & q7, node 8 & \shortstack{$0.5662$\\$[0.5477,0.5846]$} & \shortstack{$0.0241$\\$[0.0024,0.0464]$}\\
Llama3.2 L3--L8 & Same window, q14 & \shortstack{$0.8750$\\$[0.7500,0.9688]$} & \shortstack{$0.7391$\\$[0.6697,0.8063]$}\\
Llama3.2 L3--L8 & Same window, all & \shortstack{$1.0000$\\$[1.0000,1.0000]$} & \shortstack{$0.9764$\\$[0.9603,0.9869]$}\\
\end{longtable}
\endgroup

\subsection{Readable relations, known source roles and source order}
\label{sup:s7-order}
We compared direct linear relation readout with a relation composed by the evaluator from two separately decoded attributes. The three observation views used the same 512 GPT-2 L6 binding groups. Hidden receivers favored direct relation readout; recipient and future source spans favored attribute composition (Table~\ref{tab:sup-relation-domains}; Fig.~\ref{fig:sup-source-transport}D). The corresponding syntax hidden-receiver accuracies were $55.81\%$ direct and $71.09\%$ composed, preserving the task difference. These comparisons describe the readable form of a relation under fixed training budgets and regularization.

The GPT-2 L6 order control used 64 natural source windows from 60 documents and three independently calibrated map seeds (123, 321 and 567). Each window supplied its original order and four suffix permutations. Positions 0--7 and the selected feature $H_7$ were fixed; positions 8--15 retained the same token multiset in a different order. Each permutation used its actual local derivatives along the full retained chain, with the calibrated maps held fixed. Across all saved variants and seeds, the relative changes in $H_7$, the first raw adjoint and the first $\WienerG$ estimate were exactly zero; the first raw adjoint's future-energy fraction was also zero. Thus the selected feature and first raw adjoint retained their prefix support while subsequent full-position estimation could respond to the changed geometry.

For the final single-query source estimates, suffix cosine was $0.490038$ $[0.460118,0.519040]$ at matched positions and $0.881229$ $[0.846009,0.905600]$ after undoing the permutation. Permutations were averaged within each original source, then map seeds within source, before 2,000 document-bootstrap resamples. This comparison measures the estimate's alignment with suffix order under fixed prefix features and calibrated maps. The analogous Qwen2.5/Llama3.2 comparisons also showed positive permutation-alignment gains.

Source-span selection used known structural roles and factor positions. Candidate retrieval measured a different property: in the same 32 Llama3.2 P0 prefix vectors, four-way centered cosine retrieval reached $16.41\%$, while the frozen factor readouts reached $83.59\%$ and $71.09\%$. Geometric candidate ranking and factor decoding can thus separate within the same saved representation.

\begingroup\footnotesize
\setlength{\tabcolsep}{4pt}
\begin{longtable}{>{\raggedright\arraybackslash}p{.35\linewidth}>{\raggedright\arraybackslash}p{.20\linewidth}>{\raggedright\arraybackslash}p{.36\linewidth}}
\caption{Relations, source order and complementary readout rules. The first three rows give GPT-2 direct/composed relation accuracies (percent) in 512 matched groups and their paired difference (percentage points). Composition follows separate attribute decoding by the evaluator. The last two rows reuse the same 32 P0 prefix representations with a frozen factor readout and a four-way cosine retrieval rule.}\label{tab:sup-relation-domains}\\
\toprule
Measurement & Two conditions or value & Contrast / interpretation\\
\midrule
\endfirsthead
\multicolumn{3}{l}{\tablename\ \thetable\ (continued)}\\
\toprule
Measurement & Two conditions or value & Contrast / interpretation\\
\midrule
\endhead
\bottomrule\endfoot
Relation, hidden receiver & 84.62 / 78.61 & \shortstack{$6.01$\\$[3.86,8.11]$}\\
Relation, recipient source spans & 87.06 / 100.00 & \shortstack{$-12.94$\\$[-14.65,-11.23]$}\\
Relation, future source spans & 68.21 / 80.81 & \shortstack{$-12.60$\\$[-14.36,-10.84]$}\\
GPT-2 suffix cosine & 0.4900 / 0.8812 & Before / after alignment\\
Qwen2.5 permutation cosine gain & 0.5478 / 0.7102 & \shortstack{$0.1624$\\$[0.1341,0.1920]$}\\
Llama3.2 permutation cosine gain & 0.8624 / 0.9366 & \shortstack{$0.0741$\\$[0.0467,0.1022]$}\\
Llama3.2 same-prefix readout & 83.59 / 71.09 & Factor A / B (\%)\\
Llama3.2 same-prefix retrieval & 16.41 & Four-way accuracy (\%)\\
\end{longtable}
\endgroup

\subsection{Layout-specific single-point and combination adaptation}
The layout experiment used hidden-receiver features at q12--q15, extracted from attention outputs before residual addition. Direct readouts of these features crossed the training domain of a single-point baseline with that of its nested correction. The best single position was selected once on the old-layout development set. Old/new baseline and old/new correction training used equal budgets, alongside transfer of the old correction scores without retraining. Each combination had repeated-single-point and shuffled-extra-position controls. Correction training used five-fold out-of-fold baseline scores; development selection included zero correction and gains $0.25$, $0.5$ and $1$.

Each model contributed 12 panels: three depths, two tasks and two factors. GPT-2 used 64 training and 16 development groups per task. Qwen2.5/Llama3.2 used these binding budgets and 59/15 syntax groups after tokenization matching. P0 confirmation contained 256 groups per panel, except their syntax panels, which contained 252. Four corners were averaged within group before 2,000 bootstrap resamples. These intervals condition on the fitted readouts and development selections.

\begin{figure}[!htbp]
\centering
\includegraphics[width=\linewidth,height=.79\textheight,keepaspectratio]{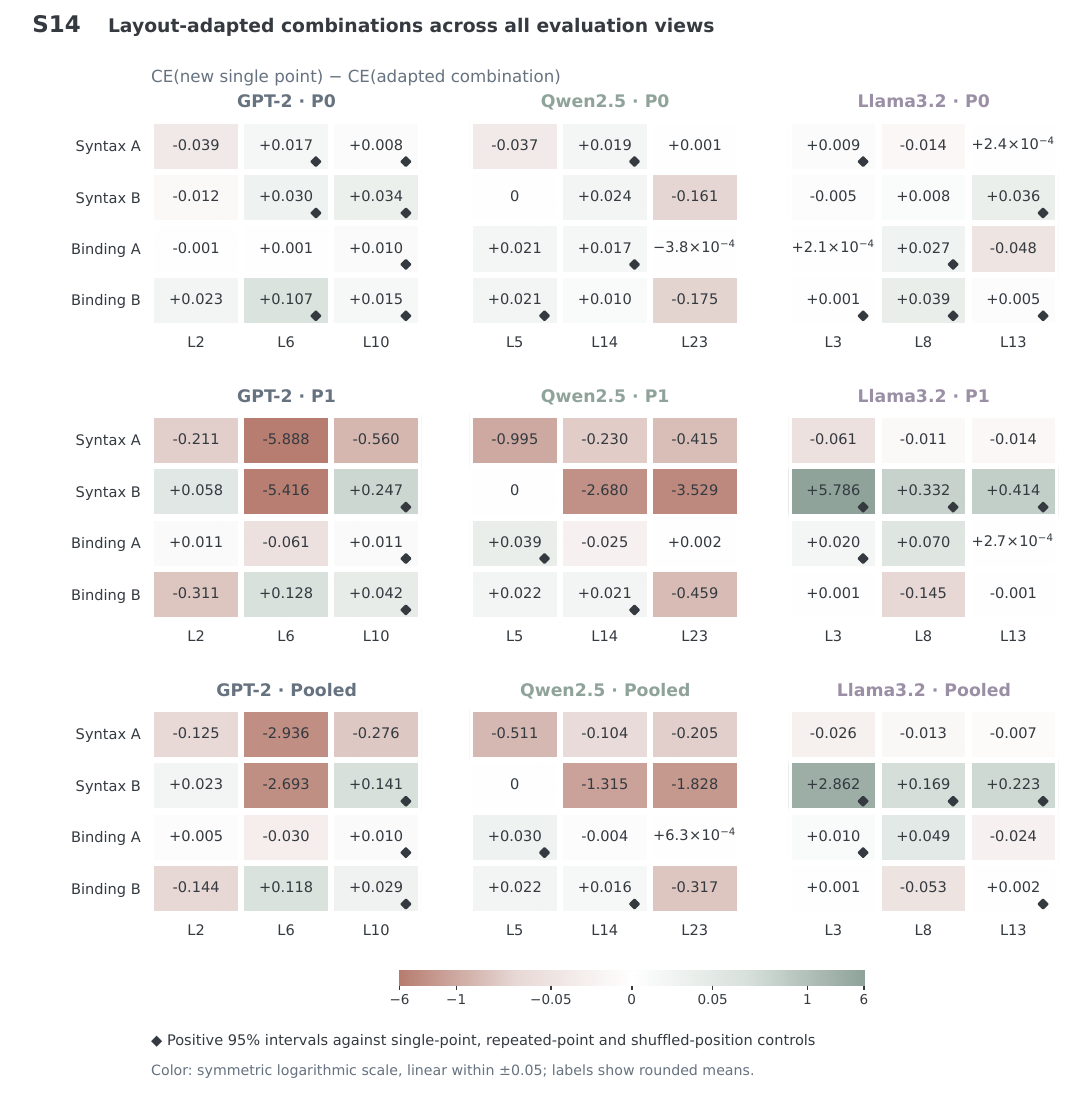}
\caption{Complete layout-conditioned combination of hidden receivers at q12--q15. Each cell is cross-entropy of the fixed new single-point baseline minus that of the adapted combination. Rows separate P0, P1 and pooled evaluation views; columns separate models. The common color bar uses a symmetric logarithmic scale centered at zero, with a linear interval of $[-0.05,0.05]$. Labels give rounded means, using scientific notation for small nonzero values. Diamonds mark positive 95\% group-bootstrap intervals against the single-point, repeated-point and shuffled-extra-position controls. The panel count is 36 per evaluation view.}
\label{fig:sup-source-layout}
\end{figure}

With the new single-point baseline held fixed, adapting the correction produced positive intervals in $10/12$, $9/12$ and $11/12$ P0 panels for GPT-2, Qwen2.5 and Llama3.2. Respectively $7/12$, $3/12$ and $6/12$ also exceeded all three controls; $2/12$, $2/12$ and $1/12$ retained negative adapted increments. Table~\ref{tab:sup-d4-counts} and Fig.~\ref{fig:sup-source-layout} retain all panels in P0, P1 and pooled evaluation views. The q12--q15 role ordering was invariant, so the adaptation concerns cross-position combination weights under the new layout.

\begingroup\footnotesize
\setlength{\tabcolsep}{4pt}
\begin{longtable}{lrrrr}
\caption{Hidden-receiver combination at q12--q15: all 36 panels in each of the P0, P1 and pooled evaluation views. Adapted increment is CE(new single point) minus CE(new-baseline/new-correction combination). Correction adaptation holds the new single point fixed and compares old versus new correction training. All-controls requires positive 95\% intervals against the single point, repeated-point and shuffled-position controls. Small nonzero estimates and interval bounds retain additional digits. Complete template/position/vocabulary strata accompany the source data.}\label{tab:sup-d4-counts}\\
\toprule
Model / depth & Task / view & Adapted increment & Correction adaptation & All controls\\
\midrule
\endfirsthead
\multicolumn{5}{l}{\tablename\ \thetable\ (continued)}\\
\toprule
Model / depth & Task / view & Adapted increment & Correction adaptation & All controls\\
\midrule
\endhead
\bottomrule\endfoot
GPT-2 L2 & SA / P0 & \shortstack{$-0.039$\\$[-0.069,-0.010]$} & \shortstack{$0.410$\\$[0.374,0.448]$} & No\\
GPT-2 L6 & SA / P0 & \shortstack{$0.017$\\$[0.008,0.025]$} & \shortstack{$0.475$\\$[0.408,0.545]$} & Yes\\
GPT-2 L10 & SA / P0 & \shortstack{$0.008$\\$[0.006,0.011]$} & \shortstack{$0.001$\\$[-0.005,0.007]$} & Yes\\
GPT-2 L2 & SB / P0 & \shortstack{$-0.012$\\$[-0.015,-0.008]$} & \shortstack{$2.434$\\$[2.242,2.619]$} & No\\
GPT-2 L6 & SB / P0 & \shortstack{$0.030$\\$[0.011,0.049]$} & \shortstack{$0.377$\\$[0.294,0.458]$} & Yes\\
GPT-2 L10 & SB / P0 & \shortstack{$0.034$\\$[0.018,0.051]$} & \shortstack{$0.477$\\$[0.381,0.566]$} & Yes\\
GPT-2 L2 & BA / P0 & \shortstack{$-0.001$\\$[-0.004,0.002]$} & \shortstack{$1.613$\\$[1.572,1.654]$} & No\\
GPT-2 L6 & BA / P0 & \shortstack{$0.001$\\$[-0.004,0.007]$} & \shortstack{$-0.003$\\$[-0.009,0.003]$} & No\\
GPT-2 L10 & BA / P0 & \shortstack{$0.010$\\$[0.008,0.012]$} & \shortstack{$0.039$\\$[0.032,0.046]$} & Yes\\
GPT-2 L2 & BB / P0 & \shortstack{$0.023$\\$[0.012,0.034]$} & \shortstack{$0.869$\\$[0.842,0.896]$} & No\\
GPT-2 L6 & BB / P0 & \shortstack{$0.107$\\$[0.087,0.127]$} & \shortstack{$3.805$\\$[3.676,3.931]$} & Yes\\
GPT-2 L10 & BB / P0 & \shortstack{$0.015$\\$[0.012,0.019]$} & \shortstack{$0.028$\\$[0.019,0.037]$} & Yes\\
GPT-2 L2 & SA / P1 & \shortstack{$-0.211$\\$[-0.259,-0.165]$} & \shortstack{$-0.252$\\$[-0.353,-0.155]$} & No\\
GPT-2 L6 & SA / P1 & \shortstack{$-5.888$\\$[-6.582,-5.158]$} & \shortstack{$-7.726$\\$[-8.680,-6.700]$} & No\\
GPT-2 L10 & SA / P1 & \shortstack{$-0.560$\\$[-0.620,-0.507]$} & \shortstack{$-0.876$\\$[-0.966,-0.788]$} & No\\
GPT-2 L2 & SB / P1 & \shortstack{$0.058$\\$[0.055,0.061]$} & \shortstack{$-1.582$\\$[-1.743,-1.414]$} & No\\
GPT-2 L6 & SB / P1 & \shortstack{$-5.416$\\$[-6.036,-4.761]$} & \shortstack{$-6.411$\\$[-7.154,-5.635]$} & No\\
GPT-2 L10 & SB / P1 & \shortstack{$0.247$\\$[0.174,0.317]$} & \shortstack{$-0.547$\\$[-0.653,-0.447]$} & Yes\\
GPT-2 L2 & BA / P1 & \shortstack{$0.011$\\$[-0.002,0.024]$} & \shortstack{$-1.186$\\$[-1.249,-1.120]$} & No\\
GPT-2 L6 & BA / P1 & \shortstack{$-0.061$\\$[-0.076,-0.048]$} & \shortstack{$-0.110$\\$[-0.127,-0.094]$} & No\\
GPT-2 L10 & BA / P1 & \shortstack{$0.011$\\$[0.008,0.013]$} & \shortstack{$-0.024$\\$[-0.029,-0.019]$} & Yes\\
GPT-2 L2 & BB / P1 & \shortstack{$-0.311$\\$[-0.355,-0.266]$} & \shortstack{$-0.798$\\$[-0.860,-0.735]$} & No\\
GPT-2 L6 & BB / P1 & \shortstack{$0.128$\\$[0.072,0.184]$} & \shortstack{$-2.789$\\$[-3.159,-2.412]$} & No\\
GPT-2 L10 & BB / P1 & \shortstack{$0.042$\\$[0.037,0.047]$} & \shortstack{$-0.013$\\$[-0.018,-0.009]$} & Yes\\
GPT-2 L2 & SA / all & \shortstack{$-0.125$\\$[-0.155,-0.096]$} & \shortstack{$0.079$\\$[0.021,0.138]$} & No\\
GPT-2 L6 & SA / all & \shortstack{$-2.936$\\$[-3.386,-2.494]$} & \shortstack{$-3.626$\\$[-4.239,-3.024]$} & No\\
GPT-2 L10 & SA / all & \shortstack{$-0.276$\\$[-0.315,-0.239]$} & \shortstack{$-0.437$\\$[-0.495,-0.380]$} & No\\
GPT-2 L2 & SB / all & \shortstack{$0.023$\\$[0.019,0.027]$} & \shortstack{$0.426$\\$[0.215,0.636]$} & No\\
GPT-2 L6 & SB / all & \shortstack{$-2.693$\\$[-3.101,-2.293]$} & \shortstack{$-3.017$\\$[-3.496,-2.534]$} & No\\
GPT-2 L10 & SB / all & \shortstack{$0.141$\\$[0.103,0.178]$} & \shortstack{$-0.035$\\$[-0.121,0.047]$} & Yes\\
GPT-2 L2 & BA / all & \shortstack{$0.005$\\$[-0.001,0.012]$} & \shortstack{$0.214$\\$[0.085,0.342]$} & No\\
GPT-2 L6 & BA / all & \shortstack{$-0.030$\\$[-0.038,-0.022]$} & \shortstack{$-0.056$\\$[-0.066,-0.047]$} & No\\
GPT-2 L10 & BA / all & \shortstack{$0.010$\\$[0.009,0.012]$} & \shortstack{$0.007$\\$[0.002,0.012]$} & Yes\\
GPT-2 L2 & BB / all & \shortstack{$-0.144$\\$[-0.172,-0.117]$} & \shortstack{$0.035$\\$[-0.049,0.113]$} & No\\
GPT-2 L6 & BB / all & \shortstack{$0.118$\\$[0.087,0.147]$} & \shortstack{$0.508$\\$[0.154,0.863]$} & No\\
GPT-2 L10 & BB / all & \shortstack{$0.029$\\$[0.026,0.032]$} & \shortstack{$0.007$\\$[0.002,0.012]$} & Yes\\
Qwen2.5 L5 & SA / P0 & \shortstack{$-0.037$\\$[-0.079,0.003]$} & \shortstack{$9.491$\\$[9.172,9.812]$} & No\\
Qwen2.5 L14 & SA / P0 & \shortstack{$0.019$\\$[0.005,0.033]$} & \shortstack{$0.877$\\$[0.775,0.986]$} & Yes\\
Qwen2.5 L23 & SA / P0 & \shortstack{$0.001$\\$[-0.018,0.022]$} & \shortstack{$0.393$\\$[0.307,0.481]$} & No\\
Qwen2.5 L5 & SB / P0 & \shortstack{$0.000$\\$[0.000,0.000]$} & \shortstack{$12.514$\\$[12.120,12.901]$} & No\\
Qwen2.5 L14 & SB / P0 & \shortstack{$0.024$\\$[-0.022,0.069]$} & \shortstack{$1.733$\\$[1.582,1.889]$} & No\\
Qwen2.5 L23 & SB / P0 & \shortstack{$-0.161$\\$[-0.231,-0.094]$} & \shortstack{$-0.188$\\$[-0.294,-0.085]$} & No\\
Qwen2.5 L5 & BA / P0 & \shortstack{$0.021$\\$[0.016,0.025]$} & \shortstack{$0.007$\\$[0.002,0.012]$} & No\\
Qwen2.5 L14 & BA / P0 & \shortstack{$0.017$\\$[0.008,0.025]$} & \shortstack{$0.365$\\$[0.311,0.418]$} & Yes\\
Qwen2.5 L23 & BA / P0 & \shortstack{$-0.00038$\\$[-0.002,0.001]$} & \shortstack{$0.002$\\$[-0.00019,0.005]$} & No\\
Qwen2.5 L5 & BB / P0 & \shortstack{$0.021$\\$[0.016,0.027]$} & \shortstack{$0.195$\\$[0.163,0.226]$} & Yes\\
Qwen2.5 L14 & BB / P0 & \shortstack{$0.010$\\$[-0.010,0.030]$} & \shortstack{$0.123$\\$[0.090,0.159]$} & No\\
Qwen2.5 L23 & BB / P0 & \shortstack{$-0.175$\\$[-0.279,-0.073]$} & \shortstack{$0.027$\\$[-0.084,0.135]$} & No\\
Qwen2.5 L5 & SA / P1 & \shortstack{$-0.995$\\$[-1.106,-0.889]$} & \shortstack{$-15.625$\\$[-16.353,-14.871]$} & No\\
Qwen2.5 L14 & SA / P1 & \shortstack{$-0.230$\\$[-0.281,-0.180]$} & \shortstack{$-2.987$\\$[-3.272,-2.709]$} & No\\
Qwen2.5 L23 & SA / P1 & \shortstack{$-0.415$\\$[-0.502,-0.327]$} & \shortstack{$-0.644$\\$[-0.799,-0.497]$} & No\\
Qwen2.5 L5 & SB / P1 & \shortstack{$0.000$\\$[0.000,0.000]$} & \shortstack{$-18.820$\\$[-19.568,-18.045]$} & No\\
Qwen2.5 L14 & SB / P1 & \shortstack{$-2.680$\\$[-2.900,-2.465]$} & \shortstack{$-4.101$\\$[-4.497,-3.700]$} & No\\
Qwen2.5 L23 & SB / P1 & \shortstack{$-3.529$\\$[-3.753,-3.282]$} & \shortstack{$-4.155$\\$[-4.394,-3.898]$} & No\\
Qwen2.5 L5 & BA / P1 & \shortstack{$0.039$\\$[0.035,0.042]$} & \shortstack{$-0.037$\\$[-0.047,-0.028]$} & Yes\\
Qwen2.5 L14 & BA / P1 & \shortstack{$-0.025$\\$[-0.041,-0.009]$} & \shortstack{$-0.686$\\$[-0.737,-0.634]$} & No\\
Qwen2.5 L23 & BA / P1 & \shortstack{$0.002$\\$[-0.003,0.007]$} & \shortstack{$-0.031$\\$[-0.040,-0.022]$} & No\\
Qwen2.5 L5 & BB / P1 & \shortstack{$0.022$\\$[0.016,0.029]$} & \shortstack{$-0.014$\\$[-0.033,0.005]$} & No\\
Qwen2.5 L14 & BB / P1 & \shortstack{$0.021$\\$[0.001,0.041]$} & \shortstack{$-0.185$\\$[-0.227,-0.146]$} & Yes\\
Qwen2.5 L23 & BB / P1 & \shortstack{$-0.459$\\$[-0.544,-0.379]$} & \shortstack{$-0.569$\\$[-0.641,-0.497]$} & No\\
Qwen2.5 L5 & SA / all & \shortstack{$-0.511$\\$[-0.584,-0.436]$} & \shortstack{$-2.942$\\$[-4.172,-1.799]$} & No\\
Qwen2.5 L14 & SA / all & \shortstack{$-0.104$\\$[-0.134,-0.077]$} & \shortstack{$-1.036$\\$[-1.269,-0.807]$} & No\\
Qwen2.5 L23 & SA / all & \shortstack{$-0.205$\\$[-0.254,-0.156]$} & \shortstack{$-0.120$\\$[-0.221,-0.024]$} & No\\
Qwen2.5 L5 & SB / all & \shortstack{$0.000$\\$[0.000,0.000]$} & \shortstack{$-2.996$\\$[-4.449,-1.662]$} & No\\
Qwen2.5 L14 & SB / all & \shortstack{$-1.315$\\$[-1.476,-1.162]$} & \shortstack{$-1.155$\\$[-1.490,-0.846]$} & No\\
Qwen2.5 L23 & SB / all & \shortstack{$-1.828$\\$[-2.023,-1.640]$} & \shortstack{$-2.152$\\$[-2.376,-1.938]$} & No\\
Qwen2.5 L5 & BA / all & \shortstack{$0.030$\\$[0.027,0.033]$} & \shortstack{$-0.015$\\$[-0.021,-0.010]$} & Yes\\
Qwen2.5 L14 & BA / all & \shortstack{$-0.004$\\$[-0.014,0.006]$} & \shortstack{$-0.160$\\$[-0.220,-0.100]$} & No\\
Qwen2.5 L23 & BA / all & \shortstack{$0.001$\\$[-0.002,0.003]$} & \shortstack{$-0.014$\\$[-0.019,-0.009]$} & No\\
Qwen2.5 L5 & BB / all & \shortstack{$0.022$\\$[0.018,0.026]$} & \shortstack{$0.090$\\$[0.070,0.111]$} & No\\
Qwen2.5 L14 & BB / all & \shortstack{$0.016$\\$[0.001,0.030]$} & \shortstack{$-0.031$\\$[-0.063,0.00049]$} & Yes\\
Qwen2.5 L23 & BB / all & \shortstack{$-0.317$\\$[-0.387,-0.249]$} & \shortstack{$-0.271$\\$[-0.346,-0.202]$} & No\\
Llama3.2 L3 & SA / P0 & \shortstack{$0.009$\\$[0.004,0.014]$} & \shortstack{$0.350$\\$[0.244,0.458]$} & Yes\\
Llama3.2 L8 & SA / P0 & \shortstack{$-0.014$\\$[-0.034,0.003]$} & \shortstack{$0.068$\\$[0.017,0.119]$} & No\\
Llama3.2 L13 & SA / P0 & \shortstack{$0.00024$\\$[-0.000021,0.001]$} & \shortstack{$0.038$\\$[0.026,0.050]$} & No\\
Llama3.2 L3 & SB / P0 & \shortstack{$-0.005$\\$[-0.034,0.023]$} & \shortstack{$2.934$\\$[2.752,3.108]$} & No\\
Llama3.2 L8 & SB / P0 & \shortstack{$0.008$\\$[-0.036,0.048]$} & \shortstack{$2.336$\\$[2.104,2.558]$} & No\\
Llama3.2 L13 & SB / P0 & \shortstack{$0.036$\\$[0.012,0.059]$} & \shortstack{$1.265$\\$[1.109,1.416]$} & Yes\\
Llama3.2 L3 & BA / P0 & \shortstack{$0.00021$\\$[-0.001,0.001]$} & \shortstack{$0.047$\\$[0.038,0.057]$} & No\\
Llama3.2 L8 & BA / P0 & \shortstack{$0.027$\\$[0.013,0.040]$} & \shortstack{$1.367$\\$[1.247,1.486]$} & Yes\\
Llama3.2 L13 & BA / P0 & \shortstack{$-0.048$\\$[-0.071,-0.028]$} & \shortstack{$0.021$\\$[-0.011,0.051]$} & No\\
Llama3.2 L3 & BB / P0 & \shortstack{$0.001$\\$[0.00027,0.002]$} & \shortstack{$0.009$\\$[0.003,0.017]$} & Yes\\
Llama3.2 L8 & BB / P0 & \shortstack{$0.039$\\$[0.029,0.050]$} & \shortstack{$0.378$\\$[0.315,0.441]$} & Yes\\
Llama3.2 L13 & BB / P0 & \shortstack{$0.005$\\$[0.004,0.007]$} & \shortstack{$1.580$\\$[1.495,1.667]$} & Yes\\
Llama3.2 L3 & SA / P1 & \shortstack{$-0.061$\\$[-0.077,-0.045]$} & \shortstack{$-5.686$\\$[-6.087,-5.271]$} & No\\
Llama3.2 L8 & SA / P1 & \shortstack{$-0.011$\\$[-0.048,0.024]$} & \shortstack{$-0.890$\\$[-1.026,-0.754]$} & No\\
Llama3.2 L13 & SA / P1 & \shortstack{$-0.014$\\$[-0.018,-0.011]$} & \shortstack{$-0.076$\\$[-0.092,-0.061]$} & No\\
Llama3.2 L3 & SB / P1 & \shortstack{$5.786$\\$[5.515,6.041]$} & \shortstack{$-0.846$\\$[-1.013,-0.679]$} & Yes\\
Llama3.2 L8 & SB / P1 & \shortstack{$0.332$\\$[0.157,0.491]$} & \shortstack{$-3.693$\\$[-4.032,-3.374]$} & Yes\\
Llama3.2 L13 & SB / P1 & \shortstack{$0.414$\\$[0.316,0.507]$} & \shortstack{$-0.568$\\$[-0.668,-0.469]$} & Yes\\
Llama3.2 L3 & BA / P1 & \shortstack{$0.020$\\$[0.017,0.022]$} & \shortstack{$-0.106$\\$[-0.118,-0.093]$} & Yes\\
Llama3.2 L8 & BA / P1 & \shortstack{$0.070$\\$[0.036,0.103]$} & \shortstack{$-0.487$\\$[-0.567,-0.410]$} & No\\
Llama3.2 L13 & BA / P1 & \shortstack{$0.00027$\\$[-0.019,0.020]$} & \shortstack{$-0.410$\\$[-0.480,-0.343]$} & No\\
Llama3.2 L3 & BB / P1 & \shortstack{$0.001$\\$[0.00033,0.002]$} & \shortstack{$-0.075$\\$[-0.079,-0.071]$} & No\\
Llama3.2 L8 & BB / P1 & \shortstack{$-0.145$\\$[-0.163,-0.127]$} & \shortstack{$-0.587$\\$[-0.680,-0.498]$} & No\\
Llama3.2 L13 & BB / P1 & \shortstack{$-0.001$\\$[-0.005,0.002]$} & \shortstack{$-1.339$\\$[-1.462,-1.216]$} & No\\
Llama3.2 L3 & SA / all & \shortstack{$-0.026$\\$[-0.035,-0.017]$} & \shortstack{$-2.637$\\$[-2.963,-2.317]$} & No\\
Llama3.2 L8 & SA / all & \shortstack{$-0.013$\\$[-0.033,0.006]$} & \shortstack{$-0.406$\\$[-0.491,-0.328]$} & No\\
Llama3.2 L13 & SA / all & \shortstack{$-0.007$\\$[-0.009,-0.005]$} & \shortstack{$-0.019$\\$[-0.030,-0.008]$} & No\\
Llama3.2 L3 & SB / all & \shortstack{$2.862$\\$[2.594,3.142]$} & \shortstack{$1.063$\\$[0.859,1.261]$} & Yes\\
Llama3.2 L8 & SB / all & \shortstack{$0.169$\\$[0.082,0.251]$} & \shortstack{$-0.648$\\$[-0.986,-0.316]$} & Yes\\
Llama3.2 L13 & SB / all & \shortstack{$0.223$\\$[0.170,0.274]$} & \shortstack{$0.358$\\$[0.237,0.479]$} & Yes\\
Llama3.2 L3 & BA / all & \shortstack{$0.010$\\$[0.008,0.011]$} & \shortstack{$-0.029$\\$[-0.039,-0.019]$} & Yes\\
Llama3.2 L8 & BA / all & \shortstack{$0.049$\\$[0.030,0.067]$} & \shortstack{$0.440$\\$[0.340,0.547]$} & No\\
Llama3.2 L13 & BA / all & \shortstack{$-0.024$\\$[-0.039,-0.010]$} & \shortstack{$-0.195$\\$[-0.238,-0.153]$} & No\\
Llama3.2 L3 & BB / all & \shortstack{$0.001$\\$[0.001,0.002]$} & \shortstack{$-0.033$\\$[-0.038,-0.028]$} & No\\
Llama3.2 L8 & BB / all & \shortstack{$-0.053$\\$[-0.066,-0.040]$} & \shortstack{$-0.104$\\$[-0.171,-0.035]$} & No\\
Llama3.2 L13 & BB / all & \shortstack{$0.002$\\$[0.000059,0.004]$} & \shortstack{$0.121$\\$[-0.025,0.267]$} & Yes\\
\end{longtable}
\endgroup

\FloatBarrier
\section{Instance geometry, response prediction and architectural working points}
\label{sup:s8-geometry}

\subsection{Joint endpoint reinjection and three distinct response metrics}
With statistical maps fixed, the original reinjection comparison used $d_H=\mathcal A(J_0)(H_1-H_0)$ and $d_T=\mathcal A(J_1)H_1-\mathcal A(J_0)H_0$. Let $M_{\mathrm{source}}$ retain only the changed source span and let $\Delta x$ be the actual input change, supported on that span. Natural-amplitude reinjection used $x'=x+0.25M_{\mathrm{source}}d$, and its native reference used $x_{\mathrm{ref}}=x+0.25\Delta x$. Source cosine compared the masked direction with $\Delta x$; descendant cosine and NMSE compared the two forward changes relative to the unedited baseline. For Qwen2.5/Llama3.2, the equal-source-norm budget rescaled the direction after masking to match $\lVert\Delta x\rVert$. The original GPT-2 fixed-input-budget control instead set the masked direction norm to $0.25\lVert M_{\mathrm{source}}x\rVert$ before the common $0.25$ injection dose; the main comparison uses natural amplitude throughout.

At natural amplitude, fixed-minus-joint response NMSE was $+0.223557$ $[0.176577,0.260912]$ in GPT-2 and $+0.056955$ $[0.049063,0.065128]$ in Qwen2.5, with original Holm-adjusted $p=0.0003$ and $0.0006$. Llama3.2 gave $-0.000894$ $[-0.003550,0.001688]$. Joint source cosine increased for both Qwen2.5 and Llama3.2. Llama3.2 binding response cosine nevertheless fell by $0.128533$ $[0.080808,0.179793]$, whereas its syntax response cosine increased by $0.043149$ $[-0.016431,0.103471]$. These separately paired outcomes distinguish source and downstream directional alignment (Table~\ref{tab:sup-r1-source}; Fig.~\ref{fig:sup-instance}A).

\Needspace{.35\textheight}
\begingroup\footnotesize
\setlength{\tabcolsep}{4pt}
\begin{longtable}{lrrr}
\caption{Original joint-endpoint reinjection. The first three rows report aggregate fixed-minus-joint response NMSE. Subsequent rows separate source and descendant-response cosines by task (S, syntax; B, binding), budget and update. All injections retain only the changed source span and use dose 0.25. Natural amplitude preserves the masked direction. Qwen2.5/Llama3.2 equal budgets match its norm to $\lVert\Delta x\rVert$; GPT-2 input-fixed budgets instead set it to $0.25\lVert M_{\mathrm{source}}x\rVert$ before the dose. Brackets are source-group 95\% intervals.}\label{tab:sup-r1-source}\\
\toprule
Model / task / budget & Update / comparison & Source cosine or metric & Response cosine / effect\\
\midrule
\endfirsthead
\multicolumn{4}{l}{\tablename\ \thetable\ (continued)}\\
\toprule
Model / task / budget & Update / comparison & Source cosine or metric & Response cosine / effect\\
\midrule
\endhead
\bottomrule\endfoot
GPT-2 & Both tasks / natural & Fixed minus joint NMSE & \shortstack{$0.2236$\\$[0.1766,0.2609]$}\\
Qwen2.5 & Both tasks / natural & Fixed minus joint NMSE & \shortstack{$0.0570$\\$[0.0491,0.0651]$}\\
Llama3.2 & Both tasks / natural & Fixed minus joint NMSE & \shortstack{$-0.0009$\\$[-0.0036,0.0017]$}\\
GPT-2 B input-fixed & Fixed J & Not separately reported & \shortstack{$-0.4736$\\$[-0.5571,-0.3955]$}\\
GPT-2 B nat. & Fixed J & Not separately reported & \shortstack{$-0.4748$\\$[-0.5579,-0.3969]$}\\
GPT-2 B input-fixed & Joint H/J & Not separately reported & \shortstack{$0.4779$\\$[0.4131,0.5375]$}\\
GPT-2 B nat. & Joint H/J & Not separately reported & \shortstack{$0.4653$\\$[0.4002,0.5268]$}\\
GPT-2 S input-fixed & Fixed J & Not separately reported & \shortstack{$0.0238$\\$[-0.0597,0.1060]$}\\
GPT-2 S nat. & Fixed J & Not separately reported & \shortstack{$0.0233$\\$[-0.0596,0.1052]$}\\
GPT-2 S input-fixed & Joint H/J & Not separately reported & \shortstack{$0.6888$\\$[0.6209,0.7481]$}\\
GPT-2 S nat. & Joint H/J & Not separately reported & \shortstack{$0.6870$\\$[0.6188,0.7467]$}\\
Qwen2.5 B equal & Fixed J & \shortstack{$0.0356$\\$[0.0248,0.0466]$} & \shortstack{$0.0323$\\$[0.0020,0.0641]$}\\
Qwen2.5 B equal & Joint H/J & \shortstack{$0.3741$\\$[0.3456,0.4022]$} & \shortstack{$0.3834$\\$[0.3520,0.4143]$}\\
Qwen2.5 B nat. & Fixed J & \shortstack{$0.0356$\\$[0.0248,0.0466]$} & \shortstack{$-0.1846$\\$[-0.2302,-0.1432]$}\\
Qwen2.5 B nat. & Joint H/J & \shortstack{$0.3741$\\$[0.3456,0.4022]$} & \shortstack{$0.3104$\\$[0.2780,0.3419]$}\\
Qwen2.5 S equal & Fixed J & \shortstack{$0.0458$\\$[0.0359,0.0551]$} & \shortstack{$0.0422$\\$[-0.0102,0.0923]$}\\
Qwen2.5 S equal & Joint H/J & \shortstack{$0.3765$\\$[0.3477,0.4045]$} & \shortstack{$0.6868$\\$[0.6443,0.7266]$}\\
Qwen2.5 S nat. & Fixed J & \shortstack{$0.0458$\\$[0.0359,0.0551]$} & \shortstack{$-0.0263$\\$[-0.0721,0.0182]$}\\
Qwen2.5 S nat. & Joint H/J & \shortstack{$0.3765$\\$[0.3477,0.4045]$} & \shortstack{$0.6306$\\$[0.5858,0.6729]$}\\
Llama3.2 B equal & Fixed J & \shortstack{$0.0063$\\$[0.0010,0.0115]$} & \shortstack{$0.1756$\\$[0.1474,0.2041]$}\\
Llama3.2 B equal & Joint H/J & \shortstack{$0.0404$\\$[0.0261,0.0559]$} & \shortstack{$0.1247$\\$[0.0950,0.1537]$}\\
Llama3.2 B nat. & Fixed J & \shortstack{$0.0063$\\$[0.0010,0.0115]$} & \shortstack{$0.1756$\\$[0.1298,0.2232]$}\\
Llama3.2 B nat. & Joint H/J & \shortstack{$0.0404$\\$[0.0261,0.0559]$} & \shortstack{$0.0470$\\$[0.0206,0.0721]$}\\
Llama3.2 S equal & Fixed J & \shortstack{$-0.0068$\\$[-0.0143,0.0006]$} & \shortstack{$0.0801$\\$[0.0446,0.1177]$}\\
Llama3.2 S equal & Joint H/J & \shortstack{$0.0262$\\$[0.0133,0.0387]$} & \shortstack{$0.0881$\\$[0.0538,0.1220]$}\\
Llama3.2 S nat. & Fixed J & \shortstack{$-0.0068$\\$[-0.0143,0.0006]$} & \shortstack{$-0.0534$\\$[-0.0963,-0.0112]$}\\
Llama3.2 S nat. & Joint H/J & \shortstack{$0.0262$\\$[0.0133,0.0387]$} & \shortstack{$-0.0102$\\$[-0.0476,0.0263]$}\\
\end{longtable}
\endgroup

\subsection{Synthetic endpoint derivative updates}
The module experiment used 32 syntax and 32 binding four-corner groups per model at GPT-2 L6, Qwen2.5 L14 and Llama3.2 L8. Each group supplied four factor flips. Eight full-chain updates and two amplitude budgets were measured. All three models used the changed-source mask and dose $0.25$ defined above, with the equal-source-norm budget matching $\lVert\Delta x\rVert$ after masking. Local-edge variants updated only a prespecified first, middle or last edge, using eight groups per task; other edges retained the original instance geometry.

\begin{figure}[!htbp]
\centering
\includegraphics[width=\linewidth,height=.79\textheight,keepaspectratio]{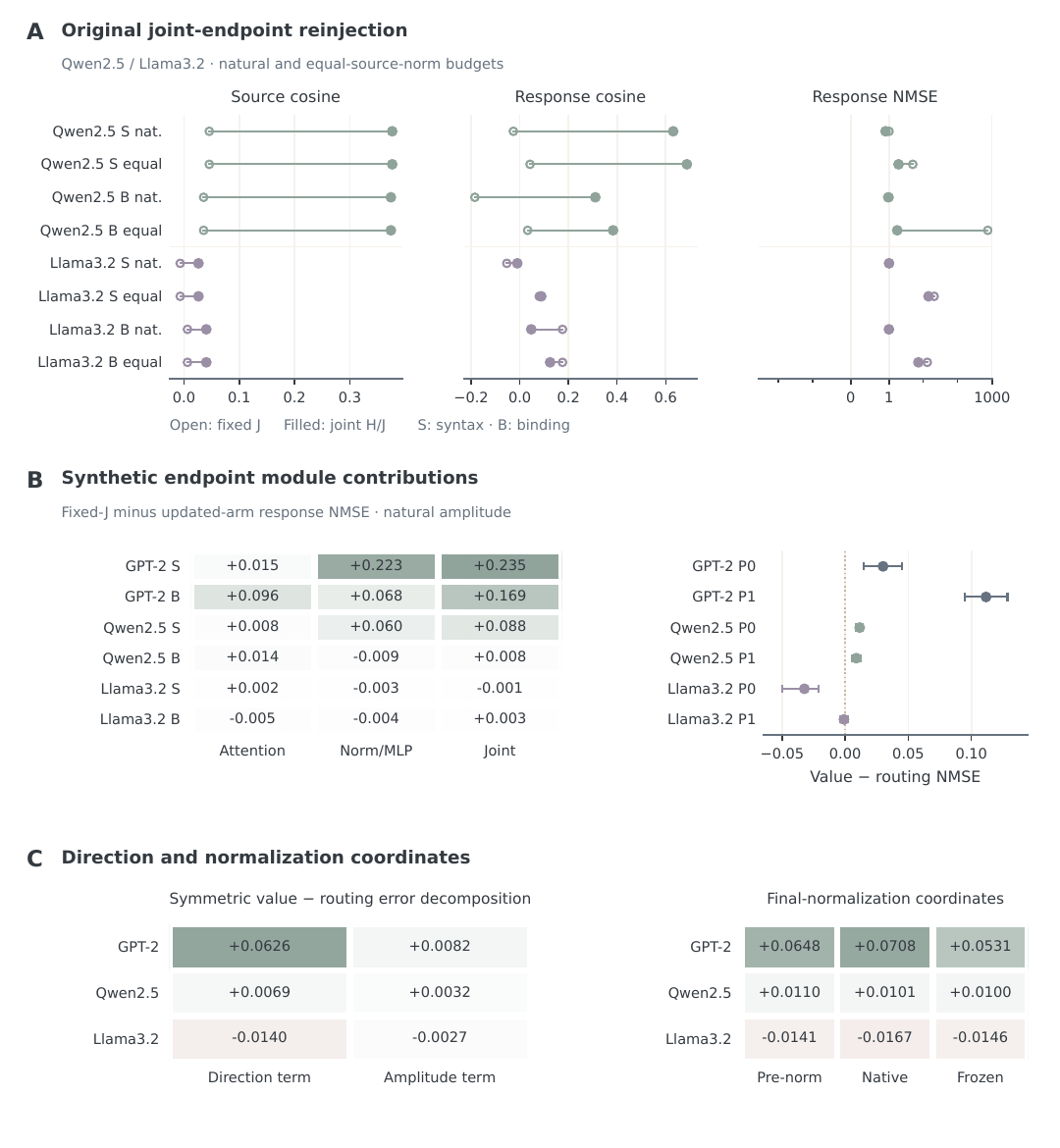}
\caption{Endpoint reinjection, synthetic module updates and layout. \textbf{A}, Original reinjection source cosine, response cosine and response NMSE in separate axes; open/filled markers denote fixed geometry/joint updates. Natural and equal-source-norm budgets are separately labeled; the NMSE axis uses a symmetric logarithmic scale with a linear region up to 1. \textbf{B}, Synthetic endpoint fixed-minus-arm NMSE by task and module, followed by layout-specific binding value-minus-routing contrasts. S/B denote syntax/binding. \textbf{C}, Direction/amplitude decomposition and final-normalization coordinates of the binding contrast. Natural forward and fixed-denominator measurements are separate conditions. Intervals are 95\% source-group bootstrap intervals.}
\label{fig:sup-instance}
\end{figure}

Attention and normalization/MLP updates were combined within the source-measurement operator. Splitting attention as $\delta(PV)=(\delta P)V+P(\delta V)$ defined a routing endpoint containing the current $V$ coefficient and a value endpoint containing the current $P$ coefficient. Synthetic source directions were reinjected into native model forward computation. Table~\ref{tab:sup-d6-directions} reports both budgets and all eight full-chain updates, retaining source cosine, response cosine and response NMSE separately. Figure~\ref{fig:sup-instance}B summarizes the task-dependent module contrast.

For binding at natural amplitude, value-minus-routing NMSE was $+0.070782$ $[0.052651,0.089391]$, $+0.010108$ $[0.008041,0.012168]$ and $-0.016694$ $[-0.026928,-0.008398]$ in GPT-2, Qwen2.5 and Llama3.2. Syntax gave $-0.05231$ $[-0.2458,0.06839]$, $+0.008886$ $[-0.001266,0.01967]$ and $+0.000141$ $[-0.006615,0.007317]$. Thus the clear natural-amplitude routing/value ordering was specific to the binding comparison.

For each branch response $r$ and reference $t$, let $\rho=\lVert r\rVert/\lVert t\rVert$ and $c=\cos(r,t)$. The symmetric value-minus-routing decomposition is
\begin{align}
\Delta E_{\mathrm{amplitude}}&=(\rho_v-\rho_r)(\rho_v+\rho_r-c_v-c_r),\\
\Delta E_{\mathrm{direction}}&=-(c_v-c_r)(\rho_v+\rho_r).
\end{align}
The terms sum to the NMSE difference per flip, before source-group averaging. Their direction terms were $+0.062568$, $+0.006861$ and $-0.014037$; amplitude terms were $+0.008215$, $+0.003247$ and $-0.002657$. The Llama3.2 direction interval was $[-0.02060,-0.00791]$, while its amplitude interval crossed zero. Approximately $84\%$ of its mean difference belonged to the directional term of this algebraic decomposition.

\subsection{Layout and final normalization}
The binding P0/P1 strata each contained 16 distinct source groups and balanced early/late queried facts. Attribute positions were $\{2,8\}$ in P0 and $\{7,11\}$ in P1; query and prediction stayed at q14/q15. Llama3.2 value-minus-routing NMSE was $-0.032367$ $[-0.049873,-0.020976]$ in P0 and $-0.001020$ $[-0.004324,0.002405]$ in P1. The P0-minus-P1 contrast was $-0.031347$ $[-0.049146,-0.019308]$. Corresponding source-cosine differences were $+0.082370$ and $-0.043444$, while response-cosine differences were $+0.344906$ and $+0.001922$. The source and response orderings therefore separate within the P1 layout.

The aggregate error signs were already present before final normalization: value-minus-routing NMSE was $+0.064810$, $+0.011011$ and $-0.014103$ for GPT-2, Qwen2.5 and Llama3.2. Freezing the baseline normalization denominator gave $+0.053058$, $+0.009958$ and $-0.014597$, respectively. These diagnostics evaluate branch and reference responses in matched output coordinates. Layout strata encompass source position, spacing and context, and the synthetic geometry updates identify their conditional effect on the extracted directions and descendant responses.

\begingroup\footnotesize
\setlength{\tabcolsep}{4pt}
\begin{longtable}{lrrrr}
\caption{Synthetic endpoint updates under both source-amplitude budgets. Each model/task contains 32 source groups; four flips are averaged before source-group bootstrap. Source cosine, descendant cosine and descendant NMSE are distinct estimands. Attention, normalization/MLP and routing/value terms update complete derivative endpoints. Local-edge variants (eight groups per task), layout contrasts and final-normalization comparisons are retained in the associated source data and text.}\label{tab:sup-d6-directions}\\
\toprule
Model / task & Budget / update & Response NMSE & Source cosine & Response cosine\\
\midrule
\endfirsthead
\multicolumn{5}{l}{\tablename\ \thetable\ (continued)}\\
\toprule
Model / task & Budget / update & Response NMSE & Source cosine & Response cosine\\
\midrule
\endhead
\bottomrule\endfoot
GPT-2 B & equal / Attention & \shortstack{$0.670$\\$[0.614,0.734]$} & \shortstack{$0.154$\\$[0.145,0.164]$} & \shortstack{$0.697$\\$[0.625,0.764]$}\\
GPT-2 B & equal / Fixed J & \shortstack{$1.321$\\$[1.227,1.417]$} & \shortstack{$0.007$\\$[0.003,0.011]$} & \shortstack{$-0.366$\\$[-0.491,-0.233]$}\\
GPT-2 B & equal / Joint H/J & \shortstack{$0.546$\\$[0.507,0.586]$} & \shortstack{$0.329$\\$[0.320,0.336]$} & \shortstack{$0.816$\\$[0.765,0.865]$}\\
GPT-2 B & equal / Norm/MLP & \shortstack{$0.837$\\$[0.757,0.926]$} & \shortstack{$0.237$\\$[0.229,0.246]$} & \shortstack{$0.513$\\$[0.418,0.607]$}\\
GPT-2 B & equal / Routing & \shortstack{$0.762$\\$[0.664,0.876]$} & \shortstack{$0.176$\\$[0.166,0.186]$} & \shortstack{$0.589$\\$[0.498,0.679]$}\\
GPT-2 B & equal / Routing + Norm & \shortstack{$0.607$\\$[0.557,0.657]$} & \shortstack{$0.329$\\$[0.321,0.337]$} & \shortstack{$0.734$\\$[0.671,0.795]$}\\
GPT-2 B & equal / Value & \shortstack{$1.175$\\$[1.073,1.286]$} & \shortstack{$-0.018$\\$[-0.024,-0.013]$} & \shortstack{$-0.015$\\$[-0.130,0.102]$}\\
GPT-2 B & equal / Value + Norm & \shortstack{$0.756$\\$[0.692,0.829]$} & \shortstack{$0.233$\\$[0.225,0.241]$} & \shortstack{$0.617$\\$[0.536,0.692]$}\\
GPT-2 B & nat. / Attention & \shortstack{$0.928$\\$[0.918,0.939]$} & \shortstack{$0.154$\\$[0.145,0.164]$} & \shortstack{$0.556$\\$[0.482,0.630]$}\\
GPT-2 B & nat. / Fixed J & \shortstack{$1.025$\\$[1.016,1.034]$} & \shortstack{$0.007$\\$[0.003,0.011]$} & \shortstack{$-0.421$\\$[-0.533,-0.299]$}\\
GPT-2 B & nat. / Joint H/J & \shortstack{$0.856$\\$[0.834,0.877]$} & \shortstack{$0.329$\\$[0.320,0.336]$} & \shortstack{$0.663$\\$[0.601,0.724]$}\\
GPT-2 B & nat. / Norm/MLP & \shortstack{$0.957$\\$[0.935,0.980]$} & \shortstack{$0.237$\\$[0.229,0.246]$} & \shortstack{$0.367$\\$[0.266,0.468]$}\\
GPT-2 B & nat. / Routing & \shortstack{$0.940$\\$[0.924,0.956]$} & \shortstack{$0.176$\\$[0.166,0.186]$} & \shortstack{$0.480$\\$[0.390,0.571]$}\\
GPT-2 B & nat. / Routing + Norm & \shortstack{$0.876$\\$[0.848,0.903]$} & \shortstack{$0.329$\\$[0.321,0.337]$} & \shortstack{$0.529$\\$[0.451,0.611]$}\\
GPT-2 B & nat. / Value & \shortstack{$1.011$\\$[1.007,1.015]$} & \shortstack{$-0.018$\\$[-0.024,-0.013]$} & \shortstack{$-0.171$\\$[-0.274,-0.064]$}\\
GPT-2 B & nat. / Value + Norm & \shortstack{$0.938$\\$[0.921,0.955]$} & \shortstack{$0.233$\\$[0.225,0.241]$} & \shortstack{$0.505$\\$[0.418,0.590]$}\\
GPT-2 S & equal / Attention & \shortstack{$3.533$\\$[1.915,5.677]$} & \shortstack{$0.205$\\$[0.189,0.221]$} & \shortstack{$0.336$\\$[0.200,0.461]$}\\
GPT-2 S & equal / Fixed J & \shortstack{$3.605$\\$[2.785,4.506]$} & \shortstack{$0.020$\\$[0.011,0.028]$} & \shortstack{$-0.076$\\$[-0.212,0.070]$}\\
GPT-2 S & equal / Joint H/J & \shortstack{$2.836$\\$[1.285,5.040]$} & \shortstack{$0.521$\\$[0.509,0.534]$} & \shortstack{$0.738$\\$[0.647,0.821]$}\\
GPT-2 S & equal / Norm/MLP & \shortstack{$2.173$\\$[1.483,2.989]$} & \shortstack{$0.486$\\$[0.473,0.499]$} & \shortstack{$0.652$\\$[0.556,0.745]$}\\
GPT-2 S & equal / Routing & \shortstack{$4.302$\\$[2.184,6.951]$} & \shortstack{$0.193$\\$[0.176,0.210]$} & \shortstack{$0.271$\\$[0.151,0.394]$}\\
GPT-2 S & equal / Routing + Norm & \shortstack{$3.004$\\$[1.311,5.390]$} & \shortstack{$0.524$\\$[0.512,0.537]$} & \shortstack{$0.729$\\$[0.641,0.807]$}\\
GPT-2 S & equal / Value & \shortstack{$3.008$\\$[2.310,3.782]$} & \shortstack{$0.061$\\$[0.049,0.073]$} & \shortstack{$0.059$\\$[-0.099,0.211]$}\\
GPT-2 S & equal / Value + Norm & \shortstack{$2.182$\\$[1.500,2.995]$} & \shortstack{$0.478$\\$[0.466,0.492]$} & \shortstack{$0.664$\\$[0.564,0.760]$}\\
GPT-2 S & nat. / Attention & \shortstack{$0.998$\\$[0.895,1.150]$} & \shortstack{$0.205$\\$[0.189,0.221]$} & \shortstack{$0.327$\\$[0.200,0.448]$}\\
GPT-2 S & nat. / Fixed J & \shortstack{$1.014$\\$[0.997,1.030]$} & \shortstack{$0.020$\\$[0.011,0.028]$} & \shortstack{$-0.068$\\$[-0.206,0.077]$}\\
GPT-2 S & nat. / Joint H/J & \shortstack{$0.779$\\$[0.573,1.052]$} & \shortstack{$0.521$\\$[0.509,0.534]$} & \shortstack{$0.729$\\$[0.640,0.804]$}\\
GPT-2 S & nat. / Norm/MLP & \shortstack{$0.791$\\$[0.688,0.911]$} & \shortstack{$0.486$\\$[0.473,0.499]$} & \shortstack{$0.628$\\$[0.532,0.722]$}\\
GPT-2 S & nat. / Routing & \shortstack{$1.051$\\$[0.921,1.251]$} & \shortstack{$0.193$\\$[0.176,0.210]$} & \shortstack{$0.278$\\$[0.156,0.405]$}\\
GPT-2 S & nat. / Routing + Norm & \shortstack{$0.806$\\$[0.597,1.078]$} & \shortstack{$0.524$\\$[0.512,0.537]$} & \shortstack{$0.708$\\$[0.623,0.784]$}\\
GPT-2 S & nat. / Value & \shortstack{$0.999$\\$[0.967,1.032]$} & \shortstack{$0.061$\\$[0.049,0.073]$} & \shortstack{$0.075$\\$[-0.077,0.227]$}\\
GPT-2 S & nat. / Value + Norm & \shortstack{$0.781$\\$[0.674,0.911]$} & \shortstack{$0.478$\\$[0.466,0.492]$} & \shortstack{$0.653$\\$[0.558,0.746]$}\\
Qwen2.5 B & equal / Attention & \shortstack{$1.432$\\$[1.276,1.602]$} & \shortstack{$0.190$\\$[0.178,0.201]$} & \shortstack{$0.372$\\$[0.326,0.420]$}\\
Qwen2.5 B & equal / Fixed J & \shortstack{$20.990$\\$[13.569,30.313]$} & \shortstack{$0.058$\\$[0.034,0.082]$} & \shortstack{$-0.078$\\$[-0.137,-0.026]$}\\
Qwen2.5 B & equal / Joint H/J & \shortstack{$2.856$\\$[2.536,3.229]$} & \shortstack{$0.355$\\$[0.306,0.402]$} & \shortstack{$0.160$\\$[0.093,0.227]$}\\
Qwen2.5 B & equal / Norm/MLP & \shortstack{$3.843$\\$[3.447,4.265]$} & \shortstack{$0.308$\\$[0.251,0.365]$} & \shortstack{$-0.020$\\$[-0.085,0.049]$}\\
Qwen2.5 B & equal / Routing & \shortstack{$1.540$\\$[1.376,1.721]$} & \shortstack{$0.220$\\$[0.210,0.230]$} & \shortstack{$0.326$\\$[0.281,0.374]$}\\
Qwen2.5 B & equal / Routing + Norm & \shortstack{$3.018$\\$[2.703,3.384]$} & \shortstack{$0.344$\\$[0.294,0.391]$} & \shortstack{$0.136$\\$[0.072,0.200]$}\\
Qwen2.5 B & equal / Value & \shortstack{$4.087$\\$[3.353,5.015]$} & \shortstack{$-0.084$\\$[-0.102,-0.067]$} & \shortstack{$0.035$\\$[-0.015,0.086]$}\\
Qwen2.5 B & equal / Value + Norm & \shortstack{$3.654$\\$[3.251,4.084]$} & \shortstack{$0.320$\\$[0.264,0.376]$} & \shortstack{$0.002$\\$[-0.068,0.074]$}\\
Qwen2.5 B & nat. / Attention & \shortstack{$0.988$\\$[0.985,0.990]$} & \shortstack{$0.190$\\$[0.178,0.201]$} & \shortstack{$0.352$\\$[0.305,0.399]$}\\
Qwen2.5 B & nat. / Fixed J & \shortstack{$1.002$\\$[1.001,1.002]$} & \shortstack{$0.058$\\$[0.034,0.082]$} & \shortstack{$-0.197$\\$[-0.274,-0.122]$}\\
Qwen2.5 B & nat. / Joint H/J & \shortstack{$0.994$\\$[0.988,1.000]$} & \shortstack{$0.355$\\$[0.306,0.402]$} & \shortstack{$0.077$\\$[0.015,0.143]$}\\
Qwen2.5 B & nat. / Norm/MLP & \shortstack{$1.011$\\$[1.005,1.016]$} & \shortstack{$0.308$\\$[0.251,0.365]$} & \shortstack{$-0.148$\\$[-0.213,-0.078]$}\\
Qwen2.5 B & nat. / Routing & \shortstack{$0.990$\\$[0.988,0.992]$} & \shortstack{$0.220$\\$[0.210,0.230]$} & \shortstack{$0.316$\\$[0.270,0.364]$}\\
Qwen2.5 B & nat. / Routing + Norm & \shortstack{$0.997$\\$[0.991,1.002]$} & \shortstack{$0.344$\\$[0.294,0.391]$} & \shortstack{$0.045$\\$[-0.015,0.110]$}\\
Qwen2.5 B & nat. / Value & \shortstack{$1.000$\\$[0.999,1.001]$} & \shortstack{$-0.084$\\$[-0.102,-0.067]$} & \shortstack{$0.012$\\$[-0.049,0.075]$}\\
Qwen2.5 B & nat. / Value + Norm & \shortstack{$1.008$\\$[1.002,1.014]$} & \shortstack{$0.320$\\$[0.264,0.376]$} & \shortstack{$-0.118$\\$[-0.187,-0.044]$}\\
Qwen2.5 S & equal / Attention & \shortstack{$2.620$\\$[2.010,3.315]$} & \shortstack{$0.206$\\$[0.190,0.222]$} & \shortstack{$0.159$\\$[0.071,0.239]$}\\
Qwen2.5 S & equal / Fixed J & \shortstack{$5.891$\\$[4.275,7.719]$} & \shortstack{$0.029$\\$[0.015,0.045]$} & \shortstack{$-0.021$\\$[-0.105,0.068]$}\\
Qwen2.5 S & equal / Joint H/J & \shortstack{$1.842$\\$[1.276,2.522]$} & \shortstack{$0.365$\\$[0.327,0.400]$} & \shortstack{$0.650$\\$[0.583,0.714]$}\\
Qwen2.5 S & equal / Norm/MLP & \shortstack{$2.268$\\$[1.661,3.013]$} & \shortstack{$0.337$\\$[0.286,0.384]$} & \shortstack{$0.554$\\$[0.473,0.631]$}\\
Qwen2.5 S & equal / Routing & \shortstack{$2.255$\\$[1.705,2.967]$} & \shortstack{$0.220$\\$[0.203,0.234]$} & \shortstack{$0.209$\\$[0.131,0.287]$}\\
Qwen2.5 S & equal / Routing + Norm & \shortstack{$1.703$\\$[1.205,2.312]$} & \shortstack{$0.367$\\$[0.330,0.403]$} & \shortstack{$0.672$\\$[0.610,0.728]$}\\
Qwen2.5 S & equal / Value & \shortstack{$4.668$\\$[3.799,5.666]$} & \shortstack{$-0.002$\\$[-0.013,0.009]$} & \shortstack{$-0.027$\\$[-0.125,0.072]$}\\
Qwen2.5 S & equal / Value + Norm & \shortstack{$2.486$\\$[1.782,3.386]$} & \shortstack{$0.331$\\$[0.281,0.378]$} & \shortstack{$0.535$\\$[0.440,0.623]$}\\
Qwen2.5 S & nat. / Attention & \shortstack{$0.992$\\$[0.983,1.001]$} & \shortstack{$0.206$\\$[0.190,0.222]$} & \shortstack{$0.141$\\$[0.057,0.220]$}\\
Qwen2.5 S & nat. / Fixed J & \shortstack{$1.001$\\$[0.998,1.003]$} & \shortstack{$0.029$\\$[0.015,0.045]$} & \shortstack{$-0.085$\\$[-0.185,0.015]$}\\
Qwen2.5 S & nat. / Joint H/J & \shortstack{$0.913$\\$[0.898,0.927]$} & \shortstack{$0.365$\\$[0.327,0.400]$} & \shortstack{$0.560$\\$[0.480,0.638]$}\\
Qwen2.5 S & nat. / Norm/MLP & \shortstack{$0.941$\\$[0.927,0.954]$} & \shortstack{$0.337$\\$[0.286,0.384]$} & \shortstack{$0.443$\\$[0.355,0.526]$}\\
Qwen2.5 S & nat. / Routing & \shortstack{$0.992$\\$[0.983,1.000]$} & \shortstack{$0.220$\\$[0.203,0.234]$} & \shortstack{$0.190$\\$[0.112,0.271]$}\\
Qwen2.5 S & nat. / Routing + Norm & \shortstack{$0.912$\\$[0.897,0.925]$} & \shortstack{$0.367$\\$[0.330,0.403]$} & \shortstack{$0.573$\\$[0.499,0.646]$}\\
Qwen2.5 S & nat. / Value & \shortstack{$1.001$\\$[0.996,1.006]$} & \shortstack{$-0.002$\\$[-0.013,0.009]$} & \shortstack{$-0.058$\\$[-0.157,0.047]$}\\
Qwen2.5 S & nat. / Value + Norm & \shortstack{$0.942$\\$[0.927,0.956]$} & \shortstack{$0.331$\\$[0.281,0.378]$} & \shortstack{$0.427$\\$[0.332,0.521]$}\\
Llama3.2 B & equal / Attention & \shortstack{$6.466$\\$[5.209,7.754]$} & \shortstack{$-0.014$\\$[-0.029,0.001]$} & \shortstack{$0.072$\\$[0.014,0.128]$}\\
Llama3.2 B & equal / Fixed J & \shortstack{$12.146$\\$[10.505,13.891]$} & \shortstack{$-0.011$\\$[-0.024,0.000]$} & \shortstack{$0.118$\\$[0.076,0.159]$}\\
Llama3.2 B & equal / Joint H/J & \shortstack{$7.391$\\$[5.951,8.867]$} & \shortstack{$0.046$\\$[0.026,0.067]$} & \shortstack{$0.110$\\$[0.060,0.155]$}\\
Llama3.2 B & equal / Norm/MLP & \shortstack{$10.259$\\$[8.909,11.777]$} & \shortstack{$-0.035$\\$[-0.050,-0.020]$} & \shortstack{$0.051$\\$[-0.009,0.115]$}\\
Llama3.2 B & equal / Routing & \shortstack{$8.073$\\$[7.035,9.365]$} & \shortstack{$-0.026$\\$[-0.047,-0.004]$} & \shortstack{$-0.010$\\$[-0.070,0.052]$}\\
Llama3.2 B & equal / Routing + Norm & \shortstack{$9.485$\\$[8.046,11.026]$} & \shortstack{$0.015$\\$[0.004,0.026]$} & \shortstack{$0.026$\\$[-0.027,0.082]$}\\
Llama3.2 B & equal / Value & \shortstack{$7.557$\\$[6.140,9.108]$} & \shortstack{$-0.007$\\$[-0.014,0.001]$} & \shortstack{$0.152$\\$[0.113,0.189]$}\\
Llama3.2 B & equal / Value + Norm & \shortstack{$7.757$\\$[6.314,9.317]$} & \shortstack{$-0.004$\\$[-0.022,0.015]$} & \shortstack{$0.116$\\$[0.068,0.162]$}\\
Llama3.2 B & nat. / Attention & \shortstack{$1.003$\\$[0.999,1.008]$} & \shortstack{$-0.014$\\$[-0.029,0.001]$} & \shortstack{$-0.058$\\$[-0.131,0.011]$}\\
Llama3.2 B & nat. / Fixed J & \shortstack{$0.998$\\$[0.997,1.000]$} & \shortstack{$-0.011$\\$[-0.024,0.000]$} & \shortstack{$0.080$\\$[-0.009,0.169]$}\\
Llama3.2 B & nat. / Joint H/J & \shortstack{$0.995$\\$[0.991,1.000]$} & \shortstack{$0.046$\\$[0.026,0.067]$} & \shortstack{$0.060$\\$[0.002,0.118]$}\\
Llama3.2 B & nat. / Norm/MLP & \shortstack{$1.002$\\$[0.999,1.007]$} & \shortstack{$-0.035$\\$[-0.050,-0.020]$} & \shortstack{$0.068$\\$[-0.002,0.138]$}\\
Llama3.2 B & nat. / Routing & \shortstack{$1.011$\\$[1.005,1.020]$} & \shortstack{$-0.026$\\$[-0.047,-0.004]$} & \shortstack{$-0.074$\\$[-0.135,-0.018]$}\\
Llama3.2 B & nat. / Routing + Norm & \shortstack{$1.009$\\$[1.002,1.020]$} & \shortstack{$0.015$\\$[0.004,0.026]$} & \shortstack{$0.018$\\$[-0.041,0.072]$}\\
Llama3.2 B & nat. / Value & \shortstack{$0.994$\\$[0.990,0.997]$} & \shortstack{$-0.007$\\$[-0.014,0.001]$} & \shortstack{$0.100$\\$[0.005,0.195]$}\\
Llama3.2 B & nat. / Value + Norm & \shortstack{$0.994$\\$[0.991,0.997]$} & \shortstack{$-0.004$\\$[-0.022,0.015]$} & \shortstack{$0.152$\\$[0.083,0.222]$}\\
Llama3.2 S & equal / Attention & \shortstack{$19.639$\\$[14.219,25.866]$} & \shortstack{$0.025$\\$[0.010,0.041]$} & \shortstack{$0.038$\\$[-0.031,0.110]$}\\
Llama3.2 S & equal / Fixed J & \shortstack{$25.406$\\$[20.835,29.932]$} & \shortstack{$-0.003$\\$[-0.016,0.012]$} & \shortstack{$0.023$\\$[-0.055,0.093]$}\\
Llama3.2 S & equal / Joint H/J & \shortstack{$21.109$\\$[15.635,27.270]$} & \shortstack{$0.019$\\$[0.004,0.034]$} & \shortstack{$0.023$\\$[-0.062,0.111]$}\\
Llama3.2 S & equal / Norm/MLP & \shortstack{$16.567$\\$[13.875,19.473]$} & \shortstack{$0.010$\\$[-0.010,0.029]$} & \shortstack{$-0.015$\\$[-0.090,0.057]$}\\
Llama3.2 S & equal / Routing & \shortstack{$17.306$\\$[12.855,21.940]$} & \shortstack{$0.032$\\$[0.017,0.049]$} & \shortstack{$-0.007$\\$[-0.069,0.062]$}\\
Llama3.2 S & equal / Routing + Norm & \shortstack{$18.488$\\$[14.273,23.257]$} & \shortstack{$0.026$\\$[0.010,0.041]$} & \shortstack{$0.006$\\$[-0.084,0.099]$}\\
Llama3.2 S & equal / Value & \shortstack{$16.216$\\$[11.324,21.718]$} & \shortstack{$0.016$\\$[0.002,0.031]$} & \shortstack{$0.004$\\$[-0.080,0.083]$}\\
Llama3.2 S & equal / Value + Norm & \shortstack{$16.701$\\$[12.419,21.407]$} & \shortstack{$0.011$\\$[-0.002,0.026]$} & \shortstack{$-0.037$\\$[-0.110,0.034]$}\\
Llama3.2 S & nat. / Attention & \shortstack{$1.007$\\$[0.998,1.016]$} & \shortstack{$0.025$\\$[0.010,0.041]$} & \shortstack{$-0.017$\\$[-0.107,0.075]$}\\
Llama3.2 S & nat. / Fixed J & \shortstack{$1.009$\\$[1.003,1.014]$} & \shortstack{$-0.003$\\$[-0.016,0.012]$} & \shortstack{$-0.111$\\$[-0.210,-0.013]$}\\
Llama3.2 S & nat. / Joint H/J & \shortstack{$1.010$\\$[0.999,1.022]$} & \shortstack{$0.019$\\$[0.004,0.034]$} & \shortstack{$-0.049$\\$[-0.163,0.063]$}\\
Llama3.2 S & nat. / Norm/MLP & \shortstack{$1.012$\\$[1.004,1.019]$} & \shortstack{$0.010$\\$[-0.010,0.029]$} & \shortstack{$-0.143$\\$[-0.238,-0.049]$}\\
Llama3.2 S & nat. / Routing & \shortstack{$1.008$\\$[1.001,1.014]$} & \shortstack{$0.032$\\$[0.017,0.049]$} & \shortstack{$-0.093$\\$[-0.178,-0.003]$}\\
Llama3.2 S & nat. / Routing + Norm & \shortstack{$1.010$\\$[1.001,1.020]$} & \shortstack{$0.026$\\$[0.010,0.041]$} & \shortstack{$-0.087$\\$[-0.204,0.030]$}\\
Llama3.2 S & nat. / Value & \shortstack{$1.008$\\$[1.001,1.015]$} & \shortstack{$0.016$\\$[0.002,0.031]$} & \shortstack{$-0.084$\\$[-0.193,0.025]$}\\
Llama3.2 S & nat. / Value + Norm & \shortstack{$1.011$\\$[1.003,1.021]$} & \shortstack{$0.011$\\$[-0.002,0.026]$} & \shortstack{$-0.121$\\$[-0.227,-0.020]$}\\
\end{longtable}
\endgroup

\subsection{Local prediction: projection, amplitude and cross-fitted scale}
The prediction analysis compared global response models with local neighborhoods selected by $H$, $J$ or $H+J$. Each four-corner group supplied four directed factor changes, from factor value 0 to 1. The $H$ observation concatenated the middle-layer attention state immediately before the earlier source factor with its change under the factor flip. The $J$ observation similarly concatenated 32 fixed bilinear measurements $u_k^{\mathsf T}Jv_k$ and their changes. Independent Rademacher vectors $u_k,v_k$ had norm one over the full $16\times C$ state; the observed derivatives ran from block 4 to attention 6 in GPT-2, block 7 to attention 14 in Qwen2.5, and block 4 to attention 8 in Llama3.2 (one-based layers). Targets were the actual attention-output changes at q12--q15, flattened across positions and channels. These predictors were fitted to observed responses from the native forward computation.

Neighborhoods used squared Euclidean distance after coordinatewise centering and standardization. Each observation branch was divided by the square root of its dimension; $H+J$ concatenated the two standardized branches and divided by $\sqrt{2}$. Standardization statistics came from GPT-2's reference groups and from Qwen2.5/Llama3.2's training groups. Development-set mean sample NMSE selected ridge coefficients in $\{0.001,0.01,0.1,1,10\}$ and neighborhood sizes in $\{128,512\}$, separately by task, branch and predictor arm; every selected local arm used 512 neighbors. The ridge penalty was $n\lambda$ for a fit on $n$ samples, and confirmation used the frozen development choices. Full selected settings, numerical floors, seeds and source-group identities accompany the source package in \texttt{methods\_support/calibration\_prediction\_protocol.json}.

GPT-2 divided each task's 512 training groups into 128 reference groups and 384 fitting groups by a fixed group hash. Local reference neighbors defined a centered source-change basis of rank 16 or 64; ridge fits on local training neighbors mapped the observation to these source coordinates and then to the descendant response. The global arm fitted the response directly on all fitting groups. Random-neighborhood controls drew reference and training neighbors independently and had their own development-selected settings. Qwen2.5/Llama3.2 fitted the descendant response directly: the global arm used all training observations, and each local fit centered its neighbors' observations and responses. Its random control retained the local arm's size and ridge while sampling training neighbors uniformly without replacement. All regression solves used FP64. Source groups remained intact across training, development and confirmation.

Original binding $H+J$ global-minus-local NMSE gains were $0.02718$ in GPT-2, $0.138323$ in Qwen2.5 and $0.065118$ in Llama3.2. GPT-2 binding $J$ alone had a negative gain of approximately $-0.00905$. In syntax, Qwen2.5 $J$ alone had an uncertain local-minus-global error of $+0.011762$, while Llama3.2 $J$ alone improved error by $0.019767$. Predictor choice and task therefore matter separately from the joint-neighborhood result.

\begin{figure}[H]
\centering
\includegraphics[width=\linewidth,height=.79\textheight,keepaspectratio]{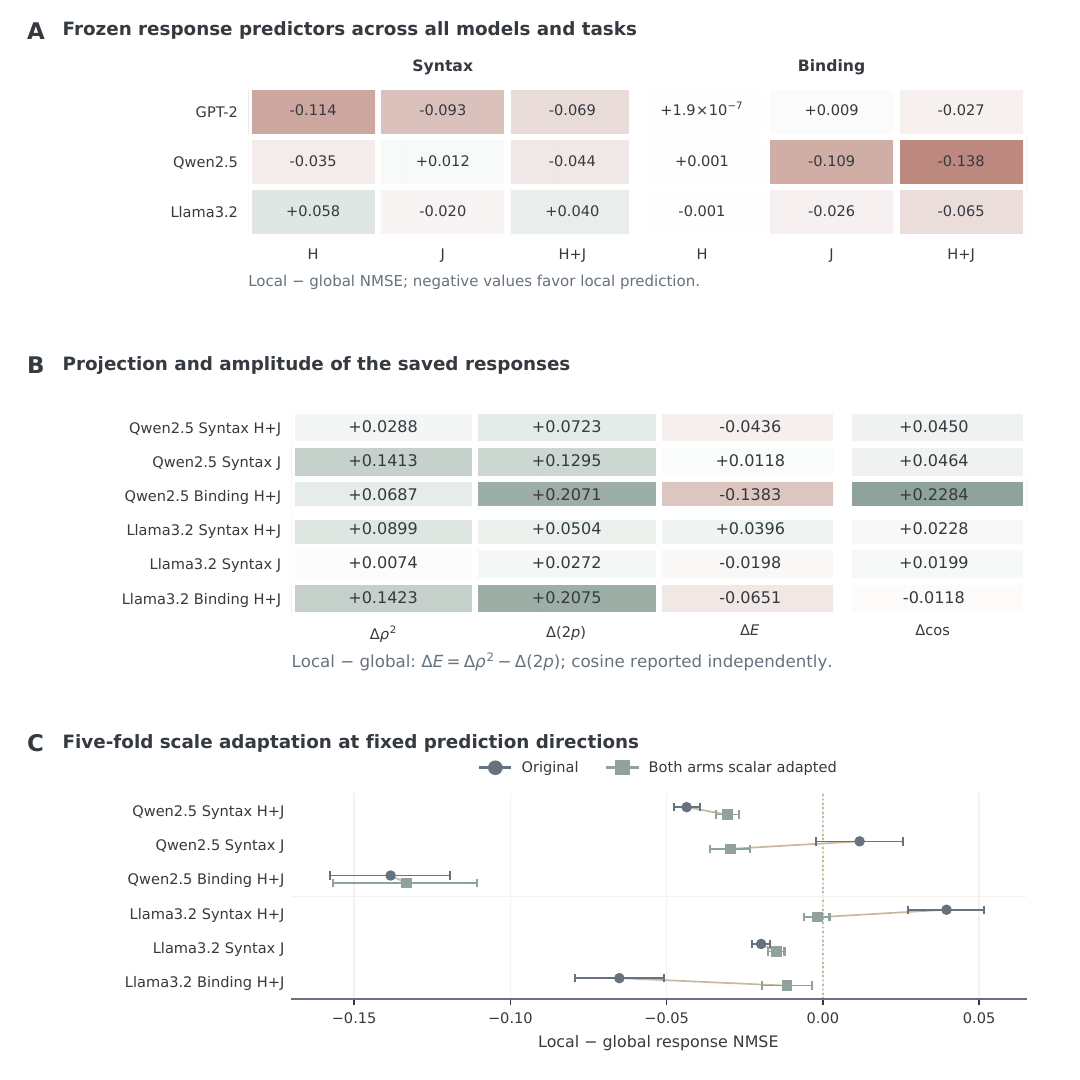}
\caption{Local response prediction separates direction from scale. \textbf{A}, Frozen H/J/H+J neighborhood comparisons across all models and tasks. Negative local-minus-global NMSE favors the local predictor. \textbf{B}, Local-minus-global changes in squared norm, twice the projection onto the native response, total NMSE and cosine. The error identity is evaluated before aggregation. \textbf{C}, Original and independently scalar-adapted local-minus-global error, with prediction directions fixed. Each arm receives one nonnegative scalar fitted outside the held-out source-group fold. Horizontal intervals are 95\% source-group bootstrap intervals with the scalars refitted in each resample.}
\label{fig:sup-local-prediction}
\end{figure}

For each saved prediction $\widehat y$ and actual response $y$, we computed
\begin{equation}
E=1+\rho^2-2p,\qquad \rho=\frac{\lVert\widehat y\rVert}{\lVert y\rVert},\quad p=\frac{\widehat y^{\mathsf T}y}{\lVert y\rVert^2}.
\end{equation}
The identity was evaluated per sample and then averaged within source groups. Llama3.2 syntax $H+J$ increased mean cosine by $0.022817$ $[0.014470,0.031320]$ while increasing error by $0.039569$; its squared-norm increase of $0.089924$ exceeded its twice-projection increase of $0.050355$. Llama3.2 binding instead reduced total error while its mean cosine changed by $-0.011775$ $[-0.018339,-0.005283]$. These two tasks jointly show why direction, projection and total error require separate measurements (Fig.~\ref{fig:sup-local-prediction}B; Table~\ref{tab:sup-c4-scale}).

Five-fold cross-fitting then held each prediction direction fixed and fitted one nonnegative scalar separately for the global and local arms. Sorted source-group identifiers were assigned cyclically to five folds. On the other four folds, the fitted scale was $\alpha=\max(0,\sum_g p_g/\sum_g\rho_g^2)$, where $p_g$ and $\rho_g^2$ are means over the factor changes within group $g$. The 9,999 source-group bootstrap resamples retained fold assignments and refitted the scales (seed 20261030). Llama3.2 syntax $H+J$ changed to local-minus-global error $-0.001754$ $[-0.005966,0.002130]$. Qwen2.5 syntax $H+J$ retained a local advantage, $-0.030534$ $[-0.034141,-0.026841]$. This supervised scale-control analysis has its own fitting budget and leaves the original frozen comparison intact (Fig.~\ref{fig:sup-local-prediction}C).

\Needspace{.25\textheight}
\begingroup\footnotesize
\setlength{\tabcolsep}{4pt}
\begin{longtable}{>{\raggedright\arraybackslash}p{.23\linewidth}>{\raggedright\arraybackslash}p{.17\linewidth}>{\raggedright\arraybackslash}p{.17\linewidth}>{\raggedright\arraybackslash}p{.17\linewidth}>{\raggedright\arraybackslash}p{.17\linewidth}}
\caption{Frozen predictor comparisons, exact response-error decomposition and scale controls. The first block reports all H/J/H+J neighborhoods and their global/local/random-neighborhood errors. The second separates squared norm, twice the response projection and cosine. The final block reports the total error after both arms receive one nonnegative scalar fitted on other source groups; bootstrap refits that scalar. These scale controls supplement the frozen original comparison.}\label{tab:sup-c4-scale}\\
\toprule
Model / task / predictor & Local $-$ global & Global NMSE & Local NMSE & Random neighbors\\
\midrule
\endfirsthead
\multicolumn{5}{l}{\tablename\ \thetable\ (continued)}\\
\toprule
Model / task / predictor & Local $-$ global & Global NMSE & Local NMSE & Random neighbors\\
\midrule
\endhead
\bottomrule\endfoot
GPT-2 Syntax H & \shortstack{$-0.1142$\\$[-0.1191,-0.1091]$} & \shortstack{$0.8601$\\$[0.8320,0.8870]$} & \shortstack{$0.7459$\\$[0.7219,0.7697]$} & \shortstack{$0.8550$\\$[0.8279,0.8812]$}\\
GPT-2 Syntax J & \shortstack{$-0.0929$\\$[-0.0998,-0.0859]$} & \shortstack{$0.8681$\\$[0.8397,0.8958]$} & \shortstack{$0.7751$\\$[0.7425,0.8078]$} & \shortstack{$0.8711$\\$[0.8428,0.8987]$}\\
GPT-2 Syntax H+J & \shortstack{$-0.0690$\\$[-0.0743,-0.0637]$} & \shortstack{$0.7869$\\$[0.7648,0.8076]$} & \shortstack{$0.7179$\\$[0.6925,0.7433]$} & \shortstack{$0.8627$\\$[0.8351,0.8896]$}\\
GPT-2 Binding H & \shortstack{$0.0000$\\$[-0.0002,0.0002]$} & \shortstack{$0.8758$\\$[0.8717,0.8798]$} & \shortstack{$0.8758$\\$[0.8718,0.8798]$} & \shortstack{$0.8803$\\$[0.8757,0.8850]$}\\
GPT-2 Binding J & \shortstack{$0.0091$\\$[0.0038,0.0144]$} & \shortstack{$0.4865$\\$[0.4655,0.5082]$} & \shortstack{$0.4956$\\$[0.4715,0.5201]$} & \shortstack{$0.8803$\\$[0.8757,0.8849]$}\\
GPT-2 Binding H+J & \shortstack{$-0.0272$\\$[-0.0319,-0.0223]$} & \shortstack{$0.5469$\\$[0.5300,0.5645]$} & \shortstack{$0.5197$\\$[0.4996,0.5400]$} & \shortstack{$0.8803$\\$[0.8757,0.8849]$}\\
Qwen2.5 Syntax H & \shortstack{$-0.0352$\\$[-0.0377,-0.0331]$} & \shortstack{$0.9997$\\$[0.9842,1.0149]$} & \shortstack{$0.9645$\\$[0.9499,0.9788]$} & \shortstack{$0.9701$\\$[0.9549,0.9848]$}\\
Qwen2.5 Syntax J & \shortstack{$0.0118$\\$[-0.0029,0.0254]$} & \shortstack{$1.0022$\\$[0.9867,1.0174]$} & \shortstack{$1.0140$\\$[0.9870,1.0410]$} & \shortstack{$1.0515$\\$[1.0275,1.0758]$}\\
Qwen2.5 Syntax H+J & \shortstack{$-0.0436$\\$[-0.0478,-0.0396]$} & \shortstack{$1.0009$\\$[0.9854,1.0161]$} & \shortstack{$0.9573$\\$[0.9394,0.9749]$} & \shortstack{$0.9674$\\$[0.9514,0.9831]$}\\
Qwen2.5 Binding H & \shortstack{$0.0011$\\$[0.0006,0.0017]$} & \shortstack{$0.9985$\\$[0.9961,1.0010]$} & \shortstack{$0.9997$\\$[0.9967,1.0026]$} & \shortstack{$1.0002$\\$[0.9973,1.0032]$}\\
Qwen2.5 Binding J & \shortstack{$-0.1093$\\$[-0.1340,-0.0843]$} & \shortstack{$0.9948$\\$[0.9921,0.9976]$} & \shortstack{$0.8855$\\$[0.8582,0.9130]$} & \shortstack{$0.9969$\\$[0.9937,1.0000]$}\\
Qwen2.5 Binding H+J & \shortstack{$-0.1383$\\$[-0.1567,-0.1199]$} & \shortstack{$0.9966$\\$[0.9941,0.9993]$} & \shortstack{$0.8583$\\$[0.8376,0.8789]$} & \shortstack{$0.9988$\\$[0.9957,1.0018]$}\\
Llama3.2 Syntax H & \shortstack{$0.0583$\\$[0.0501,0.0664]$} & \shortstack{$0.9745$\\$[0.9653,0.9838]$} & \shortstack{$1.0329$\\$[1.0160,1.0496]$} & \shortstack{$1.2587$\\$[1.2194,1.2975]$}\\
Llama3.2 Syntax J & \shortstack{$-0.0198$\\$[-0.0226,-0.0169]$} & \shortstack{$0.9416$\\$[0.9297,0.9535]$} & \shortstack{$0.9218$\\$[0.9074,0.9362]$} & \shortstack{$0.9457$\\$[0.9336,0.9579]$}\\
Llama3.2 Syntax H+J & \shortstack{$0.0396$\\$[0.0272,0.0522]$} & \shortstack{$0.9720$\\$[0.9630,0.9809]$} & \shortstack{$1.0115$\\$[0.9908,1.0322]$} & \shortstack{$1.0820$\\$[1.0579,1.1056]$}\\
Llama3.2 Binding H & \shortstack{$-0.0012$\\$[-0.0018,-0.0007]$} & \shortstack{$0.9933$\\$[0.9920,0.9946]$} & \shortstack{$0.9920$\\$[0.9903,0.9938]$} & \shortstack{$0.9965$\\$[0.9940,0.9990]$}\\
Llama3.2 Binding J & \shortstack{$-0.0259$\\$[-0.0382,-0.0135]$} & \shortstack{$0.6715$\\$[0.6464,0.6964]$} & \shortstack{$0.6456$\\$[0.6094,0.6820]$} & \shortstack{$0.6738$\\$[0.6489,0.6986]$}\\
Llama3.2 Binding H+J & \shortstack{$-0.0651$\\$[-0.0791,-0.0509]$} & \shortstack{$0.7336$\\$[0.7142,0.7528]$} & \shortstack{$0.6685$\\$[0.6361,0.7012]$} & \shortstack{$0.7359$\\$[0.7166,0.7551]$}\\
\midrule\multicolumn{5}{l}{\textit{Saved-response decomposition: local minus global}}\\
Model / task / predictor & $\Delta E$ & $\Delta\rho^2$ & $\Delta(2p)$ & $\Delta\cos$\\
Qwen2.5 Syntax H+J & \shortstack{$-0.0436$\\$[-0.0477,-0.0394]$} & \shortstack{$0.0288$\\$[0.0233,0.0343]$} & \shortstack{$0.0723$\\$[0.0656,0.0793]$} & \shortstack{$0.0450$\\$[0.0411,0.0488]$}\\
Qwen2.5 Syntax J & \shortstack{$0.0118$\\$[-0.0021,0.0253]$} & \shortstack{$0.1413$\\$[0.1337,0.1492]$} & \shortstack{$0.1295$\\$[0.1156,0.1434]$} & \shortstack{$0.0464$\\$[0.0381,0.0547]$}\\
Qwen2.5 Binding H+J & \shortstack{$-0.1383$\\$[-0.1573,-0.1191]$} & \shortstack{$0.0687$\\$[0.0640,0.0735]$} & \shortstack{$0.2071$\\$[0.1843,0.2299]$} & \shortstack{$0.2284$\\$[0.2039,0.2532]$}\\
Llama3.2 Syntax H+J & \shortstack{$0.0396$\\$[0.0275,0.0516]$} & \shortstack{$0.0899$\\$[0.0847,0.0951]$} & \shortstack{$0.0504$\\$[0.0420,0.0586]$} & \shortstack{$0.0228$\\$[0.0145,0.0313]$}\\
Llama3.2 Syntax J & \shortstack{$-0.0198$\\$[-0.0226,-0.0170]$} & \shortstack{$0.0074$\\$[0.0057,0.0091]$} & \shortstack{$0.0272$\\$[0.0232,0.0312]$} & \shortstack{$0.0199$\\$[0.0166,0.0231]$}\\
Llama3.2 Binding H+J & \shortstack{$-0.0651$\\$[-0.0795,-0.0510]$} & \shortstack{$0.1423$\\$[0.1343,0.1505]$} & \shortstack{$0.2075$\\$[0.1871,0.2278]$} & \shortstack{$-0.0118$\\$[-0.0183,-0.0053]$}\\
Qwen2.5 Syntax H+J (scaled) & \shortstack{$-0.0305$\\$[-0.0341,-0.0268]$} & Direction fixed & One scalar / arm & Five source-group folds\\
Qwen2.5 Syntax J (scaled) & \shortstack{$-0.0296$\\$[-0.0361,-0.0234]$} & Direction fixed & One scalar / arm & Five source-group folds\\
Qwen2.5 Binding H+J (scaled) & \shortstack{$-0.1332$\\$[-0.1569,-0.1106]$} & Direction fixed & One scalar / arm & Five source-group folds\\
Llama3.2 Syntax H+J (scaled) & \shortstack{$-0.0018$\\$[-0.0060,0.0021]$} & Direction fixed & One scalar / arm & Five source-group folds\\
Llama3.2 Syntax J (scaled) & \shortstack{$-0.0149$\\$[-0.0176,-0.0123]$} & Direction fixed & One scalar / arm & Five source-group folds\\
Llama3.2 Binding H+J (scaled) & \shortstack{$-0.0114$\\$[-0.0194,-0.0034]$} & Direction fixed & One scalar / arm & Five source-group folds\\
\end{longtable}
\endgroup

\Needspace{8\baselineskip}
\subsection{Architecture descriptors and numerical coordinates}
The architecture descriptors and measured coordinates in Table~\ref{tab:sup-architecture} specify candidates for controlled comparisons of conditional routing and readout. GPT-2 uses learned absolute positions and LayerNorm. Qwen2.5/Llama3.2 use rotary positions~\citep{su2024roformer} and grouped-query attention~\citep{ainslie2023gqa}; their residual blocks use RMSNorm~\citep{zhang2019rmsnorm} and SwiGLU feed-forward layers~\citep{shazeer2020glu}. Qwen2.5 has explicit Q/K/V biases, while Llama3.2 omits them. Mean projection inputs can contribute reference terms in both models.

The final prediction states of the 64 eight-condition groups (512 sentences) had mean pre-normalization residual RMS values of $22.02473$, $8.94014$ and $0.70436$ in GPT-2, Qwen2.5 and Llama3.2. The exact candidate-margin gradient had mean cosine $0.999854$, $0.999962$ and $0.999905$ with the candidate axis obtained by freezing the final normalization scale, with LayerNorm centering retained. These descriptive means locate the models at different residual scales with nearly aligned native and frozen readout axes. Table~\ref{tab:supp-clamp-numerics} directly measures the effect of the final denominator on the actual $AB$ perturbation.

At each middle and deep layer, bias size was measured relative to the bias-free projection of the native normalized input $x$: $\langle\lVert b\rVert/\lVert Wx\rVert\rangle$. Ratios were computed per state and then averaged across the query and prediction positions of all 512 sentences (1,024 states per layer). Qwen2.5's middle-layer Q bias had relative size $2.1735$, identifying a substantial learned reference direction in these states. For each paired query/key head, the attention logit is
\[
\ell_{ts}=\frac{(W_Qx_t+b_Q)^{\mathsf T}R_t^{\mathsf T}R_s(W_Kx_s+b_K)}{\sqrt{d_h}},
\]
where $R_t$ is the rotary transform at position $t$ and $d_h$ is the head dimension. In GPT-2, $R_t=I$, so the common K-bias contribution is constant across the visible keys for a fixed query and cancels in softmax. Qwen2.5's rotated bias cross-terms generally depend on relative position. V bias first supplies a constant weighted-sum offset because attention rows sum to one; this offset can affect later normalization and MLP responses through the residual baseline. These distinct entry points motivate separate Q, K and V interventions.

The short-range positional audit isolated Llama3.2's frequency scaling by retaining its head dimension and $\theta$ and replacing only the llama3 scaling rule with the unscaled RoPE spectrum. The maximum phase change was $0.01379$ rad at distance 14 and $0.01477$ rad at distance 15. This comparison quantifies the scaling rule within Llama3.2; a comparison of the full Qwen2.5 and Llama3.2 spectra also includes their different base frequencies and head dimensions.

Numerical checks verified zero/one endpoint identity, JVP/VJP adjoint consistency and summation of routing and value derivatives. Formal fixed-geometry source directions used direct $\Delta H$ evaluation, which avoids cancellation from subtracting large endpoint maps. Higher-precision sensitivity checks used FP64 weights and existing coefficients with native FP32 normalization, softmax and rotary operations. They therefore supplement the native FP32 computation as mixed-precision checks. The response replay reused saved source directions; exact error decompositions were accumulated in FP64. Llama3.2 device-placement comparisons retained cosine at least $0.999025$ with maximum absolute descendant-difference deviation $2.01\times10^{-5}$ in the tested small differences.

\Needspace{.60\textheight}
\begingroup\footnotesize
\setlength{\tabcolsep}{4pt}
\begin{longtable}{>{\raggedright\arraybackslash}p{.34\linewidth}>{\raggedright\arraybackslash}p{.19\linewidth}>{\raggedright\arraybackslash}p{.19\linewidth}>{\raggedright\arraybackslash}p{.19\linewidth}}
\caption{Architecture descriptors and measured coordinates. Final-state entries are descriptive means over 512 sentences in 64 eight-condition source groups. The cosine compares the exact candidate-margin gradient with the candidate axis under a fixed final-normalization scale; LayerNorm centering is retained. Each Q/K/V cell gives bias norm / mean relative size, computed per state using the bias-free projection of the native normalized input (1,024 query/prediction states per layer). Mixed-precision checks use FP64 weights and fixed coefficients with native FP32 normalization, softmax and rotary operations.}\label{tab:sup-architecture}\\
\toprule
Descriptor / measurement & GPT-2 & Qwen2.5 & Llama3.2\\
\midrule
\endfirsthead
\multicolumn{4}{l}{\tablename\ \thetable\ (continued)}\\
\toprule
Descriptor / measurement & GPT-2 & Qwen2.5 & Llama3.2\\
\midrule
\endhead
\bottomrule\endfoot
\multicolumn{4}{l}{\textit{Architecture}}\\*
Normalization & LayerNorm & RMSNorm & RMSNorm\\
Feed-forward module & GELU MLP & SwiGLU & SwiGLU\\
Position encoding & Learned absolute & RoPE & RoPE\\
Attention grouping & Multi-head & Grouped-query & Grouped-query\\
Explicit Q/K/V bias & Present & Present & Absent\\
\midrule\multicolumn{4}{l}{\textit{Final prediction state: descriptive means}}\\*
Residual RMS before final norm & 22.02473 & 8.94014 & 0.70436\\
Head-gradient / frozen-axis cosine & 0.999854 & 0.999962 & 0.999905\\
\midrule\multicolumn{4}{l}{\textit{Q/K/V bias: norm / mean relative size}}\\*
Middle / deep block & L6 / L10 & L14 / L23 & L8 / L13\\
Middle Q & 3.4741 / 0.0824 & 77.3878 / 2.1735 & 0.0000 / 0.0000\\
Middle K & 2.9724 / 0.0533 & 13.6338 / 0.6954 & 0.0000 / 0.0000\\
Middle V & 1.3332 / 0.0673 & 0.7056 / 0.0615 & 0.0000 / 0.0000\\
Deep Q & 6.1180 / 0.1765 & 64.4140 / 1.7425 & 0.0000 / 0.0000\\
Deep K & 2.6730 / 0.0629 & 12.2872 / 0.6200 & 0.0000 / 0.0000\\
Deep V & 0.9506 / 0.0386 & 1.2724 / 0.0784 & 0.0000 / 0.0000\\
\midrule\multicolumn{4}{l}{\textit{Numerical evaluation}}\\*
Native computation & FP32, eager & FP32, eager & FP32, eager\\
Metric decomposition & FP64 statistics & FP64 statistics & FP64 statistics\\
Derivative sensitivity & Mixed precision & Mixed precision & Mixed precision\\
Source pairing & Instance matched & Instance matched & Instance matched\\
\end{longtable}
\endgroup

\FloatBarrier

\end{document}